\documentclass[11pt]{report}

\usepackage[top=1.1in, bottom=1.2in, right=1.1in, left=1.6in]{geometry}
\usepackage{setspace}
\usepackage{graphicx}
\usepackage{amsmath}
\usepackage{amsthm}
\usepackage{amssymb}
\usepackage{parskip}
\usepackage{algorithm2e}
\usepackage{tocloft}

\usepackage{functan}

\usepackage{sidecap}

\usepackage{rotating}
\usepackage{url}

\usepackage{mathtools}

\usepackage{multirow}

\newtheorem{theorem}{Theorem}[section]
\newtheorem{lemma}{Lemma}[section]
\newtheorem{corollary}{Corollary}[section]
\newcommand{\cbar}[1]{\tilde{#1}}

\newenvironment{packed_item}{
\begin{itemize}
  \setlength{\itemsep}{1pt}
  \setlength{\parskip}{0pt}
  \setlength{\parsep}{0pt}
}{\end{itemize}}

\makeatletter
  \newcommand\tableofappendices
    {\section*{{\huge List of Appendices}}\@starttoc{toa}}
  \newcommand\l@appendix[2]
    {\par\noindent{#1}\cftdotfill{\cftdotsep}{#2}\par}
    
\newcommand{\nocontentsline}[3]{}
\newcommand{\tocless}[2]{\bgroup\let\addcontentsline=\nocontentsline#1{#2}\egroup}

\makeatother

\begin{document}

\Macro{L2}{}

\doublespacing

\pagenumbering{roman}
\input title.te
\thispagestyle{empty}
\newpage
%\pagenumbering{gobble}

% start roman numbering
\input dedication.te
\addcontentsline{toc}{chapter}{Dedication}
\newpage
\section*{Acknowledgements}

Almost all the thesis start with a thank you to the thesis advisor, why this is so. Maybe it is the time. The time one spends most with during the PhD. The time the impact from whom will last for long on shaping one's flavor of the research. My advisor, Yann LeCun, is of no double having a big influence on me. To some extent, the way we interact is like the \textit{Elastic Averaging SGD} (\textit{EASGD}) method that we have named together. During my whole PhD, I am very grateful to have the freedom to explore various research subjects. He always has patience to wait, even though sometimes there is no interesting results. It is an experiment. Sometimes, the good result is only one step away, and he has a good feeling to sense that. Sometimes, he is also very strict. Remember when I was preparing the thesis proposal, but was proposing too many directions. Everyone was worried, as the deadline was approaching. To focus and be quick. A lot of the time, the way how we think decides the question we ask and where we go. My advisor's conceptual way of thinking is still to me a mystery of arts and science.

I would also like to thank Anna Choromanska for being my collaborator and for helping me at many critical points. When we started to work on the \textit{EASGD} method, it was not very clear why we gave it a new name. The answer only became sharp when Anna questioned me countlessly why we do what we do. Luckily, we have discovered interesting answers. Chapter~\ref{sec:ch_easgd_intro}, Chapter~\ref{sec:ch_easgd_cvg} and Chapter~\ref{sec:ch_exp} of this thesis are based on our joint work.

Chapter~\ref{sec:ch_limit} is inspired by the discussion with Professor Weinan E, and his students Qianxiao Li and Cheng Tai during my visit to Princeton University. We were trying to use stochastic differential equation to analyze \textit{EASGD} method. The idea is to get the maximum insight with minimum effort. The results in Chapter~\ref{sec:ch_limit} share a similar flavor.

Chapter~\ref{sec:ch_tree} would not come into shape without the helpful discussions with Professor Jinyang Li, and her students Russell Power, Minjie Wang, Zhaoguo Wang, Christopher Mitchell and Yang Zhang. The design and the implementation of the \textit{EASGD} and \textit{EASGD Tree} are guided by their valuable experience. Also the numerical results would not be obtained without their feedback, as well as the consistent support of the NYU HPC team, in particular Shenglong Wang.

I would also like to give special thanks to Professor Rob Fergus, Margaret Wright, and David Sontag for serving in my thesis committee, as well as being my teacher during my PhD studies. Margaret was so eager to read my thesis and asked when do you expect your thesis to converge whenever I have a new version. Rob was so glad to see me when I first came to the CBLL lab, while David is always sharing his knowledge with us during the CBLL talks.

I also learn a lot from my earlier collaborators, in particular Marco Scoffier, Tom Schaul and Wan Li. Arthur Szlam, Camille Couprie, Clement Farabet, David Eigen, Dilip Krishnan, Joan Bruna, Koray Kavukcuoglu, Matt Zeiler, Olivier Henaff, Pablo Sprechmann, Pierre Sermanet, Ross Goroshin, Xiang Zhang and Y-Lan Boureau are always very mind-refreshing to be around.

I also appreciate the feedback from many senior researchers on the \textit{EASGD} method. In particular, Mark Tygert, Samy Bengio, Patrick Combettes, Peter Richtarik, Yoshua Bengio and Yuri Bakhtin.

The coach of the NYU running club, Mike Galvan, kicks me up to practice at 6:30am every Monday morning during my last year of the PhD. It gives me a consistency energy to keep running and to finish my PhD.

Almost half of my PhD time is spent at the building of Courant Institute. I have learned a lot of mathematics which is still so tremendous to me. The more you write, the less you see. There are a lot of inspiring stories that I would like to write, but maybe I should keep them in mind for the moment.

\addcontentsline{toc}{chapter}{Acknowledgements}
\newpage
\phantomsection
\addcontentsline{toc}{chapter}{Abstract}
%--abstract--
\begin{center}
\section*{Abstract}
\end{center}
We study the problem of how to distribute the training of large-scale deep learning models in the parallel computing environment. We propose a new distributed stochastic optimization method called Elastic Averaging SGD (\textit{EASGD}). We analyze the convergence rate of the \textit{EASGD} method in the synchronous scenario and compare its stability condition with the existing \textit{ADMM} method in the round-robin scheme. An asynchronous and momentum variant of the \textit{EASGD} method is applied to train deep convolutional neural networks for image classification on the \textit{CIFAR} and \textit{ImageNet} datasets. Our approach accelerates the training and furthermore achieves better test accuracy. It also requires a much smaller amount of communication than other common baseline approaches such as the \textit{DOWNPOUR} method.

We then investigate the limit in speedup of the initial and the asymptotic phase of the mini-batch \textit{SGD}, the momentum \textit{SGD}, and the \textit{EASGD} methods. We find that the spread of the input data distribution has a big impact on their initial convergence rate and stability region. We also find a surprising connection between the momentum \textit{SGD} and the \textit{EASGD} method with a negative moving average rate. A non-convex case is also studied to understand when \textit{EASGD} can get trapped by a saddle point.

Finally, we scale up the \textit{EASGD} method by using a tree structured network topology. We show empirically its advantage and challenge. We also establish a connection between the \textit{EASGD} and the \textit{DOWNPOUR} method with the classical \textit{Jacobi} and the \textit{Gauss-Seidel} method, thus unifying a class of distributed stochastic optimization methods.
%------------
\newpage
%\onehalfspacing
\tableofcontents
\newpage
\cleardoublepage
\phantomsection
\addcontentsline{toc}{chapter}{List of Figures}
\listoffigures
\newpage
\cleardoublepage
\phantomsection
\addcontentsline{toc}{chapter}{List of Tables}
\listoftables
\newpage
\cleardoublepage
\phantomsection
%\tableofappendices
%\addcontentsline{toc}{chapter}{List of Appendices} 
%\newpage
% add list of appendices here
% start regular numbering
\pagenumbering{arabic}
\chapter{Introduction}
\label{sec:ch_intro}
\section{What is the problem}

The subject of this thesis is on how to parallelize the training of large deep learning models that use a form of stochastic gradient descent (\textit{SGD})~\cite{bottou-98x}.

The classical \textit{SGD} method processes each data point sequentially on a single processor. 
As an optimization method, it often exhibits fast initial convergence toward the local optimum as compared to the batch gradient method~\cite{lecun2012efficient}.
As a learning algorithm, it often leads to better solutions in terms of the test accuracy~\cite{lecun2012efficient,hardt2015train}.

However, as the size of the dataset explodes~\cite{halevy2009unreasonable}, the amount of time taken to go through each data point sequentially becomes prohibitive. To meet this challenge, many distributed stochastic optimization methods have been proposed in literature, e.g. the mini-batch \textit{SGD}, the asynchronous \textit{SGD}, and the \textit{ADMM}-based methods.

Meanwhile, the scale and the complexity of the deep learning models is also growing to adapt to the growth of the data. There have been attempts to parallelize the training for large-scale deep learning models on thousands of CPUs, including the Google's Distbelief system~\cite{2012distbelief}.
But practical image recognition systems consist of large-scale convolutional neural networks trained on a few GPU cards sitting in a single computer~\cite{krizhevsky2012imagenet,2013arXiv1312.6229S}.
The main challenge is to devise parallel \textit{SGD} algorithms to train large-scale deep learning models that yield a significant speedup when run on multiple GPU cards across multiple machines. To date, the AlphaGo system is trained using 50 GPUs for a few weeks~\cite{silver2016mastering}.

To solve such large-scale optimization problem, one line of research is to fill-in the gap between the
stochastic gradient descent method (\textit{SGD}) and the batch gradient method~\cite{nocedal2006numerical}. The batch gradient method evaluates the gradient (in a batch) using all the data points while the stochastic gradient descent method uses only a single data point to estimate the gradient of the objective function. The mini-batch \textit{SGD}, i.e. sampling a subset of the data points to estimate the full gradient, is one possibility to fill-in this gap~\cite{cotter2011better,dekel2012optimal,NIPS2013_4894,li2014efficient}.
Larger mini-batch size reduces the variance of the stochastic gradient. Consequently, one can use a larger learning rate to gain a faster convergence in training. However, to implement it efficiently requires very skillful engineering efforts in order to minimize the communication overhead, which is difficult in particular for training large deep learning models on GPUs~\cite{NIPS2013_4894,das2016distributed}.
It is even observed that in deep learning problems, using too large mini-batch size may lead to solutions of very poor test accuracy~\cite{zhang2015staleness}.

Another possibility is to use the asynchronous stochastic gradient descent methods~\cite{doi:10.1137/S0363012995282784,langford2009slow,zinkevich2010parallelized,agarwal2011distributed,DBLPRechtRWN11,2012distbelief}.
The idea is similar to the mini-batch \textit{SGD}, i.e. to distribute the computation of the gradients to the local workers (processors) and collect them by the master (a parameter server), but asynchronously. The advantage of using asynchronous communication is that it allows local workers to communicate with the master at different time intervals, and thus it can significantly reduce the waiting time spent on the synchronization. The tradeoff is that
asynchronous behavior results in large communication delay, which can in turn slow down the convergence rate~\cite{zhang2015staleness}.

The \textit{DOWNPOUR} method belongs to the above class of asynchronous \textit{SGD} methods, and is proposed for training deep learning models~\cite{2012distbelief} . The main ingredient of the \textit{DOWNPOUR} method is to reduce the (gradient) communication overhead by running the \textit{SGD} method on each local worker for multiple steps instead of just one single step. This idea resembles the incremental gradient method~\cite{bertsekas1997new}. The merit is that each local worker can spend more time on the computation than on the communication. The disadvantage, however, is that the method is not very stable when the (gradient) communication period is large. We shall discuss this phenomenon further in Chapter~\ref{sec:ch_exp}. 

The mini-batch \textit{SGD} and the asynchronous \textit{SGD} methods that we have discussed so far can be implemented in a distributed computing environment~\cite{moritz2015sparknet,kim2016deepspark}. However, conceptually it is still centralized as there is a master server which is needed to collect all the gradients and store the latest model parameter. There's a class of the distributed optimization methods which is conceptually decentralized. It is based on the idea of consensus averaging~\cite{tsitsiklis1984distributed}. A classical problem is to compute the average value of the local clocks in a sensor network (aka. clock synchronization)~\cite{olshevsky2011convergence}. In such setting, one needs to consider how to optimize the design of the averaging network~\cite{xiao2007distributed} and to analyze the convergence rate on various networks subject to link failure~\cite{CBO9780511804458A099,duchi2012dual}. 

The \textit{ADMM} (Alternating Direction Method of Multipliers)~\cite{Boyd2011} method can also be used to solve the consensus averaging problem above. The basic idea is to decompose the objective function into several smaller ones so that each one can be solved separately in parallel and then be combined into one solution. In some sense, it is more effective because the dual Lagrangian update is used to close the primal-dual gap associated with the consensus constraints (i.e. to reach the consensus). \textit{ADMM} is also generalized to the stochastic and the asynchronous setting for solving large-scale machine learning problems~\cite{ouyang2013stochastic,icml2014c2_zhange14}.
Nevertheless, the consensus can be harder to reach,
as the oscillations from the stochastic sampling and the asynchronous behavior need to be absorbed (averaged out) by the whole system (for the constraints to be satisfied).

In this thesis, we explore another dimension of such possibility. Unlike the mini-batch \textit{SGD} and the asynchronous \textit{SGD} method, we would like to maintain the stochastic nature (oscillation) of the \textit{SGD} method as the number of workers grows. To accelerate the training of the large-scale deep learning models, in particular under communication constraints, we study instead a weak consensus reaching problem. We use a central variable to average out the noise, but we only maintain a weak coupling (consensus) between the local variables and the central variable. The goal is thus very different to the consensus averaging method and the \textit{ADMM} method.

\section{Formalizing the problem}

Consider minimizing a function $F(x)$ in a parallel computing environment~\cite{Bertsekas1989} with $p \in \mathbb{N}$ workers and a master. In this thesis we focus on the stochastic optimization problem of the following form
\begin{equation}
\min_{x} F(x):=\mathbb{E} [f(x,\xi)],
\label{eq:original}
\end{equation}
where $x$ is the model parameter to be estimated and $\xi$ is a random variable that follows the probability distribution $\mathbb{P}$ over $\Omega$ such that $F(x) = \int_{\Omega}f(x,\xi) \mathbb{P} (d\xi)$. The optimization problem in Equation~\ref{eq:original} can be reformulated as follows
\begin{equation}
\label{eq:penalty}
\min_{x^1,\ldots,x^p,\tilde{x}} \sum_{i=1}^p \mathbb{E} [f(x^i,\xi^i)] + \frac{\rho}{2}\|x^i - \tilde{x}\|^2,
\end{equation}
where each $\xi^i$ follows the same distribution $\mathbb{P}$ (thus we assume each worker can sample the entire dataset). In this thesis we refer to $x^i$'s as local variables and we refer to $\tilde{x}$ as a center variable.

The problem of the equivalence of these two objectives is studied in the literature and is known as the \textit{augmentability} or the \textit{global variable consensus} problem~\cite{hestenes1975optimization,Boyd2011}. The quadratic penalty term $\rho$ in Equation~\ref{eq:penalty} is expected to ensure that local workers will not fall into different attractors that are far away from the center variable.

We will focus on the problem of reducing the parameter communication overhead between the master and the local workers~\cite{NIPS2014_5386,2012distbelief,Yadan2013,Paine2013,onebitparallel}. The problem of data communication when the data is distributed among the workers~\cite{Bertsekas1989,Bekkerman:2011:SUM:2107736.2107740} is a more general problem and is not addressed in this work. We however emphasize that our problem setting is still highly non-trivial under the communication constraints due to the existence of many local optima~\cite{ChoromanskaHMAL15}.

We remark further a connection between the quadratic penalty term in Equation~\ref{eq:penalty} and the \textit{Moreau-Yosida} regularization~\cite{lin2015universal} in convex optimization used to smooth the non-smooth part of the objective function. When using the rectified linear units~\cite{nair2010rectified} in deep learning models, the objective function also becomes non-smooth. This suggests that it may indeed be a good idea to use such a quadratic penalty term.

\section{An overview}
We now give an overview of the main focus and results of the following chapters.

Chapter~\ref{sec:ch_easgd_intro} introduces the Elastic Averaging SGD (\textit{EASGD}) method. We first discuss the basic idea and motivate \textit{EASGD}. We then propose synchronous \textit{EASGD} and discuss its connection with the classical \textit{Jacobi} method in the numerical analysis. We further propose an asynchronous extension of \textit{EASGD} and discuss how the asynchronous \textit{EASGD} can be thought of as an perturbation of the synchronous \textit{EASGD}. We end up with combing \textit{EASGD} with the classical momentum method, notably the \textit{Nesterov}'s momentum, giving the \textit{EAMSGD}
algorithm.

Chapter~\ref{sec:ch_easgd_cvg} provides the convergence analysis of the synchronous \textit{EASGD} algorithm. The analysis is focused on the convergence of the center variable. We discuss a one-dimensional quadratic objective function first. We show that the variance of the center variable tends to zero as the number of workers grows. Meanwhile, the variance of the local variables keeps increasing. We also introduce a double averaging sequence computing the time average of the center variable and show that it is asymptotically optimal. We further extend our analysis to the strongly convex case, and discusses the tightness of the bound we have obtained compared to the quadratic case. Finally, we study the stability condition of \textit{EASGD} and \textit{ADMM} method in the round-robin scheme. We compute numerically the stability region for the \textit{ADMM} method and show that it can be unstable when the quadratic penalty term is relatively small. This is in contrast to the stability region of the \textit{EASGD} method.

Chapter~\ref{sec:ch_exp} is an empirical study on the performance of various serial and parallel stochastic optimization methods for training deep convolutional neural networks on the \textit{CIFAR} and \textit{ImageNet} dataset. We first describe the experimental setup, in particular how we preprocess and sample the dataset by the local workers in parallel. We then compare the performance of \textit{EASGD} and its momentum variant~\textit{EAMSGD} with the \textit{SGD} and \textit{DOWNPOUR} methods (including their averaging and momentum variants). We find that \textit{EASGD} is very robust when the communication period is large, which is not the case for \textit{DOWNPOUR}. Furthermore, \textit{EAMSGD} (the combination of \textit{EASGD} and Nesterov's momentum method) achieves the best performance measured both in the wallclock time and in the smallest achievable test error. The smallest achievable test error is further improved as we gradually increase the communication period and the number of processors. We provide further empirical results on the dependency of the learning rate and the communication period. To choose a proper communication period, we also give an explicit analysis in terms of the bandwidth requirement for the data communication and the parameter communication in the \textit{ImageNet} case.

Chapter~\ref{sec:ch_limit} studies the limit in speedup of various stochastic optimization methods. We would like to know what would be the limit in theory if we were given an infinite amount of processors. We study first the asymptotic phase of these methods using an additive noise model (a one-dimensional quadratic objective with Gaussian noise). The asymptotic variance can be obtained explicitly, and it can be reduced by either the mini-batch \textit{SGD} or the \textit{EASGD} method. However, the \textit{SGD} method using momentum can strictly increase this asymptotic variance. On the other hand, we seek the optimal momentum rate such that the momentum \textit{SGD} method converges the fastest. We find that this optimal momentum rate can either be positive or negative. We ask a similar question for the \textit{EASGD} method by fixing the moving rate of the center variable and then optimize the moving rate of the local variable. We find that the optimal moving (average) rate of the \textit{EASGD} method is either zero or negative. The surprising connection between these two results is that they are obtained based on a nearly same proof. We perform a similar moment analysis on the \textit{EAMSGD} method, and find (but only numerically) that the optimal moving rate can be either positive or negative, depending on the choice of the learning rate.

We then move on to study the initial phase of these methods based on a multiplicative noise model. We introduce the Gamma distribution to parametrize the spread the input data distribution. We obtain the optimal learning rate for the mini-batch \textit{SGD} method, and discuss how fast the optimal rate of convergence varies with the mini-batch size. We find that if the input data distribution has a large spread (to be made precise in Chapter~\ref{sec:ch_limit}), then one can gain more effective speedup by using mini-batch~\textit{SGD}. For the momentum \textit{SGD} method, we observe that using momentum can slow down the optimal convergence rate, but it can accelerate the convergence when the learning rate is chosen to be sub-optimal. For the \textit{EASGD} method, we observe a quite different picture to the mini-batch \textit{SGD}. There is an optimal number of workers that \textit{EASGD} method will achieve the best convergence rate. We also perform an asymptotic analysis when the number of workers is infinite, and show that the stability region can still be enlarged if the spread of the input data distribution is large.

We finally discuss a non-convex case to understand when \textit{EASGD} can get trapped by a saddle point. This is the phenomenon that we have observed in Chapter~\ref{sec:ch_exp} when the communication period is too large. We find that if the quadratic penalty term is smaller than a critical value, then the local variables can stay on both sides of a saddle point, and it is a stable configuration. This suggests that \textit{EASGD} can spend a lot of time in such configuration if the coupling between the master and the workers is too weak.

Chapter~\ref{sec:ch_tree} attempts to scale up the \textit{EASGD} method to a larger number of processors. Due to the communication constraints, we propose a tree extension of the \textit{EASGD} method. The leaf node of the tree performs the gradient descent locally and from time to time performs the elastic averaging with their parent. The root node of the tree  tracks the spatial average of the variable of its children, and in turn the average of all the leaf nodes. We perform an empirical study with two different communication schemes. The first scheme exploits the fact that faster communication can be achieved at the bottom layer (between the leaf nodes and their parent) than the upper layers. The second scheme uses a faster upward communication rate and a slower downward communication rate so that the root node can be informed of the latest information from the bottom as quick as possible. We observe that the first scheme gives better training speedup, while the second scheme gives better test accuracy. One difference compared to the asynchronous \textit{EASGD} experiment in Chapter~\ref{sec:ch_exp} is that the communication protocol between the tree nodes is fully asynchronous so as to maximize the I/O throughput.

We end up the Chapter~\ref{sec:ch_tree} by establishing a connection between the \textit{DONWPOUR} and \textit{EASGD} methods. For clarity, we focus on the synchronous scenario. These two methods can be unified by a same equation once we transform \textit{EASGD} from the Jacobi form into the Gauss-Seidel form. The difference between the two is the choice of the moving rates. A further stability analysis shows that \textit{DONWPOUR} has a very singular region for these rates which is separated from \textit{EASGD} when the number of processors is large.

The last chapter concludes the thesis with a reprise of this overview, together with some open questions and directions to follow in the future. We have also made an open source project named mpiT on github to facilitate the communication using MPI under Torch, including our implementation of \textit{DOWNPOUR} and \textit{EASGD}.

\newpage
\chapter{Elastic Averaging SGD (EASGD)}
\label{sec:ch_easgd_intro}

In this chapter we introduce the \textit{Elastic Averaging SGD} method (\textit{EASGD}) and its variants. \textit{EASGD} is motivated by the quadratic penalty
method~\cite{nocedal2006numerical}, but is re-interpreted as a parallelized extension of the averaging \textit{SGD} method~\cite{polyak1992acceleration}.
The basic idea is to let each worker maintain its own local
parameter, and the communication and coordination of work among the local workers is based on an elastic force which links the parameters they compute with a center variable stored by the master. The center variable is updated as a moving average where the average is taken in time and also in space over the parameters computed by local workers. The local variables are updated with the \textit{SGD}-based methods,
so as to keep the oscillations during the training process.

\begin{figure}[h]
\center
\includegraphics[width =0.99\textwidth]{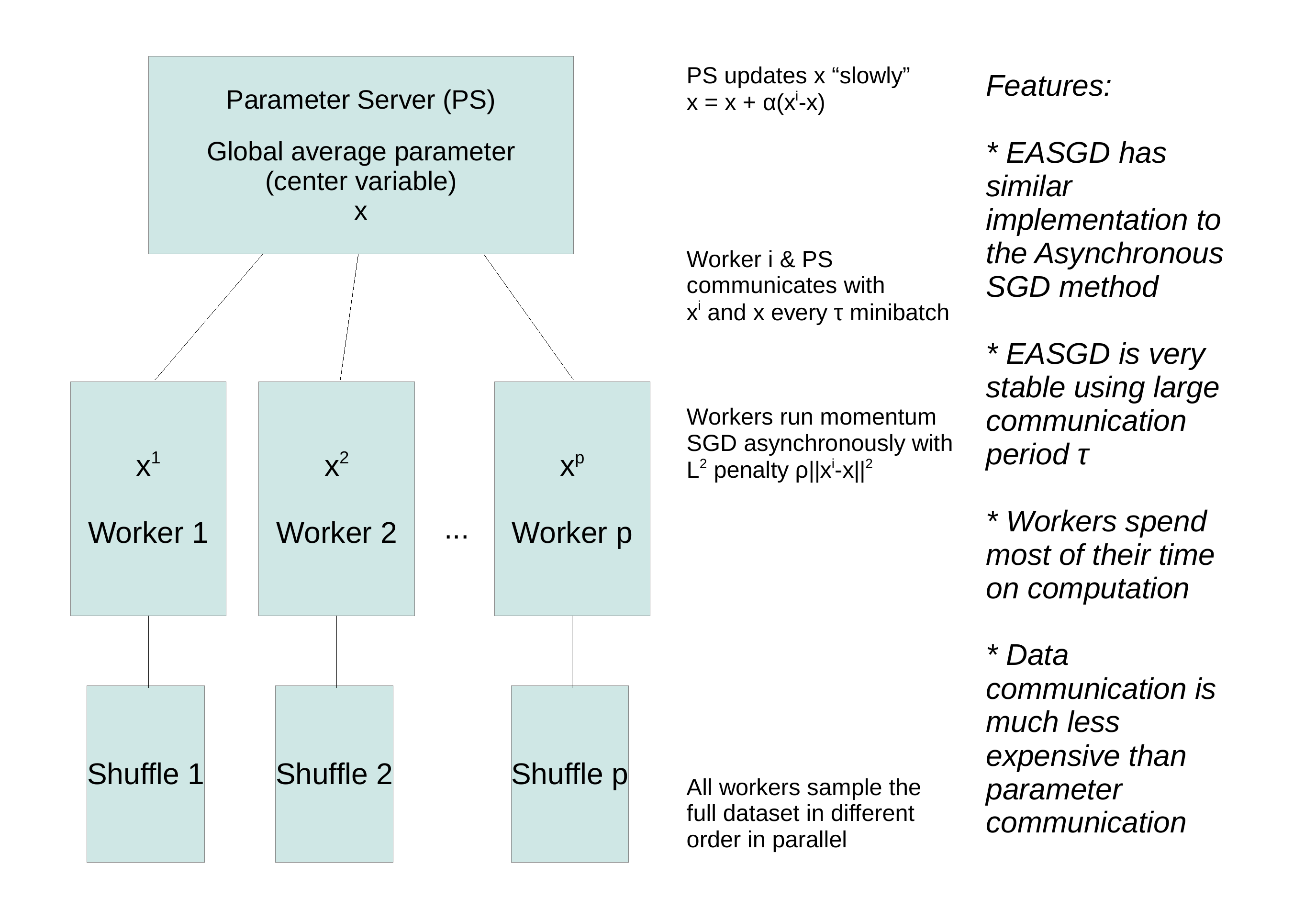}
\caption{The big picture of \textit{EASGD}.}
\label{fig:easgdbigpic}
\end{figure}

We discuss first the synchronous \textit{EASGD} in Section~\ref{sec:sync_easgd}.
Then in Section~\ref{sec:asyncEASGD} we discuss how to extend the synchronous \textit{EASGD} to the asynchronous scenario. Finally in Section~\ref{sec:momentumEASGD} we combine \textit{EASGD} with two classical momentum methods in the first-order convex optimization literature, i.e. the heavy-ball method and the \textit{Nesterov}'s momentum method, to accelerate \textit{EASGD}. The big picture is illustrated in Figure~\ref{fig:easgdbigpic}. % the local worker's \textit{SGD} update.

\section{Synchronous \textit{EASGD}}
\label{sec:sync_easgd}
The \textit{EASGD} updates captured in resp. Equation~\ref{eq:update1} and~\ref{eq:update2} are obtained by taking the gradient descent step on the objective in Equation \ref{eq:penalty} with respect to resp. variable $x^i$ and $\tilde{x}$,
\begin{eqnarray}
x^i_{t+1} &=& x^i_t - \eta (g^i_t(x_t^i) + \rho (x^i_t - \tilde{x}_t) ) \label{eq:update1}\\
\tilde{x}_{t+1} &=& \tilde{x}_t + \eta \sum_{i=1}^p \rho (x^i_t - \tilde{x}_t), \label{eq:update2}
\end{eqnarray}
where $g_t^i(x_t^i)$ denotes the stochastic gradient of $F$ with respect to $x^i$ evaluated at iteration $t$, $x^i_t$ and $\tilde{x}_t$ denote respectively the value of variables $x^{i}$ and $\tilde{x}$ at iteration $t$, and $\eta$ is the learning rate.

The update rule for the center variable $\tilde{x}$ takes the form of moving average where the average is taken over both space and time. Denote $\alpha = \eta\rho$ and $\beta=p\alpha$, then Equation~\ref{eq:update1} and~\ref{eq:update2} become
\begin{eqnarray}
x^i_{t+1} &=& x^i_t - \eta g^i_t(x_t^i) - \alpha (x^i_t - \tilde{x}_t) \label{eq:movingaverage1}\\
\tilde{x}_{t+1} &=& (1 - \beta) \tilde{x}_t + \beta \left(\frac{1}{p} \sum_{i=1}^p x^i_t\right).
\label{eq:movingaverage2}
\end{eqnarray}

Note that choosing $\beta=p \alpha$ leads to an elastic symmetry in the update rule, i.e. there exists a symmetric force equal to $\alpha(x_r^i - \tilde{x}_t)$ between the update of each $x^i$ and $\tilde{x}$.
It gives us a simple and intuitive reason for the the algorithm's stability over the \textit{ADMM} method as will be explained in Section~\ref{sec:theory_admm}. However, this relation ($\beta=p \alpha$) is by no means optimal as we shall see through our analysis in Chapter~\ref{sec:ch_limit}.

We interpret our synchronous \textit{EASGD} as an approximate model for the asynchronous \textit{EASGD}.
Thus in order to minimize the staleness~\cite{NIPS2013_4894} of the difference $x_t^i- \tilde{x}_t$ between the center and the local variable for the asynchronous \textit{EASGD} (described in Algorithm~\ref{alg:async}), the update for the master in Equation~\ref{eq:movingaverage2} involves $x_t^i$ instead of $x_{t+1}^i$. On the other hand,
we can also think of our synchronous \textit{EASGD} update rules (defined by Equation~\ref{eq:movingaverage1} and~\ref{eq:movingaverage2}) as a \textit{Jacobi} method~\cite{saad2003iterative}. We could have proposed a \textit{Gauss-Seidel} version of the synchronous \textit{EASGD} by successively updating the local and center variables with local averaging, local gradient descent, and then the global averaging. This possibility will be made precise in Chapter~\ref{sec:ch_tree}.

Note also that $\alpha = \eta \rho$, where the magnitude of $\rho$ (the quadratic penalty term in Equation \ref{eq:penalty}) represents the amount of exploration we allow in the model. In particular, small $\rho$ allows for more exploration as it allows $x^i$'s to fluctuate further from the center $\tilde{x}$.

The distinctive idea of \textit{EASGD} is to allow the local workers to perform more exploration (small $\rho$) and the master to perform exploitation. This approach differs from other settings explored in the literature~\cite{2012distbelief,DBLP:conf/icml/AzadiS14,doi:10.1137/S0363012995282784,Nedic2001381,langford2009slow,agarwal2011distributed,DBLPRechtRWN11,zinkevich2010parallelized}, and focus on how fast the center variable converges.

\section{Asynchronous \textit{EASGD}}
\label{sec:asyncEASGD}

We discussed the synchronous update of \textit{EASGD} algorithm in the previous section, where the workers update the local variables in parallel such that the $i^{\text{th}}$ worker reads the current value of the center variable and use it to update local variable $x^i$ using Equation~\ref{eq:update1}. All workers share the same global clock. The master has to wait for the $x^i$ updates from all $p$ workers before being allowed to update the value of the center variable $\tilde{x}$ according to Equation~\ref{eq:update2}.

In this section we propose its asynchronous variant. The local workers are still responsible for updating the local variables $x^i$'s, whereas the master is updating the center variable $\tilde{x}$. Each worker maintains its own clock $t^i$, which starts from $0$ and is incremented by $1$ after each stochastic gradient update of $x^{i}$ as shown in Algorithm~\ref{alg:async}. The master performs an update whenever the local workers finished $\tau$ steps of their gradient updates, where we refer to $\tau$ as the \textit{communication period}. As can be seen in Algorithm~\ref{alg:async}, whenever $\tau$ divides the local clock of the $i^{\text{th}}$ worker, the $i^{\text{th}}$ worker communicates with the master and requests the current value of the center variable $\tilde{x}$. The worker then waits until the master sends back the requested parameter value, and computes the elastic difference $\alpha (x - \tilde{x})$ (this entire procedure is captured in step a) in Algorithm~\ref{alg:async}). The elastic difference is then sent back to the master (step b) in Algorithm~\ref{alg:async}) who then updates $\tilde{x}$.

Note that the asynchronous behavior described above is partially asynchronous~\cite{Bertsekas1989}. As in the beginning of each communication period, each worker needs to read the latest parameter from the master (blocking) and then sends the elastic difference back (blocking). Although this only involves local synchronization, one can avoid this synchronization cost by using a fully asynchronous protocol such that no waiting is necessary. The fully asynchronous protocol may however increase the network traffic and the delay in the parameter communication.We shall be more precise on this fully asynchronous protocol when we discuss the \textit{EASGD Tree} algorithm in Chapter~\ref{sec:ch_tree} (Section~\ref{sec:easgd_tree}).

Recall that we have chosen the Jacobi form in our synchronous EASGD update rules (Equation~\ref{eq:movingaverage1} and~\ref{eq:movingaverage2}) as an approximate model
for the asynchronous behavior. It suggests another more efficient way to realize the partially asynchronous protocol as follows. At the beginning of each communication period, each local worker sends (non-blocking) its parameter to the master, and the master will send (non-blocking) back the elastic difference once having received that local worker's parameter. During that period of time, the local worker's computation can still make progress. At the end of that communication period (i.e. all the $\tau$ gradient updates have completed), each local worker will read (blocking) the elastic difference sent from the master, and then apply it. On the master side,
it can either sum the $p$ elastic differences altogether in one step as in the synchronous case or make an update whenever sending an elastic difference.

The communication period $\tau$ controls the frequency of the communication between every local worker and the master, and thus the trade-off between exploration and exploitation. We show demonstrate empirically in Chapter~\ref{sec:ch_exp} (Section~\ref{sec:exp_dep_tau}) that in deep learning problems, too large or too small communication period can both hurt the performance.

\noindent\makebox[\textwidth][c]{
\begin{minipage}{10cm}
\RestyleAlgo{boxruled}
\begin{algorithm}[H]
\caption{Asynchronous EASGD: \newline Processing by worker $i$ and the master}
\label{alg:async}
\begin{tabular}{l}
\textbf{Input:} learning rate $\eta$, moving rate $\alpha$,\\ $\:\:\:\:\:\:\:\:\:\:\:\:\:\:\:\:$communication period $\tau \in \mathbb{N}\:\:\:\:\:\:\:\:\:\:\:\:\:\:\:\:\:\:$\\
\textbf{Initialize:} $\tilde{x}$ is initialized randomly, $x^i = \tilde{x}$,$\:\:\:\:\:\:\:\:\:\:\:\:\:\:\:\:\:\:\:\:\:\:\:\:$\\
$\:\:\:\:\:\:\:\:\:\:\:\:\:\:\:\:\:\:\:\:\:\:\:t^i = 0$\\
\hline
\textbf{Repeat}\\
\:\:\:\:\:$x \leftarrow x^i$\\
\:\:\:\:\:\textbf{if} ($\tau $ divides $t^i $) \textbf{then}\\
\:\:\:\:\:\:\:\:\:\:\:\textbf{a)}\:$x^i \leftarrow x^i - \alpha (x - \tilde{x}) $\\
\:\:\:\:\:\:\:\:\:\:\:\textbf{b)}\:$\tilde{x} \:\:\:\!\!\leftarrow \tilde{x} \:\:\!+ \alpha (x - \tilde{x}) $\\
\:\:\:\:\:\textbf{end}\\
\:\:\:\:\:$ x^i \leftarrow x^i - \eta g^i_{t^i}(x)$\\
\:\:\:\:\:$t^i\:\:\:\:\!\!\!\!\leftarrow \:\!t^i \:\:\!\!+ 1$\\
\textbf{Until forever}
\end{tabular}
\end{algorithm}
\end{minipage}}

\section{Momentum \textit{EASGD}}
\label{sec:momentumEASGD}

The momentum EASGD (\textit{EAMSGD}) is a variant of our Algorithm~\ref{alg:async} and is captured in Algorithm~\ref{alg:asyncM}. It is based on the \textit{Nesterov}'s momentum scheme~\cite{Nesterov:2005:SMN:1058099.1058103, lan2012optimal, icml2013_sutskever13}, where the update of the local worker of the form captured in Equation~\ref{eq:update1} is replaced by the following update
\begin{eqnarray}
v_{t+1}^i &=& \delta v_t^i - \eta g^i_t(x_t^i + \delta v_t^i )\\
x^i_{t+1} &=& x^i_t + v_{t+1}^i - \eta\rho (x^i_t - \tilde{x}_t) \nonumber,
\end{eqnarray}
where $\delta$ is the momentum rate. Note that when $\delta = 0$ we recover the original \textit{EASGD} algorithm.

The idea of momentum is to accelerate the slow components in the gradient descent method.
The tradeoff is that we may slow down the components which were originally fast. We shall give an explicit example to illustrate this tradeoff in Chapter~\ref{sec:ch_limit} (Section~\ref{sec:mom_tradeoff}).

In literature, there's another well-known momentum variant called heavy-ball method (aka \textit{Polyak}'s method)~\cite{polyak1987introduction}. The analysis of its global convergence
property is still a very challenging problem in convex optimization literature~\cite{ghadimi2015global}.
If we were to combine it with \textit{EASGD}, we would have the following update
\begin{eqnarray}
v_{t+1}^i &=& \delta v_t^i - \eta g^i_t(x_t^i )\\
x^i_{t+1} &=& x^i_t + v_{t+1}^i - \eta\rho (x^i_t - \tilde{x}_t) \nonumber.
\end{eqnarray}

Note that in both cases, we do not add the momentum to the center variable. One reason is that
the momentum method has an error accumulation effect~\cite{devolder2014first}. Due to the stochastic noise in the gradient, using momentum can actually result in higher asymptotic variance (see~\cite{DBLP:journals/corr/LiTE15} and our discussion in Section~\ref{sec:msgd_lin}). The role of the center variable is indeed to reduce
the asymptotic variance.

\noindent\makebox[\textwidth][c]{
\begin{minipage}{10cm}
\RestyleAlgo{boxruled}
\begin{algorithm}[H]
\caption{Asynchronous EAMSGD: \newline Processing by worker $i$ and the master}
\label{alg:asyncM}
\begin{tabular}{l}
\textbf{Input:} learning rate $\eta$, moving rate $\alpha$,\\ $\:\:\:\:\:\:\:\:\:\:\:\:\:\:\:\:$communication period $\tau \in \mathbb{N}$,\\ $\:\:\:\:\:\:\:\:\:\:\:\:\:\:\:\:$momentum term $\delta$\\
\textbf{Initialize:} $\tilde{x}$ is initialized randomly, $x^i = \tilde{x}$,$\:\:\:\:\:\:\:\:\:\:\:\:\:\:\:\:\:\:\:\:\:\:\:\:$\\
$\:\:\:\:\:\:\:\:\:\:\:\:\:\:\:\:\:\:\:\:\:\:\:v^i = 0$, $t^i = 0$\\
\hline
\textbf{Repeat}\\
\:\:\:\:\:$x \leftarrow x^i$\\
\:\:\:\:\:\textbf{if} ($\tau $ divides $t^i $) \textbf{then}\\
\:\:\:\:\:\:\:\:\:\:\:\textbf{a)}\:$x^i \leftarrow x^i - \alpha (x - \tilde{x})$\\
\:\:\:\:\:\:\:\:\:\:\:\textbf{b)}\:$\tilde{x} \:\:\:\!\!\leftarrow \tilde{x} \:\:\!+ \alpha (x - \tilde{x}) $\\
\:\:\:\:\:\textbf{end}\\
\:\:\:\:\:$ v^i \leftarrow \delta v^i - \eta g^i_{t^i}(x + \delta v^i)$\\
\:\:\:\:\:$ x^i \leftarrow x^i + v^i$\\
\:\:\:\:\:$t^i\:\:\:\:\!\!\!\!\leftarrow \:\!t^i \:\:\!\!+ 1$\\
\textbf{Until forever}
\end{tabular}
\end{algorithm}
\end{minipage}}

\newpage
\chapter{Convergence Analysis of EASGD}
\label{sec:ch_easgd_cvg}

In this chapter, we provide the convergence analysis of the synchronous \textit{EASGD} algorithm with constant learning rate. The analysis is focused on the convergence of the center variable to the optimum. We discuss one-dimensional quadratic case first (Lemma~\ref{lem:lemm1}), then we introduce a double averaging sequence and prove that it is asymptotically optimal. For this, we provide two distinct proofs (one in Lemma~\ref{lem:lemm2} for the one-dimensional case, and the other in Lemma~\ref{lem:lemm3} for the multidimensional case). We extend the analysis to the strongly convex case as stated in Theorem~\ref{thm:strongly}. Finally, we provide stability analysis of the asynchronous \textit{EASGD} and \textit{ADMM} methods in the round-robin scheme in Section~\ref{sec:theory_admm}.

\section{Quadratic case}

Our analysis in the quadratic case extends the analysis of ASGD in~\cite{polyak1992acceleration}. Assume each of the $p$ local workers $x^i_t \in \mathbb{R}^n$ observes a noisy gradient at time $t \ge 0$ of the linear form given in Equation~\ref{eq:gradlform}.
\begin{equation}
g_t^i(x^i_t) = A x^i_t - b - \xi^i_t, \quad i \in \{1,\ldots,p\},
\label{eq:gradlform}
\end{equation}
where the matrix $A$ is positive-definite (each eigenvalue is strictly positive)
and $\{ \xi^i_t
\}$'s are i.i.d. random variables, with zero mean and positive-definite covariance matrix $\Sigma$.
Let $x^\ast $ denote the optimum solution, where $x^\ast = A^{-1} b \in \mathbb{R}^n$.

\subsection{One-dimensional case}
\label{sec:onedim}
In this section we analyze the behavior of the mean squared error (MSE) of the center variable
$\tilde{x}_t$, where this error is denoted as $\mathbb{E}[\norm{L2}{\tilde{x}_{t} - x^{*}}^2]$, as a function of $t$, $p$, $\eta$, $\alpha$ and $\beta$, where $\beta = p\alpha$. Note that the MSE error can be decomposed as (squared) bias and variance\footnote{In our notation, $\mathbb{V}$ denotes the variance.}: $\mathbb{E}[\norm{L2}{\tilde{x}_{t} - x^{*}}^2] = \norm{L2}{\mathbb{E} [\tilde{x}_{t} - x^{*}]}^2+ \mathbb{V}[\tilde{x}_{t} - x^{*}]$. For one-dimensional case ($n = 1$), we assume $A = h > 0$ and $\Sigma = \sigma^2 > 0$.

\begin{lemma}
\label{lem:lemm1}
Let $\tilde{x}_0$ and $\{x^i_0\}_{i=1,\ldots,p}$ be arbitrary constants, then
\begin{gather}
\mathbb{E} [\tilde{x}_{t} - x^\ast] =
\gamma^{t} (\tilde{x}_0 - x^\ast) + \frac{\gamma^t-\phi^t}{\gamma-\phi} \alpha u_0,
\label{eq:lem1a} \\
\mathbb{V} [\tilde{x}_{t} - x^\ast] =
\frac{p^2 \alpha^2 \eta^2 }{(\gamma-\phi)^2}
\bigg(
\frac{\gamma^2-\gamma^{2t}}{1-\gamma^2}
+\frac{\phi^2-\phi^{2t}}{1-\phi^2}
-2 \frac{\gamma \phi - (\gamma \phi)^t}{1 - \gamma \phi}\bigg)
\frac{\sigma^2}{p}, \label{eq:lem1b}
\end{gather}
where $u_0 = \sum_{i=1}^{p} ( x^i_0 - x^\ast - \frac{\alpha}{1-p\alpha-\phi} (\tilde{x}_0-x^\ast))$, $a = \eta h + (p+1)\alpha$, $c^2 = \eta h p \alpha $, $\gamma = 1-\frac{a-\sqrt{a^2-4 c^2}}{2}$, and $\phi = 1-\frac{a+\sqrt{a^2-4
c^2}}{2}$.
\end{lemma}
It follows from Lemma~\ref{lem:lemm1} that for the center variable to be stable the following has to hold
\begin{equation}
-1 < \phi < \gamma < 1.
\label{eq:condition}
\end{equation}
It can be verified that $\phi$ and $\gamma$ are the two zero-roots of the polynomial in $\lambda$: $\lambda^2 - (2-a)\lambda + (1-a+c^2)$. Recall that $\phi$ and $\lambda$ are the functions of $\eta$ and $\alpha$. Thus (see proof in Section ~\ref{sec:condition11})
\begin{itemize}
\item $\gamma < 1$ iff $c^2 > 0$ (i.e. $\eta >0$ and $ \alpha > 0$).
\item $\phi > -1$ iff $(2-\eta h)(2-p \alpha) >2 \alpha$ and $(2-\eta h) +
(2-p\alpha) > \alpha $.
\item $\phi = \gamma$ iff $a^2 = 4 c^2$ (i.e. $\eta h= \alpha =0$).
\end{itemize}

The proof the above Lemma is based on the diagonalization of the linear gradient map (this map is symmetric due to the relation $\beta = p \alpha$). The stability analysis of the asynchronous \textit{EASGD} algorithm in the round-robin scheme is similar due to this elastic symmetry.

\begin{proof}

Substituting the gradient from Equation~\ref{eq:gradlform} into the update rule used by each local worker in the synchronous \textit{EASGD} algorithm (Equation~\ref{eq:movingaverage1} and~\ref{eq:movingaverage2}) we obtain
\begin{align}
x^i_{t+1} &= x^i_{t} - \eta ( A x^i_{t} - b - \xi^i_t) - \alpha (x^i_t - \tilde{x}_t),
\label{loc:local} \\
\tilde{x}_{t+1} &= \tilde{x}_t + \sum_{i=1}^p \alpha (x^i_t - \tilde{x}_t)
\label{loc:center},
\end{align}
where $\eta$ is the learning rate, and $\alpha$ is the moving rate. Recall that $\alpha = \eta \rho$ and $A=h$.

For the ease of notation we redefine $\cbar{x}_{t}$ and $x^i_t$ as follows:
\[\cbar{x}_{t} \triangleq \cbar{x}_{t} - x^\ast \text{\:\:\:and\:\:\:} x^i_t \triangleq x^i_t - x^\ast.
\]
We prove the lemma by explicitly solving the linear equations \ref{loc:local} and
\ref{loc:center}. Let $x_{t} = (x^1_t,\ldots,x^p_t,\tilde{x}_t)^T$. We rewrite the
recursive relation captured in Equation~\ref{loc:local} and~\ref{loc:center} as simply
\[x_{t+1} = M x_t + b_t,
\]
where the drift matrix $M$ is defined as
\begin{equation*}
M=
\begin{bmatrix}
1-\alpha-\eta h & 0 & ... & 0 & \alpha \\
0 & 1-\alpha-\eta h& 0 & ... & \alpha \\
... & 0 & ... &0 & ... \\
0 & ... &0 & 1-\alpha-\eta h & \alpha \\
\alpha & \alpha & ... & \alpha & 1-p\alpha
\end{bmatrix},
\end{equation*}
and the (diffusion) vector $b_t = (\eta\xi^1_t,\ldots,\eta\xi^p_t,0)^T$.

Note that one of the eigenvalues of matrix $M$, that we call $\phi$, satisfies $(1-\alpha-\eta h-\phi)(1-p\alpha-\phi) =p\alpha^2$. The corresponding eigenvector is $(1,1,\ldots,1,-\frac{p\alpha}{1-p\alpha-\phi})^T$. Let $u_t$ be the
projection of $x_{t}$ onto this eigenvector. Thus $u_t = \sum_{i=1}^{p} ( x^i_t - \frac{\alpha}{1-p\alpha-\phi} \cbar{x}_t)$. Let furthermore $\xi_t = \sum_{i=1}^p \xi_t^i$. Therefore we have
\begin{gather}
u_{t+1} = \phi u_t + \eta \xi_t. \label{eq:proj1}
\end{gather}
By combining Equation \ref{loc:center} and \ref{eq:proj1} as follows
\begin{align*}
\tilde{x}_{t+1} &= \tilde{x}_t + \sum_{i=1}^p \alpha (x^i_t - \tilde{x}_t) = (1-p\alpha) \tilde{x}_t + \alpha (u_t + \frac{p\alpha}{1-p\alpha-\phi} \tilde{x}_t )\\
&= (1-p\alpha+\frac{p\alpha^2}{1-p\alpha-\phi}) \tilde{x}_t + \alpha u_t = \gamma \cbar{x}_t + \alpha u_t,
\end{align*}
where the last step results from the following relations: $\frac{p\alpha^2}{1-p\alpha-\phi} = 1-\alpha-\eta h-\phi $ and $\phi + \gamma = 1-\alpha-\eta h + 1-p\alpha $.
Thus we obtained
\begin{gather}
\cbar{x}_{t+1} = \gamma \cbar{x}_t + \alpha u_t. \label{eq:proj2}
\end{gather}

Based on Equation \ref{eq:proj1} and \ref{eq:proj2}, we can then expand $u_t$ and $\cbar{x}_t$
recursively,
\begin{gather}
u_{t+1} = \phi^{t+1} u_0 + \phi^{t} (\eta \xi_0) + \ldots + \phi^{0} (\eta \xi_t),
\label{eq:lem1c} \\
\cbar{x}_{t+1} = \gamma^{t+1} \cbar{x}_0 + \gamma^{t} (\alpha u_0) + \ldots +
\gamma^{0} (\alpha u_t). \label{eq:lem1d}
\end{gather}
Substituting $u_0, u_1, \dots, u_t$, each given through Equation \ref{eq:lem1c}, into Equation~\ref{eq:lem1d} we obtain
\begin{gather}
\cbar{x}_t = \gamma^t \cbar{x}_0 +
\frac{\gamma^t-\phi^t}{\gamma-\phi} \alpha u_0 +
\alpha \eta \sum_{l=1}^{t-1} \frac{\gamma^{t-l}-\phi^{t-l}}{\gamma-\phi} \xi_{l-1} .
\label{eq:lem1e}
\end{gather}

To be more specific, the Equation \ref{eq:lem1e} is obtained by interchanging the order of summation,
\begin{align*}
\cbar{x}_{t+1} &= \gamma^{t+1} \cbar{x}_0 + \sum_{i=0}^t \gamma^{t-i} (\alpha u_i) \\
&= \gamma^{t+1} \cbar{x}_0 +
\sum_{i=0}^t \gamma^{t-i} (\alpha (\phi^i u_0 + \sum_{l=0}^{i-1} \phi^{i-1-l} \eta \xi_l )) \\
			 &= \gamma^{t+1} \cbar{x}_0 +
\sum_{i=0}^t \gamma^{t-i} \phi^i (\alpha u_0) +
\sum_{l=0}^{t-1} \sum_{i=l+1}^{t} \gamma^{t-i} \phi^{i-1-l} (\alpha \eta \xi_l) \\
			 &= \gamma^{t+1} \cbar{x}_0 +
\frac{\gamma^{t+1}-\phi^{t+1}}{\gamma - \phi} (\alpha u_0) +
\sum_{l=0}^{t-1} \frac{\gamma^{t-l}-\phi^{t-l}}{\gamma-\phi} (\alpha \eta \xi_l).
\end{align*}
Since the random variables $\xi_l$ are i.i.d, we may sum the variance term by term as follows
\begin{align}
\sum_{l=0}^{t-1} \bigg( \frac{\gamma^{t-l}-\phi^{t-l}}{\gamma-\phi} \bigg)^2 &=
\sum_{l=0}^{t-1} \frac{\gamma^{2(t-l)} - 2\gamma^{t-l}\phi^{t-l} +\phi^{2(t-l)}}{(\gamma-\phi)^2} \nonumber\\
&= \frac{1}{(\gamma-\phi)^2} \bigg( \frac{\gamma^2-\gamma^{2(t+1)}}{1-\gamma^2}
- 2 \frac{\gamma \phi - (\gamma \phi)^{t+1}}{1-\gamma \phi }
+ \frac{\phi^2-\phi^{2(t+1)}}{1-\phi^2}
\bigg).
\label{eq:new}
\end{align}

Note that $\mathbb{E}[\xi_t] = \sum_{i=1}^p \mathbb{E} [\xi_t^i] = 0$ and $\mathbb{V}[\xi_t] =
\sum_{i=1}^p \mathbb{V} [\xi_t^i] = p\sigma^2$. These two facts, the equality in Equation~\ref{eq:lem1e} and Equation~\ref{eq:new} can then be used to compute $\mathbb{E}[\tilde{x}_t]$ and $\mathbb{V}[\tilde{x}_t]$ as given in Equation~\ref{eq:lem1a} and~\ref{eq:lem1b} in Lemma~\ref{lem:lemm1}.
\end{proof}

\subsubsection{Visualizing Lemma~\ref{lem:lemm1}}

\begin{figure}[htp!]
\begin{center}
\centerline{\includegraphics[trim=115 65 115
35,clip,width=1.0\columnwidth]{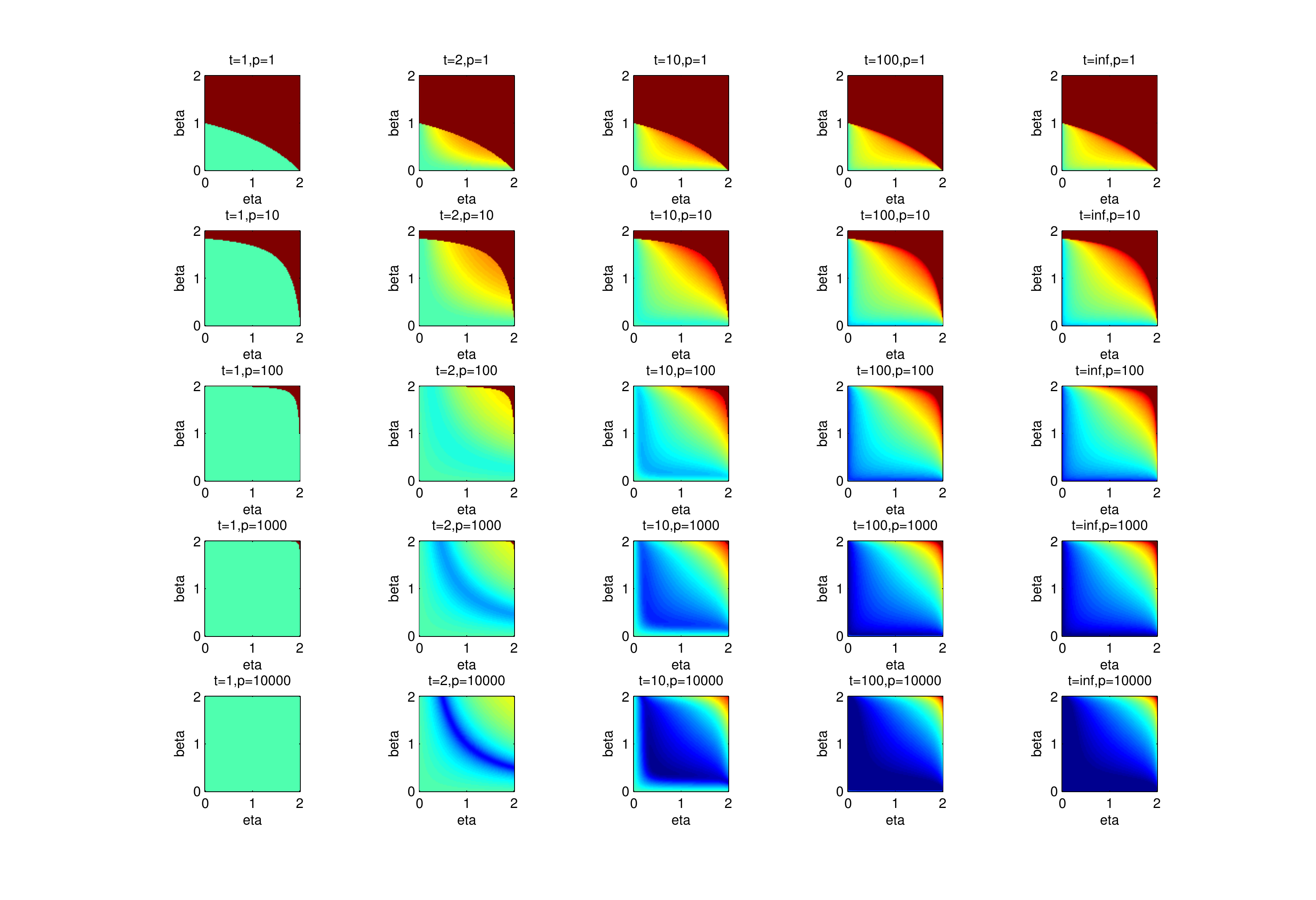}}
\caption{Theoretical mean squared error (MSE) of the center $\tilde{x}$ in the quadratic case, with
various choices of the learning rate $\eta$ (horizontal within each block), and the
moving rate $\beta=p\alpha$ (vertical within each block),
the number of processors $p=\{1,10,100,1000,10000\}$ (vertical across blocks), and the time steps $t=\{1,2,10,100,\infty\}$ (horizontal across blocks). The MSE is plotted in log
scale, ranging from $10^{-3}$ to $10^{3}$ (from deep blue to red). The dark red (i.e. on the upper-right corners) indicates divergence.}
\label{fig:easgd_grid}
\end{center}
\end{figure}

In Figure \ref{fig:easgd_grid}, we illustrate the dependence of MSE on $\beta$, $\eta$ and the number of processors $p$ over time $t$. We consider the large-noise setting where $\tilde{x}_0 = x^i_0= 1$, $h=1$ and $\sigma = 10$. The MSE error is color-coded such that the deep blue color corresponds to the MSE equal to $10^{-3}$, the green color corresponds to the MSE equal to $1$, the red color corresponds to MSE equal to $10^3$ and the dark red color corresponds to the divergence of algorithm \textit{EASGD} (condition in Equation~\ref{eq:condition} is then violated). The plot shows that
we can achieve significant variance reduction by increasing the number of local workers $p$.
This effect is less sensitive to the choice of $\beta$ and $\eta$ for large $p$.

\subsubsection{Condition in Equation~\ref{eq:condition}}
\label{sec:condition11}
We are going to show that
\begin{itemize}
\item $\gamma < 1$ iff $c^2 > 0$ (i.e. $\eta >0$ and $ \beta > 0$).
\item $\phi > -1$ iff $(2-\eta h)(2-\beta) >2\beta/p$ and $(2-\eta h) +
(2-\beta) > \beta/p $.
\item $\phi = \gamma$ iff $a^2 = 4 c^2$ (i.e. $\eta h= \beta =0$).
\end{itemize}

Recall that $a = \eta h + (p+1)\alpha$, $c^2 = \eta h p \alpha $, $\gamma = 1-\frac{a-\sqrt{a^2-4 c^2}}{2}$, $\phi = 1-\frac{a+\sqrt{a^2-4 c^2}}{2}$, and $\beta = p \alpha$.
We have
\begin{itemize}
\item
$\gamma < 1 \Leftrightarrow \frac{a-\sqrt{a^2-4 c^2}}{2} > 0 \Leftrightarrow a > \sqrt{a^2-4 c^2} \Leftrightarrow a^2 > a^2-4 c^2 \Leftrightarrow c^2 > 0 $.
\item
$\phi > -1 \Leftrightarrow 2 > \frac{a+\sqrt{a^2-4 c^2}}{2} \Leftrightarrow 4-a > \sqrt{a^2-4 c^2}
\Leftrightarrow 4-a > 0, (4-a)^2 > a^2-4 c^2 \Leftrightarrow 4-a > 0, 4 - 2a + c^2 > 0
\Leftrightarrow 4 > \eta h + \beta + \alpha , 4 - 2(\eta h + \beta + \alpha) + \eta h \beta > 0$.
\item $\phi = \gamma \Leftrightarrow \sqrt{a^2-4 c^2} = 0 \Leftrightarrow a^2 = 4c^2$.
\end{itemize}

The next corollary is a consequence of Lemma~\ref{lem:lemm1}. As the number of workers $p$ grows, the averaging property of the \textit{EASGD} can be characterized as follows

\begin{corollary}
Let the \textit{Elastic Averaging relation} $\beta = p \alpha$ and the condition \ref{eq:condition} hold,
then
\begin{gather*}
\lim_{p \rightarrow \infty}\lim_{t \rightarrow \infty} p\mathbb{E} [ (\tilde{x}_{t} - x^{*})^2] = \frac{\beta \eta h}{(2-\beta)(2-\eta h)} \cdot
\frac{2 -\beta -\eta h + \beta \eta h}{\beta + \eta h - \beta \eta h } \cdot
\frac{\sigma^2}{h^2}.
\end{gather*}
\label{cor:lemm1cor}
\end{corollary}

\begin{proof}

Note that when $\beta$ is fixed, $\lim_{p\rightarrow\infty}a = \eta h+ \beta$ and $c^2 = \eta h \beta$. Then $\lim_{p\rightarrow\infty} \phi = \min(1 - \beta,1 - \eta h)$ and $\lim_{p\rightarrow\infty} \gamma= \max(1 - \beta,1 - \eta h)$. Also note that using Lemma~\ref{lem:lemm1} we obtain
\begin{align*}
\lim_{t \rightarrow \infty} \mathbb{E} [(\tilde{x}_{t} - x^{*})^2] &=
\frac{\beta^2 \eta^2}{(\gamma-\phi)^2} \bigg( \frac{\gamma^2}{1-\gamma^2} +
\frac{\phi^2}{1-\phi^2} - \frac{2 \gamma \phi}{1- \gamma \phi} \bigg) \frac{\sigma^2}{p} \\
&=
\frac{\beta^2 \eta^2}{(\gamma-\phi)^2}
\bigg( \frac{\gamma^2(1-\phi^2)(1-\phi \gamma) + \phi^2(1-\gamma^2)(1-\phi \gamma) - 2\gamma \phi (1-\gamma^2)(1-\phi^2) }{(1-\gamma^2)(1-\phi^2)(1- \gamma \phi)} \bigg ) \frac{\sigma^2}{p} \\
&= \frac{\beta^2 \eta^2}{(\gamma-\phi)^2}
\bigg( \frac{ (\gamma-\phi)^2(1+\gamma \phi) }{(1-\gamma^2)(1-\phi^2)(1- \gamma \phi)} \bigg ) \frac{\sigma^2}{p} \\
&= \frac{\beta ^2 \eta^2}{(1-\gamma^2)(1-\phi^2)} \cdot
\frac{1+\gamma \phi}{1-\gamma \phi} \cdot
\frac{\sigma^2}{p}.
\end{align*}

Corollary~\ref{cor:lemm1cor} is obtained by plugging in the limiting values of $\phi$ and $\gamma$.
\end{proof}

The crucial point of Corollary \ref{cor:lemm1cor} is that the MSE in the limit $t \rightarrow \infty$ is in the order of $1/p$ which implies that as the number of processors $p$ grows, the MSE will decrease for the \textit{EASGD} algorithm. Also note that the smaller the $\beta$ is (recall that $\beta = p\alpha = p\eta\rho$), the more exploration is allowed (small $\rho$) and simultaneously the smaller the MSE is.

The next lemma (Lemma~\ref{lem:lemm2}) shows that \textit{EASGD} algorithm achieves the highest possible rate of convergence when we consider the double averaging sequence (similarly to~\cite{polyak1992acceleration})
$\{z_1,z_2,\dots\}$ defined as below
\begin{equation}
z_{t+1} = \frac{1}{t+1} \sum_{k=0}^{t} \tilde{x}_k.
\label{eq:doubleseq}
\end{equation}

\begin{lemma}[Weak convergence]
\label{lem:lemm2}
If the condition in Equation~\ref{eq:condition} holds, then the
normalized double averaging sequence defined in Equation~\ref{eq:doubleseq} converges weakly to the
normal distribution with zero mean and variance $\sigma^2/p h^2$,
\begin{gather}
\sqrt{t} (z_{t} - x^\ast) \rightharpoonup \mathcal{N}(0,\frac{\sigma^2}{p h^2}),
\quad t\rightarrow \infty.
\end{gather}
\end{lemma}

\begin{proof}
As in the proof of Lemma~\ref{lem:lemm1}, for the ease of notation we redefine $\cbar{x}_{t}$ and $x^i_t$ as follows:
\[\cbar{x}_{t} \triangleq \cbar{x}_{t} - x^\ast \text{\:\:\:and\:\:\:} x^i_t \triangleq x^i_t - x^\ast.
\]
Also recall that $\{ \xi^i_t \}$'s are \textit{i.i.d.} random variables (noise) with zero mean and the same positive definite covariance matrix $\Sigma \succ 0$. We are interested in the asymptotic behavior of the double averaging sequence $\{z_1,z_2,\dots\}$ defined as
\begin{gather}
z_{t+1} = \frac{1}{t+1} \sum_{k=0}^{t} \cbar{x}_k.
\end{gather}

Recall the Equation~\ref{eq:lem1e} from the proof of Lemma~\ref{lem:lemm1} (for the convenience it is provided below):
\begin{gather*}
\cbar{x}_k = \gamma^k \cbar{x}_0 + \alpha u_0 \frac{\gamma^k-\phi^k}{\gamma-\phi}+\alpha \eta
\sum_{l=1}^{k-1} \frac{\gamma^{k-l}-\phi^{k-l}}{\gamma-\phi} \xi_{l-1},
\end{gather*}
where $\xi_t = \sum_{i=1}^p \xi_t^i$.
Therefore
\begin{eqnarray*}
\sum_{k=0}^t \cbar{x}_k \!\!\!&=&\!\!\! \frac{1-\gamma^{t+1}}{1-\gamma} \cbar{x}_0 + \alpha u_0 \frac{1}{\gamma-\mu} \bigg(\frac{1-\gamma^{t+1}}{1-\gamma} - \frac{1-\phi^{t+1}}{1-\phi} \bigg) + \alpha \eta \sum_{l=1}^{t-1} \sum_{k=l+1}^{t} \frac{\gamma^{k-l}-\phi^{k-l}}{\gamma-\phi} \xi_{l-1} \\
&=&\!\!\! O(1) + \alpha \eta \sum_{l=1}^{t-1} \frac{1}{\gamma - \phi}
\bigg (\gamma \frac{1-\gamma^{t-l}}{1-\gamma} - \phi \frac{1-\phi^{t-l}}{1-\phi} \bigg) \xi_{l-1}
\end{eqnarray*}

Note that the only non-vanishing term (in weak convergence) of $1/\sqrt{t}\sum_{k=0}^t \cbar{x}_k$ as $t\rightarrow \infty$ is
\begin{align}
\frac{1}{\sqrt{t}} \alpha \eta \sum_{l=1}^{t-1} \frac{1}{\gamma - \phi}
\bigg ( \frac{\gamma}{1-\gamma} - \frac{\phi}{1-\phi} \bigg) \xi_{l-1}.
\label{eq:normalized}
\end{align}
Also recall that $\mathbb{V}[\xi_{l-1}] = p \sigma^2$ and
\begin{align*}
\frac{1}{\gamma - \phi}
\bigg ( \frac{\gamma}{1-\gamma} - \frac{\phi}{1-\phi} \bigg) = \frac{1}{(1-\gamma)(1-\phi) } = \frac{1}{\eta h p \alpha} .
\end{align*}
Therefore the expression in Equation~\ref{eq:normalized} is asymptotically normal with zero mean and variance $\sigma^2/p h^2$.
\end{proof}

\subsection{Generalization to multidimensional case}
The asymptotic variance in the Lemma~\ref{lem:lemm2} is optimal
with any
fixed $\eta$ and $\beta$ for which Equation~\ref{eq:condition} holds. The next lemma (Lemma~\ref{lem:lemm3}) extends the result in Lemma~\ref{lem:lemm2} to the multi-dimensional setting.

\begin{lemma}[Weak convergence]
\label{lem:lemm3}

Let $h$ denotes the largest eigenvalue of $A$. If $(2-\eta h)(2-\beta) >2 \beta/p$, $(2-\eta h) +
(2-\beta) > \beta/p $, $\eta>0$ and $\beta>0$,
then the
normalized double averaging sequence converges weakly to the
normal distribution with zero mean and the covariance matrix $V=A^{-1} \Sigma (A^{-1})^T$,
\begin{gather}
\sqrt{tp} (z_{t}- x^\ast) \rightharpoonup \mathcal{N}(0,V),
\quad t\rightarrow \infty.
\end{gather}
\end{lemma}

\begin{proof}
Since $A$ is symmetric, one can use the proof technique of Lemma~\ref{lem:lemm2} to prove Lemma~\ref{lem:lemm3} by diagonalizing the matrix $A$. This diagonalization essentially generalizes Lemma~\ref{lem:lemm1} to the multidimensional case. We will not go into the details of this proof as we will provide a simpler way to look at the system. As in the proof of Lemma~\ref{lem:lemm1} and Lemma~\ref{lem:lemm2}, for the ease of notation we redefine $\cbar{x}_{t}$ and $x^i_t$ as follows:
\[\cbar{x}_{t} \triangleq \cbar{x}_{t} - x^\ast \text{\:\:\:and\:\:\:} x^i_t \triangleq x^i_t - x^\ast.
\]
Let the spatial average of the local parameters at time $t$ be denoted as $y_t$ where
$y_t = \frac{1}{p} \sum_{i=1}^p x_t^i$, and let the average noise
be denoted as $\xi_t$, where $\xi_t = \frac{1}{p} \sum_{i=1}^p \xi_t^i$. Equations~\ref{loc:local} and \ref{loc:center} can then be reduced to the following
\begin{eqnarray}
	y_{t+1} &=& y_{t} - \eta ( Ay_t - \xi_t)
+ \alpha ( \cbar{x}_t - y_t ), \label{eq:reduced1}\\
	\cbar{x}_{t+1} &=& \cbar{x}_t + \beta ( y_t - \cbar{x}_t).
\label{eq:reduced2}
\end{eqnarray}

We focus on the case where the learning rate $\eta$ and the moving rate $\alpha$ are kept constant over time\footnote{As a side note, notice that the center parameter $\cbar{x}_t$ is
tracking the spatial average $y_t$ of the local parameters with a
non-symmetric spring in Equation~\ref{eq:reduced1} and ~\ref{eq:reduced2}. To be more precise note that the update on $y_{t+1}$ contains $ ( \cbar{x}_t - y_t )$ scaled by $\alpha$, whereas the update on $\cbar{x}_{t+1}$ contains $-( \cbar{x}_t - y_t )$ scaled by $\beta$. Since $\alpha = \beta / p$ the impact of the center $\tilde{x}_{t+1}$ on the spatial local average $y_{t+1}$ becomes more negligible as $p$ grows.}. Recall $\beta = p \alpha$ and $\alpha = \eta \rho$.

Let's introduce the block notation $U_t = (y_t,\cbar{x}_t)$, $\Xi_t = (\eta \xi_t,0)$, $M = I - \eta L$
and
\begin{align*}
L =
\left(
\begin{array}{cc}
A + \frac{\alpha}{\eta} I & - \frac{\alpha}{\eta}I \\
- \frac{\beta}{\eta} I & \frac{\beta}{\eta} I \end{array}
\right).
\end{align*}
From Equations~\ref{eq:reduced1} and~\ref{eq:reduced2} it follows that $U_{t+1} = M U_t + \Xi_t$. Note that this linear system has a degenerate noise $\Xi_t$ which prevents us from directly applying results of~\cite{polyak1992acceleration}. Expanding this recursive relation and summing by parts, we have
\begin{eqnarray*}
\sum_{k=0}^t U_k &=& M^0 U_0 + \\
			 &&M^1 U_0 + M^0 \Xi_0+ \\
			 &&M^2 U_0 + M^1 \Xi_0+ M^0 \Xi_1 + \\
			 &&... \\
			 &&M^t U_0 + M^{t-1} \Xi_0+ \cdots + M^0 \Xi_{t-1} .
\end{eqnarray*}

By Lemma~\ref{lem:additional}, $\|M\|_2 < 1$ and thus
\begin{align*}
	M^0 + M^1 + \cdots + M^t + \cdots = (I-M)^{-1} = \eta^{-1} L^{-1}.
\end{align*}

Since $A$ is invertible, we get
\begin{align*}
	L^{-1} =
\left(
\begin{array}{cc}
A^{-1} & \frac{\alpha}{\beta} A^{-1} \\
A^{-1} & \frac{\eta}{\beta} + \frac{\alpha}{\beta} A^{-1} \end{array}
\right),
\end{align*}
thus
\begin{align*}
	\frac{1}{\sqrt{t}} \sum_{k=0}^t U_k =
\frac{1}{\sqrt{t}} U_0 +
		\frac{1}{\sqrt{t}} \eta L^{-1} \sum_{k=1}^{t} \Xi_{k-1} -
		\frac{1}{\sqrt{t}} \sum_{k=1}^{t} M^{k+1} \Xi_{k-1}.
\end{align*}

Note that the only non-vanishing term of $\frac{1}{\sqrt{t}} \sum_{k=0}^t U_k$ is
$\frac{1}{\sqrt{t}} (\eta L)^{-1} \sum_{k=1}^{t} \Xi_{k-1} $, thus
in weak convergence we have
\begin{align}
\frac{1}{\sqrt{t}}	(\eta L)^{-1} \sum_{k=1}^{t} \Xi_{k-1} \rightharpoonup \mathcal{N}
\bigg( \left( \begin{array}{c} 0 \\ 0 \end{array} \right) , \left( \begin{array}{cc} V & V \\ V & V \end{array} \right) \bigg),
\end{align}
where $V=A^{-1} \Sigma (A^{-1})^T$.
\end{proof}

\begin{lemma}
If the following conditions hold:
\begin{eqnarray*}
(2-\eta h)(2-p\alpha) &>& 2\alpha\\
(2-\eta h) + (2-p\alpha) &>& \alpha\\
\eta &>& 0\\
\alpha &>& 0
\end{eqnarray*}
then $\|M\|_2 < 1$.
\label{lem:additional}
\end{lemma}
\begin{proof}
The eigenvalue $\lambda$ of $M$ and the (non-zero) eigenvector $(y,z)$ of $M$ satisfy
\begin{equation}
	M \left( \begin{array}{c} y \\ z \end{array} \right) =
\lambda \left( \begin{array}{c} y \\ z \end{array} \right) .
\label{eq:oneM}
\end{equation}
Recall that
\begin{align}
M = I - \eta L =
\left(
\begin{array}{cc}
I - \eta A -\alpha I & \alpha I \\
\beta I & I-\beta I \end{array}
\right).
\label{eq:twoM}
\end{align}
From the Equations~\ref{eq:oneM} and~\ref{eq:twoM} we obtain
\begin{eqnarray}
\left\{
	\begin{array}{ll}
		y - \eta A y - \alpha y + \alpha z = \lambda y \\
		\beta y + (1-\beta) z = \lambda z
	\end{array} \right..
\label{eq:bundle}
\end{eqnarray}

Since $(y,z)$ is assumed to be non-zero, we can write $z = \beta y / (\lambda + \beta -1)$.
Then the Equation~\ref{eq:bundle} can be reduced to
\begin{eqnarray}
	\eta A y = (1-\alpha -\lambda) y + \frac{\alpha \beta}{\lambda + \beta - 1} y .
\label{eq:mapping}
\end{eqnarray}
Thus $y$ is the eigenvector of $A$. Let $\lambda_A$ be the eigenvalue of matrix $A$ such that $ A y = \lambda_A y$. Thus based on Equation~\ref{eq:mapping} it follows that
\begin{equation}
\eta \lambda_A = (1-\alpha -\lambda) + \frac{\alpha \beta}{\lambda + \beta - 1}.
\label{eq:intermediate}
\end{equation}
Equation~\ref{eq:intermediate} is equivalent to
\begin{eqnarray}
	\lambda^2 - (2-a)\lambda + (1-a+c^2) = 0,
\end{eqnarray}
where $a = \eta \lambda_A + (p+1)\alpha$, $c^2 = \eta \lambda_A p \alpha $. It follows from the
condition in Equation~\ref{eq:condition} that $-1 < \lambda < 1$ iff $\eta > 0$, $\beta > 0$,
$(2-\eta \lambda_A)(2-\beta) >2\beta/p$ and $(2-\eta \lambda_A) +(2-\beta) > \beta/p $.
Let $h$ denote the maximum eigenvalue of $A$ and note that $2-\eta \lambda_A \ge 2-\eta h$.
This implies that the condition of our lemma is sufficient.
\end{proof}

As in Lemma~\ref{lem:lemm2}, the asymptotic covariance in the Lemma~\ref{lem:lemm3} is optimal, i.e. meets the Fisher information lower-bound. The fact that this asymptotic covariance matrix $V$ does not contain any term involving $\rho$ is quite remarkable, since
the penalty term $\rho$ does have an impact on the condition number of the Hessian in Equation \ref{eq:penalty}.

\section{Strongly convex case}
We now extend the above proof ideas to analyze the strongly convex case, in which the noisy gradient $g_t^i(x) = \nabla{F}(x) - \xi^i_t$ has the regularity that there exists some $0 < \mu \le L$,
for which $ \mu \norm{L2}{x-y}^2 \le \scalprod{L2}{\nabla{F}(x)- \nabla{F}(y)}{x-y} \le L \norm{L2}{x-y}^2$ holds uniformly for any $x\in \mathbb{R}^d,y \in \mathbb{R}^d$. The noise $\{ \xi^i_t
\}$'s is assumed to be i.i.d. with zero mean and bounded variance $\mathbb{E}[\norm{L2}{\xi^i_t}^2] \le \sigma^2$.

\begin{theorem}
Let $a_t = \mathbb{E} \norm{L2}{\frac{1}{p} \sum_{i=1}^p x^i_t -x^\ast}^2$,
$b_t= \frac{1}{p} \sum_{i=1}^p \mathbb{E} \norm{L2}{ x^i_t -x^\ast}^2$,
$c_t = \mathbb{E} \norm{L2}{\tilde{x}_t-x^\ast}^2 $, $\gamma_1=2\eta \frac{\mu L}{\mu+L}$ and $\gamma_2 = 2 \eta L (1-\frac{2 \sqrt{\mu L}}{\mu+L})$. If $0 \le \eta \le \frac{2}{\mu+L}(1-\alpha)$, $0 \le \alpha<1$ and $0 \le \beta \le 1$ then
\begin{gather*}
\begin{pmatrix}
a_{t+1} \\
b_{t+1} \\
c_{t+1} \\
\end{pmatrix} \le
\begin{pmatrix}
1-\gamma_1-\gamma_2-\alpha & \gamma_2 & \alpha \\
0 & 1-\gamma_1-\alpha & \alpha \\
\beta & 0 & 1-\beta
\end{pmatrix}
\begin{pmatrix}
a_{t} \\
b_{t} \\
c_{t} \\
\end{pmatrix}
+
\begin{pmatrix}
\eta^2 \frac{\sigma^2}{p} \\
\eta^2 \sigma^2 \\
0 \\
\end{pmatrix}
.
\end{gather*}
\label{thm:strongly}
\end{theorem}

\begin{proof}
The idea of the proof is based on the point of view in Lemma~\ref{lem:lemm3}, i.e. how close
the center variable $\tilde{x}_t$ is to the spatial average of the local variables $y_t = \frac{1}{p} \sum_{i=1}^p x^i_t$. To further simplify the notation, let the noisy gradient be $\nabla f^i_{t,\xi}=g_t^i(x_t^i) = \nabla{F}(x_t^i) - \xi^i_t$, and $\nabla f^i_{t} = \nabla{F}(x_t^i)$ be its deterministic part.
Then \textit{EASGD} updates can be rewritten as follows,
\begin{eqnarray}
	x_{t+1}^i &=& x_t^i - \eta \nabla f^i_{t,\xi} - \alpha(x^i_t - \tilde{x}_t), \label{eq:strong_updates1} \\
	\cbar{x}_{t+1} &=& \cbar{x}_t + \beta ( y_t - \cbar{x}_t). \label{eq:strong_updates2}
\end{eqnarray}
We have thus the update for the spatial average,
\begin{eqnarray}
	y_{t+1} &=& y_{t} - \eta \frac{1}{p} \sum_{i=1}^p \nabla f^i_{t,\xi} - \alpha (y_t - \tilde{x}_t).
\label{eq:strong_updates3}
\end{eqnarray}
The idea of the proof is to bound the distance $\norm{L2}{\tilde{x}_t-x^\ast}^2$ through $\norm{L2}{y_t-x^\ast}^2$ and $\frac{1}{p} \sum_i^p \norm{L2}{x^i_t-x^\ast}^2$. We start from the following estimate for the strongly convex function~\cite{nesterov2004introductory},
\begin{eqnarray*}
	\scalprod{L2}{\nabla{F}(x)-\nabla{F}(y)}{x-y} \ge \frac{\mu L}{\mu+L} \norm{L2}{x-y}^2+
\frac{1}{\mu+L} \norm{L2}{\nabla{F}(x)-\nabla{F}(y)}^2.
\end{eqnarray*}

Since $\nabla{f( x^\ast)}=0$, we have
\begin{eqnarray}
	 \scalprod{L2}{ \nabla f^i_{t}}{x^i_t - x^\ast} \ge \frac{\mu L}{\mu+L} \norm{L2}{x^i_t - x^\ast}^2+ \frac{1}{\mu+L} \norm{L2}{\nabla f^i_{t}}^2. \label{eq:est0}
\end{eqnarray}

From Equation~\ref{eq:strong_updates1} the following relation holds,
\begin{eqnarray}
	\norm{L2}{x^i_{t+1}-x^\ast}^2 &=& \norm{L2}{x^i_{t}-x^\ast}^2
+ \eta^2 \norm{L2}{ \nabla f^i_{t,\xi} } ^2 + \alpha^2 \norm{L2}{x^i_t - \tilde{x}_t} ^2 \nonumber \\
&-&2\eta \scalprod{L2}{ \nabla f^i_{t,\xi}}{x^i_t - x^\ast}
-2 \alpha \scalprod{L2}{x^i_t - \tilde{x}_t}{x^i_t - x^\ast} \nonumber \\
&+&2 \eta \alpha \scalprod{L2}{ \nabla f^i_{t,\xi}}{ x^i_t - \tilde{x}_t} .
\label{eq:est1a}
\end{eqnarray}
By the cosine rule ($2
\scalprod{L2}{a-b}{c-d} = \norm{L2}{a-d}^2 - \norm{L2}{a-c}^2 + \norm{L2}{c-b}^2 - \norm{L2}{d-b}^2 $), we have
\begin{eqnarray}
	2 \scalprod{L2}{x^i_t - \tilde{x}_t}{x^i_t - x^\ast} = \norm{L2}{x^i_t-x^\ast}^2 +
		\norm{L2}{x^i_t-\tilde{x}_t}^2 - \norm{L2}{\tilde{x}_t - x^\ast}^2 .
\label{eq:est1b}
\end{eqnarray}
By the Cauchy-Schwarz inequality, we have
\begin{eqnarray}
\scalprod{L2}{ \nabla f^i_{t}}{ x^i_t - \tilde{x}_t} \le \norm{L2}{ \nabla f^i_{t} } \norm{L2}{ x^i_t - \tilde{x}_t}.
\label{eq:est1c}
\end{eqnarray}
Combining the above estimates in Equations~\ref{eq:est0},~\ref{eq:est1a},~\ref{eq:est1b},~\ref{eq:est1c}, we obtain
\begin{eqnarray}
\norm{L2}{x^i_{t+1}-x^\ast}^2 &\le& \norm{L2}{x^i_{t}-x^\ast}^2
+ \eta^2 \norm{L2}{ \nabla f^i_{t} - \xi^i_t } ^2 + \alpha^2 \norm{L2}{x^i_t - \tilde{x}_t}^2 \nonumber \\
&-&2\eta \bigg( \frac{\mu L}{\mu+L} \norm{L2}{x^i_t - x^\ast}^2+ \frac{1}{\mu+L} \norm{L2}{\nabla f^i_{t}}^2 \bigg) + 2\eta \scalprod{L2}{ \xi^i_t}{x^i_t - x^\ast} \nonumber \\
&-&\alpha \big ( \norm{L2}{x^i_t-x^\ast}^2 +
			\norm{L2}{x^i_t-\tilde{x}_t}^2 - \norm{L2}{\tilde{x}_t - x^\ast}^2 \big) \nonumber \\
&+& 2 \eta \alpha \norm{L2}{ \nabla f^i_{t} } \norm{L2}{ x^i_t - \tilde{x}_t}
- 2\eta \alpha \scalprod{L2}{ \xi^i_t}{x^i_t-\tilde{x}_t} \label{eq:est1}.
\end{eqnarray}

Choosing $0 \le \alpha<1$, we can have this upper-bound for the terms $\alpha^2 \norm{L2}{x^i_t - \tilde{x}_t}^2 - \alpha \norm{L2}{x^i_t-\tilde{x}_t}^2 + 2 \eta \alpha \norm{L2}{ \nabla f^i_{t} } \norm{L2}{ x^i_t - \tilde{x}_t} = -\alpha(1-\alpha) \norm{L2}{x^i_t - \tilde{x}_t}^2 +2 \eta \alpha \norm{L2}{ \nabla f^i_{t} } \norm{L2}{ x^i_t - \tilde{x}_t} \le \frac{\eta^2 \alpha }{1-\alpha} \norm{L2}{ \nabla f^i_{t} }^2 $ by applying $-ax^2+bx \le \frac{b^2}{4a}$ with $x= \norm{L2}{ x^i_t - \tilde{x}_t}$. Thus we can further
bound Equation~\ref{eq:est1} with
\begin{eqnarray}
\norm{L2}{x^i_{t+1}-x^\ast}^2 &\le& (1-2\eta \frac{\mu L}{\mu + L} - \alpha) \norm{L2}{x^i_{t}-x^\ast}^2 + (\eta^2 + \frac{\eta^2 \alpha }{1-\alpha} -\frac{2 \eta}{\mu+L} ) \norm{L2}{ \nabla f^i_{t}} ^2 \nonumber \\
&-& 2 \eta^2 \scalprod{L2}{ \nabla f^i_{t} }{ \xi^i_t }
+2\eta \scalprod{L2}{ \xi^i_t }{ x^i_t - x^\ast} - 2\eta \alpha \scalprod{L2}{ \xi^i_t }{ x^i_t - \tilde{x}_t } \label{eq:est2a} \\
&+& \eta^2 \norm{L2}{ \xi^i_t }^2 + \alpha \norm{L2}{ \tilde{x}_t - x^\ast}^2 \label{eq:est2b}
\end{eqnarray}

As in Equation~\ref{eq:est2a} and ~\ref{eq:est2b}, the noise $\xi^i_t$ is zero mean ($\mathbb{E} \xi^i_t = 0$) and the variance of the noise $\xi^i_t$ is bounded ($\mathbb{E} \norm{L2}{ \xi^i_t }^2 \le \sigma^2 $), if $\eta$ is chosen small enough such that
$\eta^2 + \frac{\eta^2 \alpha }{1-\alpha} -\frac{2 \eta}{\mu+L} \le 0$, then
\begin{eqnarray}
\mathbb{E} \norm{L2}{x^i_{t+1}-x^\ast}^2 &\le& (1-2\eta \frac{\mu L}{\mu + L} - \alpha)
\mathbb{E} \norm{L2}{x^i_{t}-x^\ast}^2 + \eta^2 \sigma^2 + \alpha \mathbb{E} \norm{L2}{ \tilde{x}_t - x^\ast}^2 .
\label{eq:key1}
\end{eqnarray}

Now we apply similar idea to estimate $\norm{L2}{y_t-x^\ast}^2$. From Equation~\ref{eq:strong_updates3} the following relation holds,
\begin{eqnarray}
\norm{L2}{y_{t+1}-x^\ast}^2 &=& \norm{L2}{y_{t}-x^\ast}^2
+ \eta^2 \norm{L2}{ \frac{1}{p} \sum_{i=1}^p \nabla f^i_{t,\xi}} ^2 + \alpha^2 \norm{L2}{y_t - \tilde{x}_t} ^2 \nonumber \\
&-&2\eta \scalprod{L2}{ \frac{1}{p} \sum_{i=1}^p \nabla f^i_{t,\xi} }{y_t - x^\ast}
-2 \alpha \scalprod{L2}{y_t - \tilde{x}_t}{y_t - x^\ast} \nonumber \\
&+&2 \eta \alpha \scalprod{L2}{ \frac{1}{p} \sum_{i=1}^p \nabla f^i_{t,\xi} }{ y_t - \tilde{x}_t} .
\label{eq:est3}
\end{eqnarray}

By $\scalprod{L2}{\frac{1}{p} \sum_{i=1}^p a_i}{\frac{1}{p} \sum_{j=1}^p b_j}=\frac{1}{p}
\sum_{i=1}^p \scalprod{L2}{a_i}{b_i} - \frac{1}{p^2} \sum_{i>j} \scalprod{L2}{a_i-a_j}{b_i-b_j}$,
we have
\begin{eqnarray}
\scalprod{L2}{ \frac{1}{p} \sum_{i=1}^p \nabla f^i_{t} }{y_t - x^\ast} =
\frac{1}{p}
\sum_{i=1}^p \scalprod{L2}{ \nabla f^i_{t} }{x_t^i - x^\ast} - \frac{1}{p^2} \sum_{i>j} \scalprod{L2}{ \nabla f^i_{t} - \nabla f^j_{t} }{x_t^i - x_t^j }.
\label{eq:est3a}
\end{eqnarray}
By the cosine rule, we have
\begin{eqnarray}
2 \scalprod{L2}{y_t - \tilde{x}_t}{y_t - x^\ast} = \norm{L2}{y_t-x^\ast}^2 +
\norm{L2}{y_t-\tilde{x}_t}^2 - \norm{L2}{\tilde{x}_t - x^\ast}^2.
\label{eq:est3b}
\end{eqnarray}

Denote $\xi_t = \frac{1}{p} \sum_{i=1}^p \xi^i_t $, we can rewrite Equation~\ref{eq:est3} as
\begin{eqnarray}
\norm{L2}{y_{t+1}-x^\ast}^2 &=& \norm{L2}{y_{t}-x^\ast}^2
+ \eta^2 \norm{L2}{ \frac{1}{p} \sum_{i=1}^p \nabla f^i_{t} - \xi_t } ^2 + \alpha^2 \norm{L2}{y_t - \tilde{x}_t} ^2 \nonumber \\
&-&2\eta \scalprod{L2}{ \frac{1}{p} \sum_{i=1}^p \nabla f^i_{t} - \xi_t }{y_t - x^\ast}
-2 \alpha \scalprod{L2}{y_t - \tilde{x}_t}{y_t - x^\ast} \nonumber \\
&+&2 \eta \alpha \scalprod{L2}{ \frac{1}{p} \sum_{i=1}^p \nabla f^i_{t} - \xi_t }{ y_t - \tilde{x}_t} .
\label{eq:est3c}
\end{eqnarray}

By combining the above Equations~\ref{eq:est3a},~\ref{eq:est3b} with ~\ref{eq:est3c}, we obtain
\begin{eqnarray}
\norm{L2}{y_{t+1}-x^\ast}^2 &=& \norm{L2}{y_{t}-x^\ast}^2
+ \eta^2 \norm{L2}{ \frac{1}{p} \sum_{i=1}^p \nabla f^i_{t} - \xi_t } ^2 + \alpha^2 \norm{L2}{y_t - \tilde{x}_t} ^2 \nonumber \\
&-& 2\eta \bigg(\frac{1}{p}
\sum_{i=1}^p \scalprod{L2}{ \nabla f^i_{t} }{x_t^i - x^\ast} - \frac{1}{p^2} \sum_{i>j} \scalprod{L2}{ \nabla f^i_{t} - \nabla f^j_{t} }{x_t^i - x_t^j } \bigg) \\
&+& 2 \eta\scalprod{L2}{ \xi_t }{y_t - x^\ast}
- \alpha ( \norm{L2}{y_t-x^\ast}^2 +
\norm{L2}{y_t-\tilde{x}_t}^2 - \norm{L2}{\tilde{x}_t - x^\ast}^2 ) \nonumber \\
&+&2 \eta \alpha \scalprod{L2}{ \frac{1}{p} \sum_{i=1}^p \nabla f^i_{t} - \xi_t }{ y_t - \tilde{x}_t} .
\label{eq:est3d}
\end{eqnarray}
Thus it follows from Equation~\ref{eq:est0} and~\ref{eq:est3d} that
\begin{eqnarray}
\norm{L2}{y_{t+1}-x^\ast}^2 &\le& \norm{L2}{y_{t}-x^\ast}^2
+ \eta^2 \norm{L2}{ \frac{1}{p} \sum_{i=1}^p \nabla f^i_{t} - \xi_t } ^2 + \alpha^2 \norm{L2}{y_t - \tilde{x}_t} ^2 \nonumber \\
&-& 2\eta \frac{1}{p}
\sum_{i=1}^p \bigg( \frac{\mu L}{\mu+L} \norm{L2}{x^i_t - x^\ast}^2+ \frac{1}{\mu+L} \norm{L2}{\nabla f^i_{t}}^2 \bigg) \nonumber \\
&+& 2\eta \frac{1}{p^2} \sum_{i>j} \scalprod{L2}{ \nabla f^i_{t} - \nabla f^j_{t} }{x_t^i - x_t^j } \nonumber \\
&+& 2 \eta \scalprod{L2}{ \xi_t }{y_t - x^\ast}
- \alpha ( \norm{L2}{y_t-x^\ast}^2 +
\norm{L2}{y_t-\tilde{x}_t}^2 - \norm{L2}{\tilde{x}_t - x^\ast}^2 ) \nonumber \\
&+&2 \eta \alpha \scalprod{L2}{ \frac{1}{p} \sum_{i=1}^p \nabla f^i_{t} - \xi_t }{ y_t - \tilde{x}_t} .
\label{eq:est3e}
\end{eqnarray}

Recall $y_t = \frac{1}{p} \sum_{i=1}^p x^i_t$, we have the following bias-variance relation,
\begin{eqnarray}
\frac{1}{p} \sum_{i=1}^p \norm{L2}{x^i_t - x^\ast}^2 &=& \frac{1}{p} \sum_{i=1}^p \norm{L2}{x^i_t - y_t}^2 + \norm{L2}{ y_t - x^\ast}^2 =
\frac{1}{p^2} \sum_{i>j} \norm{L2}{x^i_t - x^j_t}^2 + \norm{L2}{ y_t - x^\ast}^2,
\nonumber \\
\frac{1}{p} \sum_{i=1}^p \norm{L2}{ \nabla f^i_{t}}^2 &=&
\frac{1}{p^2} \sum_{i>j} \norm{L2}{ \nabla f^i_{t} - \nabla f^j_{t}}^2 + \norm{L2}{ \frac{1}{p} \sum_{i=1}^p \nabla f^i_{t} }^2.
\label{eq:est3f}
\end{eqnarray}

By the Cauchy-Schwarz inequality, we have
\begin{eqnarray}
	\frac{\mu L}{\mu+L} \norm{L2}{x^i_t - x^j_t}^2
+ \frac{1}{\mu+L} \norm{L2}{ \nabla f^i_{t} - \nabla f^j_{t}}^2 \ge
\frac{2\sqrt{\mu L}}{\mu+L} \scalprod{L2}{ \nabla f^i_{t} - \nabla f^j_{t} }{x_t^i - x_t^j } .
\label{eq:est3g}
\end{eqnarray}

Combining the above estimates in Equations~\ref{eq:est3e},~\ref{eq:est3f},~\ref{eq:est3g}, we obtain
\begin{eqnarray}
\norm{L2}{y_{t+1}-x^\ast}^2 &\le& \norm{L2}{y_{t}-x^\ast}^2
+ \eta^2 \norm{L2}{ \frac{1}{p} \sum_{i=1}^p \nabla f^i_{t} - \xi_t } ^2 + \alpha^2 \norm{L2}{y_t - \tilde{x}_t} ^2 \nonumber \\
&-& 2\eta \bigg( \frac{\mu L}{\mu+L} \norm{L2}{y_t - x^\ast}^2+ \frac{1}{\mu+L} \norm{L2}{ \frac{1}{p} \sum_{i=1}^p \nabla f^i_{t} }^2 \bigg) \nonumber \\
&+& 2 \eta \bigg(1-\frac{2\sqrt{\mu L}}{\mu+L}\bigg) \frac{1}{p^2} \sum_{i>j} \scalprod{L2}{ \nabla f^i_{t} - \nabla f^j_{t} }{x_t^i - x_t^j } \nonumber \\
&+& 2 \eta \scalprod{L2}{ \xi_t }{y_t - x^\ast}
- \alpha ( \norm{L2}{y_t-x^\ast}^2 +
\norm{L2}{y_t-\tilde{x}_t}^2 - \norm{L2}{\tilde{x}_t - x^\ast}^2 ) \nonumber \\
&+&2 \eta \alpha \scalprod{L2}{ \frac{1}{p} \sum_{i=1}^p \nabla f^i_{t} - \xi_t }{ y_t - \tilde{x}_t} .
\label{eq:est4}
\end{eqnarray}

Similarly if $0 \le \alpha<1$, we can have this upper-bound for the terms $\alpha^2 \norm{L2}{y_t - \tilde{x}_t}^2 - \alpha \norm{L2}{y_t-\tilde{x}_t}^2 + 2 \eta \alpha \norm{L2}{ \frac{1}{p} \sum_{i=1}^p \nabla f^i_{t} } \norm{L2}{y_t - \tilde{x}_t} \le \frac{\eta^2 \alpha }{1-\alpha} \norm{L2}{ \frac{1}{p} \sum_{i=1}^p \nabla f^i_{t}}^2 $ by applying $-ax^2+bx \le \frac{b^2}{4a}$ with $x= \norm{L2}{ y_t - \tilde{x}_t}$. Thus we have the following bound for the Equation~\ref{eq:est4}
\begin{eqnarray}
\norm{L2}{y_{t+1}-x^\ast}^2 &\le&
(1-2\eta \frac{\mu L}{\mu + L} - \alpha) \norm{L2}{y_{t}-x^\ast}^2
+ (\eta^2 + \frac{\eta^2 \alpha }{1-\alpha} -\frac{2 \eta}{\mu+L} )
\norm{L2}{ \frac{1}{p} \sum_{i=1}^p \nabla f^i_{t} } ^2 \nonumber \\
&-& 2 \eta^2 \scalprod{L2}{ \frac{1}{p} \sum_{i=1}^p \nabla f^i_{t} }{ \xi_t }
+2\eta \scalprod{L2}{ \xi_t }{y_t - x^\ast} - 2\eta \alpha \scalprod{L2}{ \xi_t }{y_t - \tilde{x}_t } \nonumber \\
&+& 2 \eta \bigg(1-\frac{2\sqrt{\mu L}}{\mu+L}\bigg) \frac{1}{p^2} \sum_{i>j} \scalprod{L2}{ \nabla f^i_{t} - \nabla f^j_{t} }{x_t^i - x_t^j } \nonumber \\
&+& \eta^2 \norm{L2}{\xi_t }^2 + \alpha \norm{L2}{ \tilde{x}_t - x^\ast}^2.
\label{eq:est5}
\end{eqnarray}
Since $ \frac{2\sqrt{\mu L}}{\mu+L} \le 1 $, we need also bound the nonlinear term
$\scalprod{L2}{ \nabla f^i_{t} - \nabla f^j_{t} }{x_t^i - x_t^j } \le L \norm{L2}{ x_t^i - x_t^j}^2$.
Recall the bias-variance relation
$\frac{1}{p} \sum_{i=1}^p \norm{L2}{x^i_t - x^\ast}^2 =
\frac{1}{p^2} \sum_{i>j} \norm{L2}{x^i_t - x^j_t}^2 + \norm{L2}{ y_t - x^\ast}^2$.
The key observation is that if $\frac{1}{p} \sum_{i=1}^p \norm{L2}{x^i_t - x^\ast}^2$ remains bounded,
then larger variance $ \sum_{i>j} \norm{L2}{x^i_t - x^j_t}^2$ implies smaller bias $ \norm{L2}{ y_t - x^\ast}^2$. Thus this nonlinear term can be compensated.

Again choose $\eta$ small enough such that
$\eta^2 + \frac{\eta^2 \alpha }{1-\alpha} -\frac{2 \eta}{\mu+L} \le 0$ and take expectation
in Equation~\ref{eq:est5},
\begin{eqnarray}
\mathbb{E} \norm{L2}{y_{t+1}-x^\ast}^2 &\le&
(1-2\eta \frac{\mu L}{\mu + L} - \alpha) \mathbb{E} \norm{L2}{y_{t}-x^\ast}^2 \nonumber \\
&+& 2 \eta L \bigg(1-\frac{2\sqrt{\mu L}}{\mu+L}\bigg) \bigg( \frac{1}{p} \sum_{i=1}^p \mathbb{E} \norm{L2}{x^i_t - x^\ast}^2 - \mathbb{E} \norm{L2}{y_{t}-x^\ast}^2 \bigg) \nonumber \\
&+& \eta^2 \frac{\sigma^2}{p} + \alpha \mathbb{E} \norm{L2}{ \tilde{x}_t - x^\ast}^2.
\label{eq:key2}
\end{eqnarray}

As for the center variable in Equation~\ref{eq:strong_updates2}, we apply simply the convexity of the norm $\norm{L2}{\cdot}^2$ to obtain
\begin{eqnarray}
\norm{L2}{\tilde{x}_{t+1} - x^\ast}^2 \le (1-\beta) \norm{L2}{\cbar{x}_t - x^\ast}^2 + \beta \norm{L2}{ y_t - x^\ast}^2 .
\label{eq:key3}
\end{eqnarray}

Combining the estimates from Equations~\ref{eq:key1},~\ref{eq:key2},~\ref{eq:key3}, and denote
$a_t = \mathbb{E} \norm{L2}{y_t -x^\ast}^2 $,
$b_t= \frac{1}{p} \sum_{i=1}^p \mathbb{E} \norm{L2}{ x^i_t -x^\ast}^2$,
$c_t = \mathbb{E} \norm{L2}{\tilde{x}_t-x^\ast}^2 $, $\gamma_1=2\eta \frac{\mu L}{\mu+L}$,
$\gamma_2 = 2 \eta L (1-\frac{2 \sqrt{\mu L}}{\mu+L})$, then
\begin{gather*}
\begin{pmatrix}
a_{t+1} \\
b_{t+1} \\
c_{t+1} \\
\end{pmatrix} \le
\begin{pmatrix}
1-\gamma_1-\gamma_2-\alpha & \gamma_2 & \alpha \\
0 & 1-\gamma_1-\alpha & \alpha \\
\beta & 0 & 1-\beta
\end{pmatrix}
\begin{pmatrix}
a_{t} \\
b_{t} \\
c_{t} \\
\end{pmatrix}
+
\begin{pmatrix}
\eta^2 \frac{\sigma^2}{p} \\
\eta^2 \sigma^2 \\
0 \\
\end{pmatrix}
,
\end{gather*}
as long as $0\le \beta \le 1$, $0 \le \alpha < 1$ and $\eta^2 + \frac{\eta^2 \alpha }{1-\alpha} -\frac{2 \eta}{\mu+L} \le 0$, i.e. $0 \le \eta \le \frac{2}{\mu+L}(1-\alpha).$
\end{proof}

The above theorem captures the bias-variance tradeoff of the
spatial average $\frac{1}{p} \sum_{i=1}^p x^i_t $ of
the local variables (the $a_t$), with respect to the
averaged mean squared error of each local variable (the $b_t$). The center variable
$\tilde{x}_t$ is tracking $\frac{1}{p} \sum_{i=1}^p x^i_t $ over time (the $c_t$).

To get an upper bound on the rate of convergence for $\tilde{x}_t$, we need
to assume the matrix $M$ to be positive, and its spectral norm to be smaller than one.
Here
\begin{gather*}
M= \begin{pmatrix}
1-\gamma_1-\gamma_2-\alpha & \gamma_2 & \alpha \\
0 & 1-\gamma_1-\alpha & \alpha \\
\beta & 0 & 1-\beta
\end{pmatrix} .
\end{gather*}

We have three eigenvalues of $M$ as follows:
\begin{align*}
\lambda_1 &= 1-\alpha-\gamma_1-\gamma_2, \\
\lambda_2 &= 1 + \frac{1}{2} ( - \alpha - \beta - \gamma_1 + \sqrt{ (\alpha + \beta + \gamma_1)^2 - 4 \beta \gamma_1 }), \\
\lambda_3 &= 1 + \frac{1}{2} ( - \alpha - \beta - \gamma_1 - \sqrt{ (\alpha + \beta + \gamma_1)^2 - 4 \beta \gamma_1 }).
\end{align*}

Under the conditions of the above theorem ($0 \le \eta \le \frac{2}{\mu+L}(1-\alpha)$, $0 \le \alpha<1$ and $0 \le \beta \le 1$), we still need to assume $\lambda_1 \ge 0$ so that $M$ is positive. Since $\gamma_1=2\eta \frac{\mu L}{\mu+L} \ge 0 $ and $\gamma_2 = 2 \eta L (1-\frac{2 \sqrt{\mu L}}{\mu+L}) \ge 0$, we
deduce that $\lambda_1 \le 1$. We can also verify that $\lambda_3 \le \lambda_2 \le 1$. Thus for the stability we
only need $\lambda_3 \ge -1$.

For $\lambda_1 > 0$, we get the condition $0<\eta < \frac{1-\alpha}{\frac{2\mu L}{\mu+L} + 2 L (1- \frac{2 \sqrt{ \mu L} }{\mu + L})}$.
For $\lambda_3 > -1$, we have the condition $\eta < \frac{\mu + L }{\mu L } (1 - \frac{\alpha}{2-\beta})$.
When $\mu = L$, these two conditions mean $0 < \eta < \frac{1-\alpha}{L}$ and $0 < \eta < \frac{2}{L} (1 - \frac{\alpha}{2-\beta})$. On the other hand, when $\mu=0$, we have $0<\eta < \frac{1-\alpha}{ 2 L }$.
In either case, our method operates in the under-damping (no oscillations) region.

With the above conditions, we can now ask what is the asymptotic variance of $a_t$, $b_t$ and $c_t$.
By solving the fixed point equation $(a_\infty,b_\infty,c_\infty)' = M (a_\infty,b_\infty,c_\infty)' + (\eta^2 \frac{\sigma^2}{p} , \eta^2 \sigma^2 , 0)'$, we obtain
\begin{align*}
a_\infty = c_\infty &= \frac{\alpha/p+\gamma_1/p+\gamma_2}{\gamma_1(\alpha+\gamma_1+\gamma_2)} \eta^2 \sigma^2, \\
b_\infty &= \frac{\alpha/p+\gamma_1+\gamma_2}{\gamma_1(\alpha+\gamma_1+\gamma_2)} \eta^2 \sigma^2.
\end{align*}

If $\mu = L$, then $\gamma_2 = 0$, we indeed get the asymptotic variance $c_\infty$ of order $\sigma^2/p$. This order matches our quadratic case analysis above.
However, if $\mu << L$, then $\gamma_1$ will be close to zero, and we don't see in this upper bound
the benefit of variance reduction by increasing $p$ (number of workers). It would be interesting to find a non-quadratic example such that this can actually happen.

\section{Stability of \textit{EASGD} and \textit{ADMM}}
\label{sec:theory_admm}

In this section we study the stability of the asynchronous \textit{EASGD} and \textit{ADMM} methods in the round-robin scheme~\cite{langford2009slow}.
We first state the updates of both algorithms in this setting, and then we study their stability. We will show that in the one-dimensional quadratic case, \textit{ADMM} algorithm can exhibit chaotic behavior,
leading to exponential divergence. The analytic condition for the \textit{ADMM} algorithm to be stable is still unknown, while for the \textit{EASGD} algorithm it is very simple.

In our setting, the \textit{ADMM} method~\cite{Boyd2011,icml2014c2_zhange14,ouyang2013stochastic} involves solving the following minimax problem\footnote{The convergence analysis in~\cite{icml2014c2_zhange14} is based on the assumption that ``At any master iteration, updates from the workers have the same probability of arriving at the master.", which
is not satisfied in the round-robin scheme.},
\begin{equation}
\max_{\lambda^1,\ldots,\lambda^p} \min_{x^1,\ldots,x^p,\tilde{x}}
\sum_{i=1}^p F(x^i) - \lambda^i (x^i-\tilde{x}) + \frac{\rho}{2}\|x^i - \tilde{x}\|^2,
\end{equation}
where $\lambda^{i}$'s are the Lagrangian multipliers. The resulting updates of the \textit{ADMM} algorithm in the round-robin scheme are given next. Let $t\ge0$ be a global clock. At each $t$, we linearize the function $F(x^i) $ with $ F(x^i_t)+\scalprod{L2}{\nabla{F}(x^i_t)}{x^i-x^i_t}+\frac{1}{2\eta} \norm{L2}{x^i-x^i_t}^2$ as in~\cite{ouyang2013stochastic}. The updates become
\begin{eqnarray}
\lambda^{i}_{t+1} &=&
\left\{ \begin{array}{ll}
\lambda^{i}_t - (x^{i}_t- \tilde{x}_t) & \mbox{if $\mod(t,p) = i-1$};\\
\lambda^i_t & \mbox{if $\mod(t,p) \ne i-1$}.\end{array} \right. \label{eq:admm_update1} \\
x^{i}_{t+1} &=&
\left\{ \begin{array}{ll}
\frac{x^i_t - \eta \nabla{F}(x^{i}_t)+ \eta \rho (\lambda^i_{t+1} + \tilde{x}_t)}{1+\eta \rho}
& \mbox{if $\mod(t,p) = i-1$};\\
x^i_t & \mbox{if $\mod(t,p) \ne i-1$}.\end{array} \right. \label{eq:admm_update2}\\
\tilde{x}_{t+1} &=& \frac{1}{p} \sum_{i=1}^p ( x^i_{t+1} - \lambda^i_{t+1}). \label{eq:admm_update3}
\end{eqnarray}
Each local variable $x^i$ is periodically updated (with period $p$). First, the Lagrangian multiplier $\lambda^{i}$ is updated with the dual ascent update
as in Equation \ref{eq:admm_update1}. It is followed by the gradient descent update of the local variable
as given in Equation \ref{eq:admm_update2}. Then the center variable $\tilde{x}$ is updated
with the most recent values of all the local variables and Lagrangian multipliers as in Equation \ref{eq:admm_update3}.
Note that since the step size for the dual ascent update is chosen to be $\rho$ by convention~\cite{Boyd2011,icml2014c2_zhange14,ouyang2013stochastic}, we have
re-parametrized the Lagrangian multiplier to be $\lambda^{i}_{t} \leftarrow \lambda^{i}_{t}/\rho$ in
the above updates.

The \textit{EASGD} algorithm in the round-robin scheme is defined similarly and is given below
\begin{eqnarray}
x^{i}_{t+1} &=&
\left\{ \begin{array}{ll}
x^i_t - \eta \nabla{F}(x^{i}_t) - \alpha (x_t^i - \tilde{x}_t)
& \mbox{if $\mod(t,p) = i-1$};\\
x^i_t & \mbox{if $\mod(t,p) \ne i-1$}.\end{array} \right. \label{eq:easgd_update1}\\
\tilde{x}_{t+1} &=& \tilde{x}_{t} + \sum_{i: \mod(t,p) = i-1} \alpha ( x^i_{t} - \tilde{x}_{t}). \label{eq:easgd_update2}
\end{eqnarray}
At time $t$, only the $i$-th local worker (whose index $i-1$ equals $t$ modulo $p$) is activated,
and performs the update in Equations~\ref{eq:easgd_update1} which is followed by the master update given in Equation~\ref{eq:easgd_update2}.

We will now focus on the one-dimensional quadratic case without noise, i.e.
\[F(x) = \frac{x^2}{2},\quad x \in \mathbb{R}.\]

For the \textit{ADMM} algorithm, let the state of the (dynamical) system at time $t$ be $s_t=(\lambda_t^1,x_t^1,\ldots,\lambda_t^p,x_t^p,\tilde{x}_t) \in \mathbb{R}^{2p+1}$. The local worker $i$'s updates in Equations~\ref{eq:admm_update1},~\ref{eq:admm_update2}, and~\ref{eq:admm_update3} are composed of
three linear maps which can be written as $s_{t+1}=(F^i_3 \circ F^i_2 \circ F^i_1) (s_t)$. For simplicity,
we will only write them out below for the case when $i=1$ and $p=2$:
\small{\[
F_1^1\!\!=\!\! \left( \begin{array}{ccccc}
1 & -1 & 0 & 0 & 1 \\
0 & 1& 0 & 0 & 0 \\
0 & 0 & 1 &0 &0 \\
0 & 0 & 0 &1 &0 \\
0 & 0 & 0 &0 &1\\
\end{array} \right)\!\!,
F_2^1\!\!=\!\! \left( \begin{array}{ccccc}
1 & 0 & 0 & 0 & 0 \\
\frac{\eta \rho}{1+\eta \rho} & \frac{1-\eta}{1+\eta\rho} & 0 & 0 & \frac{\eta\rho}{ 1+\eta\rho} \\
0 & 0 & 1 &0 &0 \\
0 & 0 & 0 &1 &0 \\
0 & 0 & 0 &0 &1\\
\end{array} \right)\!\!,
F_3^1\!\!=\!\! \left( \begin{array}{ccccc}
1 & 0 & 0 & 0 & 0 \\
0 & 1 & 0 & 0 & 0 \\
0 & 0 & 1 &0 &0 \\
0 & 0 & 0 &1 &0 \\
-\frac{1}{p} & \frac{1}{p} & -\frac{1}{p} &\frac{1}{p} &0\\
\end{array} \right)\!\!.
\]}

\normalsize
For each of the $p$ linear maps, it's possible to find a simple condition such that
each map, where the $i^{\text{th}}$ map has the form $F^i_3 \circ F^i_2 \circ F^i_1$, is stable (the absolute value of the eigenvalues of the map are smaller or equal to one).
However, when these non-symmetric maps are composed one after another as follows $\mathcal{F}=F^p_3 \circ F^p_2 \circ F^p_1 \circ \ldots \circ F^1_3 \circ F^1_2 \circ F^1_1$,
the resulting map $\mathcal{F}$ can become unstable!
(more precisely, some eigenvalues of the map can sit outside
the unit circle in the complex plane).

\begin{figure}[h]
\center
\includegraphics[trim=0in 0in 0in 0in,clip,width = 5in]{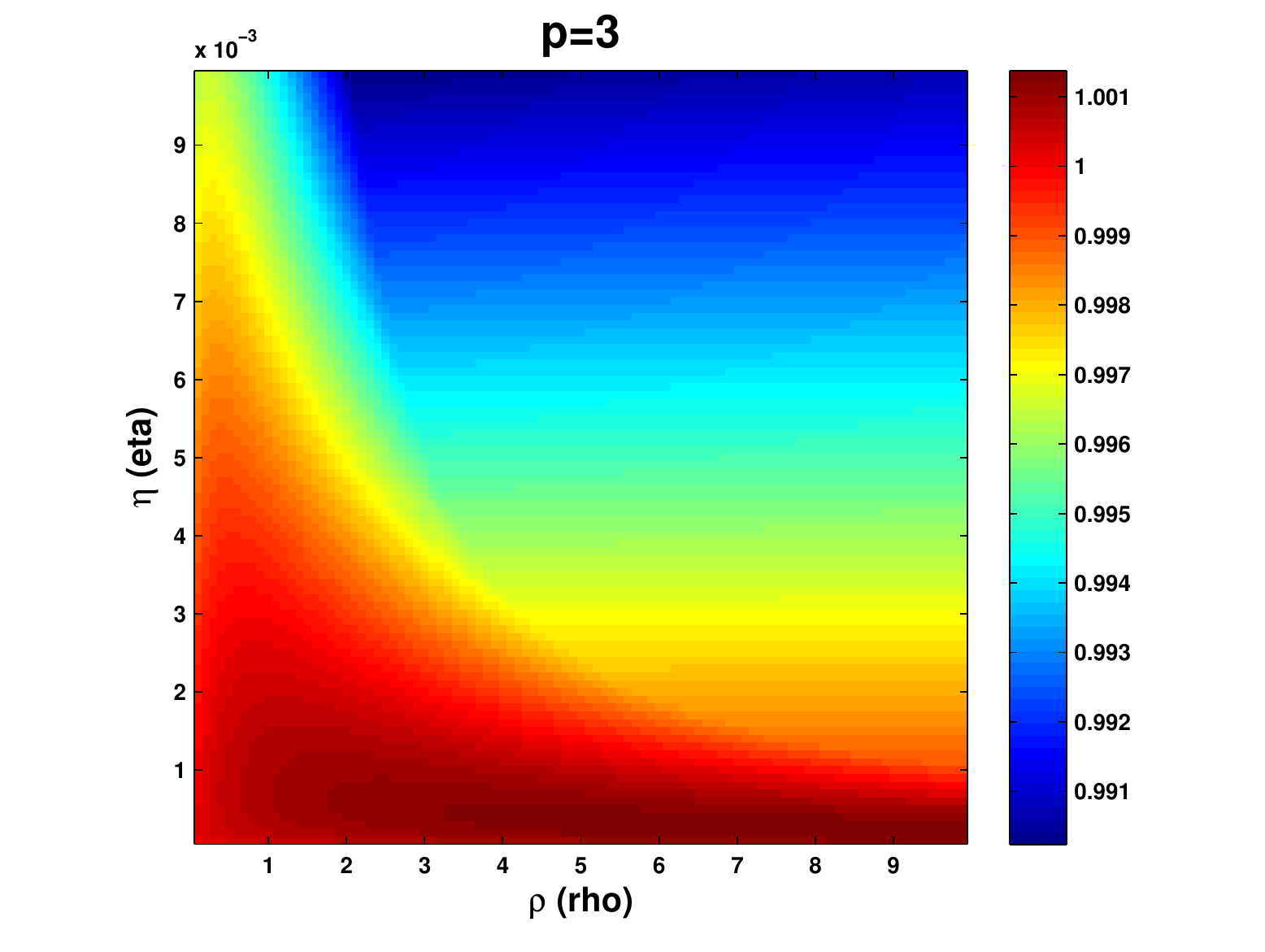} \\
\includegraphics[trim=0in 0in 0in 0in,clip,width = 5in]{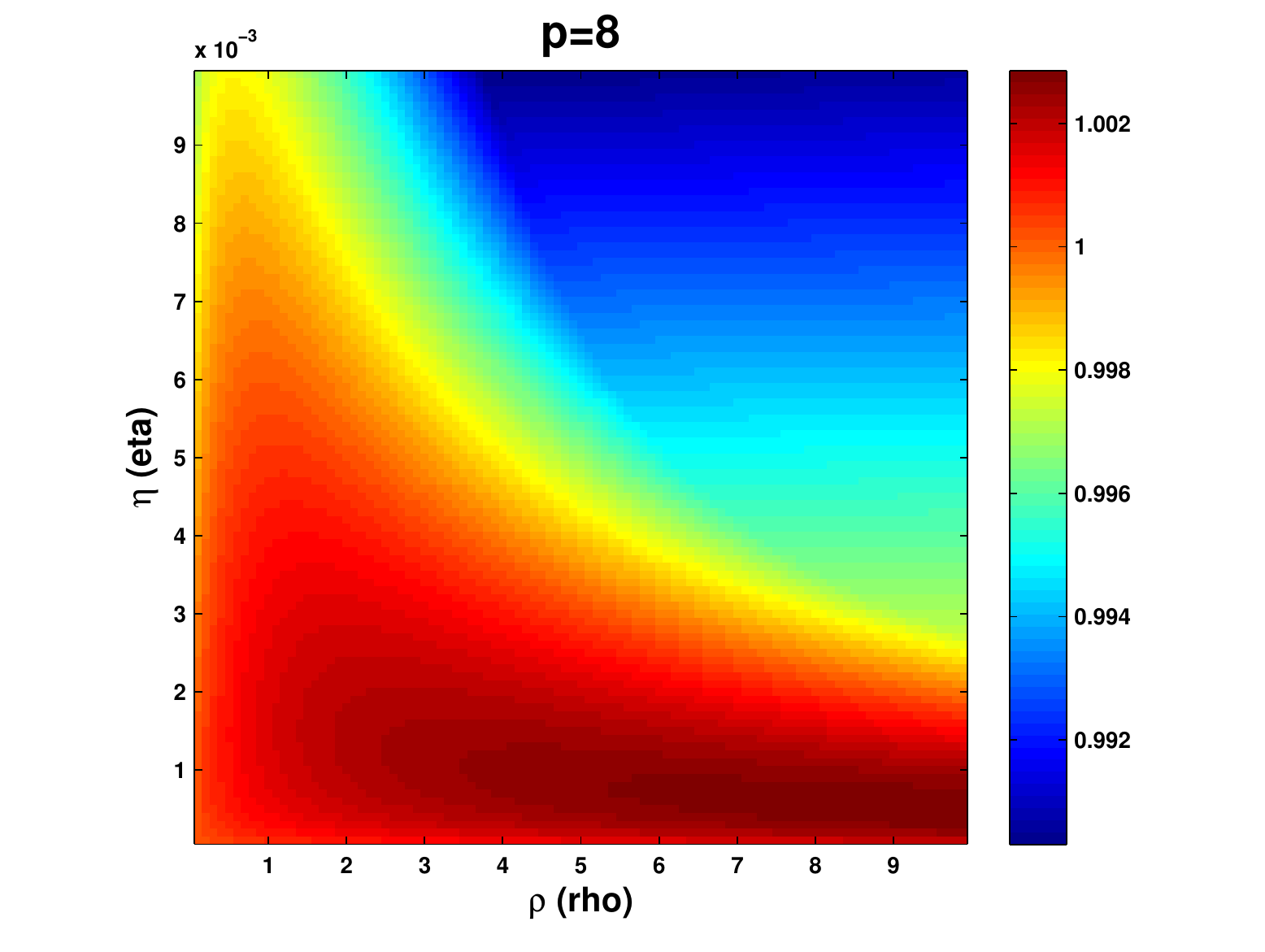}
\caption{The largest absolute eigenvalue of the linear map $\mathcal{F}=F^p_3 \circ F^p_2 \circ F^p_1 \circ \ldots \circ F^1_3 \circ F^1_2 \circ F^1_1$ as a function of $\eta \in (0,10^{-2})$ and $\rho \in (0,10)$ when $p=3$ and $p=8$.
To simulate the chaotic behavior of the \textit{ADMM} algorithm, one may pick $\eta=0.001$ and $\rho=2.5$
and initialize the state $s_0$ either randomly or with $\lambda^i_0=0,x^i_0=\tilde{x}_0=1000, \forall i=1,\ldots,p$. \textit{Figure should be read in color}.}
\label{fig:admm_fig}
\end{figure}

\begin{figure}[h]
\center
\includegraphics[trim=0in 0in 0in 0in,clip,width = 6in]{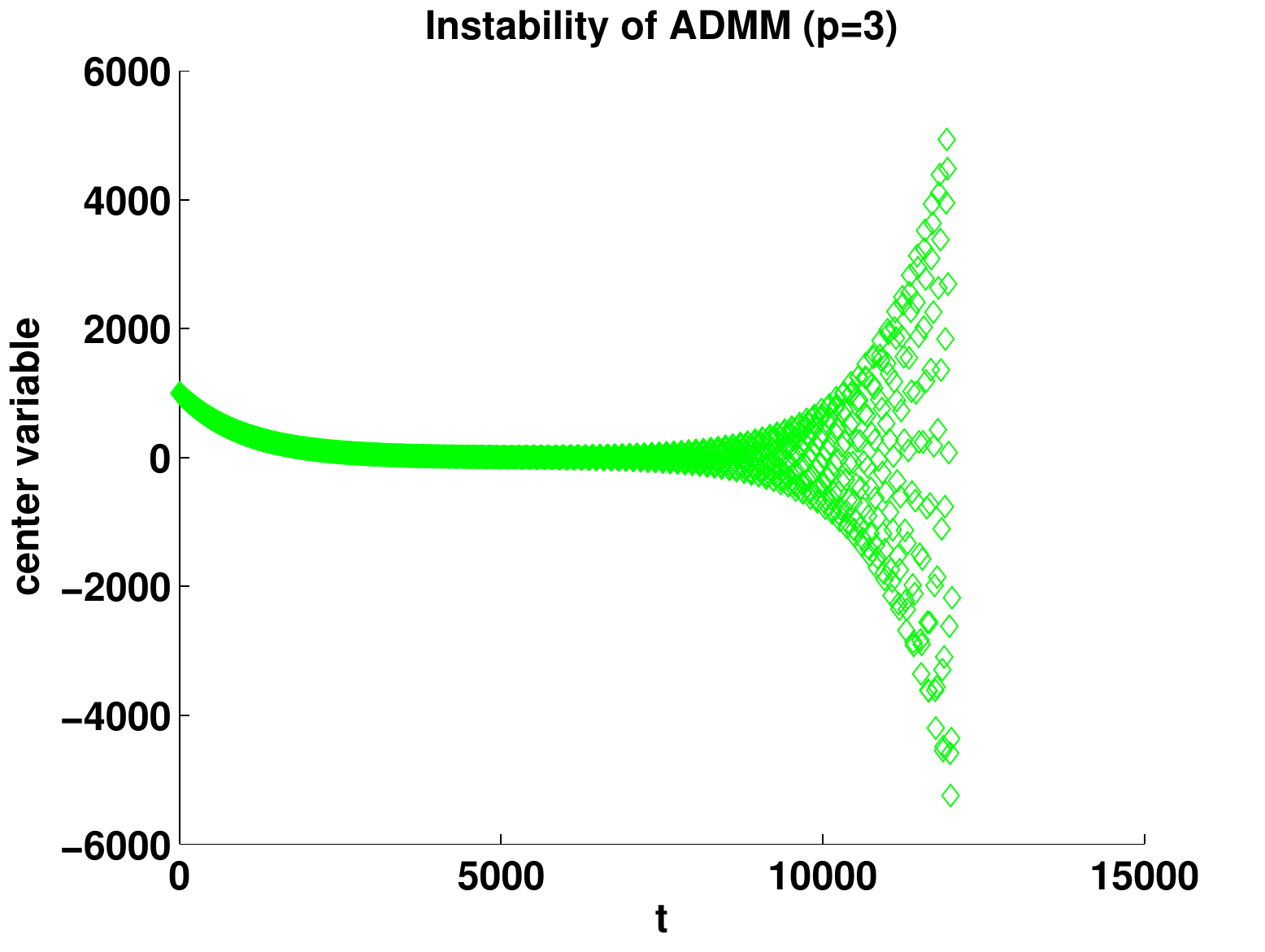}
\caption{Instability of \textit{ADMM} in the round-robin scheme. Pick $p=3$, $\eta=0.001$, $\rho=2.5$
and initialize the state $s_0$ with $\lambda^i_0=0,x^i_0=\tilde{x}_0=1000, \forall i=1,\ldots,p$. The x-axis is the time step $t$, the y-axis is the (one-dimensional) value of the center variable $\tilde{x}_t$.}
\label{fig:quod_admmchaos3}
\end{figure}

We now present the numerical conditions for which the \textit{ADMM} algorithm becomes unstable in the round-robin scheme for $p=3$ and $p=8$, by computing the largest absolute eigenvalue of the map $\mathcal{F}$.
Figure \ref{fig:admm_fig} summarizes the obtained result. We also illustrate this unstable behavior in Figure \ref{fig:quod_admmchaos3}

On the other hand, the \textit{EASGD} algorithm involves composing only symmetric linear maps due to
the elasticity. Let the state of the (dynamical) system at time $t$ be $s_t=(x^1_t,\ldots,x^p_t,\tilde{x}_t) \in \mathbb{R}^{p+1}$. The activated local worker $i$'s update in Equation~\ref{eq:easgd_update1} and the master update in Equation~\ref{eq:easgd_update2} can be written as $s_{t+1}=F^i(s_t)$.
In case of $p=2$, the map $F^1$ and $F^2$ are defined as follows
\[
F^1\!\!=\!\! \left( \begin{array}{ccc}
1-\eta-\alpha & 0 & \alpha \\
0 & 1& 0 \\
\alpha & 0 &1-\alpha \\
\end{array} \right)\!\!, \quad
F^2\!\!=\!\! \left( \begin{array}{ccc}
1 & 0 & 0 \\
0 & 1-\eta-\alpha & \alpha \\
0 & \alpha &1-\alpha \\
\end{array} \right)\!\!
\]
For the composite map $F^p \circ \ldots \circ F^1 $ to be stable, the condition that needs to be satisfied is actually the same for each $i$, and is furthermore independent of $p$ (since each linear map $F^i$ is symmetric). It essentially involves the stability of the $2\times 2$ matrix \[\left( \begin{array}{cc}
1-\eta-\alpha & \alpha \\
\alpha &1-\alpha
\end{array} \right), \]
whose two (real) eigenvalues $\lambda$ satisfy $(1-\eta-\alpha-\lambda)(1-\alpha-\lambda)=\alpha^2$. The resulting stability condition ($|\lambda| \le 1$) is simple and given as
\[0 \le \eta \le 2, \quad 0 \le \alpha \le \frac{4-2 \eta}{4-\eta}.\]

\newpage
\chapter{Performance in Deep Learning}
\label{sec:ch_exp}

In this chapter, we compare empirically the performance in deep learning of asynchronous \textit{EASGD} and \textit{EAMSGD} with the parallel method \textit{DOWNPOUR} and the sequential method \textit{SGD}, as well as their averaging and momentum variants.

All the parallel comparator methods are listed below\footnote{We have compared asynchronous \textit{ADMM}~\cite{icml2014c2_zhange14} with \textit{EASGD} in our setting as well, the performance is nearly the same. However, \textit{ADMM}'s momentum variant is not as stable when using large communication period $\tau$.}:
\begin{packed_item}
\item \textit{DOWNPOUR}~\cite{2012distbelief}, the detail and the pseudo-code of the implementation are described in Section~\ref{sec:addpcode} (Algorithm~\ref{alg:asyncD}).
\item \textit{Momentum DOWNPOUR} (\textit{MDOWNPOUR}), where the Nesterov's momentum scheme is applied to the master's update (note it is unclear how to apply it to the local workers or for the case when $\tau>1$). The pseudo-code is described in Section~\ref{sec:addpcode} (Algorithms~\ref{alg:asyncMDworker} and~\ref{alg:asyncMDmaster}).
\item A method that we call \textit{ADOWNPOUR}, where we compute the average over time of the center variable $\tilde{x}$ as follows: $z_{t+1} = (1-\alpha_{t+1}) z_t + \alpha_{t+1} \tilde{x}_t$,
and $\alpha_{t+1} = \frac{1}{t+1}$ is a moving rate, and $z_0 = \tilde{x}_0$. The $t$ denotes the master clock, which is initialized to $0$ and incremented every time the center variable $\tilde{x}$ is updated.
\item A method that we call \textit{MVADOWNPOUR}, where we compute the moving average of the center variable $\tilde{x}$ as follows: $z_{t+1} = (1-\alpha) z_t + \alpha \tilde{x}_t$,
and the moving rate $\alpha$ was chosen to be constant, and $z_0 = \tilde{x}_0$. The $t$ denotes the master clock and is defined in the same way as for the \textit{ADOWNPOUR} method.
\end{packed_item}
All the sequential comparator methods ($p=1$) are listed below:
\begin{packed_item}
\item \textit{SGD}~\cite{bottou-98x} with constant learning rate $\eta$.
\item \textit{Momentum SGD} (\textit{MSGD})~\cite{icml2013_sutskever13} with constant (Nesterov's) momentum rate $\delta$.
\item \textit{ASGD}~\cite{polyak1992acceleration} with moving rate $\alpha_{t+1} = \frac{1}{t+1}$.
\item \textit{MVASGD}~\cite{polyak1992acceleration} with moving rate $\alpha$ set to a constant.
\end{packed_item}

We perform experiments on two benchmark datasets: CIFAR-10 (we refer to it as \textit{CIFAR})\footnote{Downloaded from \url{http://www.cs.toronto.edu/~kriz/cifar.html}.} and ImageNet ILSVRC 2013 (we refer to it as \textit{ImageNet})\footnote{Downloaded from \url{http://image-net.org/challenges/LSVRC/2013}.}. We focus on the image classification task with deep convolutional neural networks. We first explain the experimental setup in Section~\ref{sec:exp_setup1} and then present the main experimental results in Section~\ref{sec:exp_result1}. We present further experimental results in Section~\ref{sec:exp_result2} and discuss the effect of the averaging, the momentum, the learning rate, the communication period, the data and parameter communication tradeoff, and finally the speedup.

\section{Experimental setup}
\label{sec:exp_setup1}
For all our experiments we use a GPU-cluster interconnected with InfiniBand. Each node has $4$ Titan GPU processors where each local worker corresponds to one GPU processor. The center variable of the master is stored and updated on the centralized parameter server~\cite{2012distbelief}. Our implementation is available at \url{https://github.com/sixin-zh/mpiT}.

To describe the architecture of the convolutional neural network, we will first introduce a notation. Let $(c,x,y)$ denotes the size of the input image to each layer, where $c$ is the number of color channels and $(x,y)$ denotes the horizontal and the vertical dimension of the input. Let $C$ denotes the fully-connected convolutional operator and let $R$ denotes the rectified linear non-linearity (relu, c.f.~\cite{nair2010rectified}), $P$ denotes the max pooling operator, $L$ denotes the linear operator and $D$ denotes the dropout operator with rate equal to $0.5$ and $S$ denotes the the softmax nonlinearity. We use the cross-entropy loss for the classification.

For the \textit{ImageNet} experiment we use the similar approach to~\cite{2013arXiv1312.6229S} with the following $11$-layer convolutional neural network:
$(3,221,221)
\xrightarrow[(7,7,2,2)]{C,R}
(96,108,108)
\xrightarrow[(3,3,3,3)]{P}
(96,36,36)
\xrightarrow[(5,5,1,1)]{C,R}
(256,32,32)
\xrightarrow[(2,2,2,2)]{P}
(256,16,16)
\xrightarrow[(3,3,1,1)]{C,R}
(384,14,14)
\xrightarrow[(2,2,1,1)]{C,R} \\
(384,13,13)
\xrightarrow[(2,2,1,1)]{C,R}
(256,12,12)
\xrightarrow[(2,2,2,2)]{P}
(256,6,6)
\xrightarrow[0.5]{L,R,D}
(4096,1,1)
\xrightarrow[0.5]{L,R,D} \\
(4096,1,1)
\xrightarrow[]{L,S}
(1000,1,1).$

For the \textit{CIFAR} experiment we use the similar approach to~\cite{DBLPWanZZLF13} with the following $7$-layer convolutional neural network:
$(3,28,28)\xrightarrow[(5,5,1,1)]{C,R}(64,24,24)\xrightarrow[(2,2,2,2)]{P}
(64,12,12)\xrightarrow[(5,5,1,1)]{C,R}(128,8,8)\xrightarrow[(2,2,2,2)]{P}
(128,4,4)\xrightarrow[(3,3,1,1)]{C,R}(64,2,2)\xrightarrow[0.5]{L,R,D}(256,1,1)\xrightarrow[]{L,S}(10,1,1)$.

Note that the numbers below the rightarrow of the C and P operator represent the kernel size (first horizontal and then vertical), the stride size (first horizontal and then vertical) and the padding size (if exists, first horizontal and then vertical) on each of the two sides of the image. The number below the rightarrow of the D operator emphasizes the dropout rate 0.5~\cite{JMLR:v15:srivastava14a}.

In our experiments, all the methods we run use the same initial parameter chosen randomly, except that we set all the biases to zero for \textit{CIFAR} case and to 0.1 for \textit{ImageNet} case. This parameter is used to initialize the master and all the local workers\footnote{On the contrary, initializing the local workers and the master with different random seeds 'traps' the algorithm in the symmetry breaking phase.}. We add $l_2$-regularization $\frac{\lambda}{2}\norm{}{x}^2$ to the loss function $F(x)$. For \textit{ImageNet} we use $\lambda = 10^{-5}$ and for \textit{CIFAR} we use $\lambda = 10^{-4}$. We also compute the stochastic gradient using mini-batches of sample size $128$.

\subsubsection{Data preprocessing}

For the \textit{ImageNet} experiment, we re-size each RGB image so that the smallest dimension is $256$ pixels. We also re-scale each pixel value to the interval $[0,1]$. We then extract random crops (and their horizontal flips) of size $3\times 221\times 221$ pixels and present these to the network in mini-batches of size $128$.

For the \textit{CIFAR} experiment, we use the original RGB image of size $3 \times 32 \times 32$. As before, we re-scale each pixel value to the interval $[0,1]$. We then extract random crops (and their horizontal flips) of size $3\times 28\times 28$ pixels and present these to the network in mini-batches of size $128$.

The training and test loss and the test error are only computed from the center patch ($3 \times 28 \times 28$) for the \textit{CIFAR} experiment and the center patch ($3 \times 221 \times 221$) for the \textit{ImageNet} experiment.

\subsubsection{Data prefetching (Sampling the dataset by the local workers in parallel)}
We will now explain precisely how the dataset is sampled by each local worker as uniformly and efficiently as possible. The general parallel data loading scheme on a single machine is as follows: we use $k$ CPUs, where $k=8$, to load the data in parallel. Each data loader reads from the memory-mapped (mmap) file a chunk of $c$ raw images (preprocessing was described in the previous subsection) and their labels (for \textit{CIFAR} $c = 512$ and for \textit{ImageNet} $c = 64$). For the \textit{CIFAR}, the mmap file of each data loader contains the entire dataset whereas for \textit{ImageNet}, each mmap file of each data loader contains different $1/k$ fractions of the entire dataset. A chunk of data is always sent by one of the data loaders to the first worker who requests the data. The next worker requesting the data from the same data loader will get the next chunk. Each worker requests in total $k$ data chunks from $k$ different data loaders and then process them before asking for new data chunks. Notice that each data loader cycles\footnote{Its advantage is observed in~\cite{bottou-slds-open-problem-2009}.} through the data in the mmap file, sending consecutive chunks to the workers in order in which it receives requests from them. When the data loader reaches the end of the mmap file, it selects the address in memory uniformly at random from the interval $[0,s]$, where $s = (\textsf{number of images in the mmap file} \text{\:\:modulo\:\:} \textsf{mini-batch size})$, and uses this address to start cycling again through the data in the mmap file. After the local worker receives the $k$ data chunks from the data loaders, it shuffles them and divides it into mini-batches of size $128$.

\section{Experimental results}
\label{sec:exp_result1}

For all experiments in this section we use \textit{EASGD} with $\beta = 0.9$ and $\alpha = \beta / p$,
for all momentum-based methods we set the momentum term $\delta = 0.99$ and finally for \textit{MVADOWNPOUR} we set the moving rate to $\alpha = 0.001$. We start with the experiment on \textit{CIFAR} dataset with $p=4$ local workers running on a single computing node.

For all the methods, we examined the communication periods from the following set $\tau = \{1,4,16,64\}$. For each method we examined a wide range of learning rates. The learning rates explored in all experiments are summarized in Table~\ref{tab:learningrate},~\ref{tab:learningrate2} and~\ref{tab:learningrate3}. The \textit{CIFAR} experiment was run $3$ times independently from the same random initialization and for each method we report its best performance measured by the smallest achievable test error.

\begin{table}[p]
\center
\setlength{\tabcolsep}{4pt}
\caption{Learning rates explored for each method shown in Figure~\ref{fig:CIFAR_tau1},~\ref{fig:CIFAR_tau4},~\ref{fig:CIFAR_tau16} and~\ref{fig:CIFAR_tau64} (\textit{CIFAR} experiment).}
\begin{tabular}{c|c|}
\cline{2-2}
\multirow{2}{*}{} & $\eta$\\
\hline
\multicolumn{1}{|c|}{\multirow{1}{*}{EASGD}}& $\{0.05,0.01,0.005\}$\\
\hline
\multicolumn{1}{|c|}{\multirow{1}{*}{EAMSGD}}& $\{0.01,0.005,0.001\}$\\
\hline
\multicolumn{1}{|c|}{\multirow{1}{*}{DOWNPOUR}}& \\
\multicolumn{1}{|c|}{\multirow{1}{*}{ADOWNPOUR}}& $\{0.005,0.001,0.0005\}$\\
\multicolumn{1}{|c|}{\multirow{1}{*}{MVADOWNPOUR}}& \\
\hline
\multicolumn{1}{|c|}{\multirow{1}{*}{MDOWNPOUR}}& $\{0.00005,0.00001,0.000005\}$\\
\hline
\hline
\multicolumn{1}{|c|}{\multirow{1}{*}{SGD, ASGD, MVASGD}}& $\{0.05,0.01,0.005\}$\\
\hline
\multicolumn{1}{|c|}{\multirow{1}{*}{MSGD}}& $\{0.001,0.0005,0.0001\}$\\
\hline
\end{tabular}
\label{tab:learningrate}
\end{table}

\begin{table}[p]
\center
\setlength{\tabcolsep}{4pt}
\caption{Learning rates explored for each method shown in Figure~\ref{fig:CIFAR2_p4},~\ref{fig:CIFAR2_p8} and~\ref{fig:CIFAR2_p16} (\textit{CIFAR} experiment).}
\begin{tabular}{c|c|}
\cline{2-2}
\multirow{2}{*}{} & $\eta$\\
\hline
\multicolumn{1}{|c|}{\multirow{1}{*}{EASGD}}& $\{0.05,0.01,0.005\}$\\
\hline
\multicolumn{1}{|c|}{\multirow{1}{*}{EAMSGD}}& $\{0.01,0.005,0.001\}$\\
\hline
\multicolumn{1}{|c|}{\multirow{1}{*}{DOWNPOUR}}& $\{0.005,0.001,0.0005\}$\\
\hline
\multicolumn{1}{|c|}{\multirow{1}{*}{MDOWNPOUR}}& $\{0.00005,0.00001,0.000005\}$\\
\hline
\hline
\multicolumn{1}{|c|}{\multirow{1}{*}{SGD, ASGD, MVASGD}}& $\{0.05,0.01,0.005\}$\\
\hline
\multicolumn{1}{|c|}{\multirow{1}{*}{MSGD}}& $\{0.001,0.0005,0.0001\}$\\
\hline
\end{tabular}
\label{tab:learningrate2}
\end{table}

\begin{table}[p]
\center
\setlength{\tabcolsep}{4pt}
\caption{Learning rates explored for each method shown in Figure~\ref{fig:ImageNet_p4} and~\ref{fig:ImageNet_p8} (\textit{ImageNet} experiment).}
\begin{tabular}{c|c|}
\cline{2-2}
\multirow{2}{*}{} & $\eta$\\
\hline
\multicolumn{1}{|c|}{\multirow{1}{*}{EASGD}}& $0.1$\\
\hline
\multicolumn{1}{|c|}{\multirow{1}{*}{EAMSGD}}& $0.001$\\
\hline
\multicolumn{1}{|c|}{\multirow{1}{*}{DOWNPOUR}}& for $p=4$: $0.02$\\
\multicolumn{1}{|c|}{\multirow{1}{*}{}}& for $p=8$: $0.01$\\
\hline
\hline
\multicolumn{1}{|c|}{\multirow{1}{*}{SGD, ASGD, MVASGD}}& $0.05$\\
\hline
\multicolumn{1}{|c|}{\multirow{1}{*}{MSGD}}& $0.0005$\\
\hline
\end{tabular}
\label{tab:learningrate3}
\end{table}

\begin{figure}[p]
\center
% $\tau = 1$\\
\includegraphics[trim=0cm 0cm 0cm 0cm,clip,width = 3in]{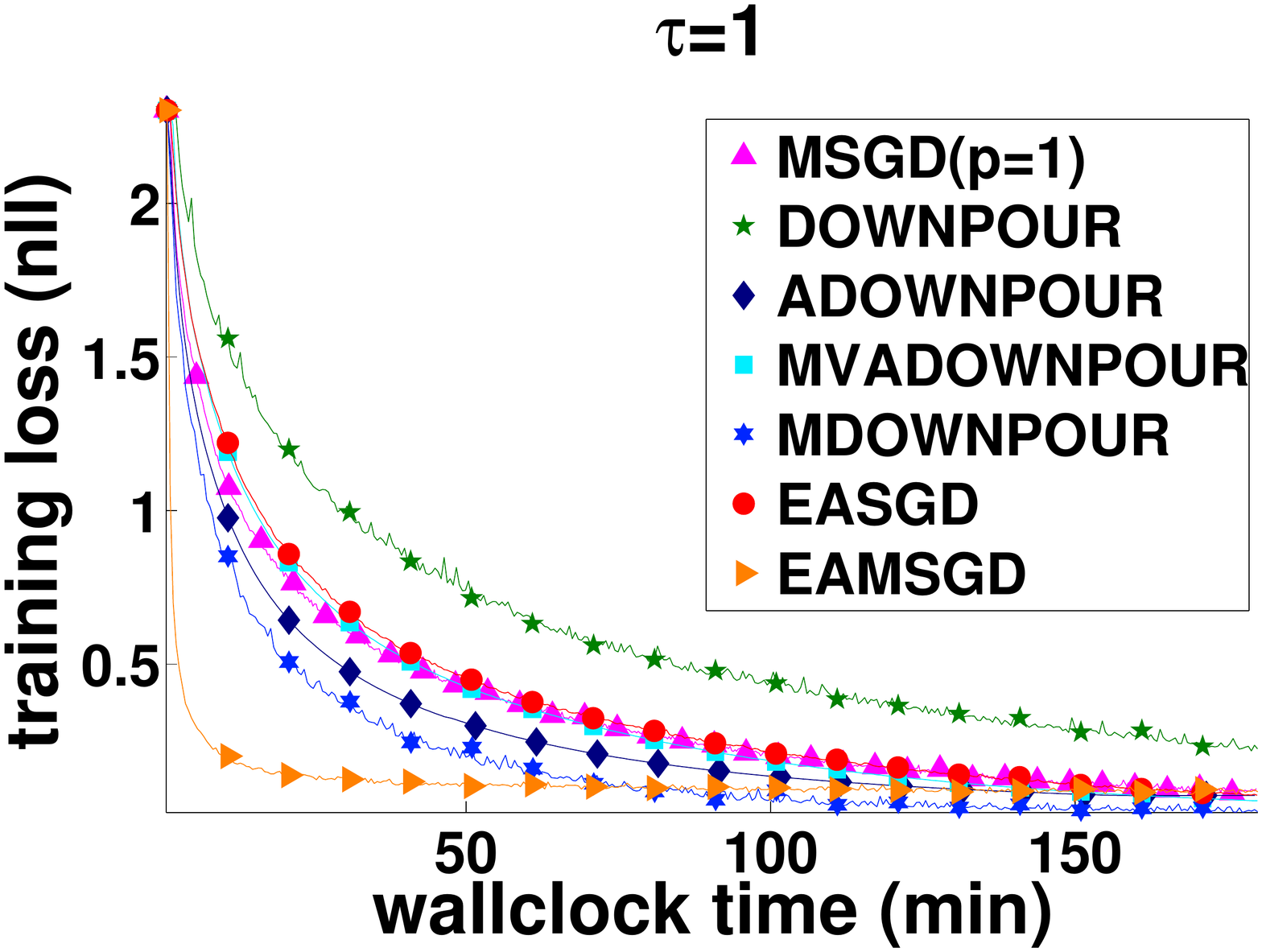}
\hspace{-0.3in}\includegraphics[trim=0cm 0cm 0cm 0cm,clip,width = 3in]{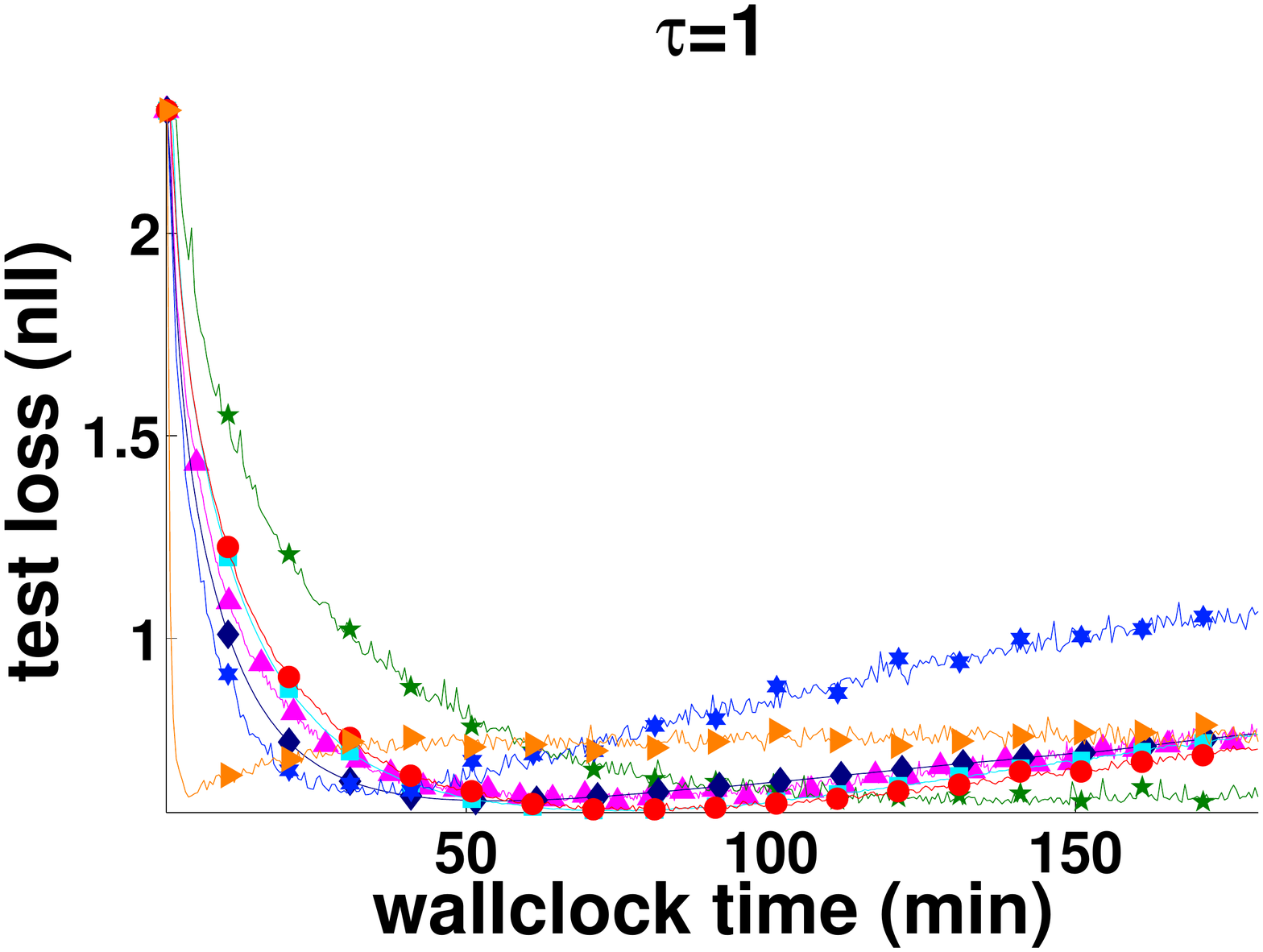} \\
\includegraphics[trim=0cm 0cm 0cm 0cm,clip,width = 3in]{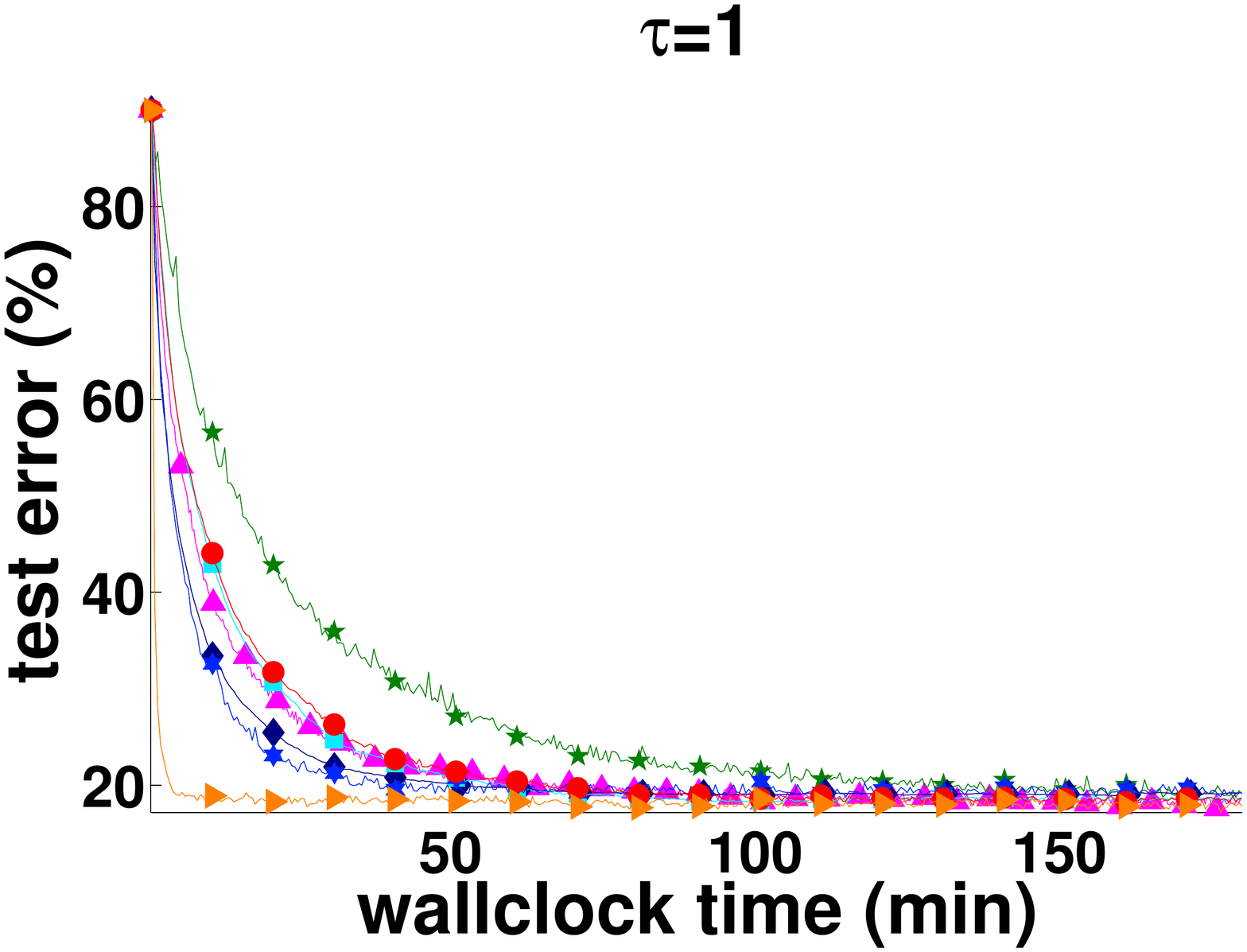}
\hspace{-0.3in}\includegraphics[trim=0cm 0cm 0cm 0cm,clip,width = 3in]{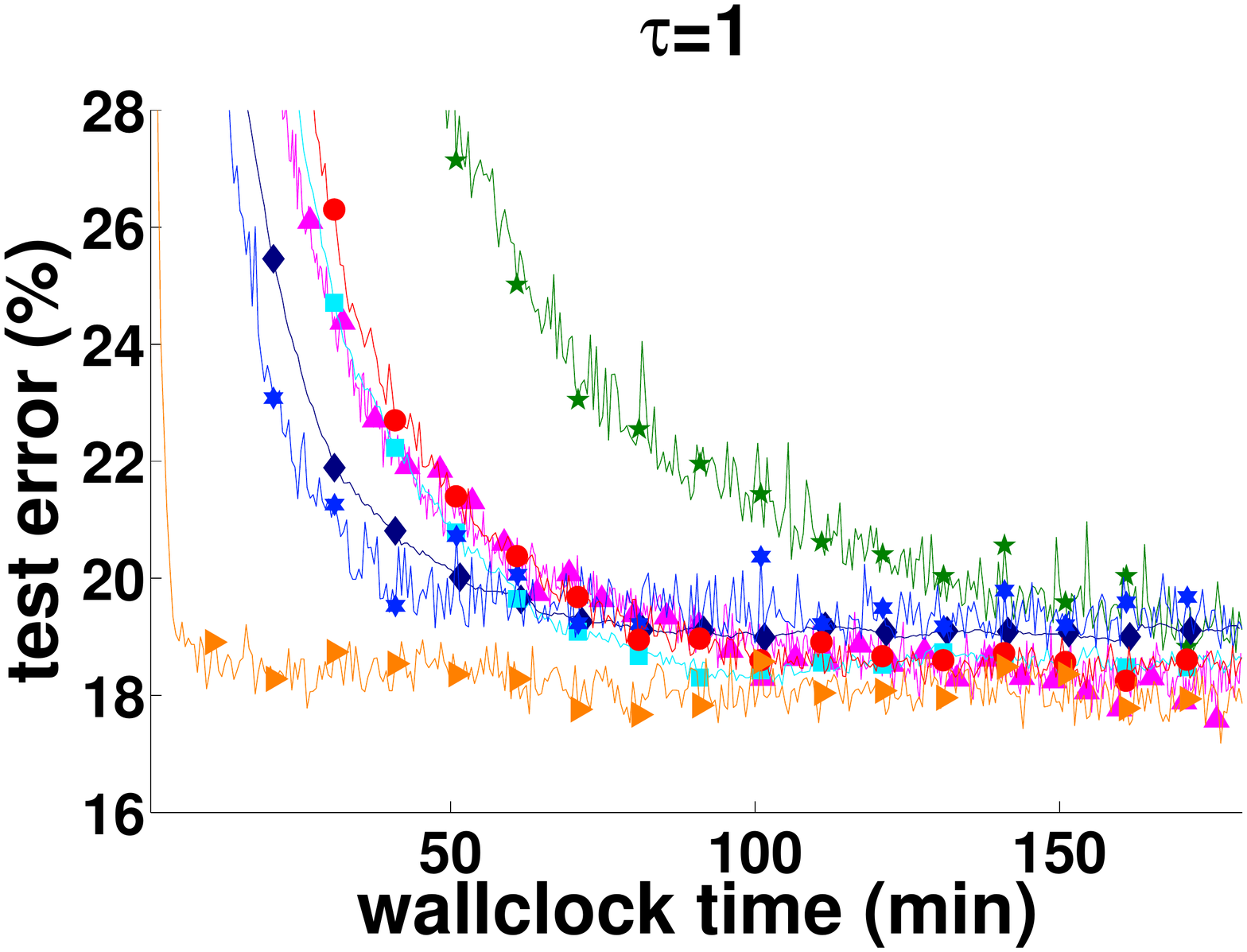}\\
\caption{Training and test loss and the test error for the center variable versus a wallclock time for communication period $\tau=1$ on \textit{CIFAR} dataset with the $7$-layer convolutional neural network. $p=4$.}
\label{fig:CIFAR_tau1}
\end{figure}

\begin{figure}[p]
\center
% $\tau = 4$\\
\includegraphics[trim=0cm 0cm 0cm 0cm,clip,width = 3in]{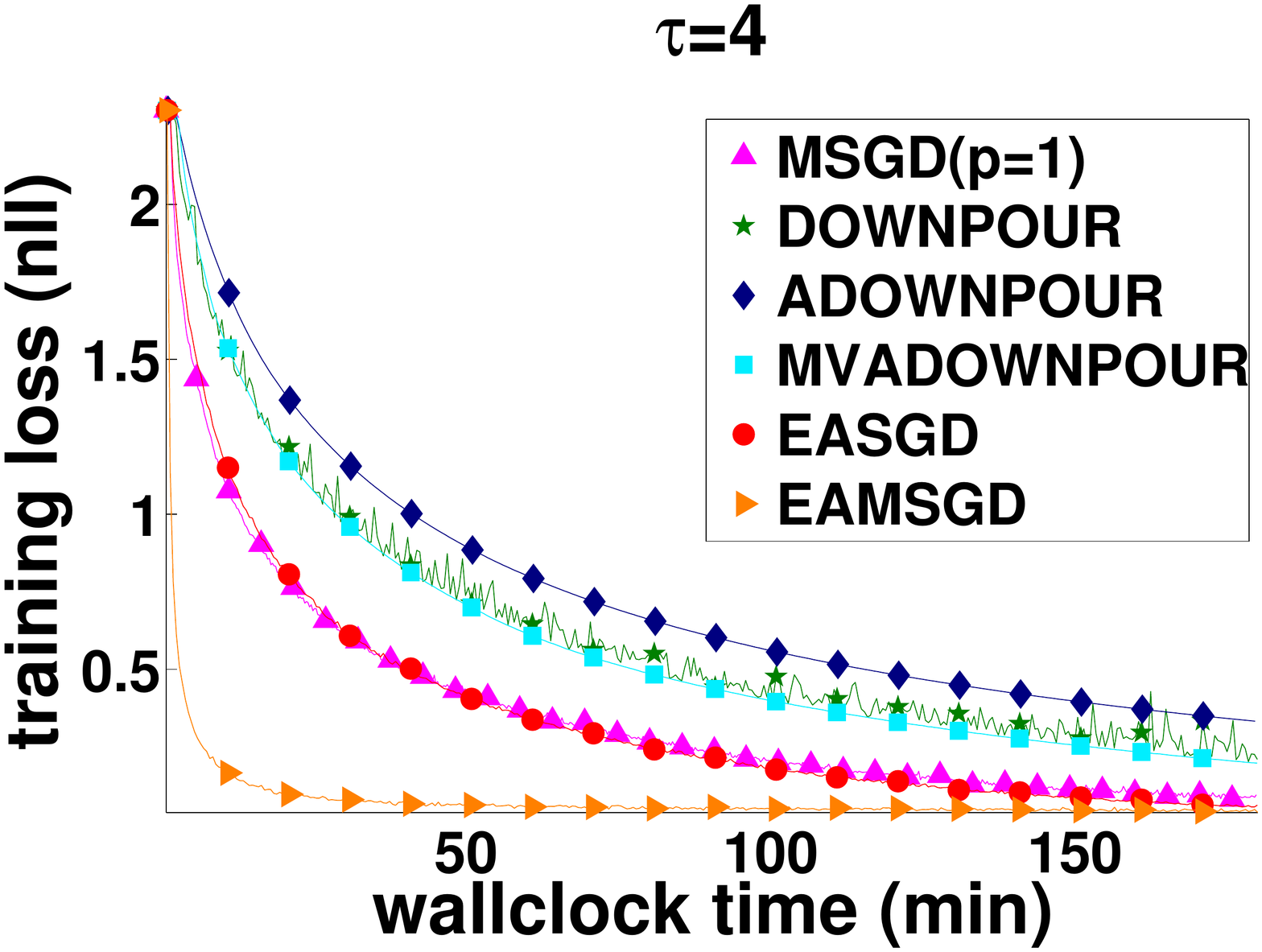}
\hspace{-0.3in}\includegraphics[trim=0cm 0cm 0cm 0cm,clip,width = 3in]{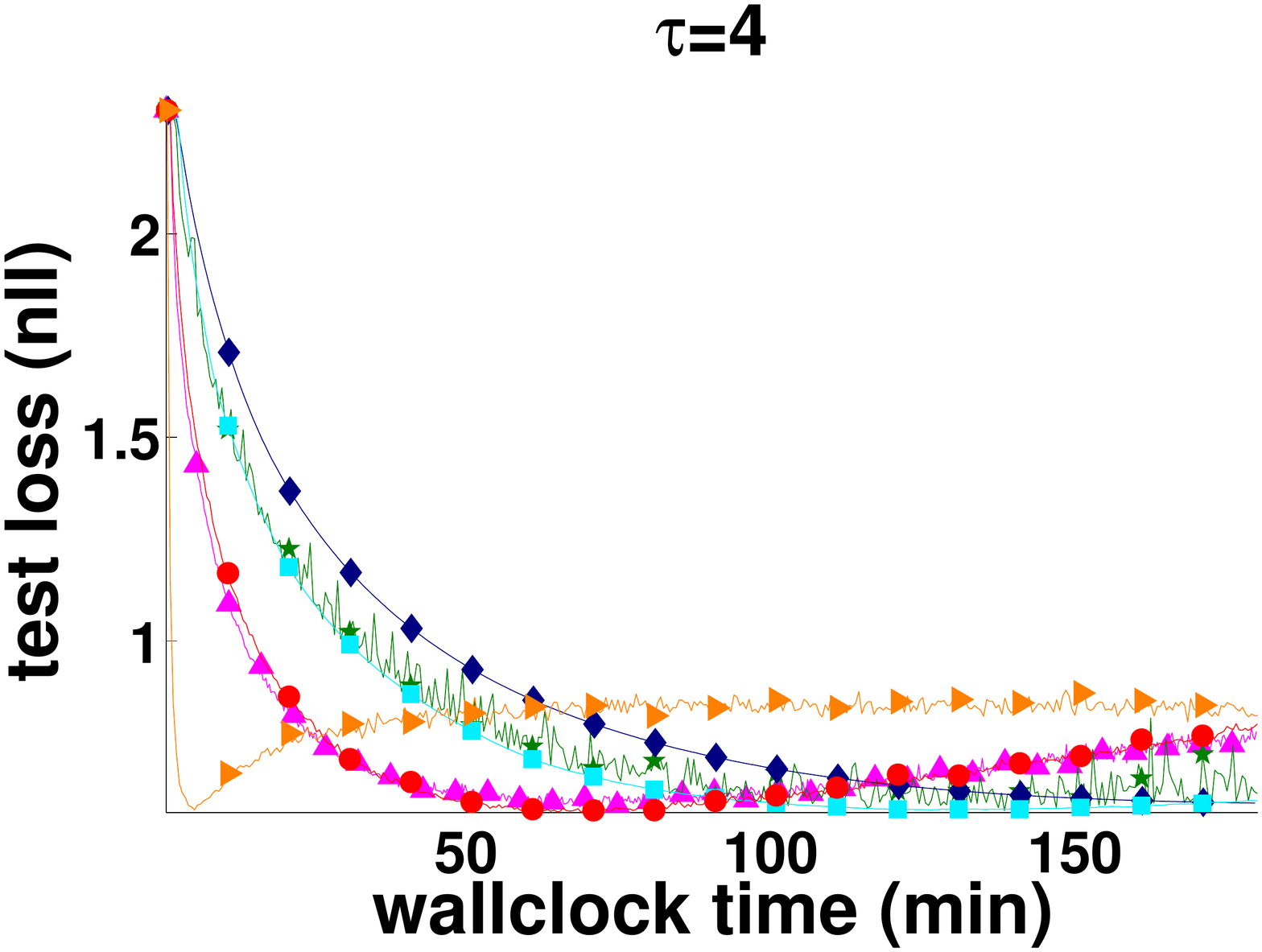} \\
\includegraphics[trim=0cm 0cm 0cm 0cm,clip,width = 3in]{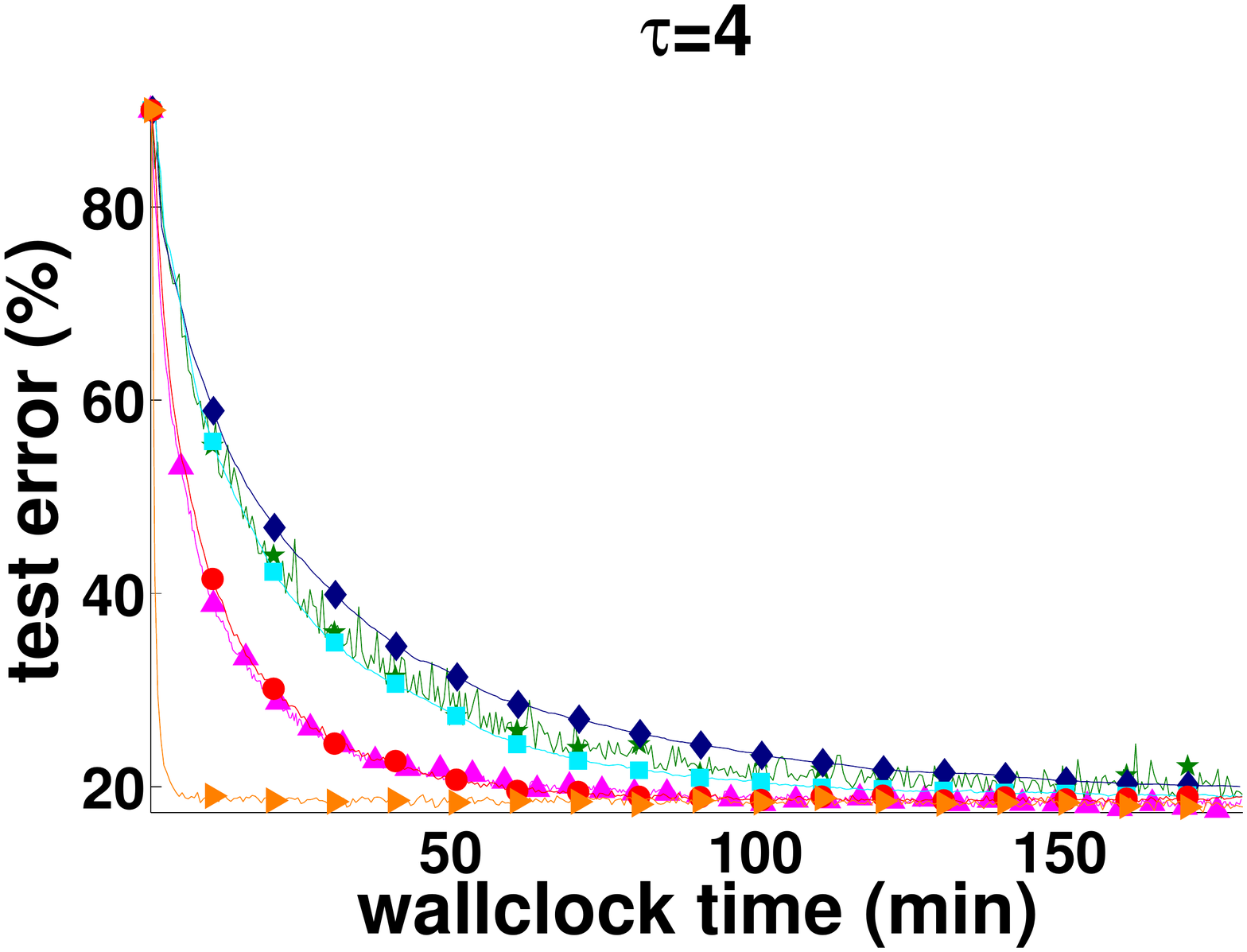}
\hspace{-0.3in}\includegraphics[trim=0cm 0cm 0cm 0cm,clip,width = 3in]{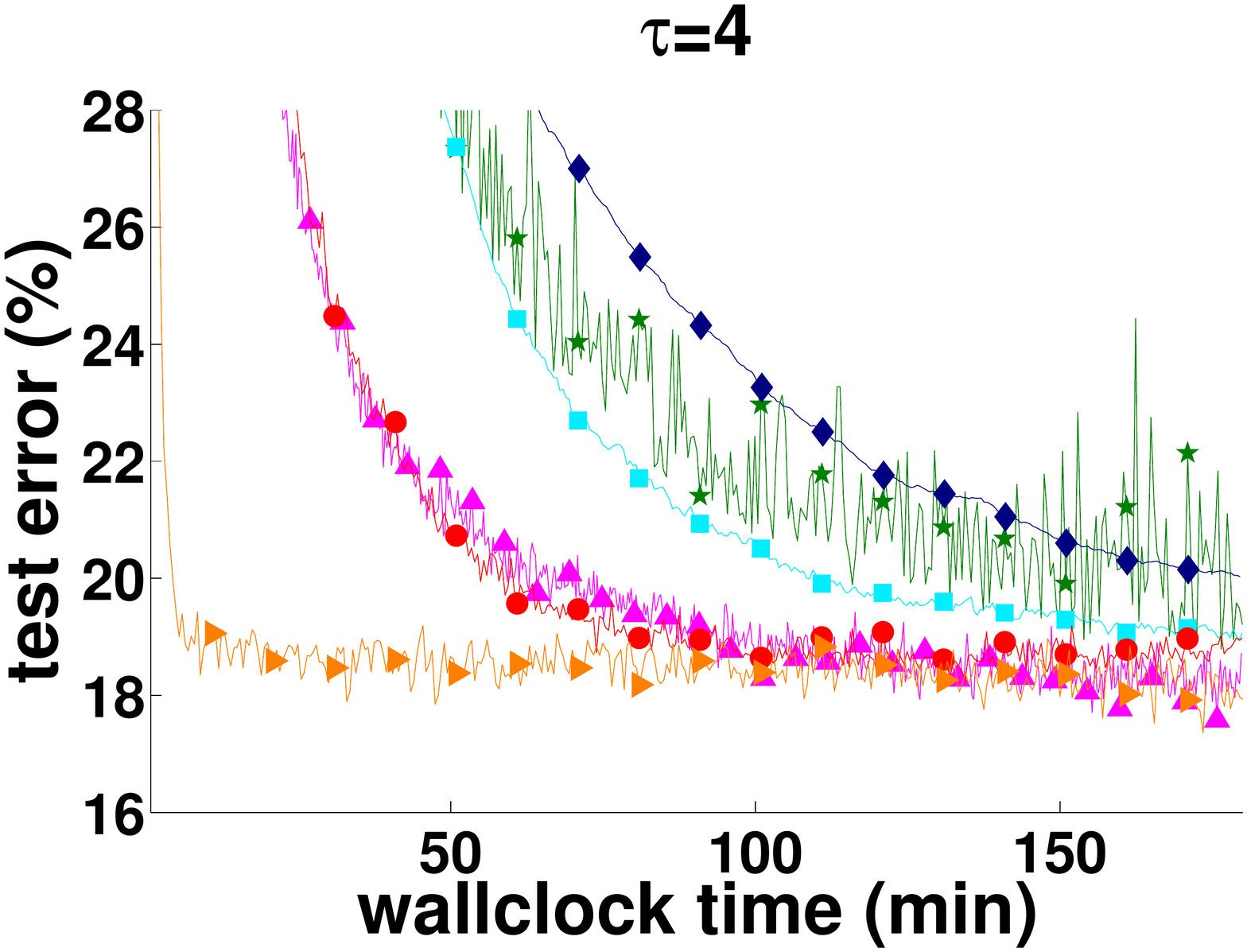}\\
\caption{Training and test loss and the test error for the center variable versus a wallclock time for communication period $\tau=4$ on \textit{CIFAR} dataset with the $7$-layer convolutional neural network. $p=4$.}
\label{fig:CIFAR_tau4}
\end{figure}

\begin{figure}[p]
\center
% $\tau = 16$\\
\includegraphics[trim=0cm 0cm 0cm 0cm,clip,width = 3in]{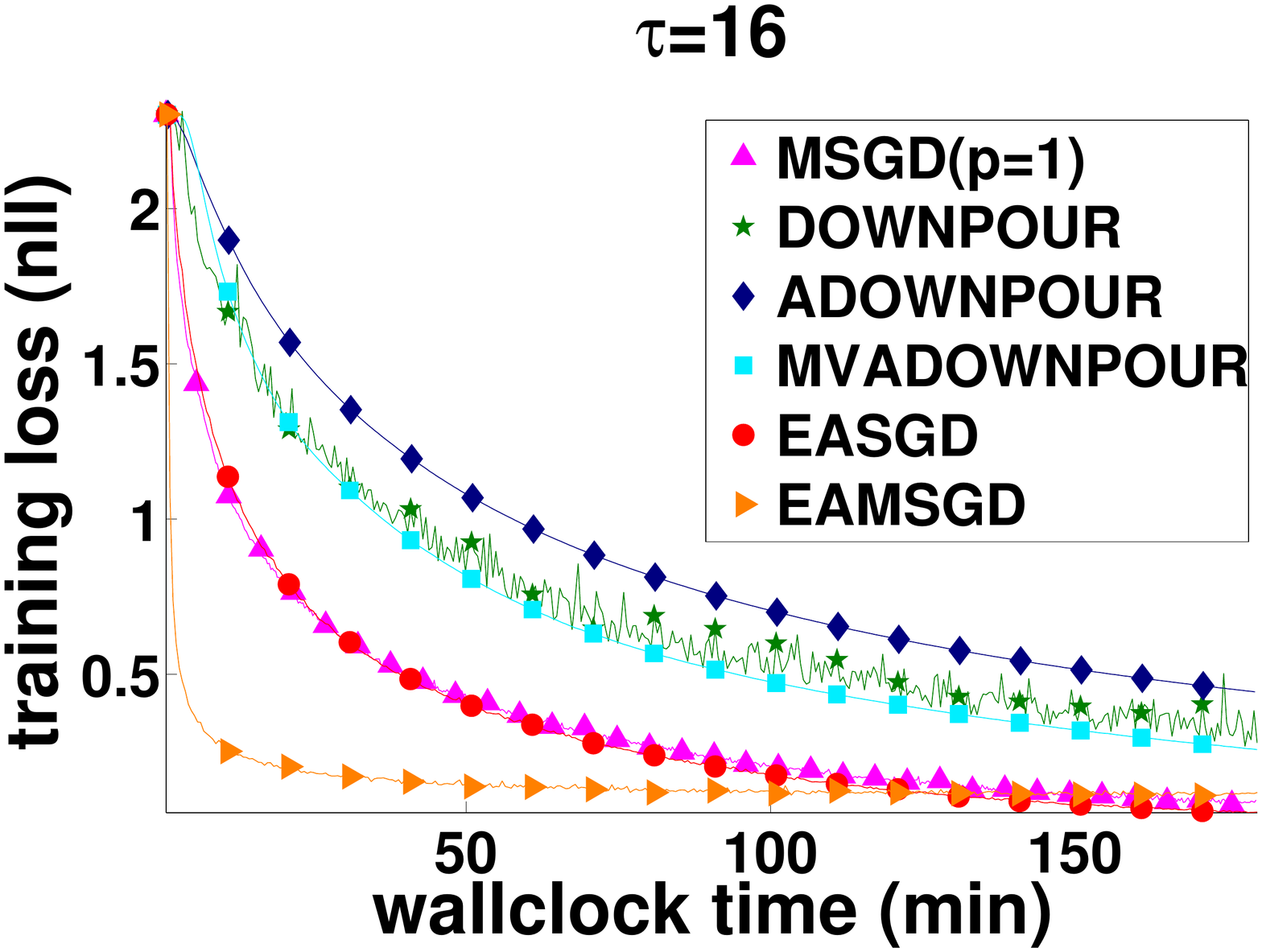}
\hspace{-0.3in}\includegraphics[trim=0cm 0cm 0cm 0cm,clip,width = 3in]{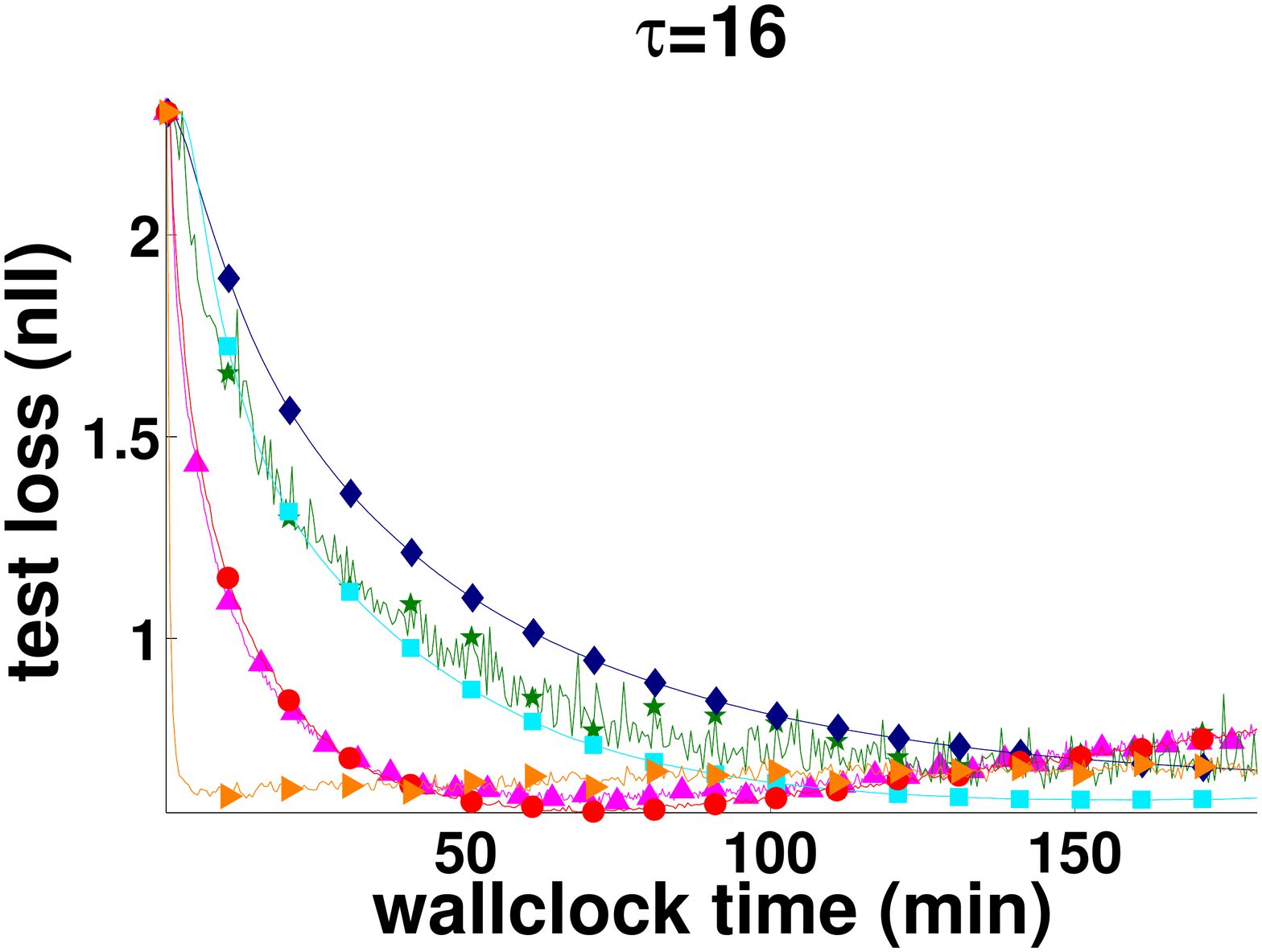} \\
\includegraphics[trim=0cm 0cm 0cm 0cm,clip,width = 3in]{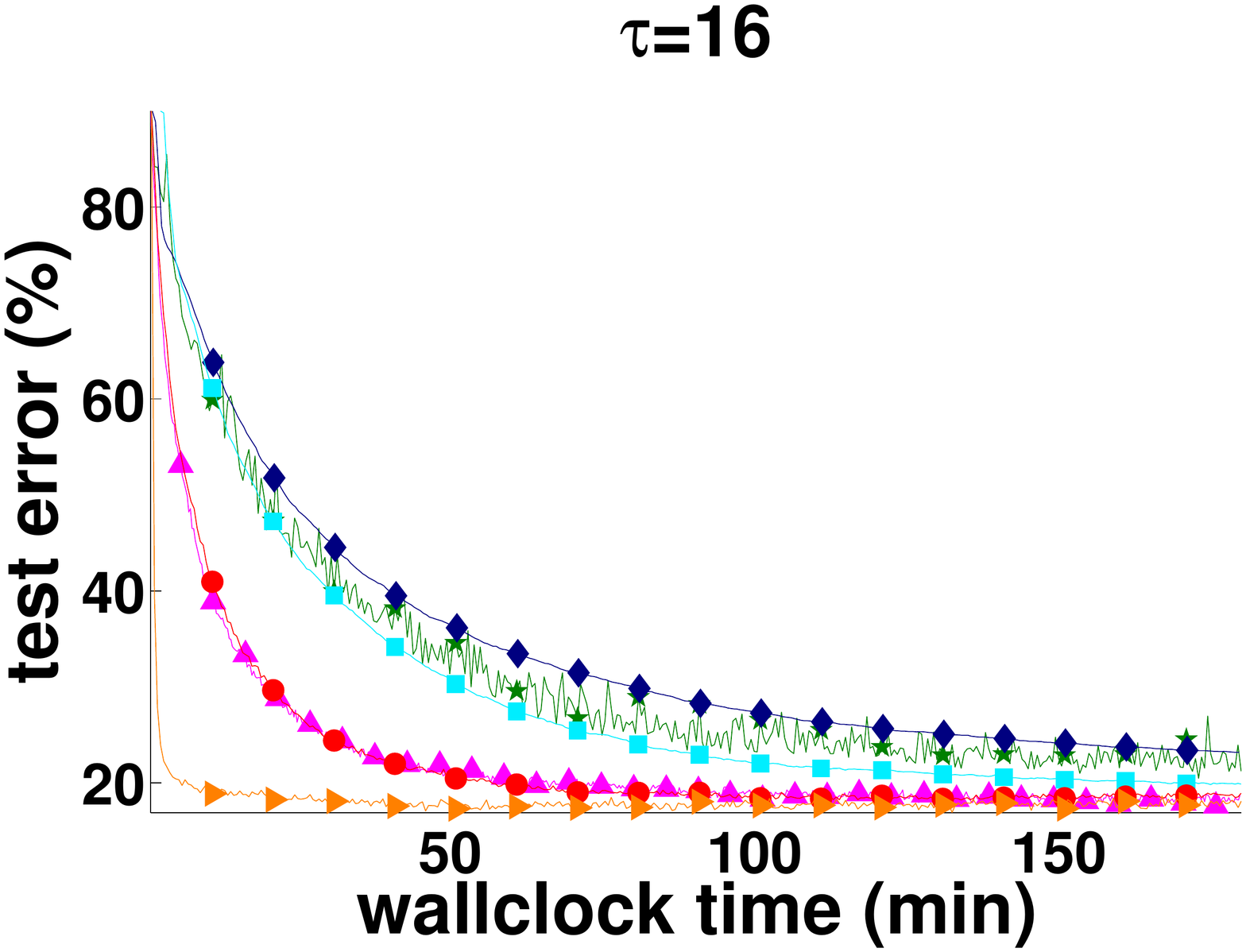}
\hspace{-0.3in}\includegraphics[trim=0cm 0cm 0cm 0cm,clip,width = 3in]{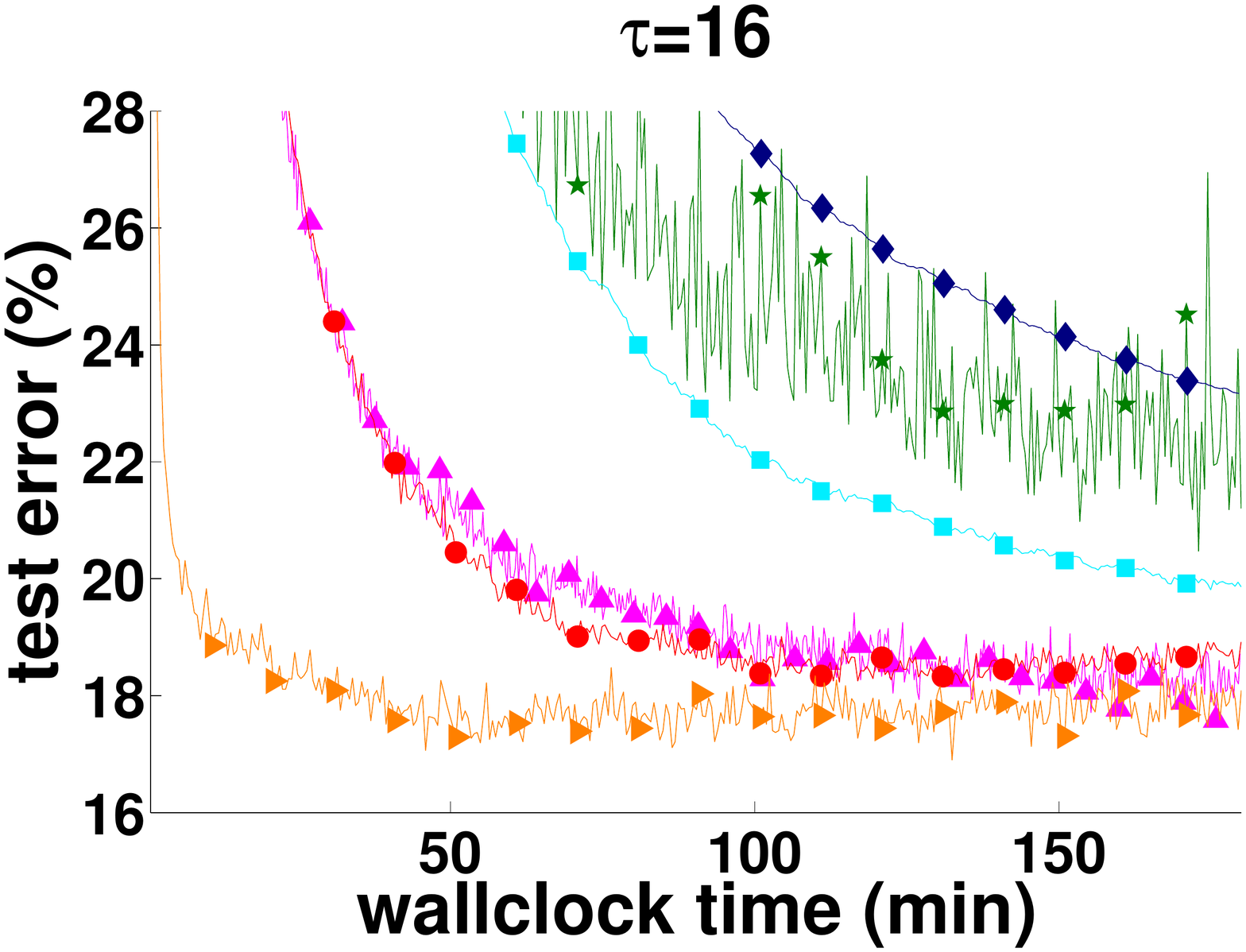}\\
\caption{Training and test loss and the test error for the center variable versus a wallclock time for communication period $\tau=16$ on \textit{CIFAR} dataset with the $7$-layer convolutional neural network. $p=4$.}
\label{fig:CIFAR_tau16}
\end{figure}

\begin{figure}[p]
\center
% $\tau = 64$\\
\includegraphics[trim=0cm 0cm 0cm 0cm,clip,width = 3in]{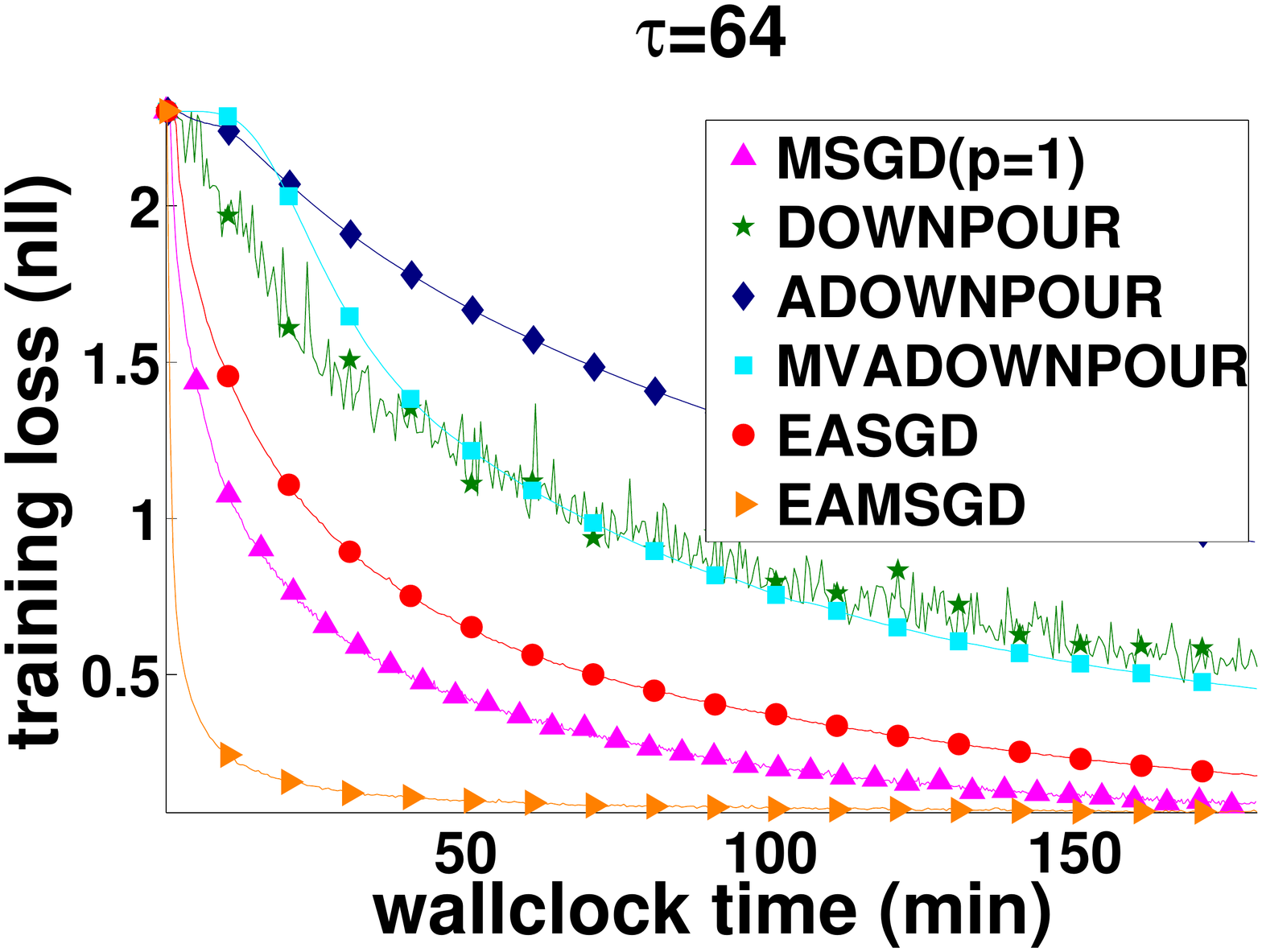}
\hspace{-0.3in}\includegraphics[trim=0cm 0cm 0cm 0cm,clip,width = 3in]{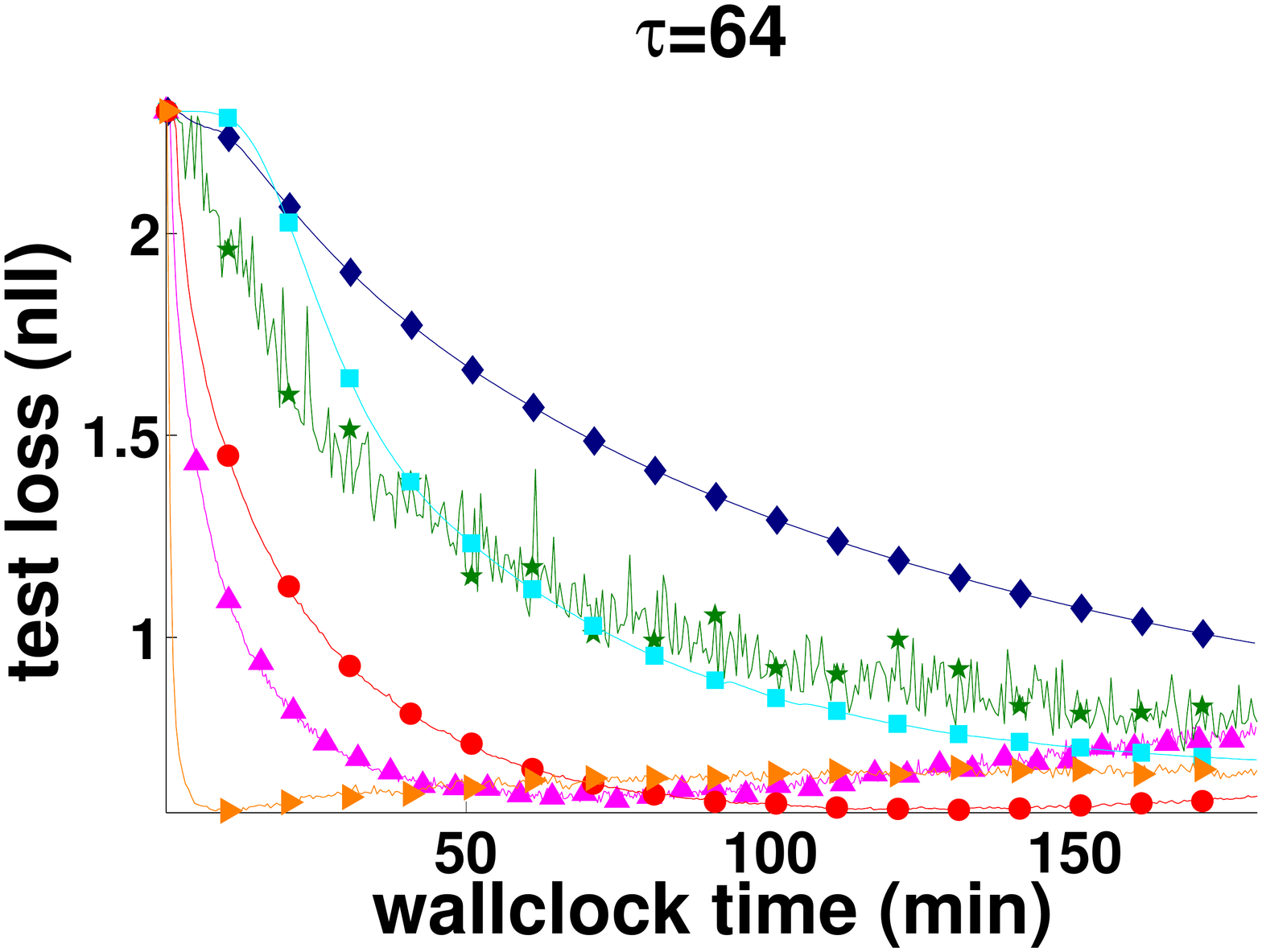} \\
\includegraphics[trim=0cm 0cm 0cm 0cm,clip,width = 3in]{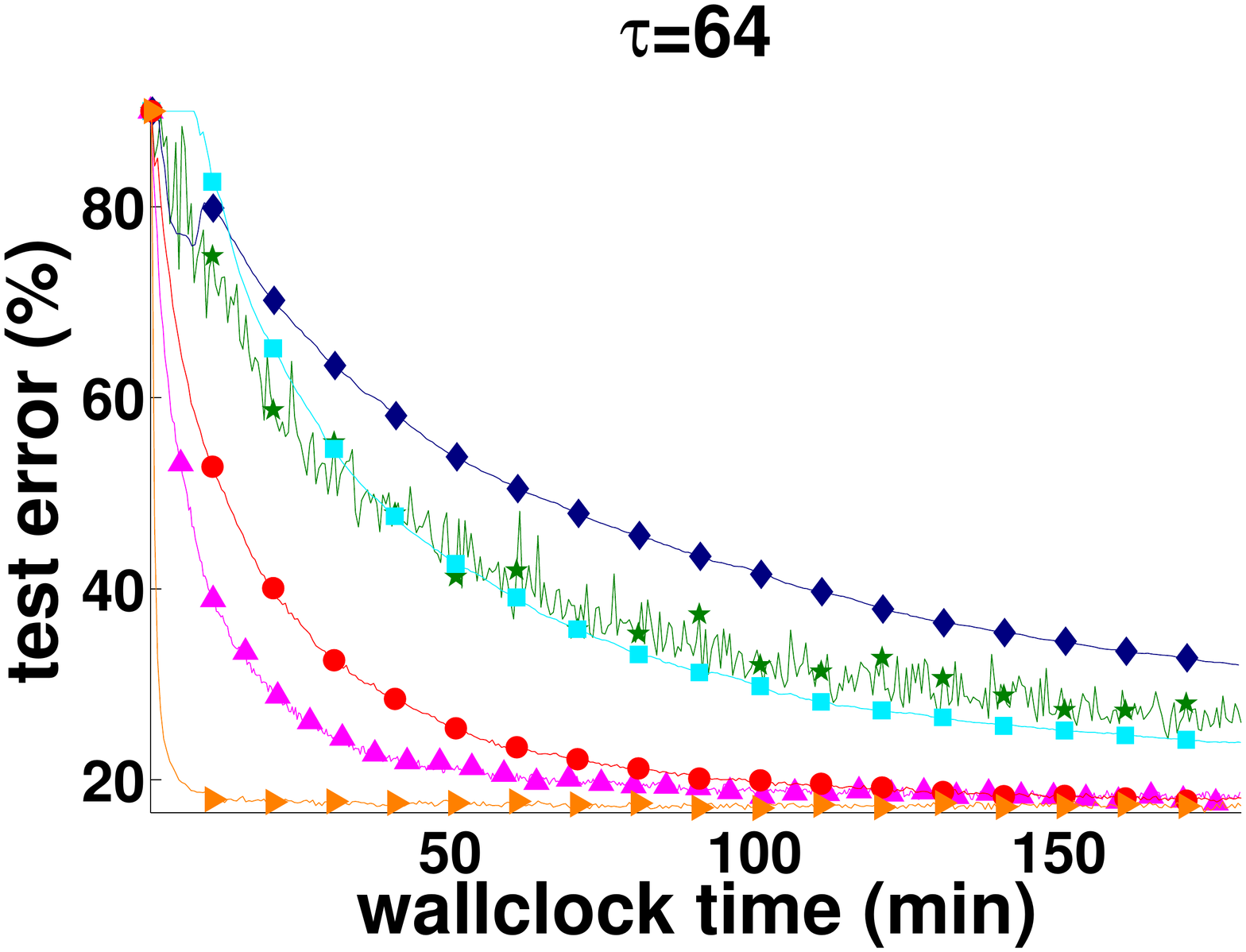}
\hspace{-0.3in}\includegraphics[trim=0cm 0cm 0cm 0cm,clip,width = 3in]{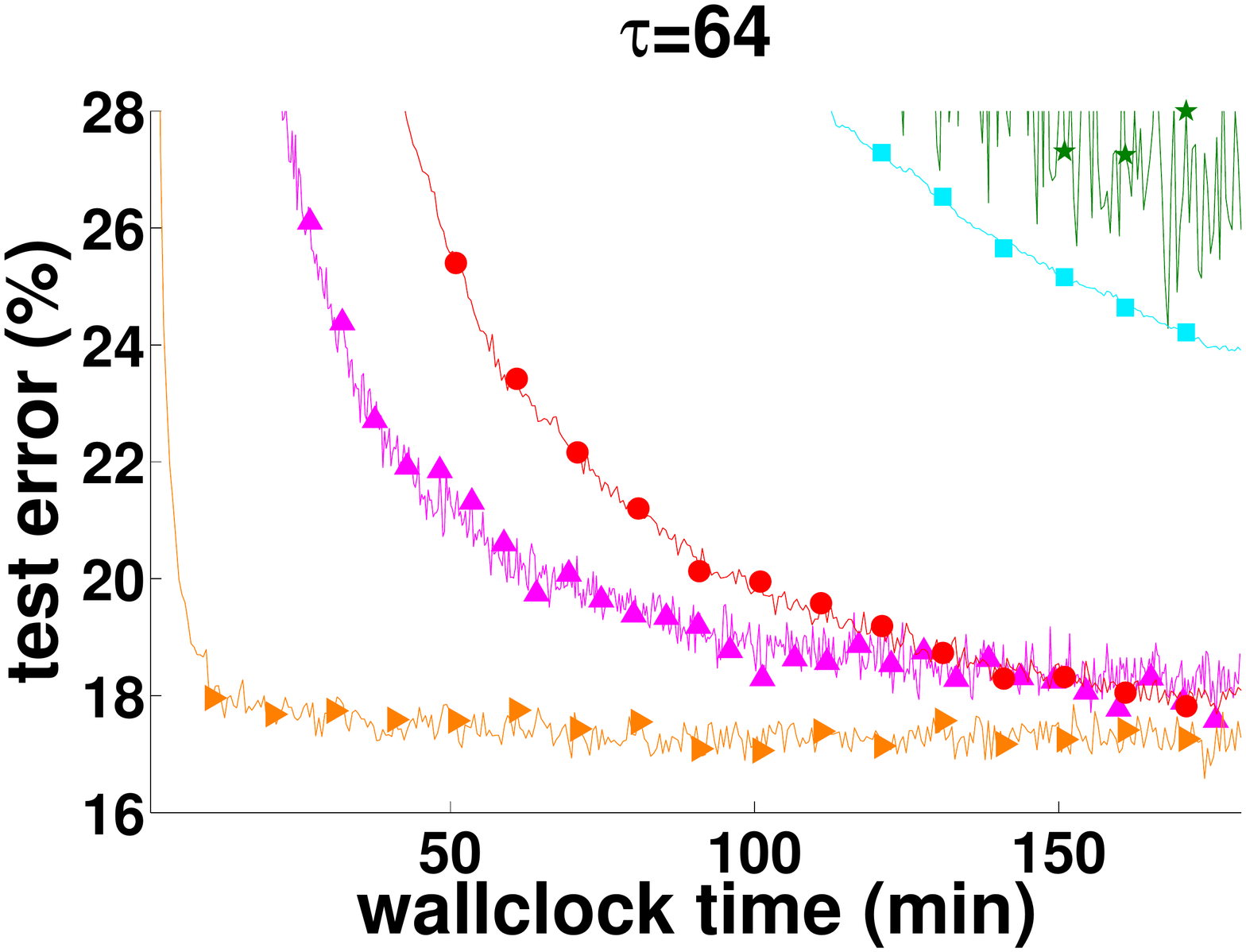}\\
\caption{Training and test loss and the test error for the center variable versus a wallclock time for communication period $\tau=64$ on \textit{CIFAR} dataset with the $7$-layer convolutional neural network. $p=4$.}
\label{fig:CIFAR_tau64}
\end{figure}

From the results in Figure~\ref{fig:CIFAR_tau1},~\ref{fig:CIFAR_tau4},~\ref{fig:CIFAR_tau16} and~\ref{fig:CIFAR_tau64}, we conclude that all \textit{DOWNPOUR}-based methods achieve their best performance (test error) for small $\tau$ ($\tau \in \{1,4\}$), and become highly unstable for $\tau \in \{16,64\}$. While \textit{EAMSGD} significantly outperforms comparator methods for all values of $\tau$ by having faster convergence. It also finds better-quality solution measured by the test error and this advantage becomes more significant for $\tau \in \{16,64\}$.
Note that the tendency to achieve better test performance with larger $\tau$ is also characteristic for the \textit{EASGD} algorithm. We remark that if the stochastic gradient is sparse, \textit{DOWNPOUR} empirically performs well with large communication period~\cite{DBLP:journals/corr/FengXZ15}.

\begin{figure}[p]
\center
% $p = 4$\\
\includegraphics[trim=0cm 0cm 0cm 0cm,clip,width = 3in]{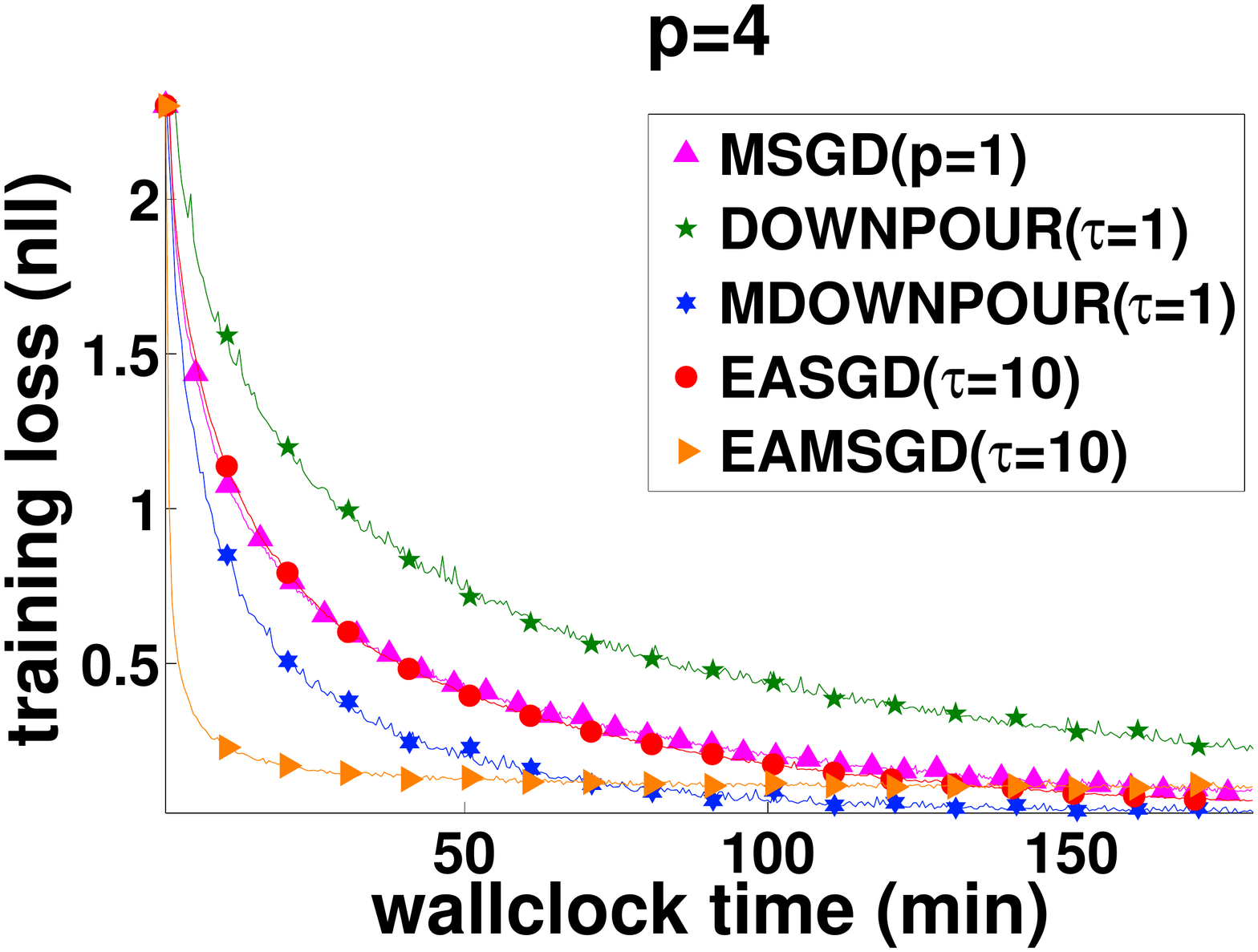}
\hspace{-0.3in}\includegraphics[trim=0cm 0cm 0cm 0cm,clip,width = 3in]{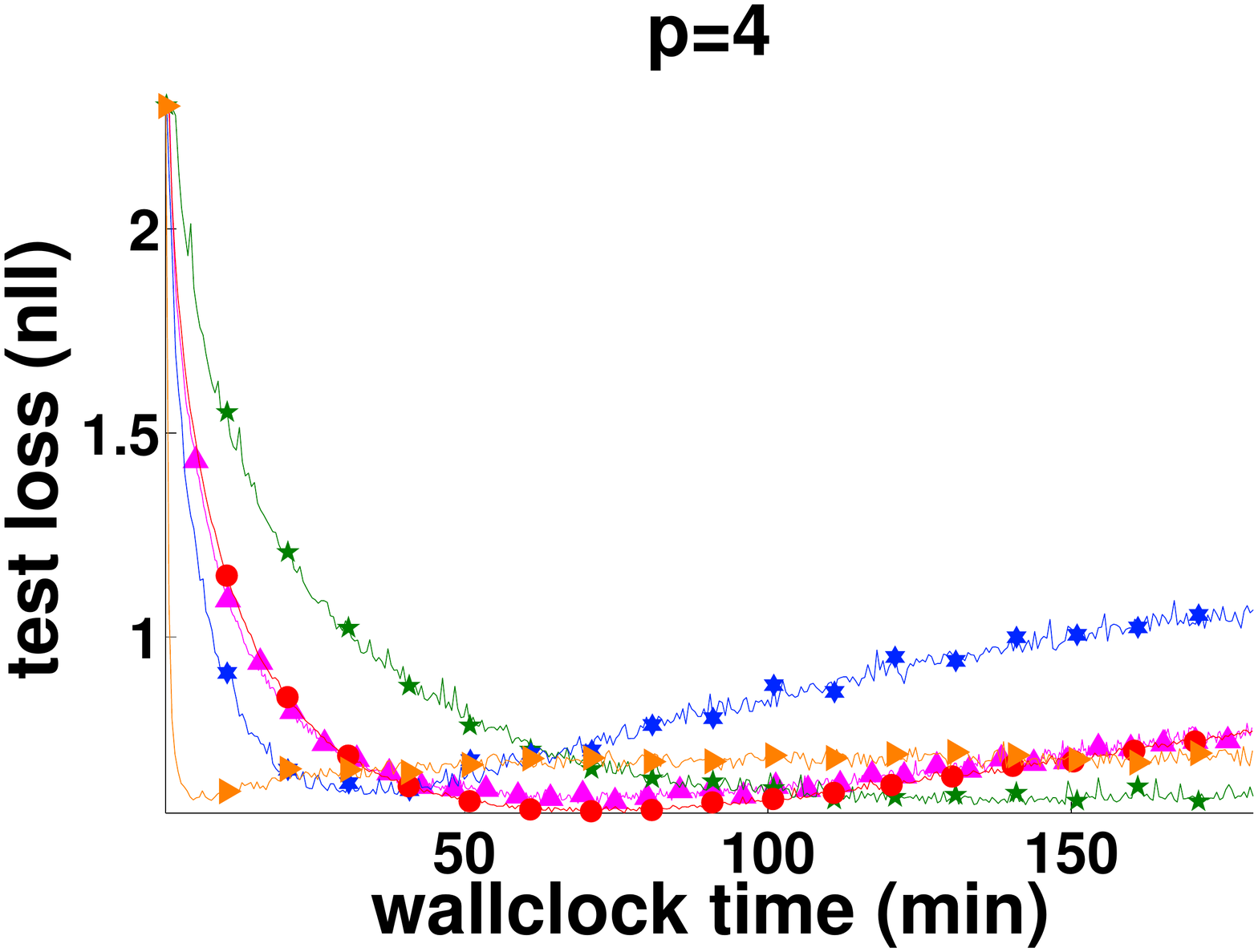} \\
\includegraphics[trim=0cm 0cm 0cm 0cm,clip,width = 3in]{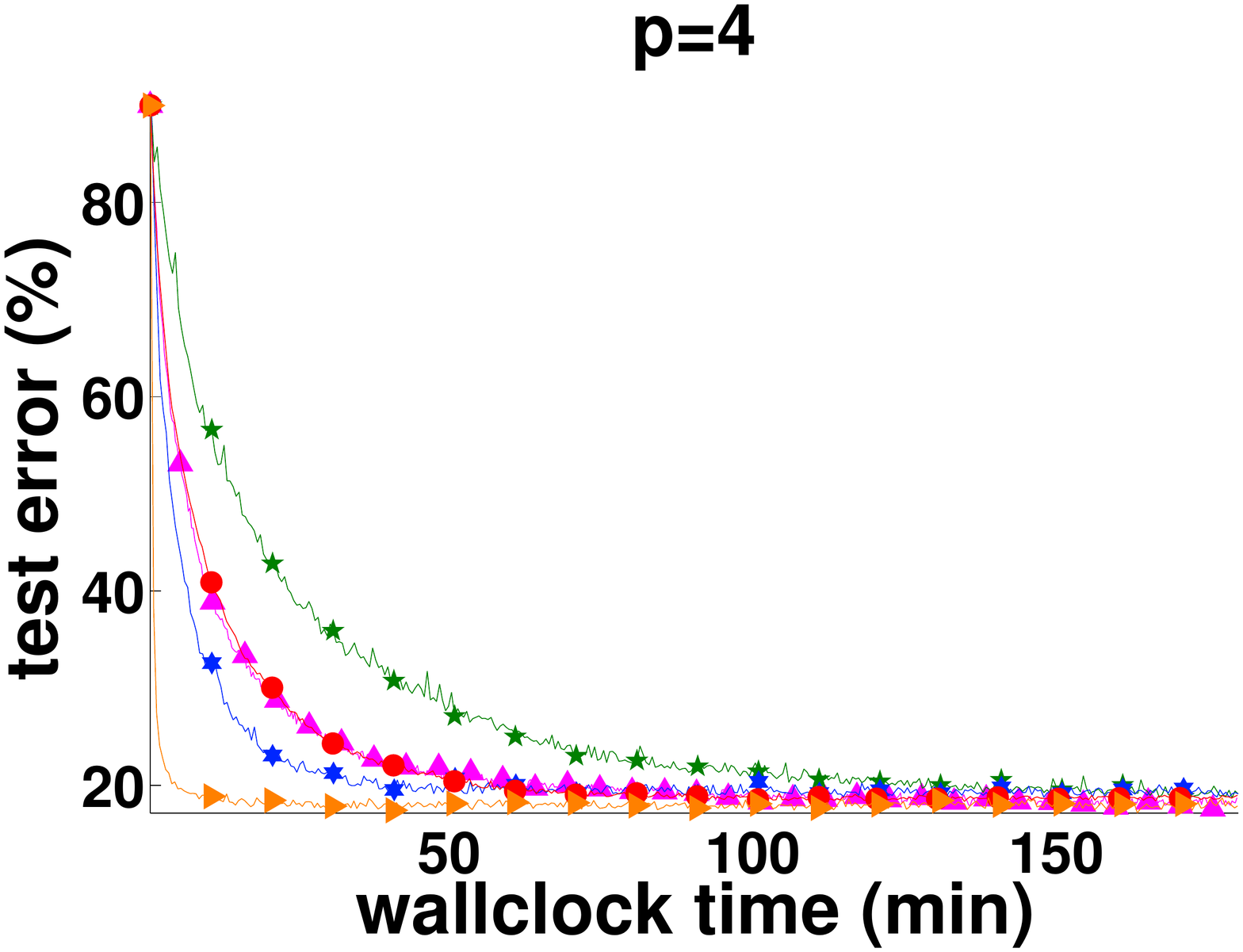}
\hspace{-0.3in}\includegraphics[trim=0cm 0cm 0cm 0cm,clip,width = 3in]{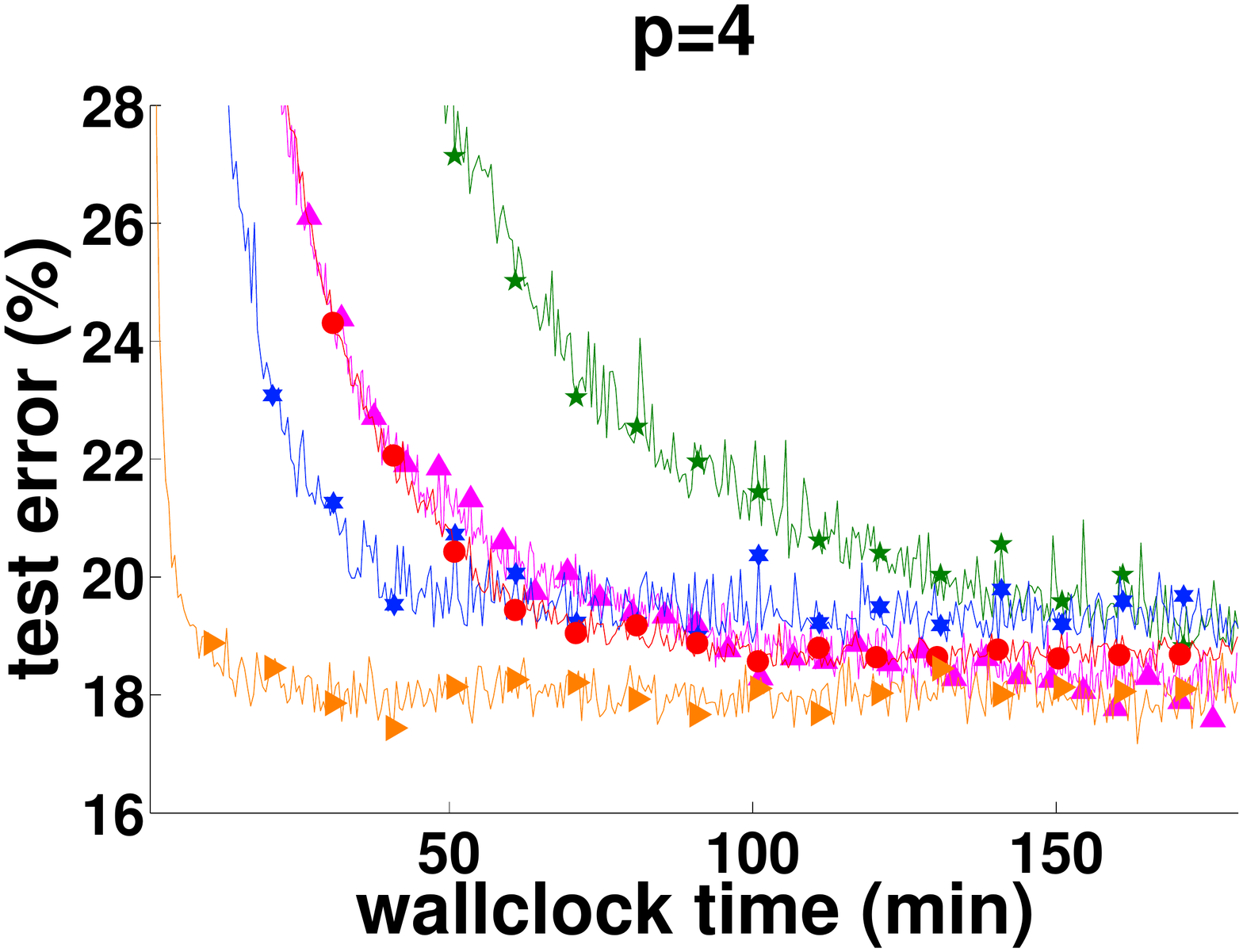}\\
\caption{Training and test loss and the test error for the center variable versus a wallclock time with the number of local workers $p=4$ for parallel methods on \textit{CIFAR} with the $7$-layer convolutional neural network.}
\label{fig:CIFAR2_p4}
\end{figure}

\begin{figure}[p]
\center
% $p = 8$\\
\includegraphics[trim=0cm 0cm 0cm 0cm,clip,width = 3in]{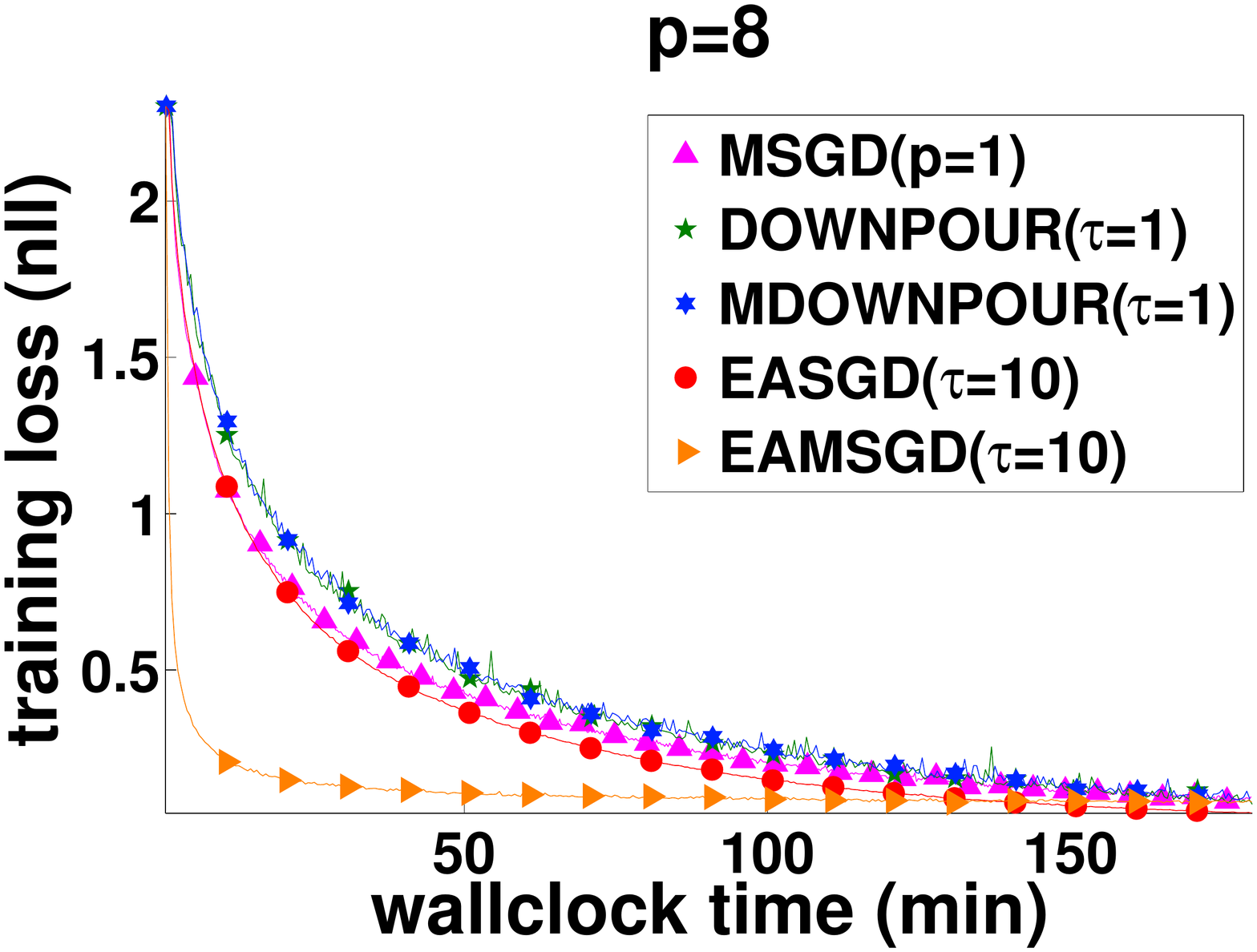}
\hspace{-0.3in}\includegraphics[trim=0cm 0cm 0cm 0cm,clip,width = 3in]{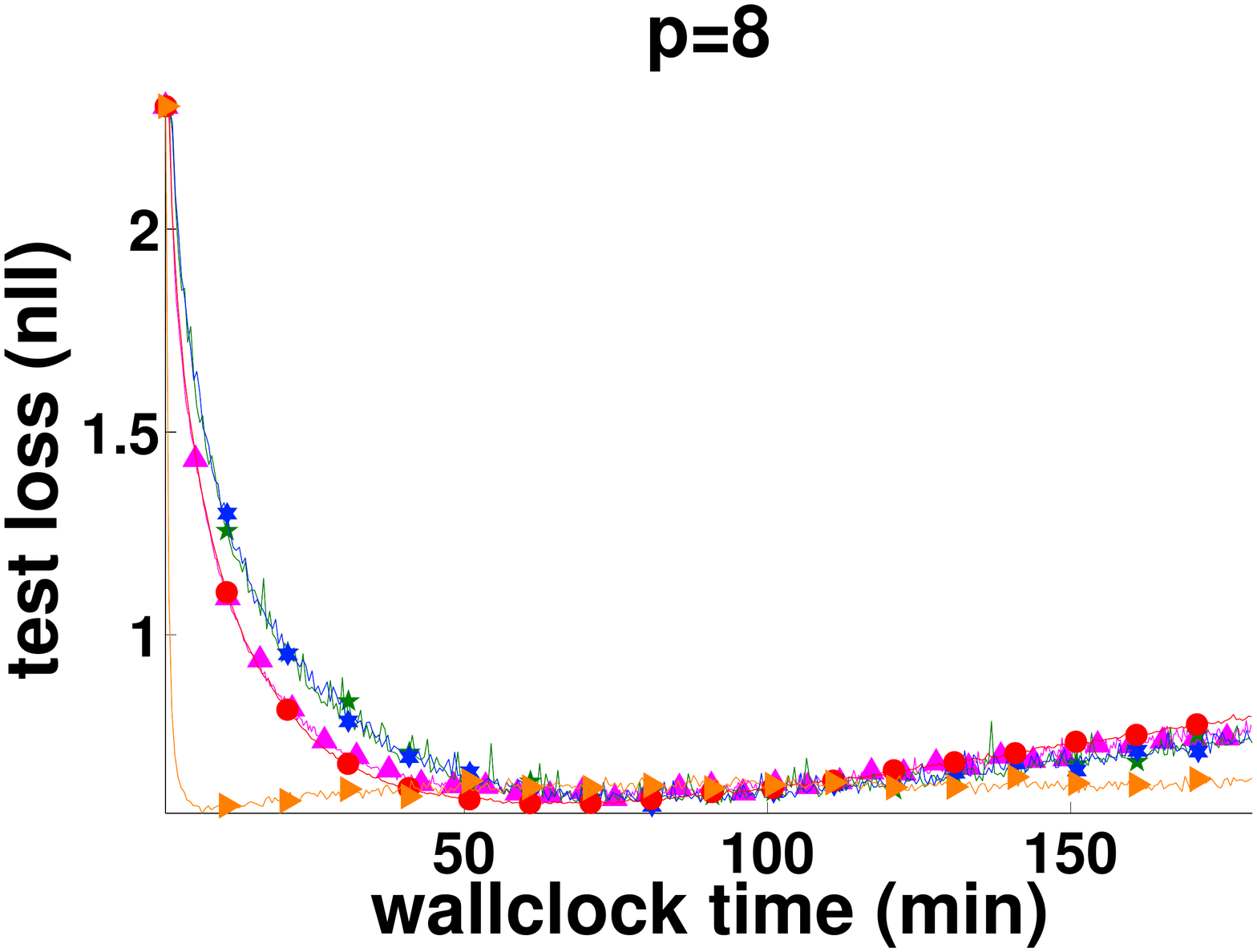} \\
\includegraphics[trim=0cm 0cm 0cm 0cm,clip,width = 3in]{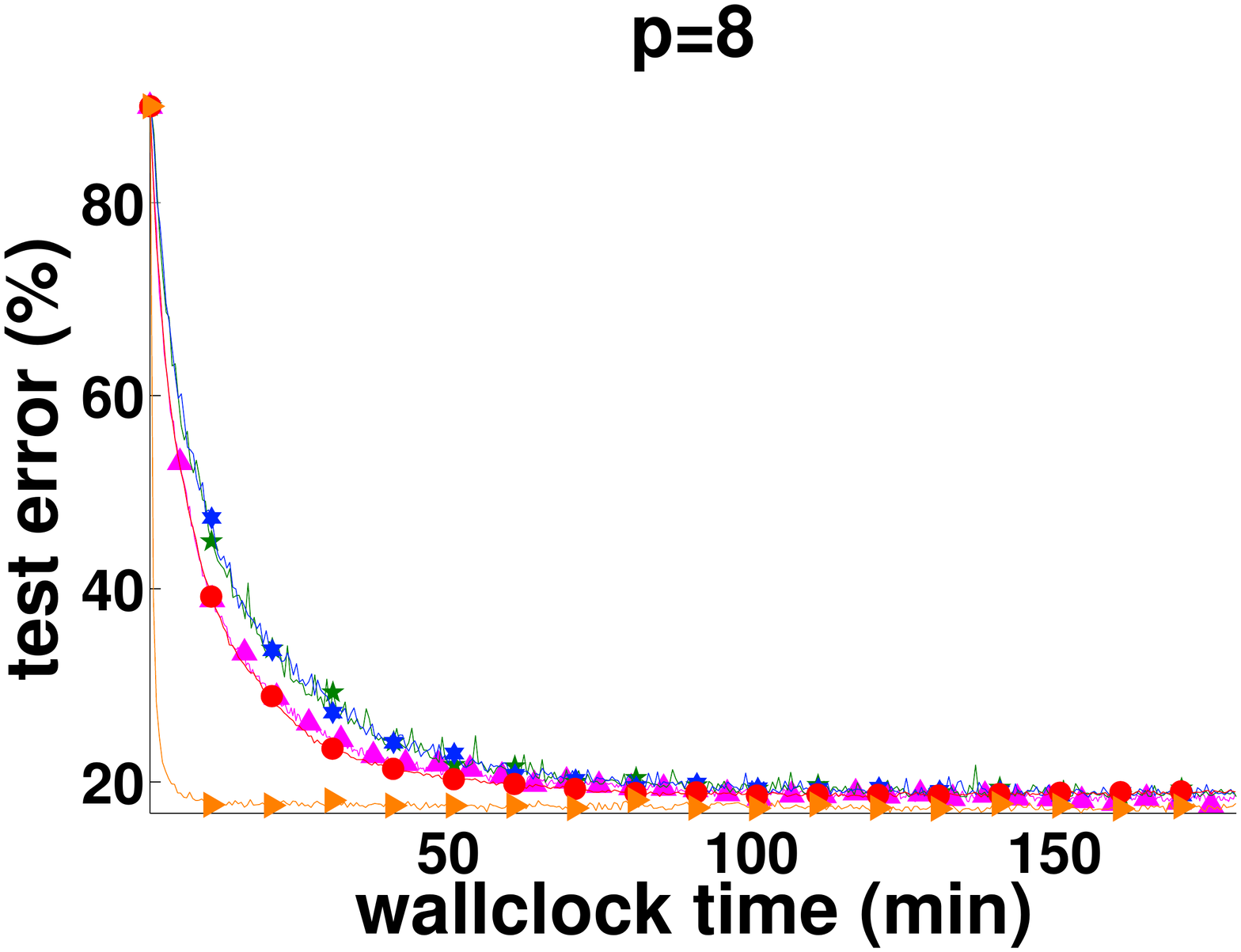}
\hspace{-0.3in}\includegraphics[trim=0cm 0cm 0cm 0cm,clip,width = 3in]{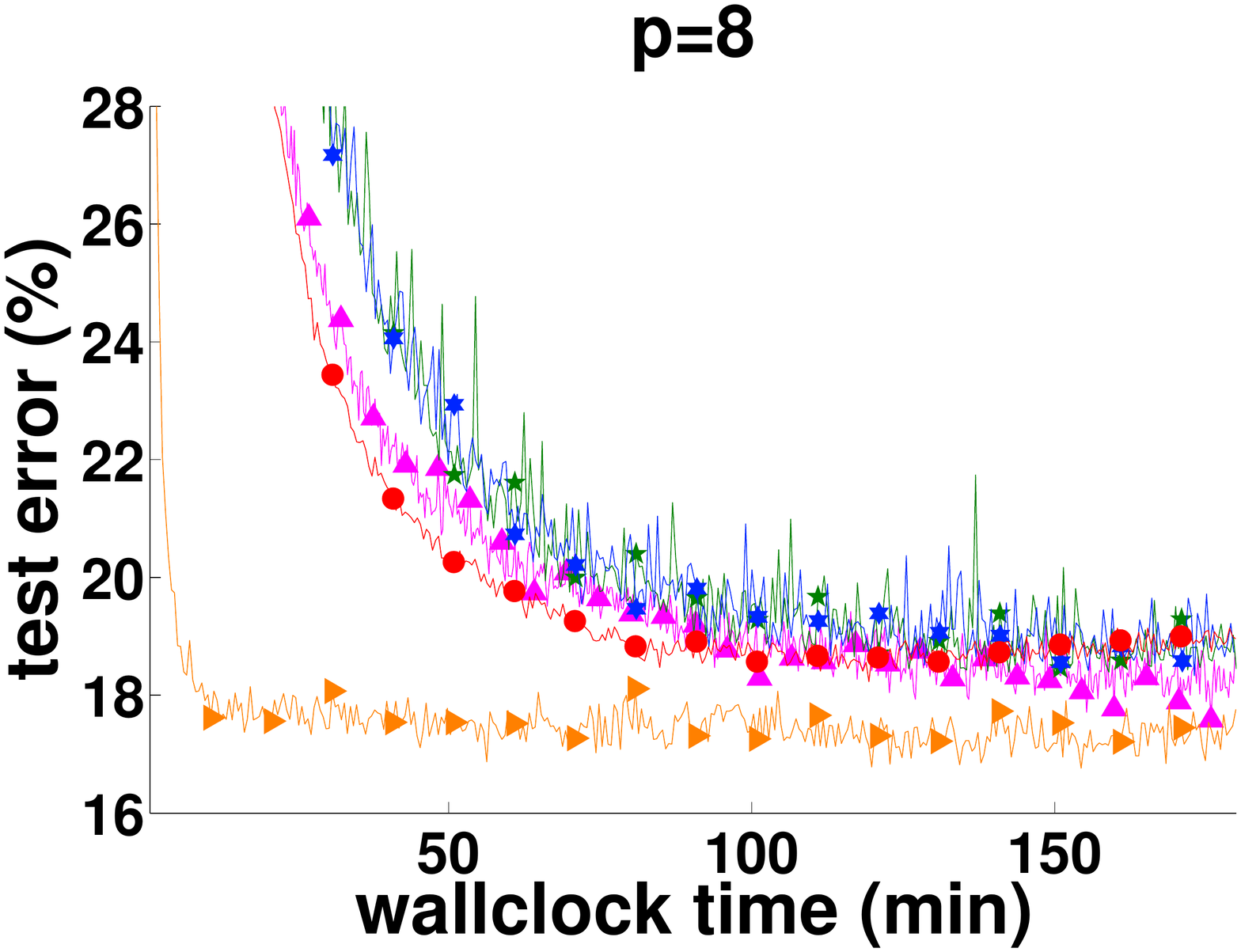}\\
\caption{Training and test loss and the test error for the center variable versus a wallclock time with the number of local workers $p=8$ for parallel methods on \textit{CIFAR} with the $7$-layer convolutional neural network.}
\label{fig:CIFAR2_p8}
\end{figure}

\begin{figure}[p]
\center
% $p = 16$\\
\includegraphics[trim=0cm 0cm 0cm 0cm,clip,width = 3in]{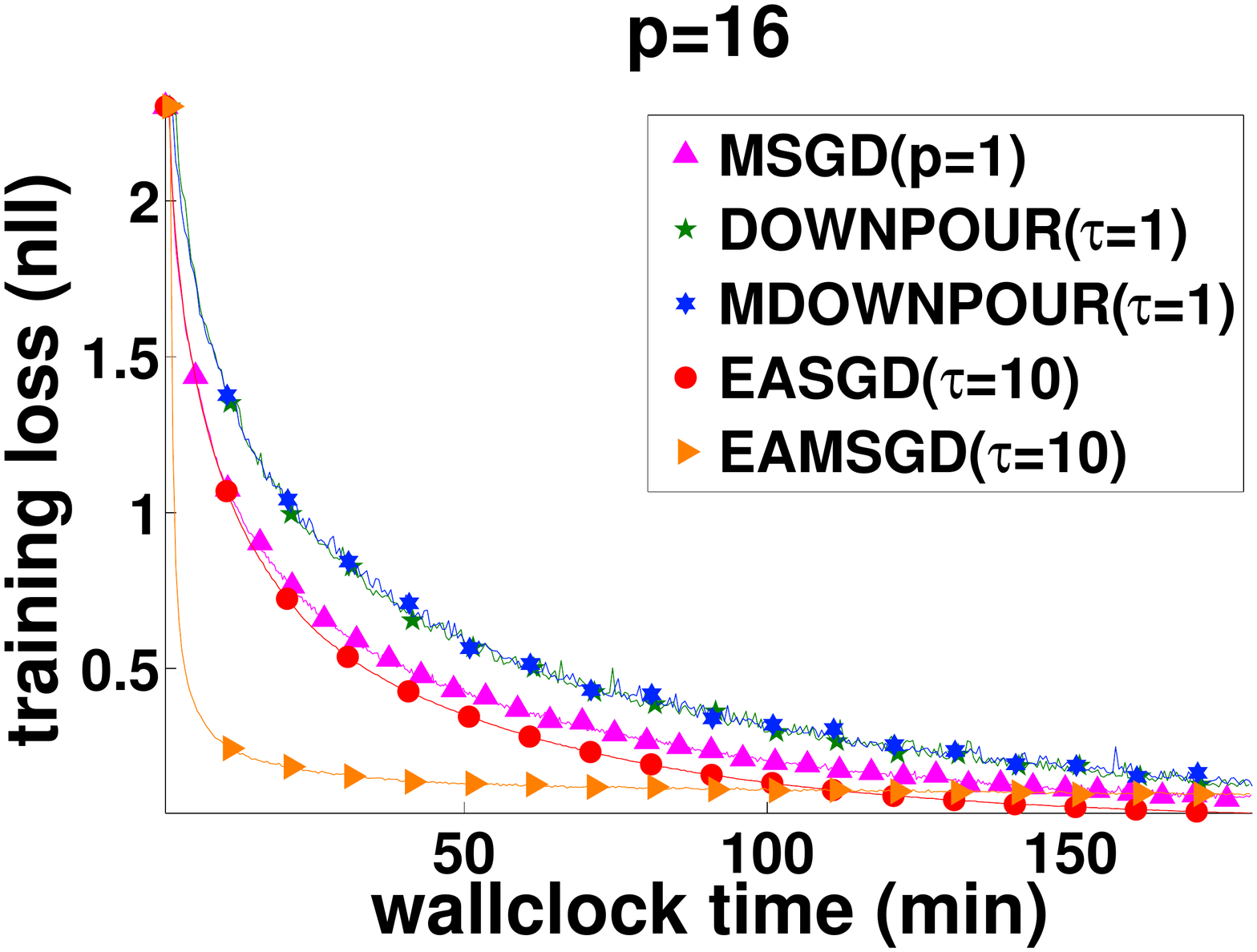}
\hspace{-0.3in}\includegraphics[trim=0cm 0cm 0cm 0cm,clip,width = 3in]{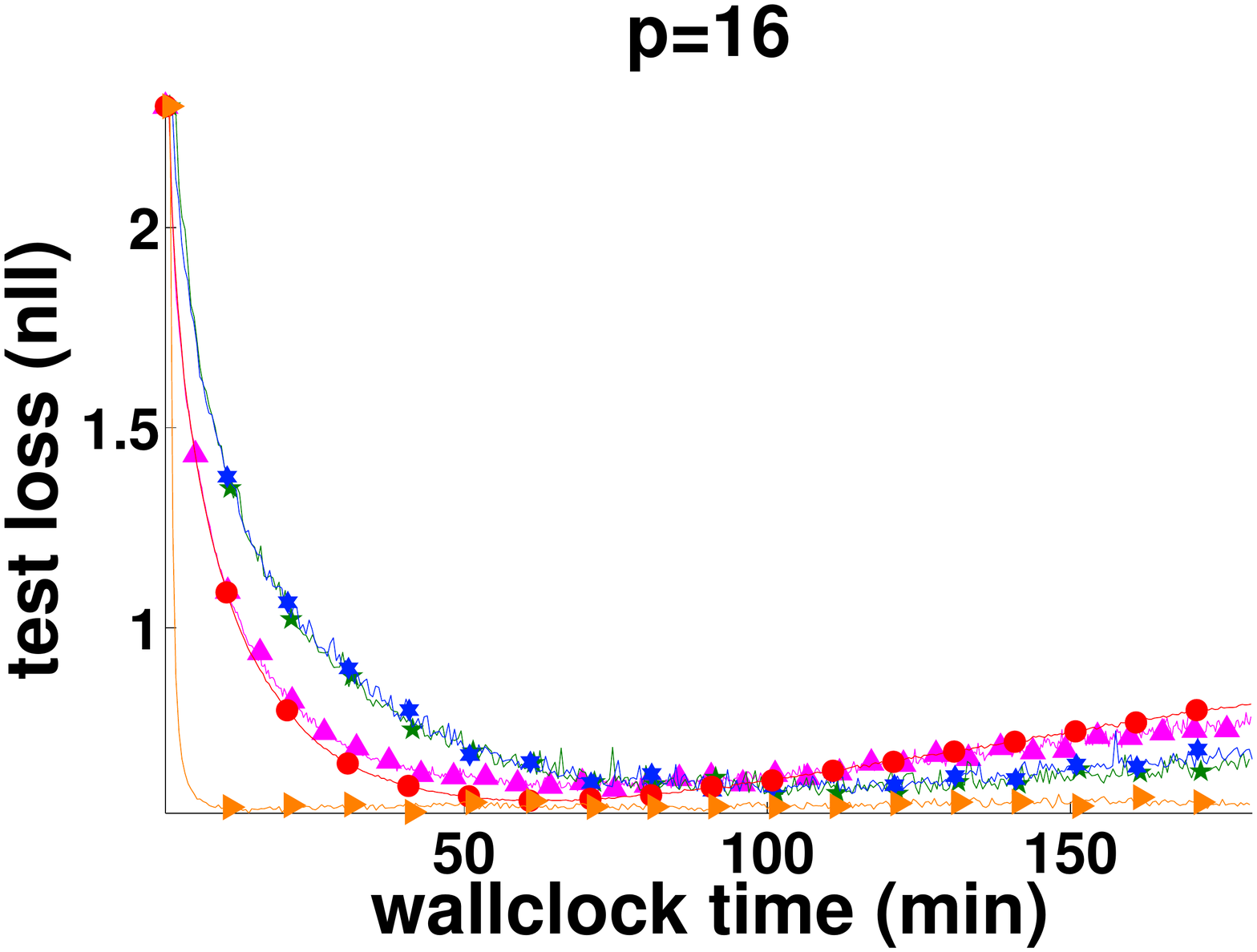} \\
\includegraphics[trim=0cm 0cm 0cm 0cm,clip,width = 3in]{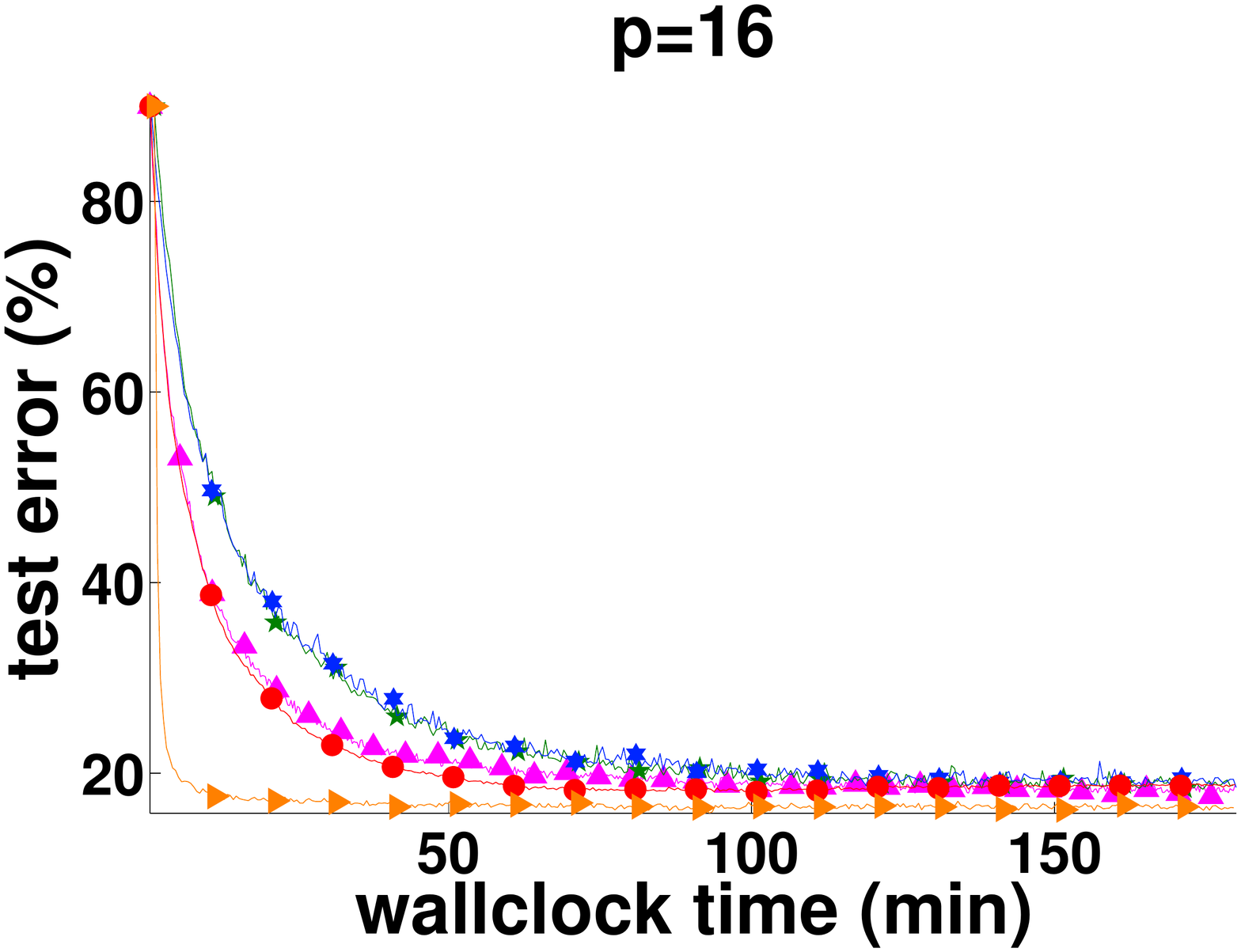}
\hspace{-0.3in}\includegraphics[trim=0cm 0cm 0cm 0cm,clip,width = 3in]{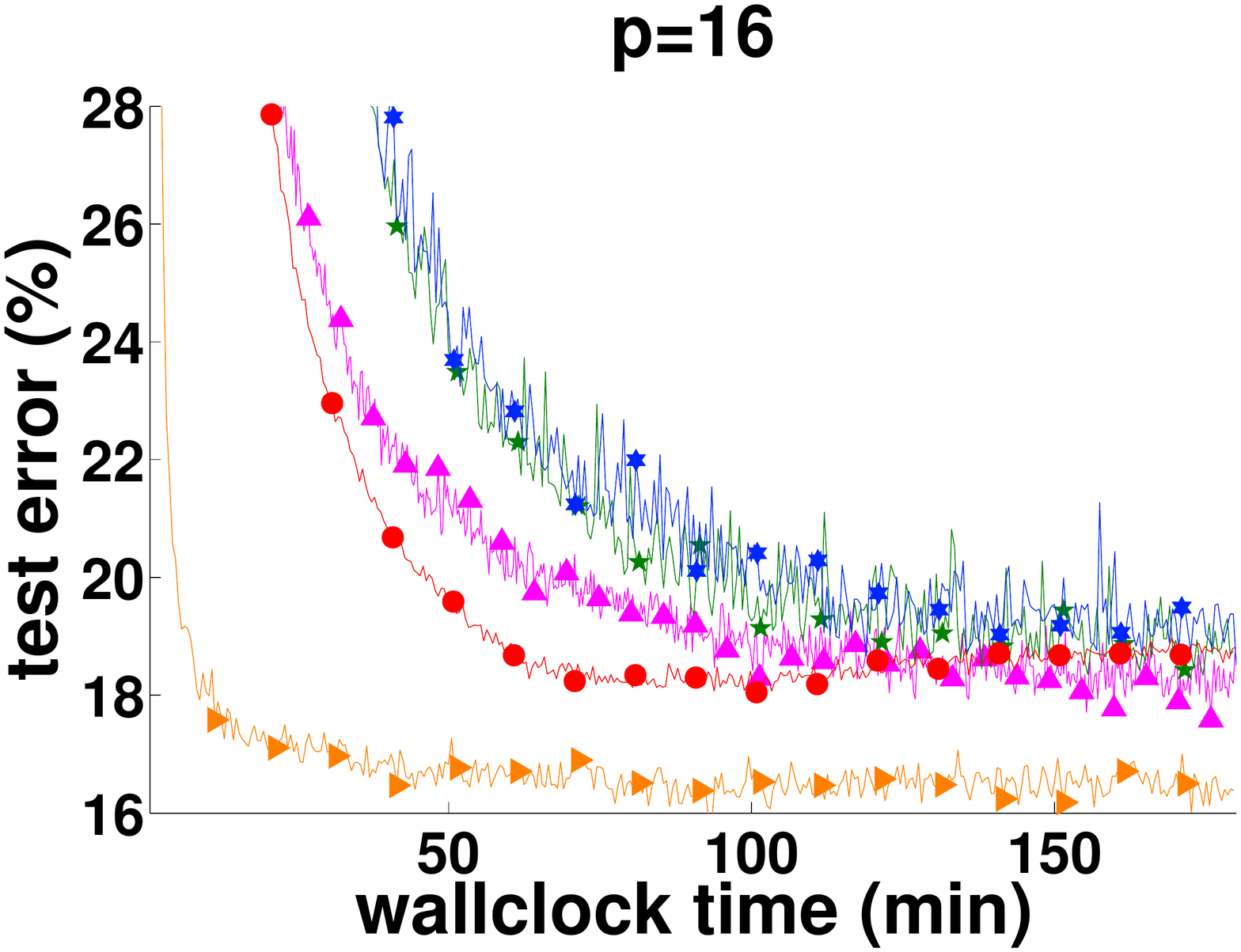}\\
\caption{Training and test loss and the test error for the center variable versus a wallclock time with the number of local workers $p=16$ for parallel methods on \textit{CIFAR} with the $7$-layer convolutional neural network.}
\label{fig:CIFAR2_p16}
\end{figure}

\begin{figure}[p]
\center
%$p = 4$\\
\includegraphics[trim=0cm 0cm 0cm 0cm,clip,width = 3in]{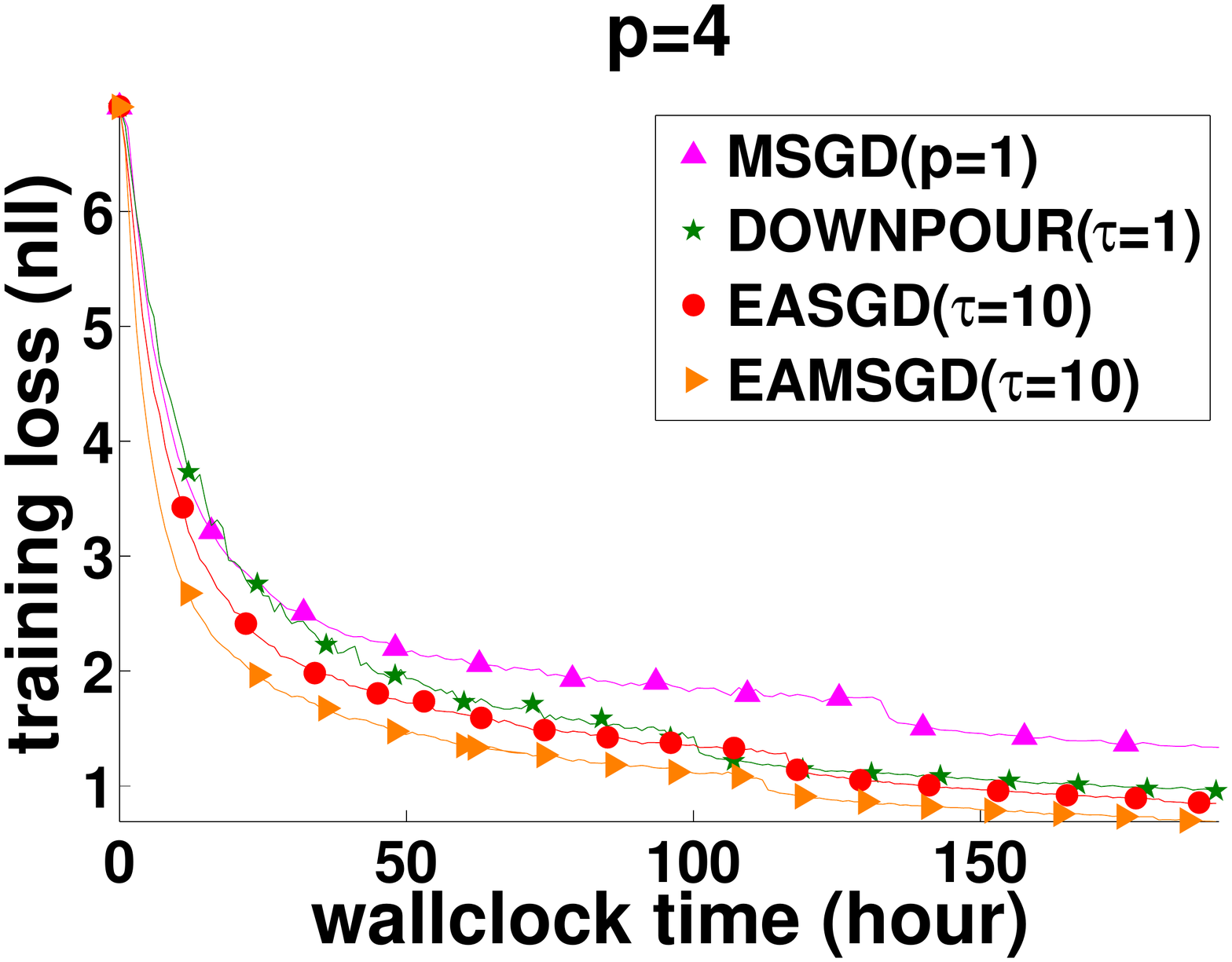}
\hspace{-0.3in}\includegraphics[trim=0cm 0cm 0cm 0cm,clip,width = 3in]{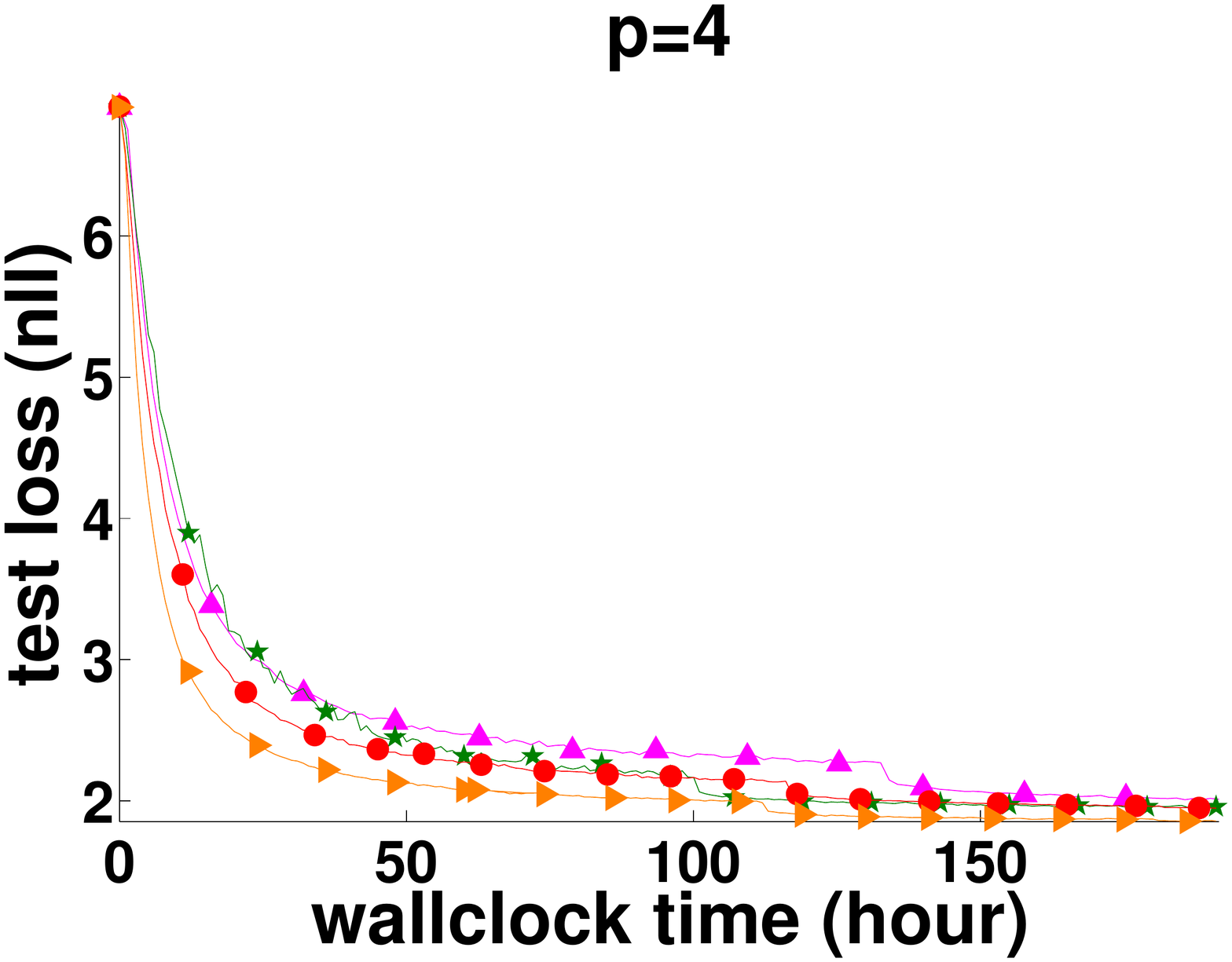} \\
\includegraphics[trim=0cm 0cm 0cm 0cm,clip,width = 3in]{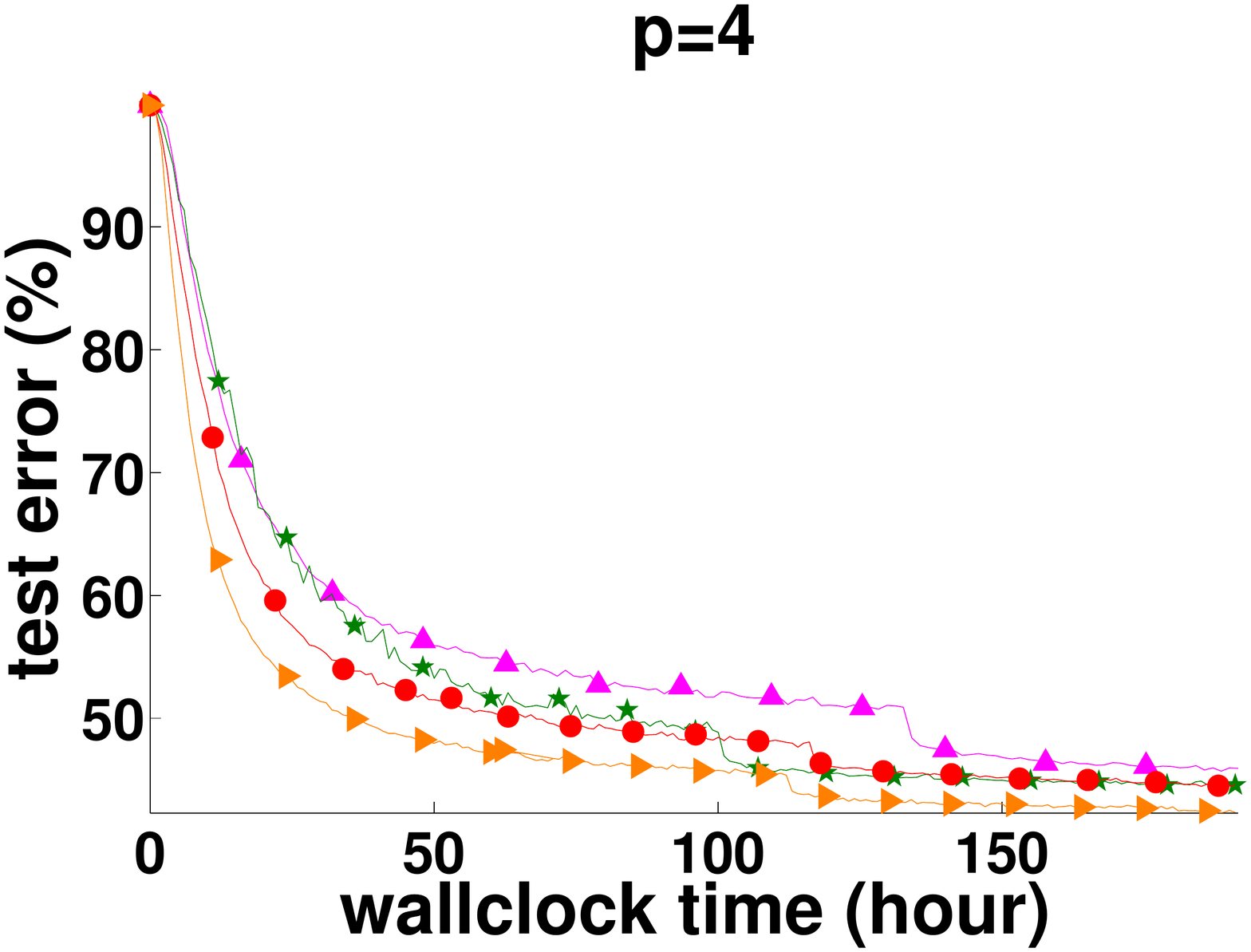}
\hspace{-0.3in}\includegraphics[trim=0cm 0cm 0cm 0cm,clip,width = 3in]{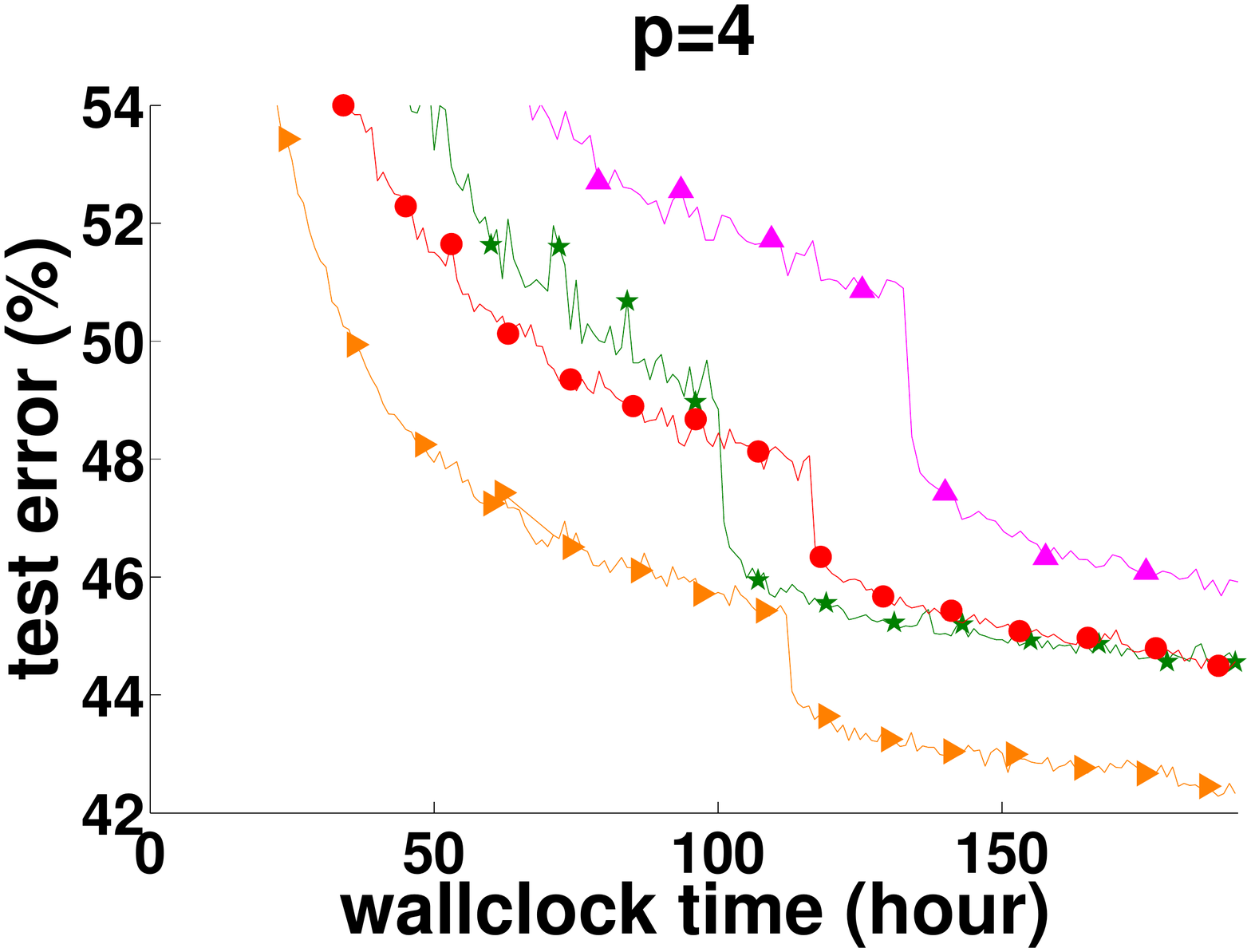}\\
\caption{Training and test loss and the test error for the center variable versus a wallclock time with the number of local workers $p=4$ on \textit{ImageNet} with the $11$-layer convolutional neural network.}
\label{fig:ImageNet_p4}
\end{figure}

\begin{figure}[p]
\center
%$p = 8$\\
\includegraphics[trim=0cm 0cm 0cm 0cm,clip,width = 3in]{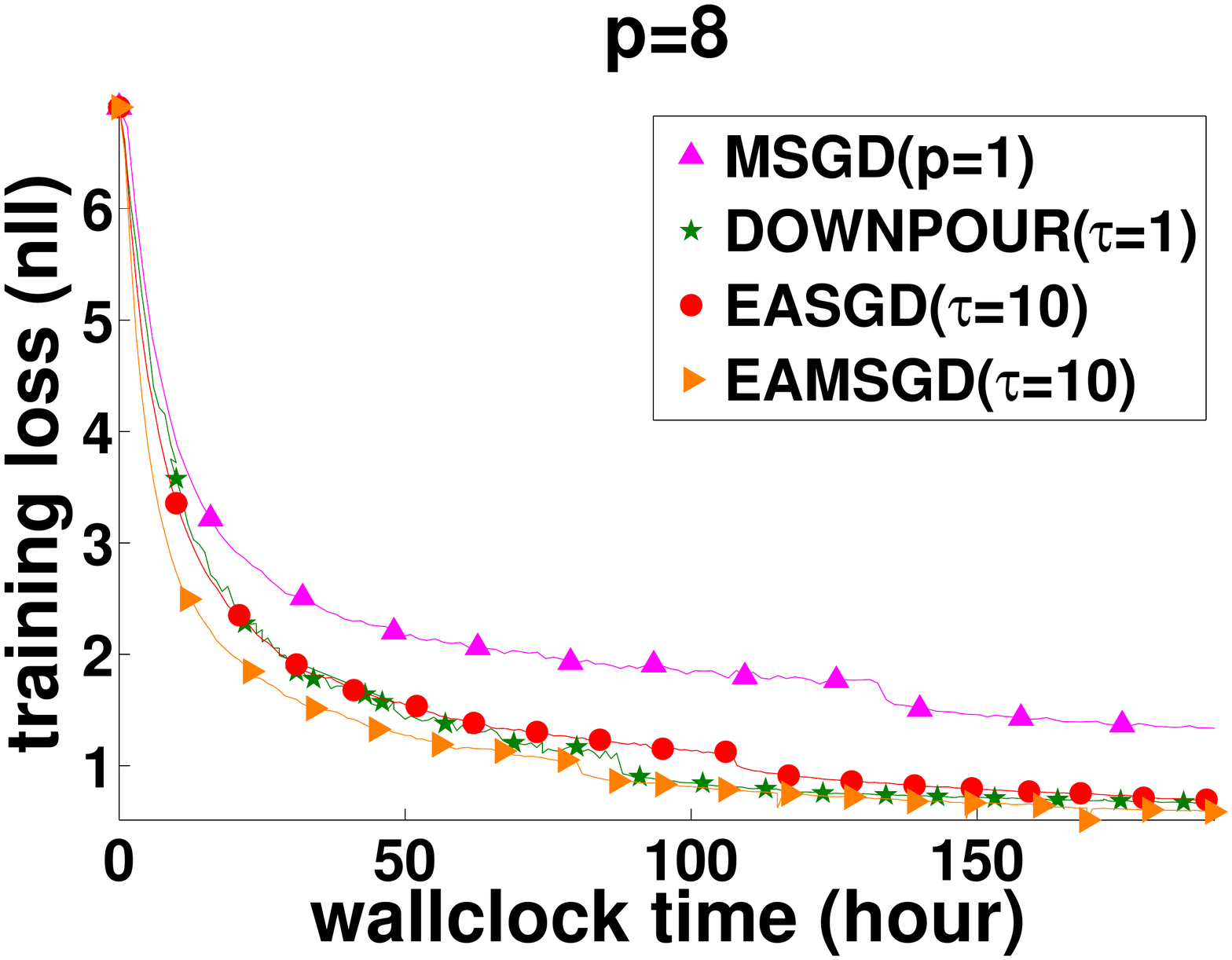}
\hspace{-0.3in}\includegraphics[trim=0cm 0cm 0cm 0cm,clip,width = 3in]{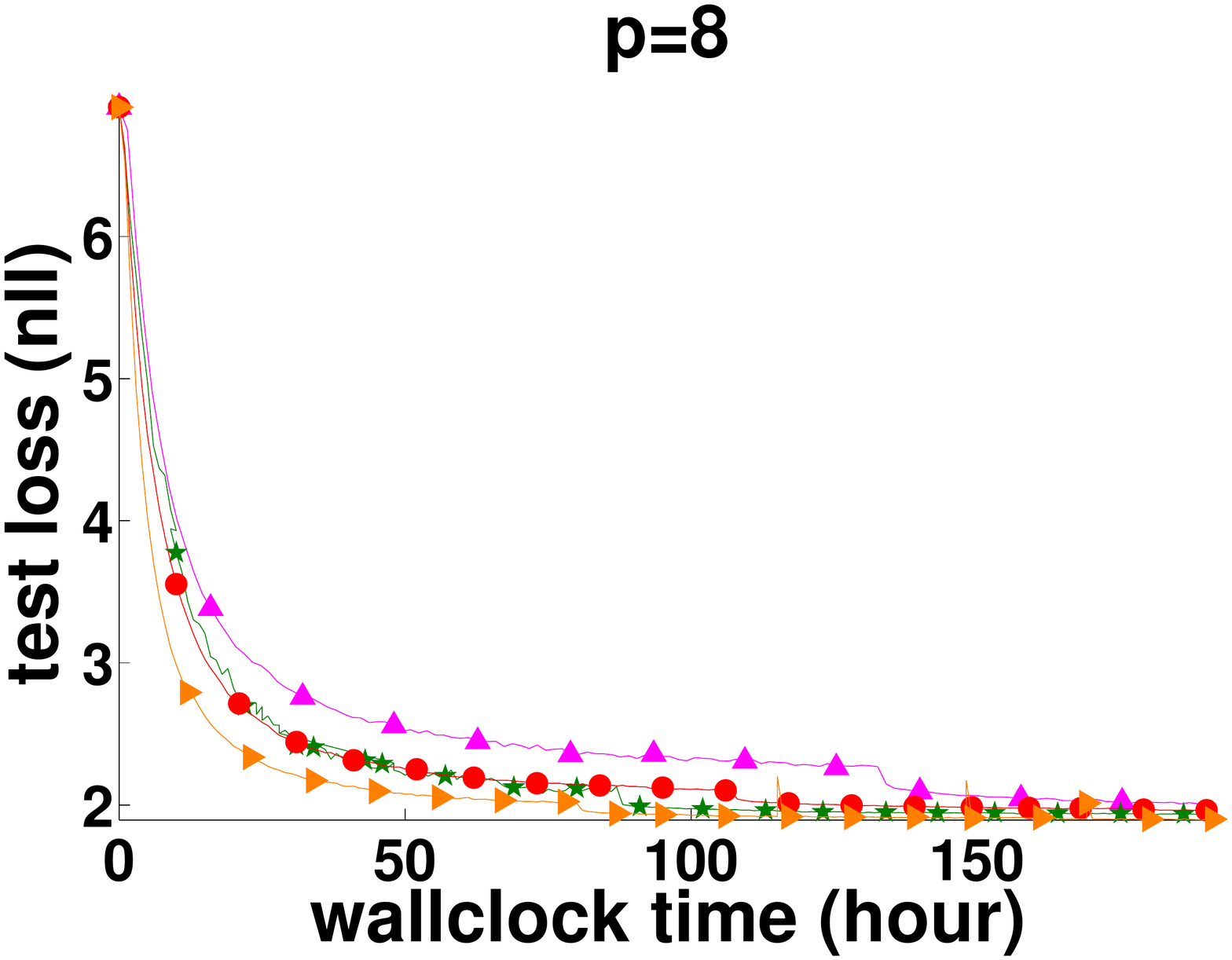} \\
\includegraphics[trim=0cm 0cm 0cm 0cm,clip,width = 3in]{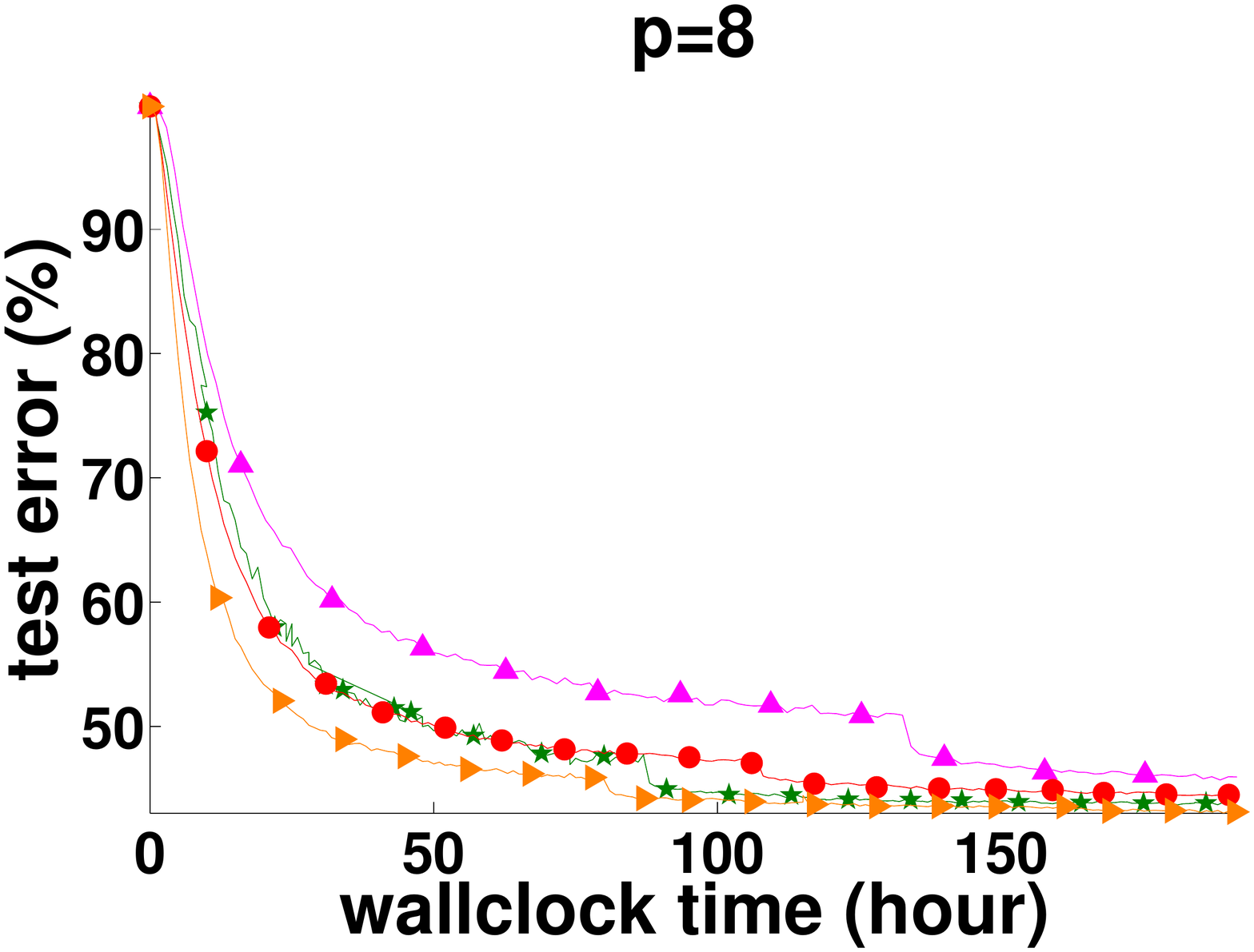}
\hspace{-0.3in}\includegraphics[trim=0cm 0cm 0cm 0cm,clip,width = 3in]{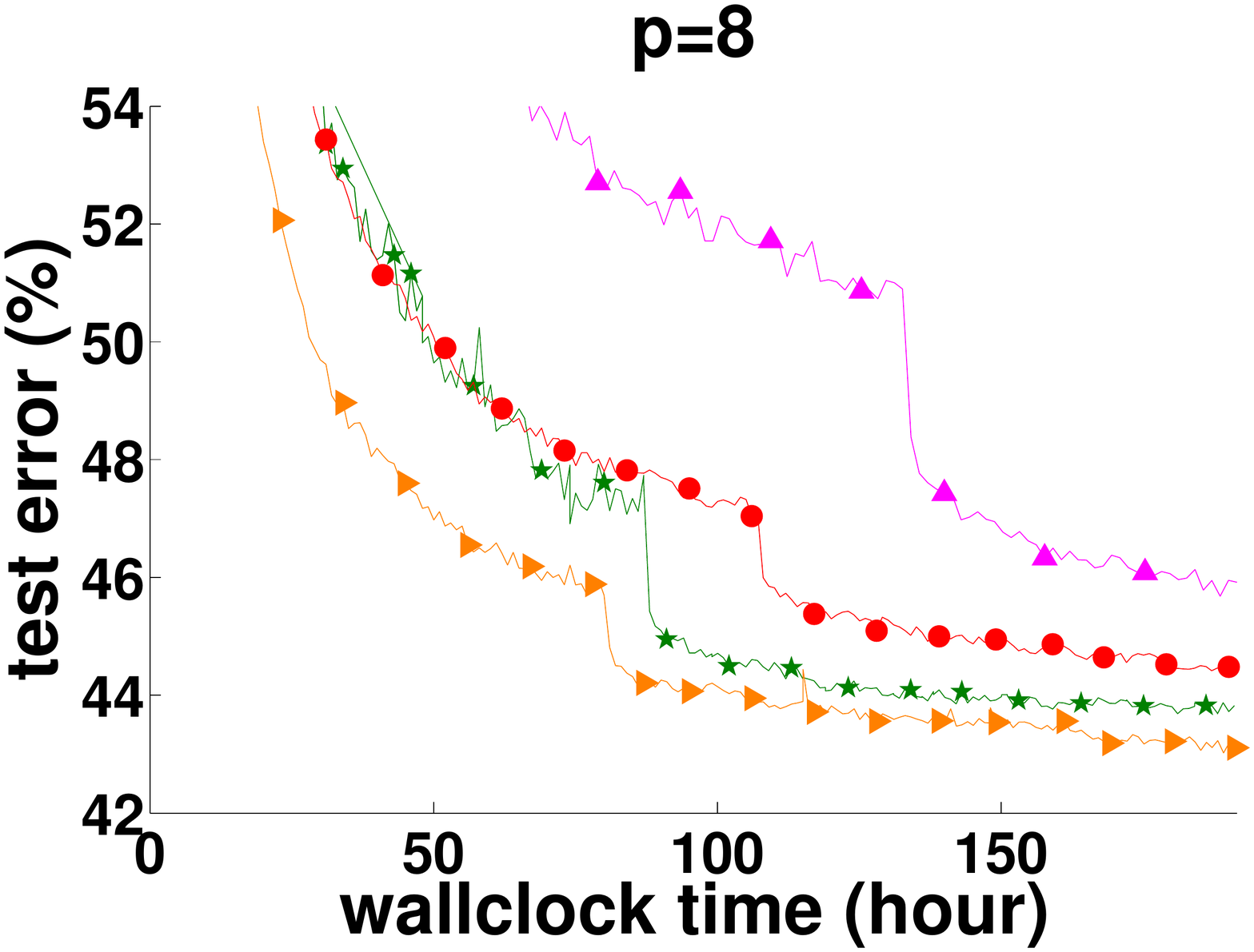}\\
\caption{Training and test loss and the test error for the center variable versus a wallclock time with the number of local workers $p=8$ on \textit{ImageNet} with the $11$-layer convolutional neural network.}
\label{fig:ImageNet_p8}
\end{figure}

We next explore different number of local workers $p$ from the set $p = \{4,8,16\}$ for the \textit{CIFAR} experiment, and $p = \{4,8\}$ for the \textit{ImageNet} experiment\footnote{For the \textit{ImageNet} experiment, the training loss is measured on a subset of the training data of size 50,000.}. For the \textit{ImageNet} experiment we report the results of one run with the best setting we have found. \textit{EASGD} and \textit{EAMSGD} were run with $\tau =10$ whereas \textit{DOWNPOUR} and \textit{MDOWNPOUR} were run with $\tau = 1$.

For the \textit{CIFAR} experiment, the results are in Figure~\ref{fig:CIFAR2_p4},~\ref{fig:CIFAR2_p8} and~\ref{fig:CIFAR2_p16}. \textit{EAMSGD} achieves significant accelerations compared to other methods, e.g. the relative speedup for $p=16$ (the best comparator method is then \textit{MSGD}) to achieve the test error $21\%$ equals $11.1$.
It's noticeable that the smallest achievable test error by either \textit{EASGD} or \textit{EAMSGD} decreases with larger $p$. This can potentially be explained by the fact that larger $p$ allows for more exploration of the parameter space. In the next section, we discuss further the trade-off between exploration and exploitation as a function of the learning rate (section~\ref{sec:deplr}) and the communication period (section~\ref{sec:exp_dep_tau}).

For the \textit{ImageNet} experiment, the results are in Figure~\ref{fig:ImageNet_p4} and~\ref{fig:ImageNet_p8}.
The difficulty in this task is that we need to manually reduce the learning rate, otherwise the training loss will stagnate. Thus our initial learning rate is decreased twice over time, by a factor of $5$ and then $2$, when we observe that the online predictive loss~\cite{cesa2004generalization} stagnates.
\textit{EAMSGD} again achieves significant accelerations compared to other methods, e.g. the relative speedup for $p=8$ (the best comparator method is then \textit{DOWNPOUR}) to achieve the test error $49\%$ equals $1.8$, and simultaneously it reduces the communication overhead (\textit{DOWNPOUR} uses communication period $\tau = 1$ and \textit{EAMSGD} uses $\tau = 10$).
However, there's an annealing effect here in the sense that depending on the time the learning rate is reduced, the final test performance can be quite different. This makes the performance comparison difficult to define. In general, this is also a difficulty in comparing the NP-hard problem solvers.

\section{Further discussion and understanding}
\label{sec:exp_result2}

\subsection{Comparison of \textit{SGD}, \textit{ASGD}, \textit{MVASGD} and \textit{MSGD}}
\begin{figure}[p]
\center
\includegraphics[trim=0cm 0cm 0cm 0cm,clip,width = 3in]{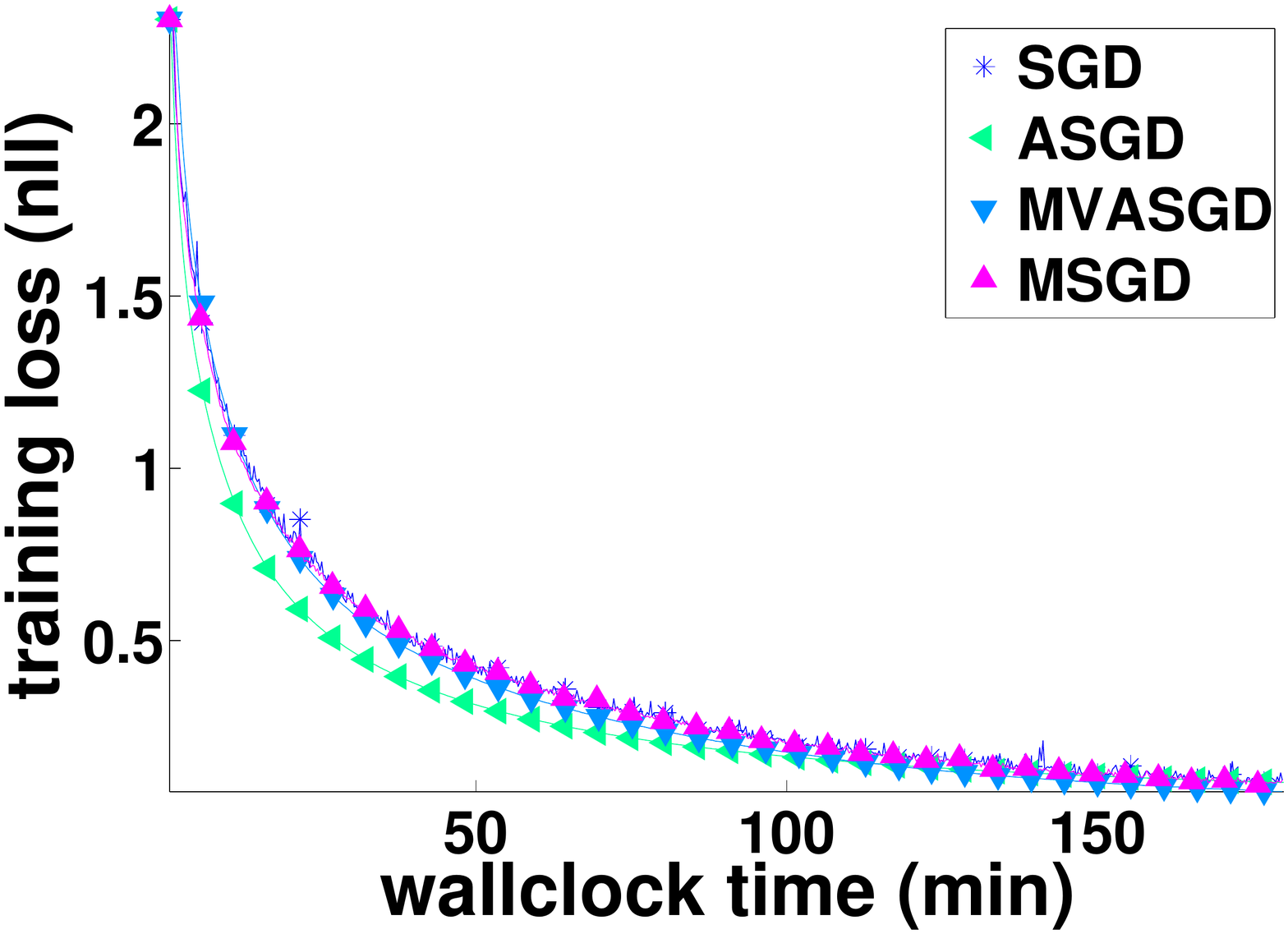}
\hspace{-0.3in}\includegraphics[trim=0cm 0cm 0cm 0cm,clip,width = 3in]{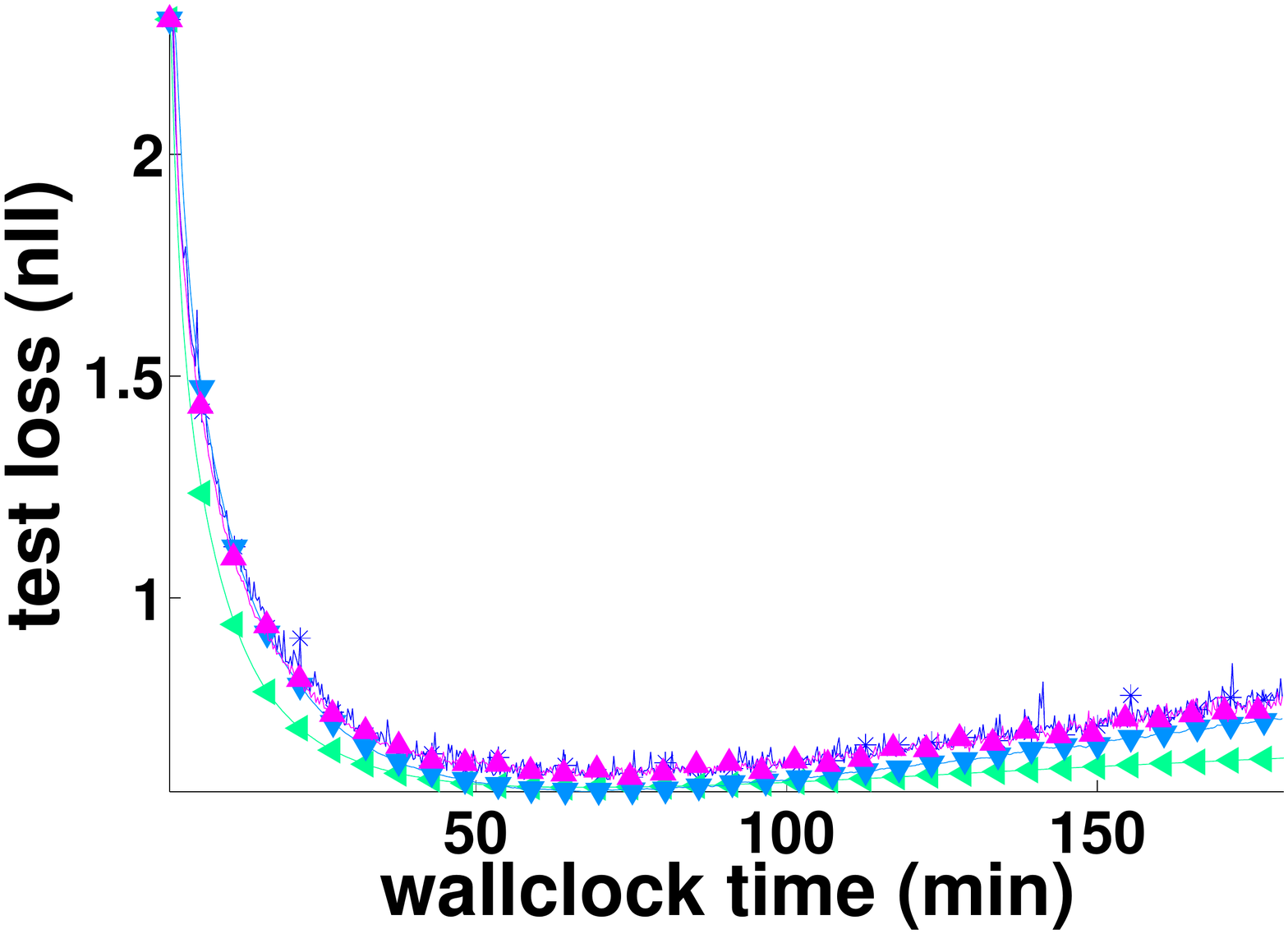} \\
\includegraphics[trim=0cm 0cm 0cm 0cm,clip,width = 3in]{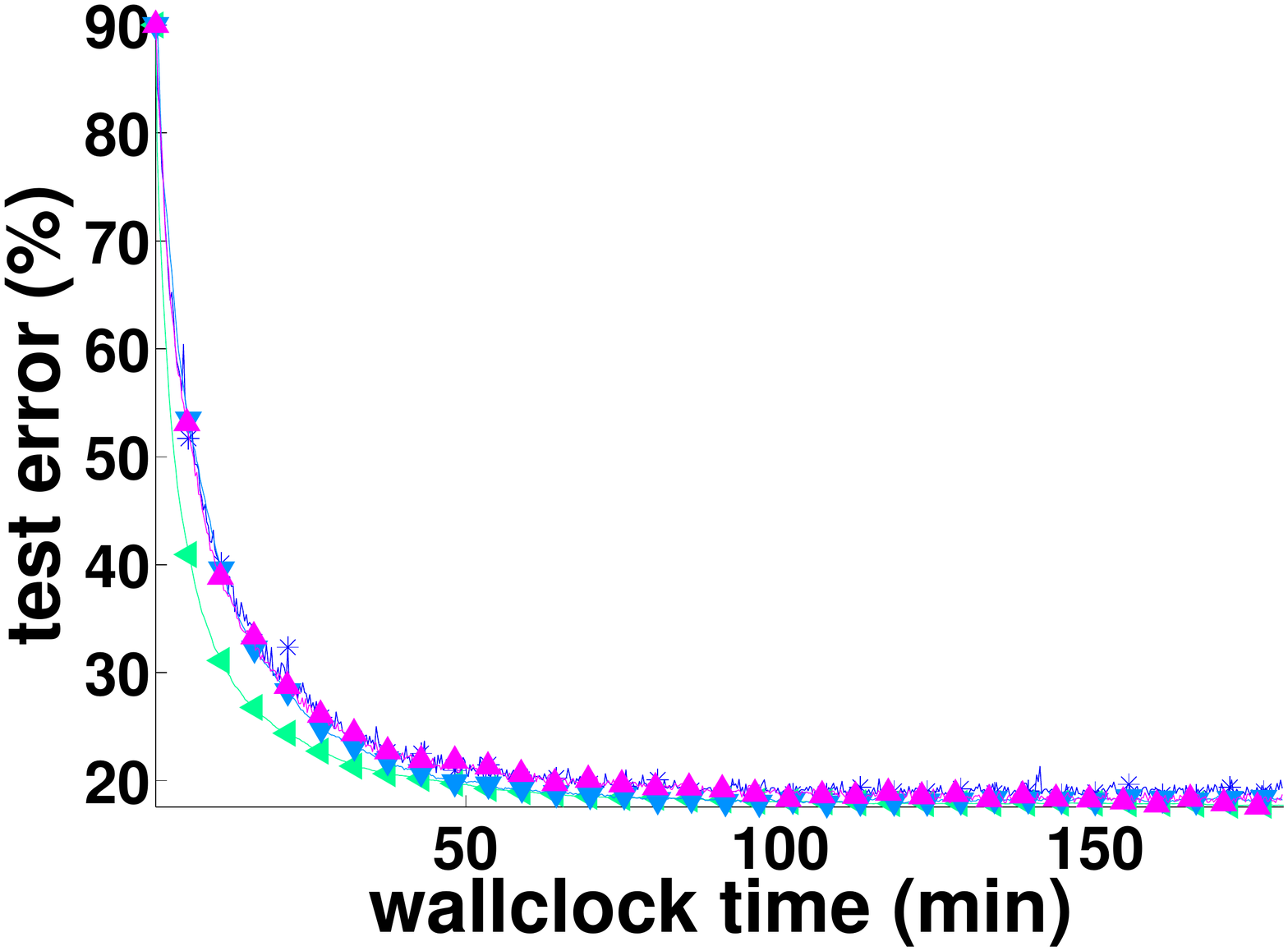}
\hspace{-0.3in}\includegraphics[trim=0cm 0cm 0cm 0cm,clip,width = 3in]{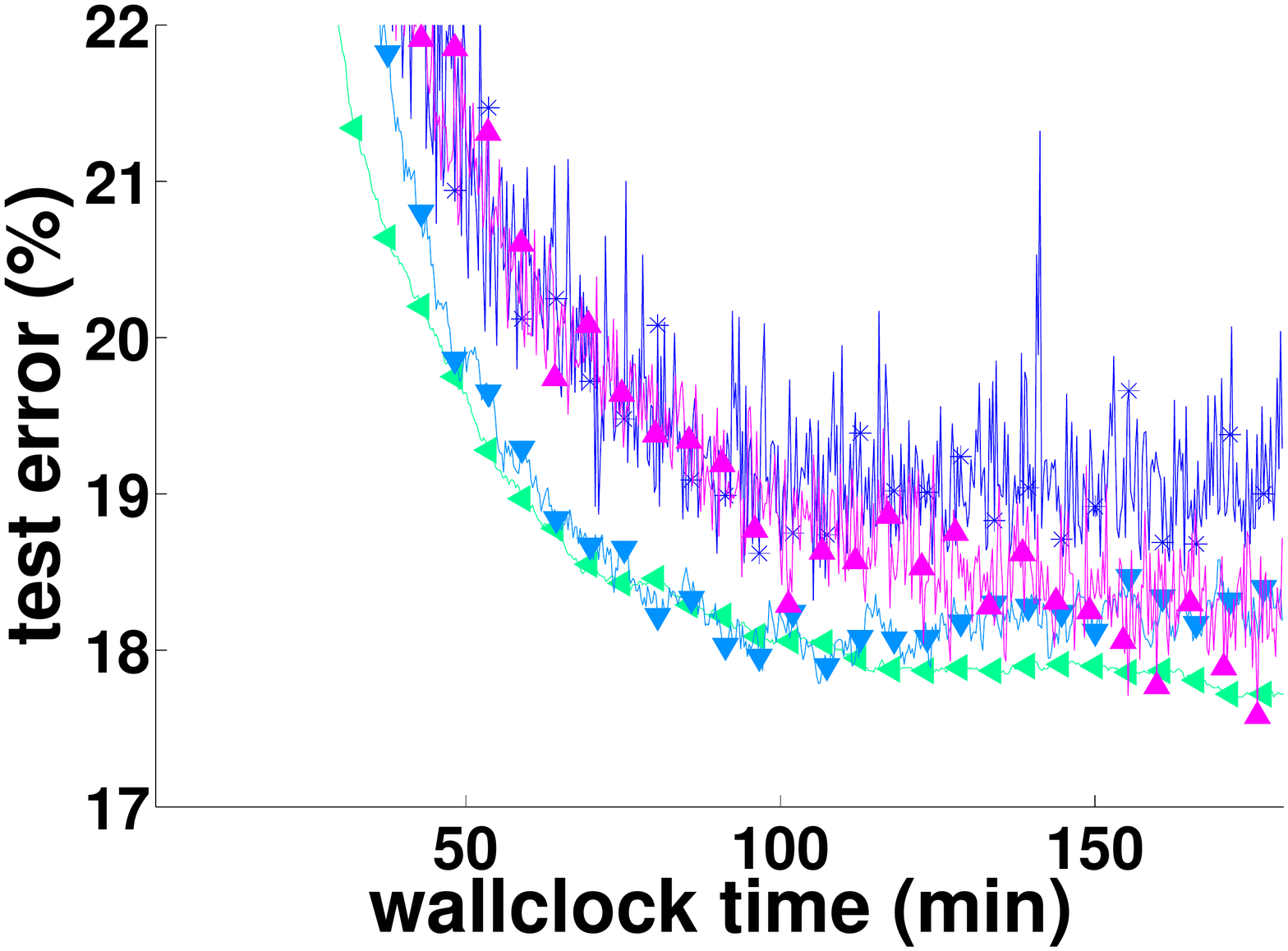}
\caption{Convergence of the training and test loss (negative log-likelihood) and the test error (original and zoomed) computed for the center variable as a function of wallclock time for \textit{SGD}, \textit{ASGD}, \textit{MVASGD} and \textit{MSGD} ($p = 1$) on the \textit{CIFAR} experiment.}
\label{fig:CIFAR_supp}
\end{figure}

For comparison we also report the performance of \textit{MSGD} which outperformed \textit{SGD}, \textit{ASGD} and \textit{MVASGD} on the test dataset. Recall that the way we compare the performance
between different methods is based on the smallest achievable test error. Since the test dataset is fixed
a prior, we may have the tendency to overfit this test dataset. Indeed, as we shall see in the Figure~\ref{fig:CIFAR_supp}.
One could use cross-validation to remedy this, we however emphasize that the point here is not to seek the best possible test accuracy, but to see all the possibilities that we can find, i.e. the richness of the dynamics arising from the neural network. We are aware that we could not exhaust all the possibilities. % It is a widely open question regarding the statistical property of the optimization solutions in neural networks.

\begin{figure}[p]
\center
\includegraphics[trim=0cm 0cm 0cm 0cm,clip,width = 3in]{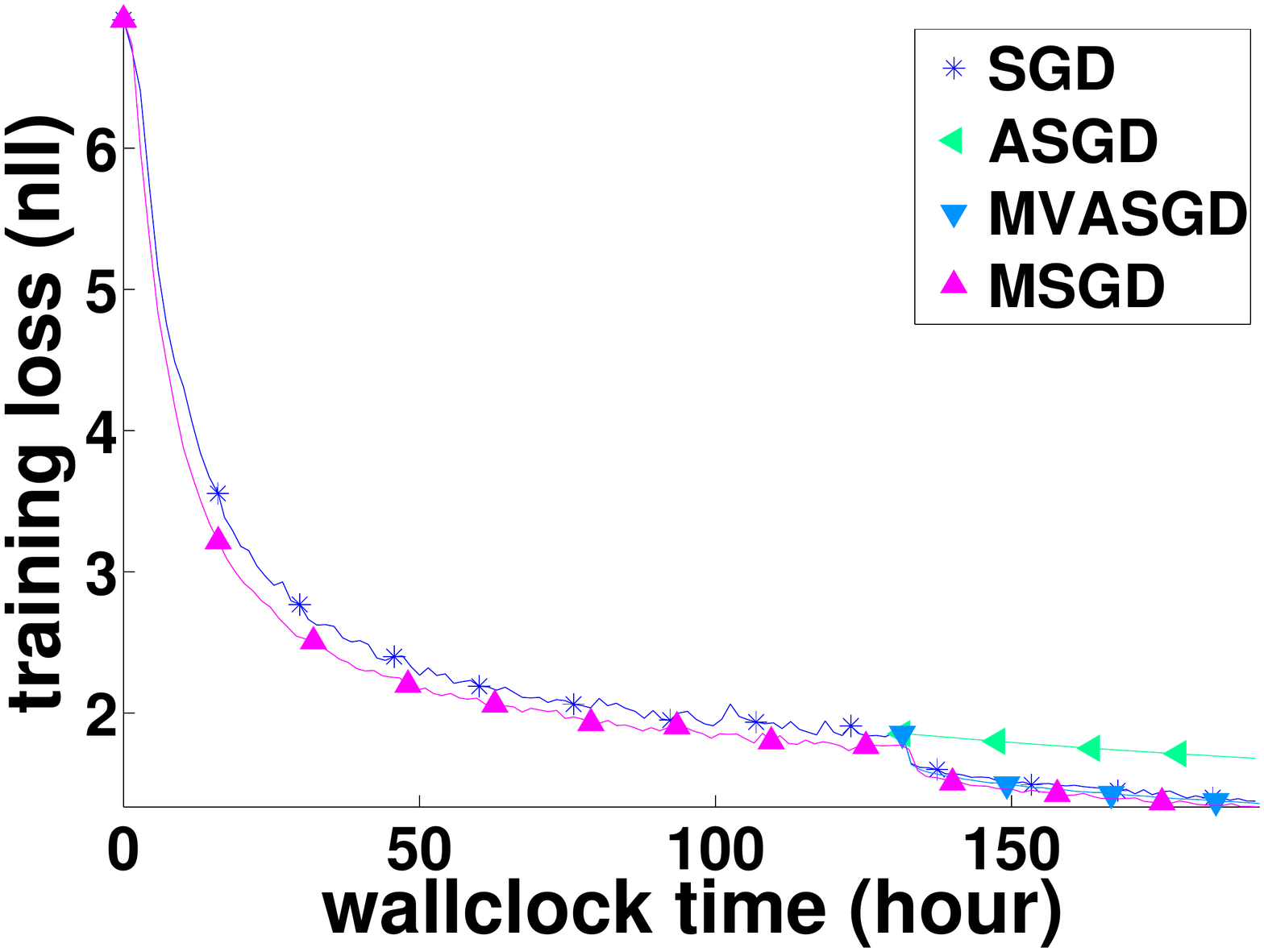}
\hspace{-0.3in}\includegraphics[trim=0cm 0cm 0cm 0cm,clip,width = 3in]{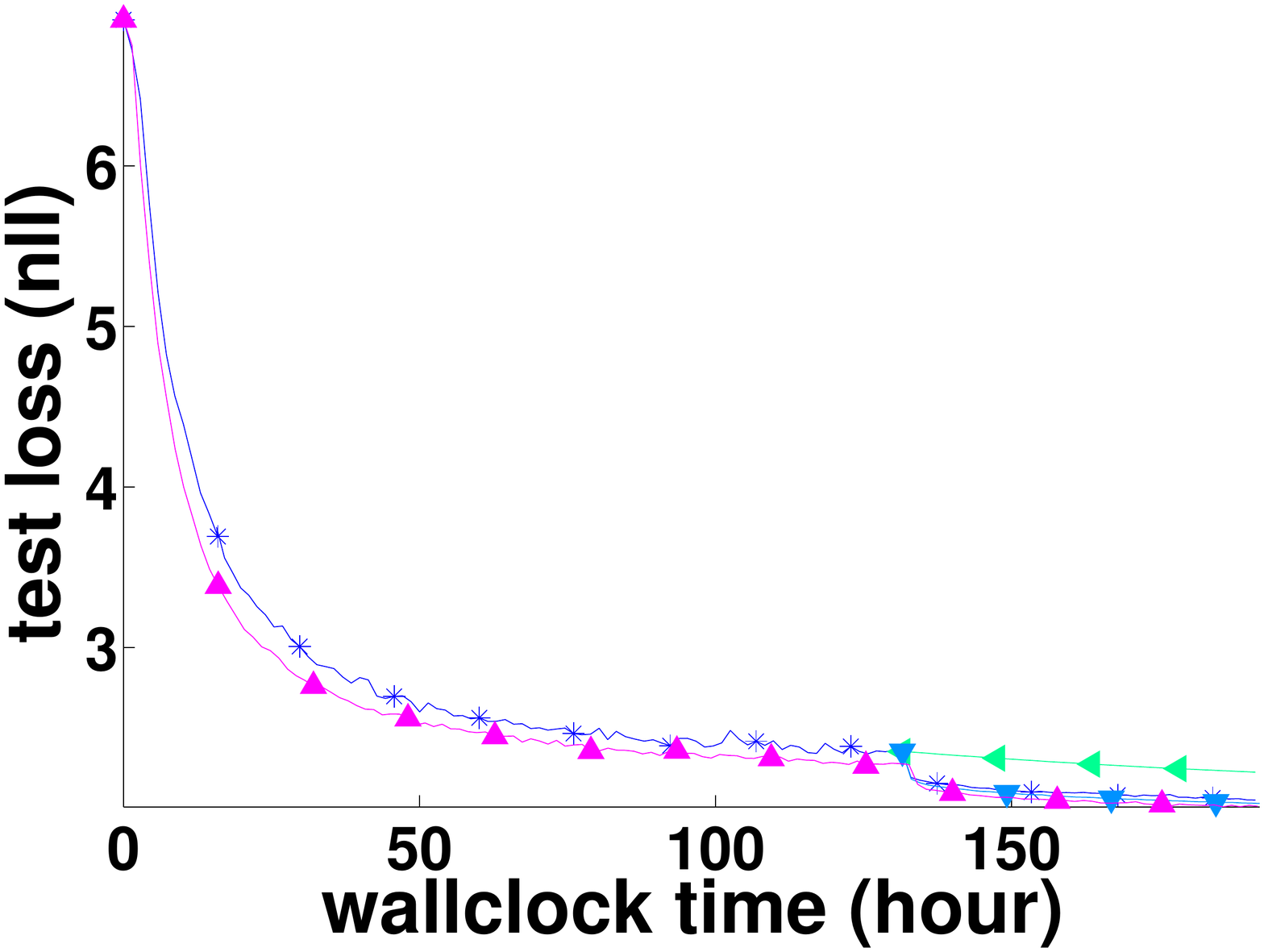} \\
\includegraphics[trim=0cm 0cm 0cm 0cm,clip,width = 3in]{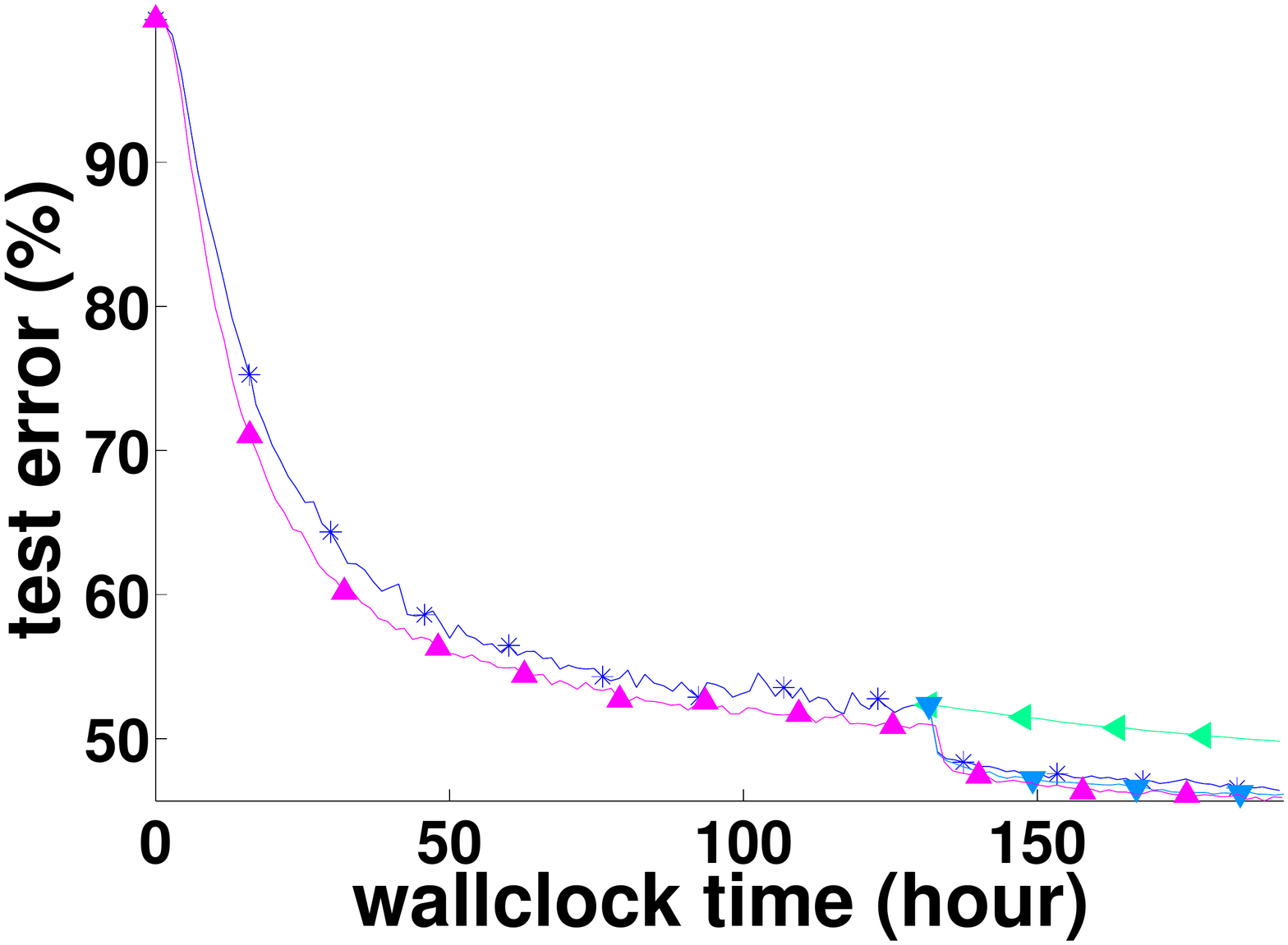}
\hspace{-0.3in}\includegraphics[trim=0cm 0cm 0cm 0cm,clip,width = 3in]{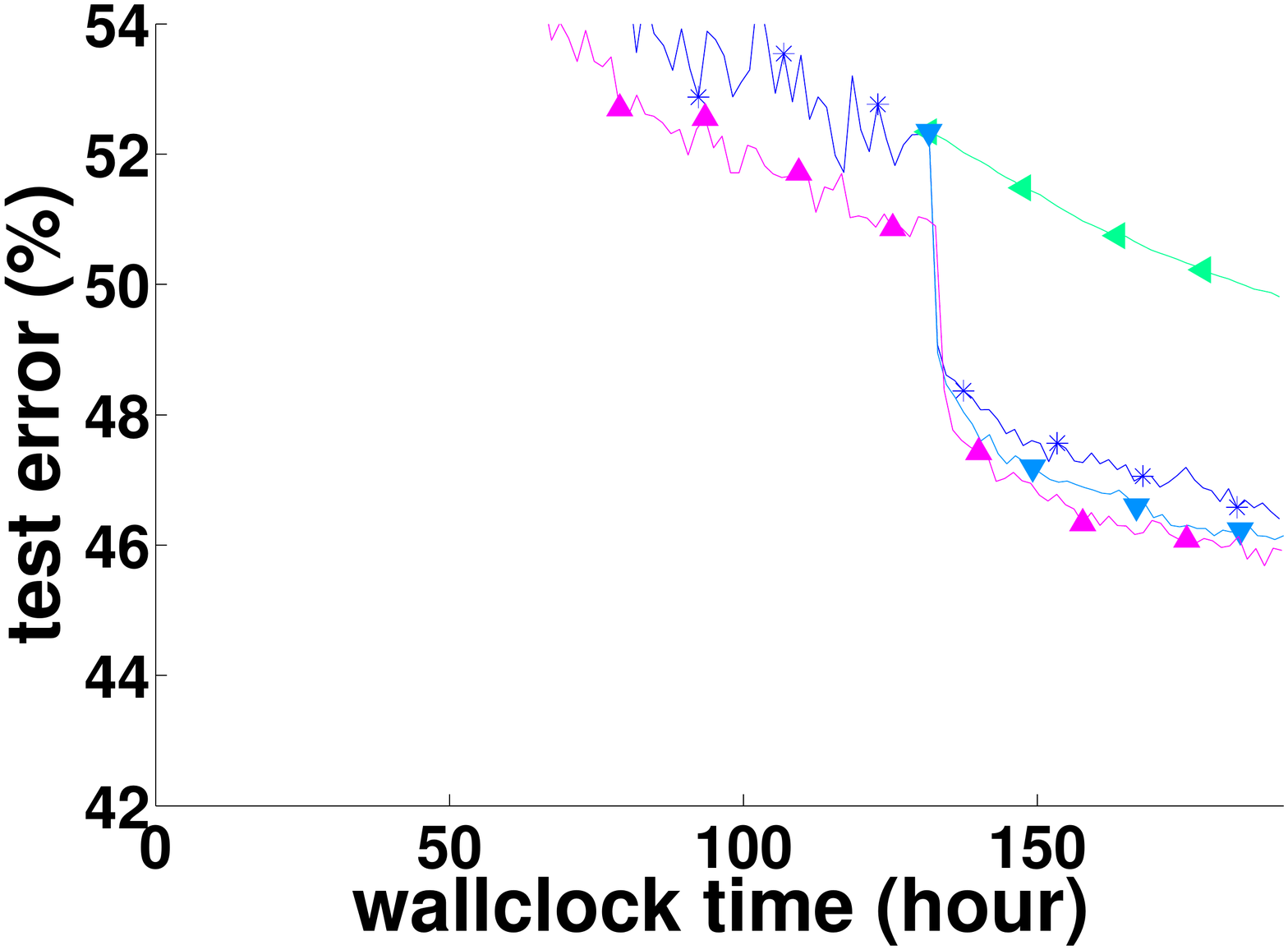}
\caption{Convergence of the training and test loss (negative log-likelihood) and the test error (original and zoomed) computed for the center variable as a function of wallclock time for \textit{SGD}, \textit{ASGD}, \textit{MVASGD} and \textit{MSGD} ($p = 1$) on the \textit{ImageNet} experiment.}
\label{fig:ImageNet_supp}
\end{figure}

Figure~\ref{fig:CIFAR_supp} shows the convergence of the training and test loss (negative log-likelihood) and the test error computed for the center variable as a function of wallclock time for \textit{SGD}, \textit{ASGD}, \textit{MVASGD} and \textit{MSGD} ($p=1$) on the \textit{CIFAR} experiment. We observe that the final test performance of \textit{ASGD} and \textit{MSGD} are quite close to each other. But \textit{ASGD} is much faster from the beginning. This explains why we do not see much speedup of the \textit{EASGD} method (e.g. in Figure~\ref{fig:CIFAR_tau1},~\ref{fig:CIFAR_tau4},~\ref{fig:CIFAR_tau16}) and can sometimes be even slower (e.g. in Figure~\ref{fig:CIFAR_tau64}). It is caused by the sensitivity of the test performance to the choice of the learning rate. We shall discuss this phenomenon further in the next section~\ref{sec:deplr}.

Figure~\ref{fig:ImageNet_supp} shows the convergence of the training and test loss (negative log-likelihood) and the test error computed for the center variable as a function of wallclock time for \textit{SGD}, \textit{ASGD}, \textit{MVASGD} and \textit{MSGD} ($p=1$) on the \textit{ImageNet} experiment.
Note that for all \textit{CIFAR} experiments we always start the averaging for the $ADOWNPOUR$ and $ASGD$ methods from the very beginning of each experiment. But for the \textit{ImageNet} experiments we start the averaging for the \textit{ASGD} and \textit{MVASGD} at the first time when we reduce the learning rate. We have tried to start the averaging from the right beginning, both of the training and test performance are poor and they look very similar to the \textit{ASGD} curve in Figure~\ref{fig:ImageNet_supp}. The big difference between \textit{ASGD} and \textit{MVASGD} is quite striking and worth further study.

\subsection{Dependence of the learning rate}
\label{sec:deplr}
\begin{figure}[p]
\center
\includegraphics[trim=0cm 0cm 0cm 0cm,clip,width = 3in]{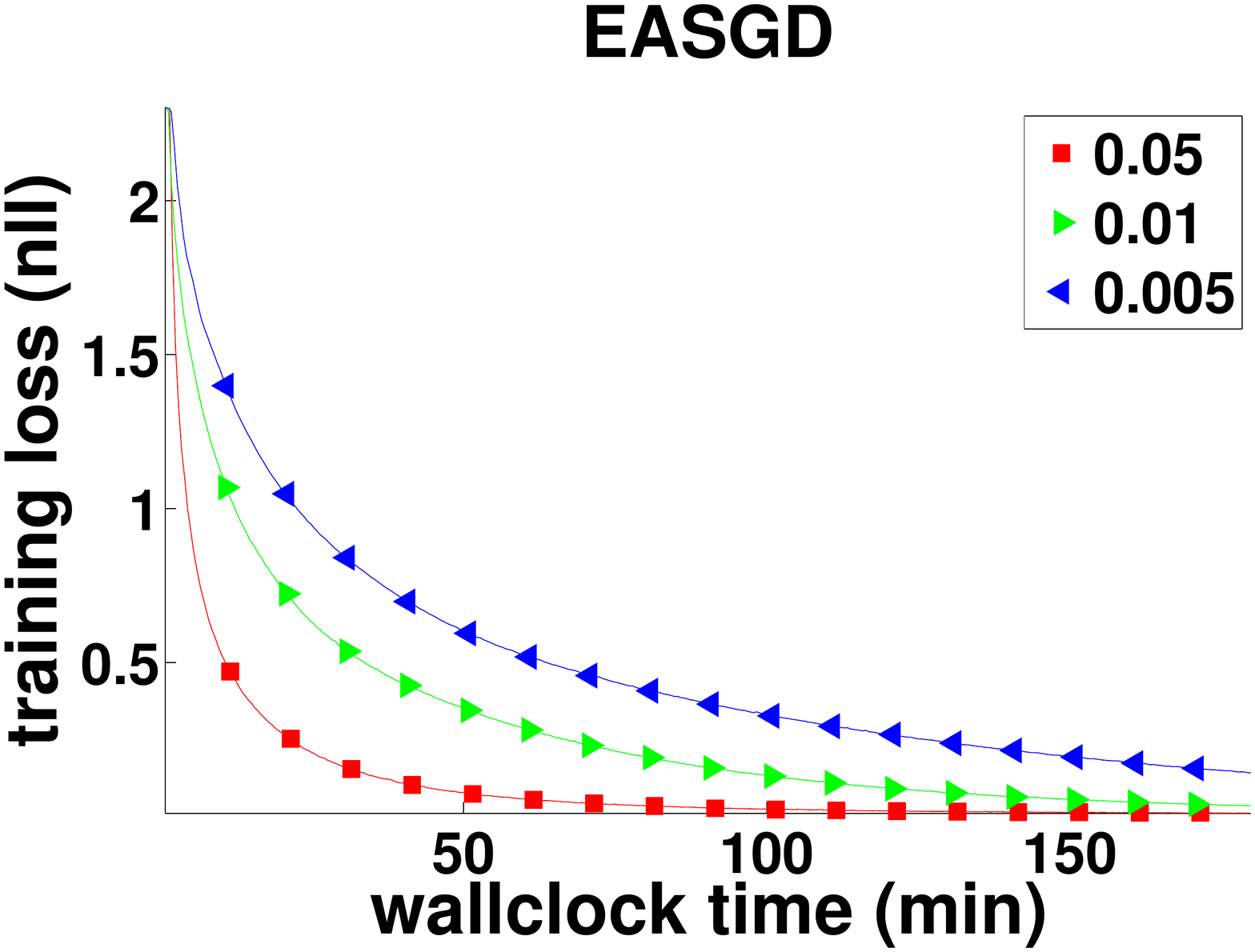}
\hspace{-0.3in}\includegraphics[trim=0cm 0cm 0cm 0cm,clip,width = 3in]{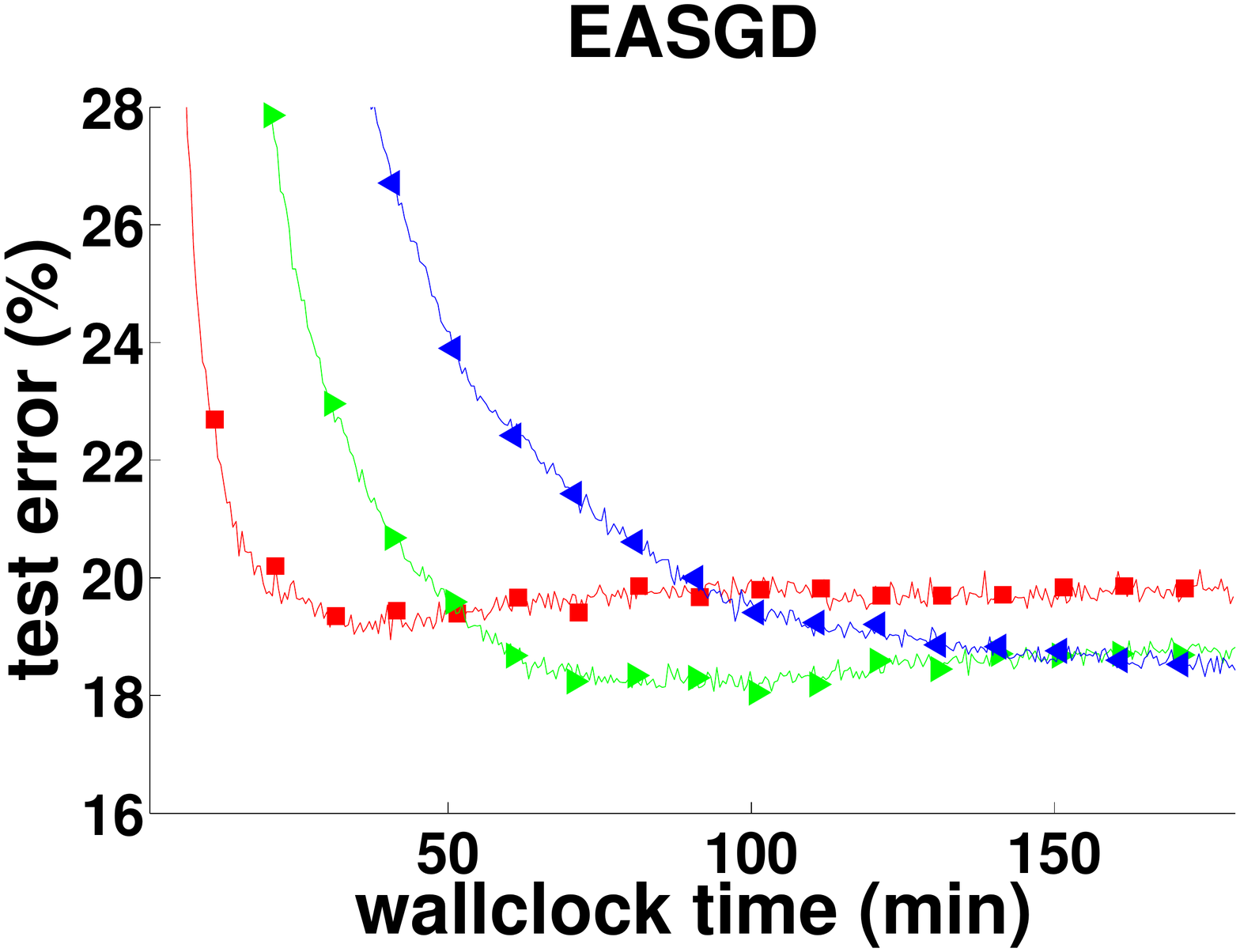} \\
\includegraphics[trim=0cm 0cm 0cm 0cm,clip,width = 3in]{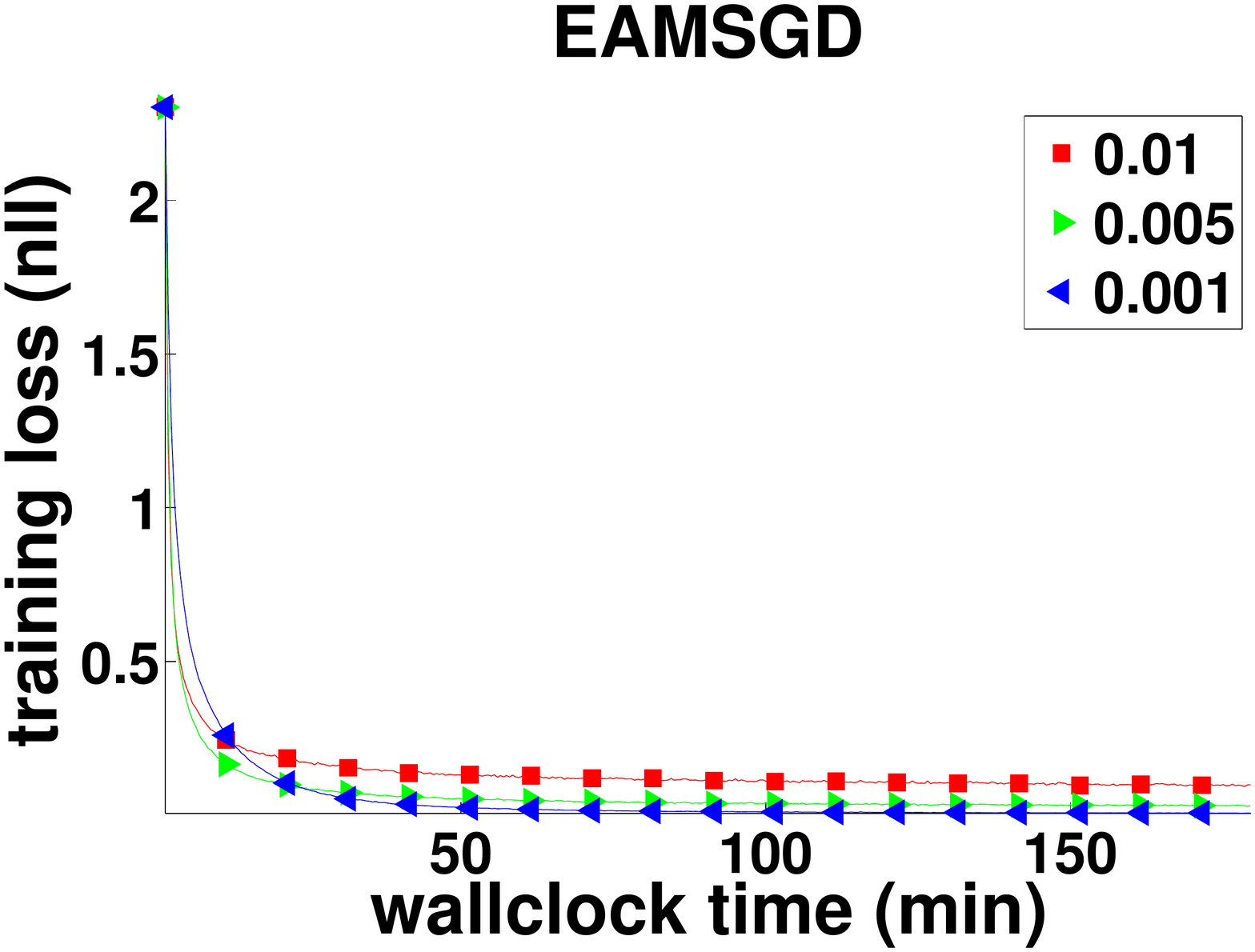}
\hspace{-0.3in}\includegraphics[trim=0cm 0cm 0cm 0cm,clip,width = 3in] {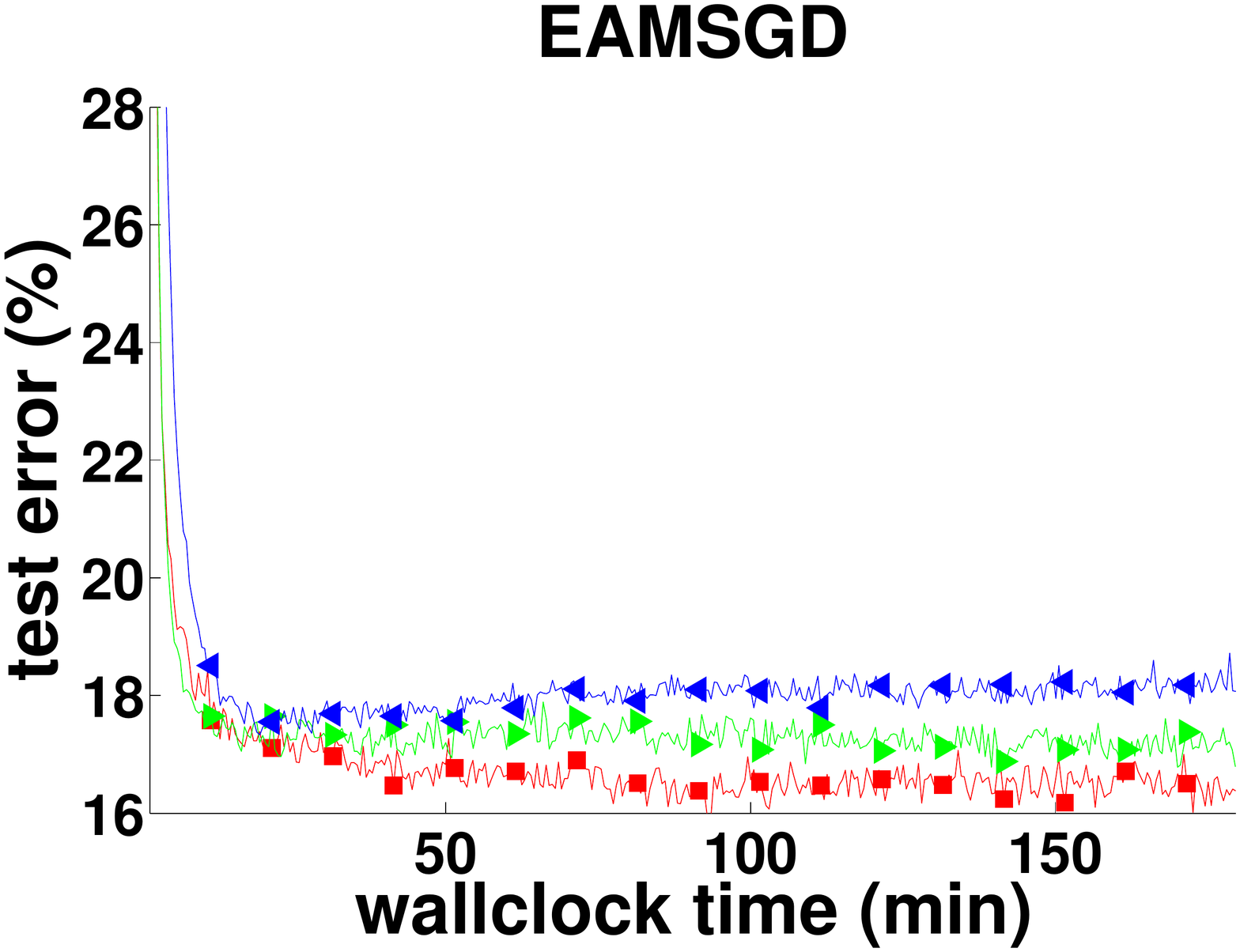}
\caption{Convergence of the training loss (negative log-likelihood, original) and the test error (zoomed) computed for the center variable as a function of wallclock time for \textit{EAMSGD} and \textit{EASGD} run with different values of $\eta$ on the \textit{CIFAR} experiment. $p=16$, $\tau = 10$.}
\label{fig:CIFAR3}
\end{figure}

This section discusses the dependence of the trade-off between exploration and exploitation on the learning rate. We compare the performance of respectively \textit{EAMSGD} and \textit{EASGD} for different learning rates $\eta$ when $p=16$ and $\tau = 10$ on the \textit{CIFAR} experiment. We observe in Figure~\ref{fig:CIFAR3} that higher learning rates $\eta$ lead to better test performance for the \textit{EAMSGD} algorithm which potentially can be justified by the fact that they sustain higher fluctuations of the local workers. We conjecture that higher fluctuations lead to more exploration and simultaneously they also impose higher regularization. This picture however seems to be opposite for the \textit{EASGD} algorithm for which larger learning rates hurt the performance of the method and lead to overfitting. Interestingly in this experiment for both \textit{EASGD} and \textit{EAMSGD} algorithm, the learning rate for which the best training performance was achieved simultaneously led to the worst test performance.

\subsection{Dependence of the communication period}
\label{sec:exp_dep_tau}
\begin{figure}[p]
\center
\includegraphics[trim=0cm 0cm 0cm 0cm,clip,width = 3in]{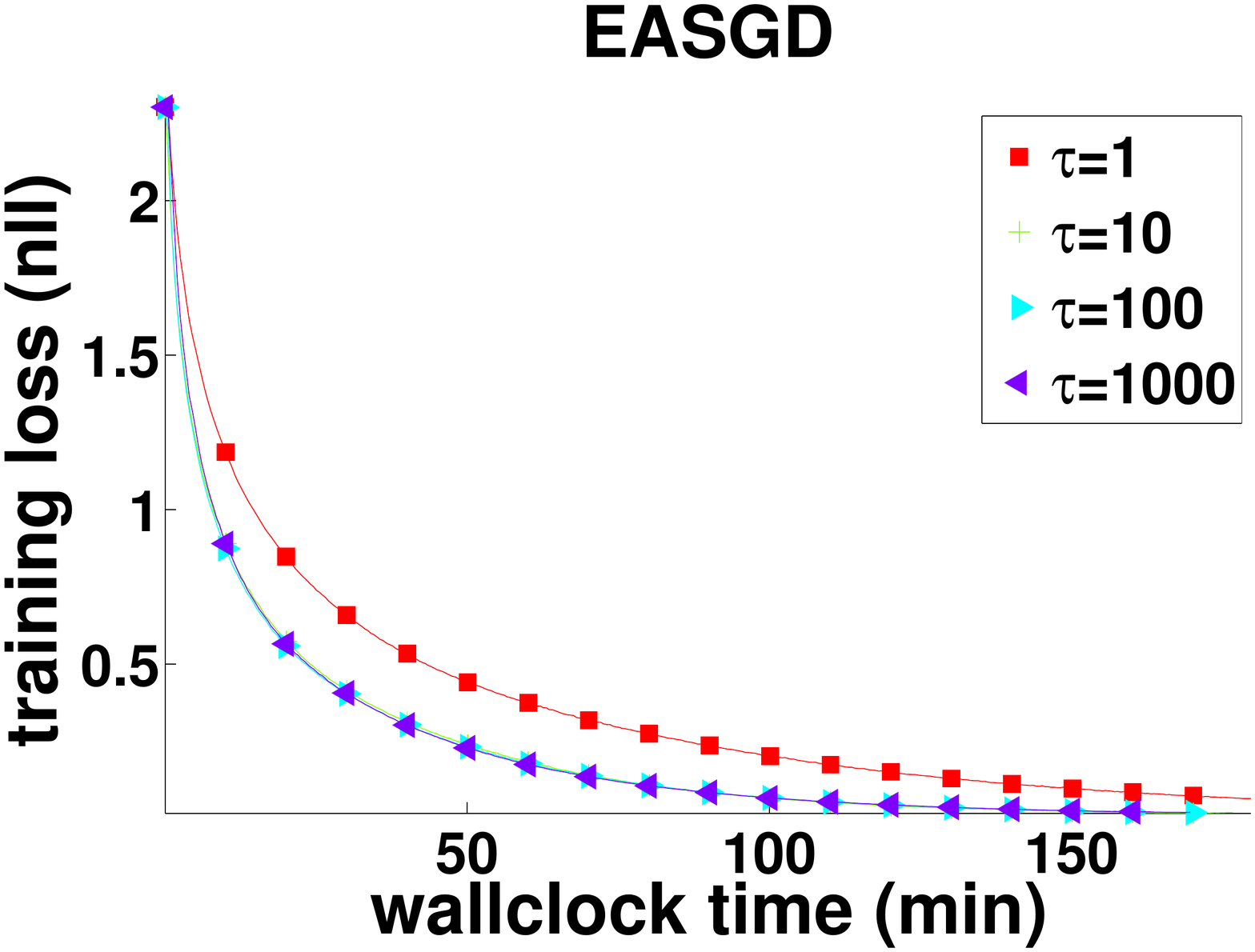}
\hspace{-0.3in}\includegraphics[trim=0cm 0cm 0cm 0cm,clip,width = 3in]{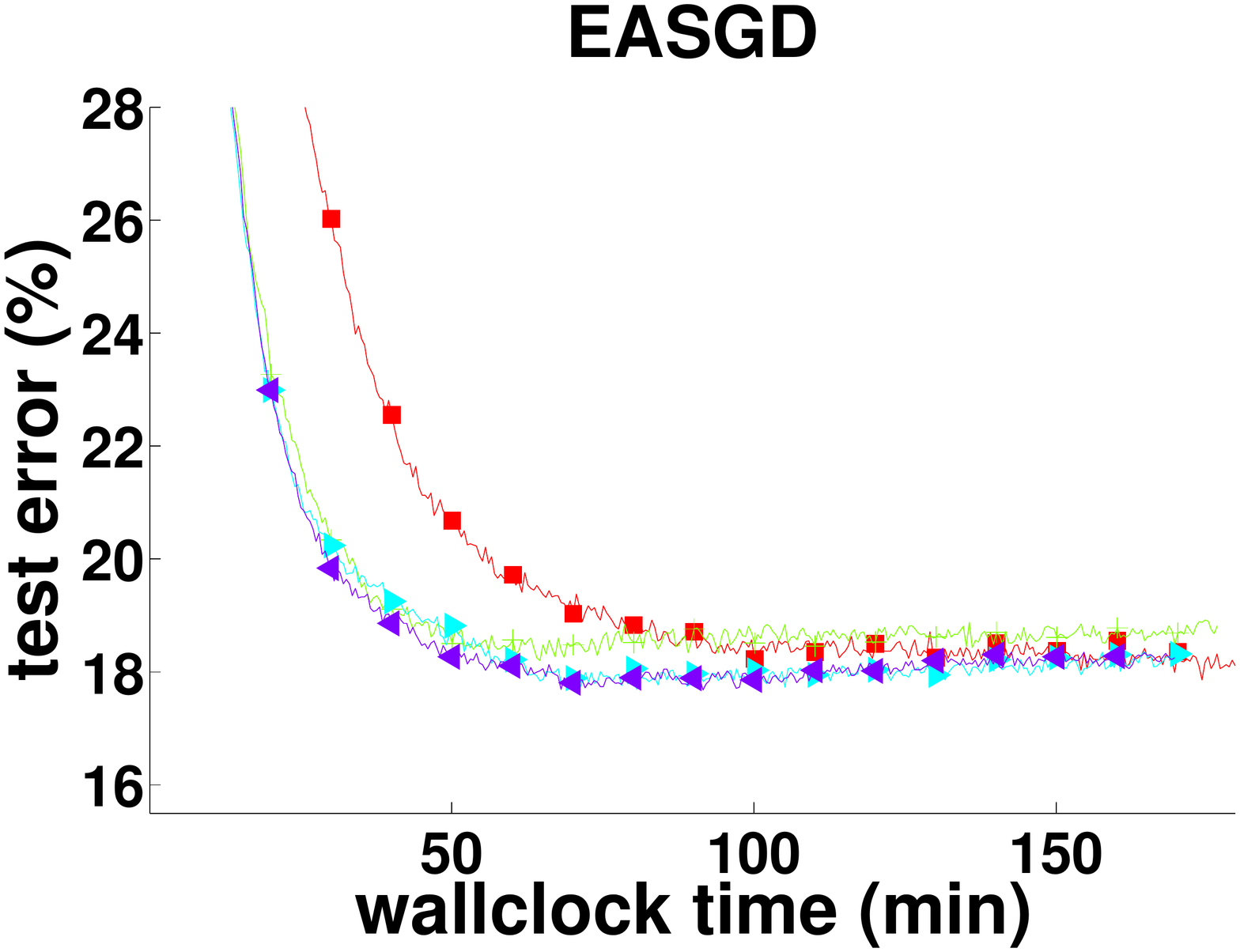} \\
\includegraphics[trim=0cm 0cm 0cm 0cm,clip,width = 3in]{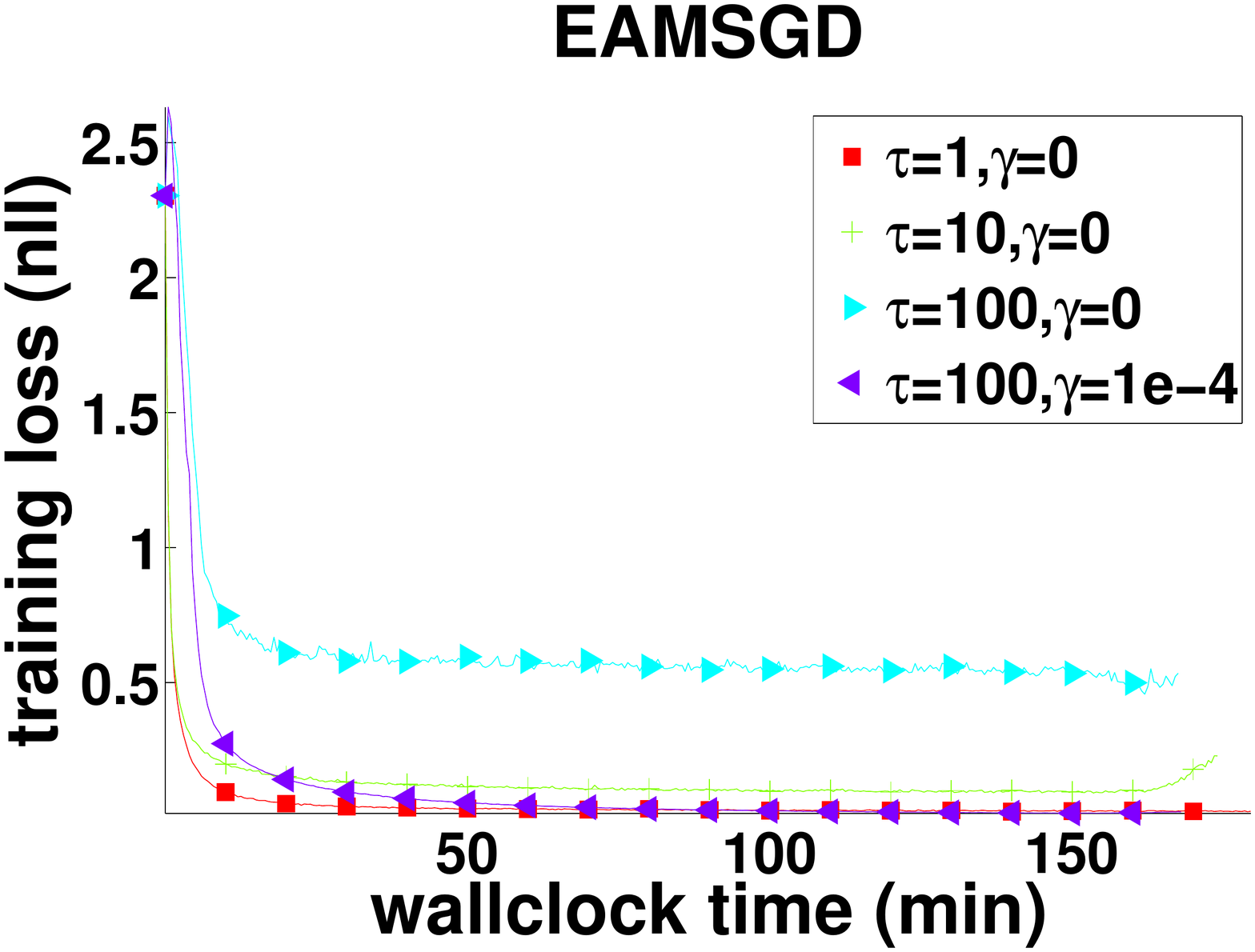}
\hspace{-0.3in}\includegraphics[trim=0cm 0cm 0cm 0cm,clip,width = 3in]{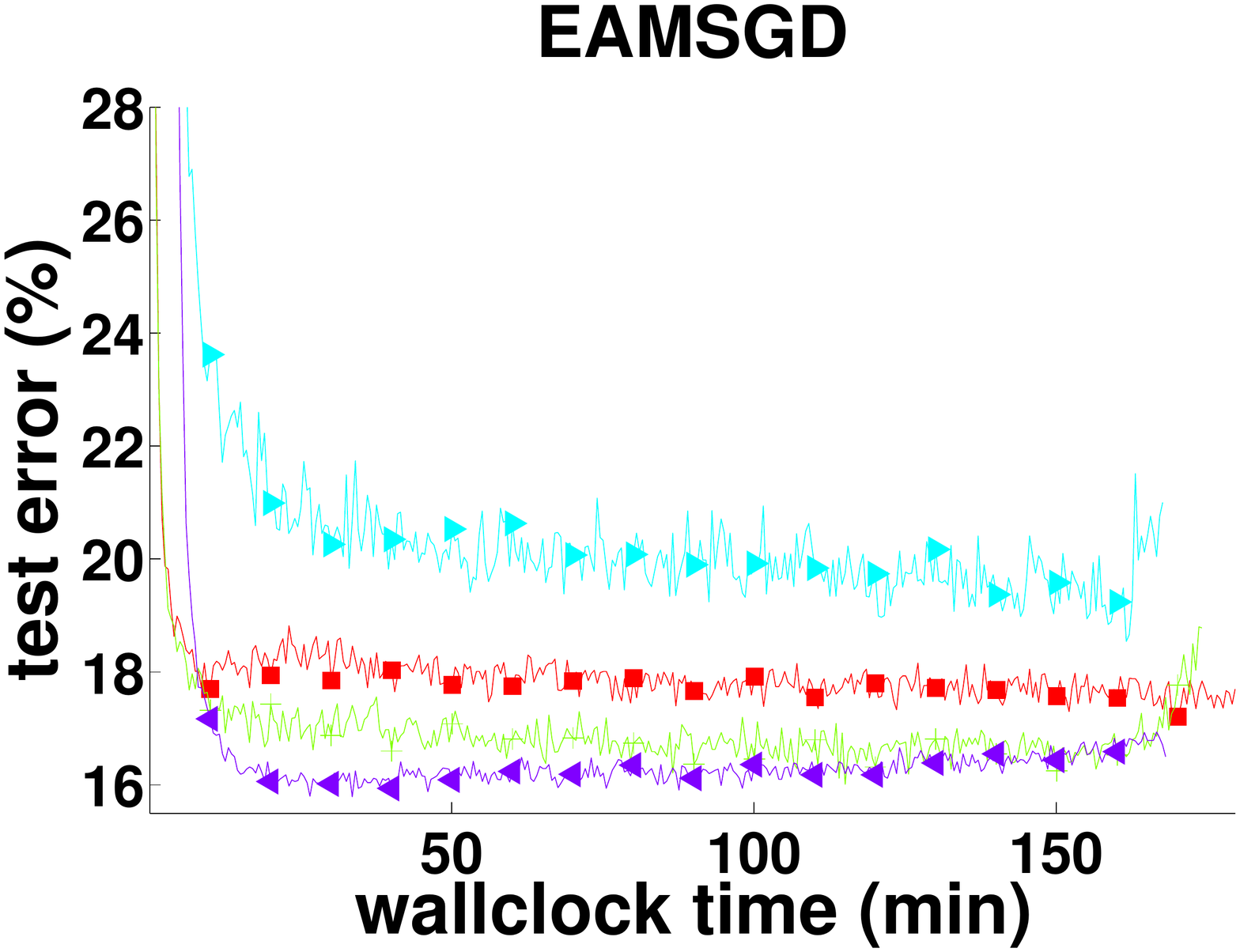}
\caption{Convergence of the training loss (negative log-likelihood, original) and the test error (zoomed) computed for the center variable as a function of wallclock time for \textit{EASGD} and \textit{EAMSGD}
($p = 16,\eta=0.01,\beta=0.9,\delta=0.99$) on the \textit{CIFAR} experiment with various communication period $\tau$ and learning rate decay $\gamma$. The learning rate is decreased gradually over time based each local worker's own clock $t$ with $\eta_t= \eta/(1+\gamma t)^{0.5}$.}
\label{fig:CIFAR3_supp}
\end{figure}

This section discusses the dependence of the trade-off between exploration and exploitation on the communication period. We observe in Figure~\ref{fig:CIFAR3_supp} that \textit{EASGD} algorithm exhibits very similar convergence behavior when $\tau=1$ up to even $\tau=1000$ for the \textit{CIFAR} experiment, whereas \textit{EAMSGD} can get trapped at a quite high energy level (of the objective) when $\tau=100$. This trapping behavior is due to the non-convexity of the objective function. It can be avoided by gradually decreasing the learning rate, i.e. increasing the penalty term $\rho$ (recall $\alpha = \eta \rho$), as shown in Figure~\ref{fig:CIFAR3_supp}. In contrast, the \textit{EASGD} algorithm does not seem to get trapped by any saddle point at all along its trajectory.

The performance\footnote{Compared to all earlier results, the experiment in this section is re-run three times with a new random seed and with faster cuDNN package on two Tesla K80 nodes (\url{developer.nvidia.com/cuDNN} and \url{github.com/soumith/cudnn.torch}).
Also to clarify, the random initialization we use is by default in Torch's implementation. All our methods are implemented in Torch (\url{torch.ch}). The Message Passing Interface implementation MVAPICH2 (\url{mvapich.cse.ohio-state.edu}) is used for the GPU-CPU communication.} of \textit{EASGD} being less sensitive to the communication period compared to \textit{EAMSGD} is another striking observation.

It's also very important to notice the tail behavior of the asynchronous \textit{EASGD} method, i.e. what would happen if some local worker had finished the gradient updates and stopped the communication with the master.
In the \textit{EAMSGD} case in Figure~\ref{fig:CIFAR3_supp}, we see that the final training loss and test error can both become worse. This is due to the situation that some of the local workers have stopped earlier than the others, so that the averaging effect on the center variable is diminished.

\subsection{The tradeoff between data and parameter communication}

In addition, we report in Table~\ref{tab:comptime} the breakdown of the total running time for \textit{EASGD} when $\tau=10$ (the time breakdown for \textit{EAMSGD} is almost identical) and \textit{DOWNPOUR} when $\tau = 1$ into computation time, data loading time and parameter communication time. For the \textit{CIFAR} experiment the reported time corresponds to processing $400 \times 128$ data samples whereas for the \textit{ImageNet} experiment it corresponds to processing $1024 \times 128$ data samples. For $\tau=1$ and $p \in \{8,16\}$ we observe that the communication time accounts for significant portion of the total running time whereas for $\tau=10$ the communication time becomes negligible compared to the total running time (recall that based on previous results \textit{EASGD} and \textit{EAMSGD} achieve best performance with larger $\tau$ which is ideal in the setting when communication is time-consuming).

\begin{table}[t]
\setlength{\tabcolsep}{1.75pt}
\begin{minipage}{.5\linewidth}
\begin{center}
\begin{tabular}{c|c|c|c|c|}
\cline{2-5}
\multirow{1}{*}{} & $p=1$ & $p=4$ & $p=8$ & $p=16$\\
\hline
\multicolumn{1}{|c|}{\multirow{1}{*}{$\tau = 1$}} & $12/1/0$ & $11/2/3$ & $11/2/5$ & $11/2/9$\\
\hline
\hline
\multicolumn{1}{|c|}{\multirow{1}{*}{$\tau = 10$}} & NA & $11/2/1$ & $11/2/1$ & $12/2/1$\\
\hline
\end{tabular}
\end{center}
\end{minipage}
\begin{minipage}{.5\linewidth}
\begin{center}
\begin{tabular}{c|c|c|c|}
\cline{2-4}
\multirow{1}{*}{} & $p=1$ & $p=4$ & $p=8$\\
\hline
\multicolumn{1}{|c|}{\multirow{1}{*}{$\tau = 1$}} & $1248/20/0$ & $1323/24/173$ & $1239/61/284$\\
\hline
\hline
\multicolumn{1}{|c|}{\multirow{1}{*}{$\tau = 10$}} & NA & $1254/58/7$ & $1266/84/11$\\
\hline
\end{tabular}
\end{center}
\end{minipage}
\caption{Approximate computation time, data loading time and parameter communication time [sec] for \textit{DOWNPOUR} (top line for $\tau=1$) and \textit{EASGD} (the time breakdown for \textit{EAMSGD} is almost identical) (bottom line for $\tau=10$). Left time corresponds to \textit{CIFAR} experiment and right table corresponds to \textit{ImageNet} experiment. The computation and the communication time may have some overlap, due to the MPI implementation.}
\label{tab:comptime}
\end{table}

We shall now examine the data communication cost in detail. Let's focus on the \textit{ImageNet} case.
Based on the Table~\ref{tab:comptime}, for each single GPU ($p=1,\tau=1$), it takes around 1248 seconds to process 1024 mini-batches of size 128. This is approximately processing one mini-batch per second. Each mini-batch consists of $128 \times 3 \times 221 \times 221$ pixels. If each pixel value is represented by one byte, then each mini-batch is around 18 MB. As the whole dataset has around 1,300,000 images, it would take 174 GB. In fact, we have compressed the images as JPEG, so that the whole dataset is only around 36 GB. Thus we gain
a compression ratio around 1/5, i.e. we can assume each mini-batch size is 18/5 = 3.6 MB.
The required data communication rate is thus 3.6 MB/sec. On the other hand, the $\textit{DOWNPOUR}$ method
with $\tau=1$ requires communicating the whole model parameter per mini-batch.
As the parameter size is around
233 MB, we need the network bandwidth at least 233 MB/sec per local worker (we have not even accounted for the gradient communication, which can double this cost). The parameter communication cost is thus at least 66 times of the data communication cost. In the \textit{ImageNet} case, having access to the full dataset by the local workers is indeed a good tradeoff.

\subsection{Time speed-up}
\begin{figure}[p]
\center
\includegraphics[trim=0cm 0cm 0cm 0cm,clip,width = 3in]{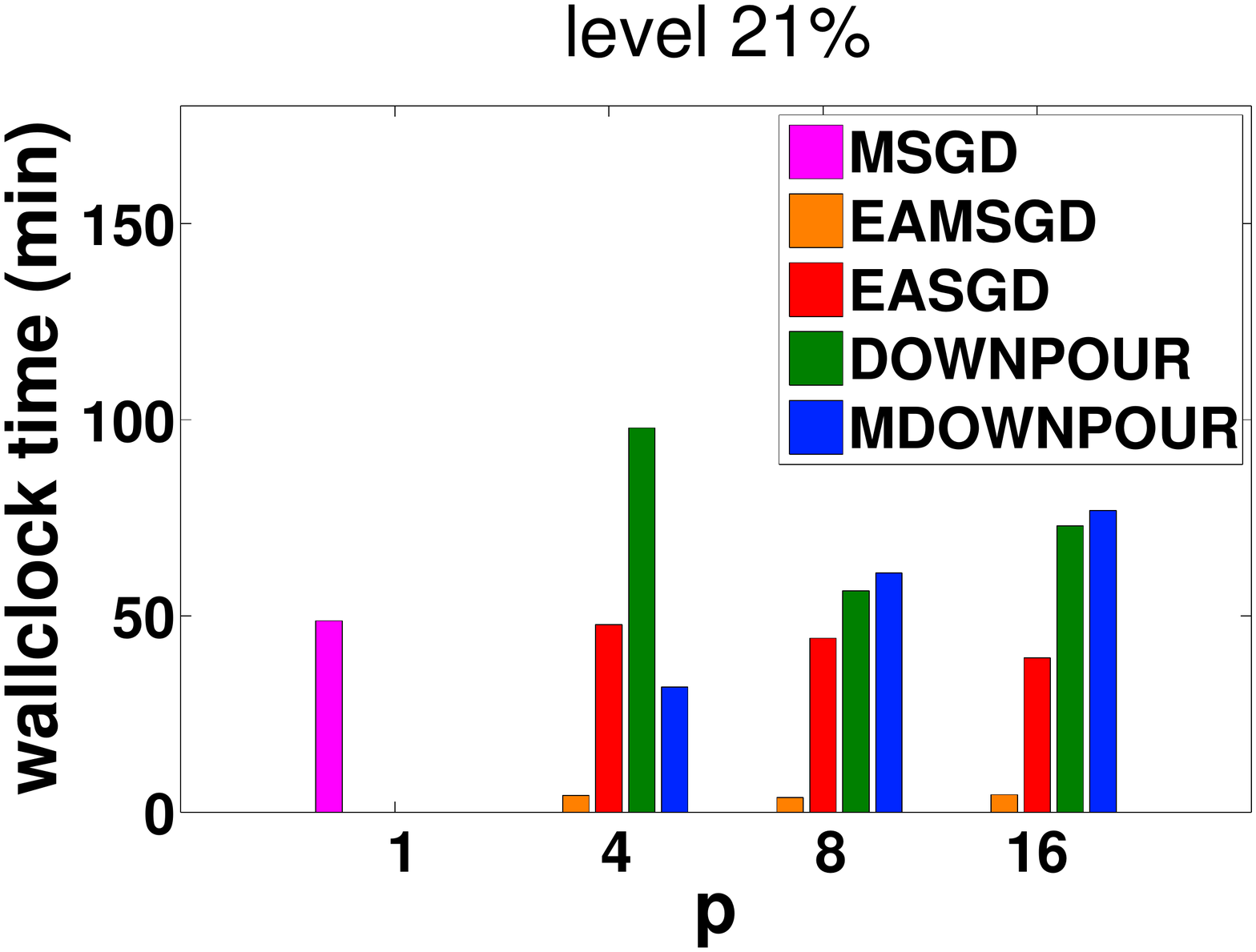}
\hspace{-0.3in}\includegraphics[trim=0cm 0cm 0cm 0cm,clip,angle=0,width = 3in]{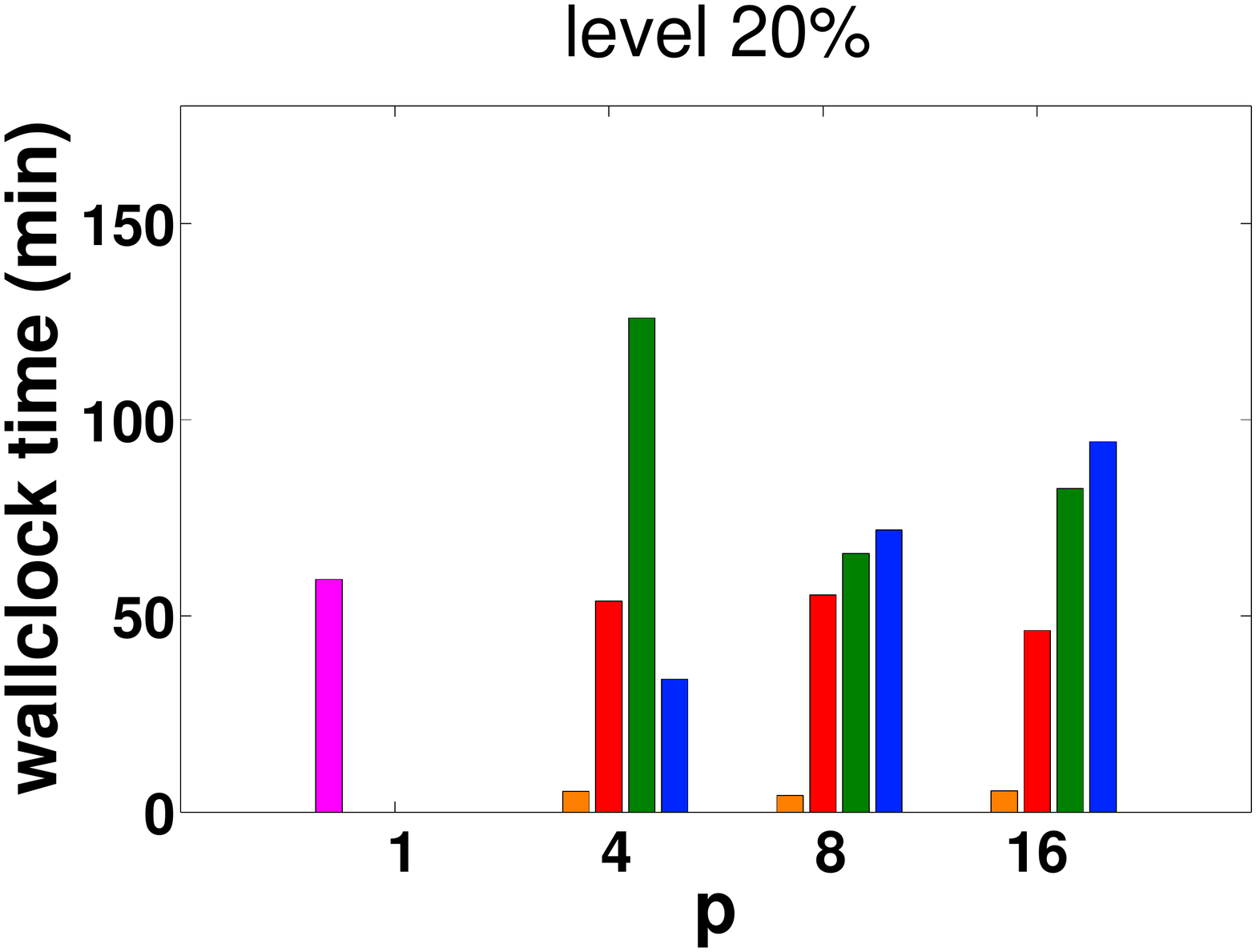} \\
\includegraphics[trim=0cm 0cm 0cm 0cm,clip,angle=0,width = 3in]{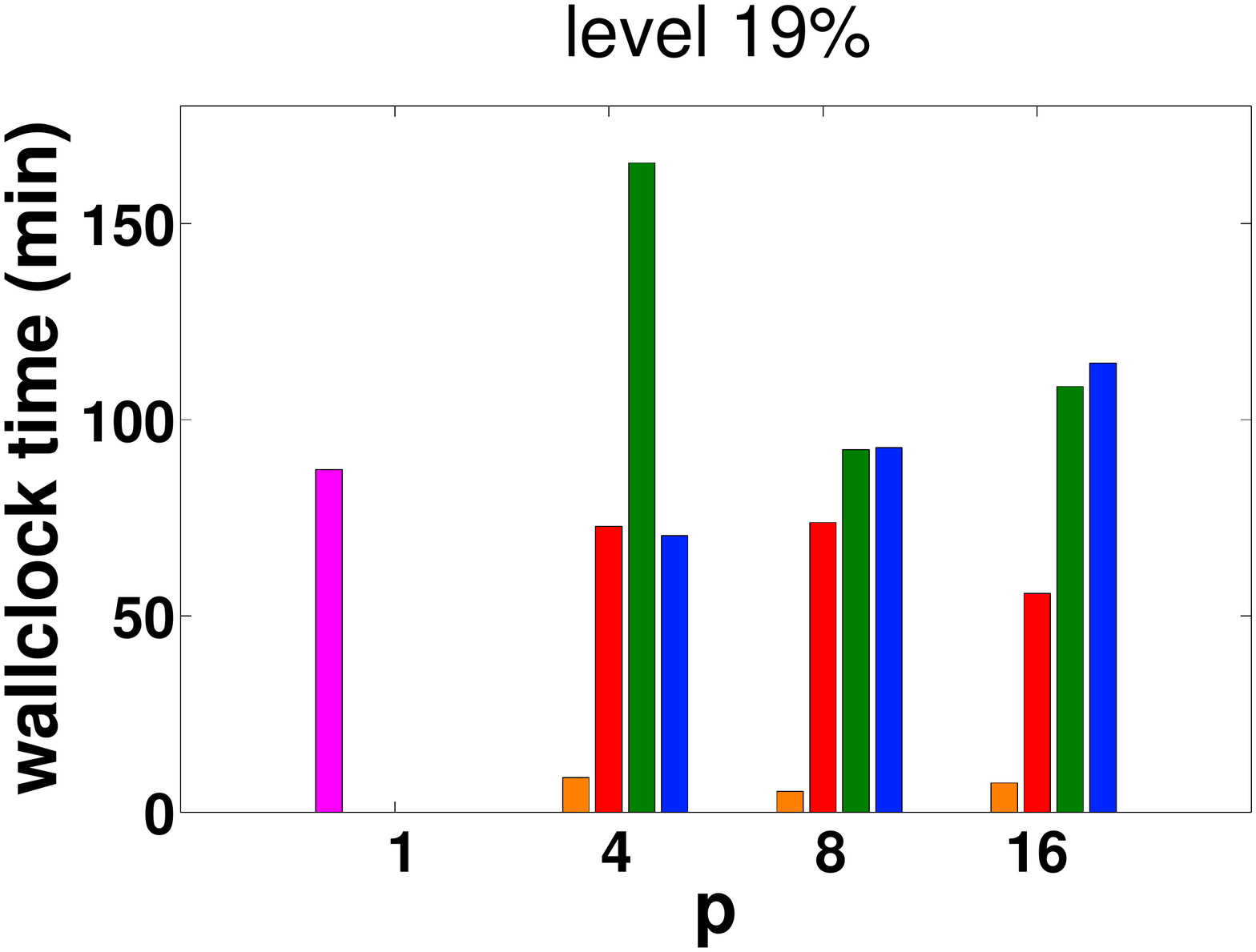}
\hspace{-0.3in}\includegraphics[trim=0cm 0cm 0cm 0cm,clip,angle=0,width = 3in]{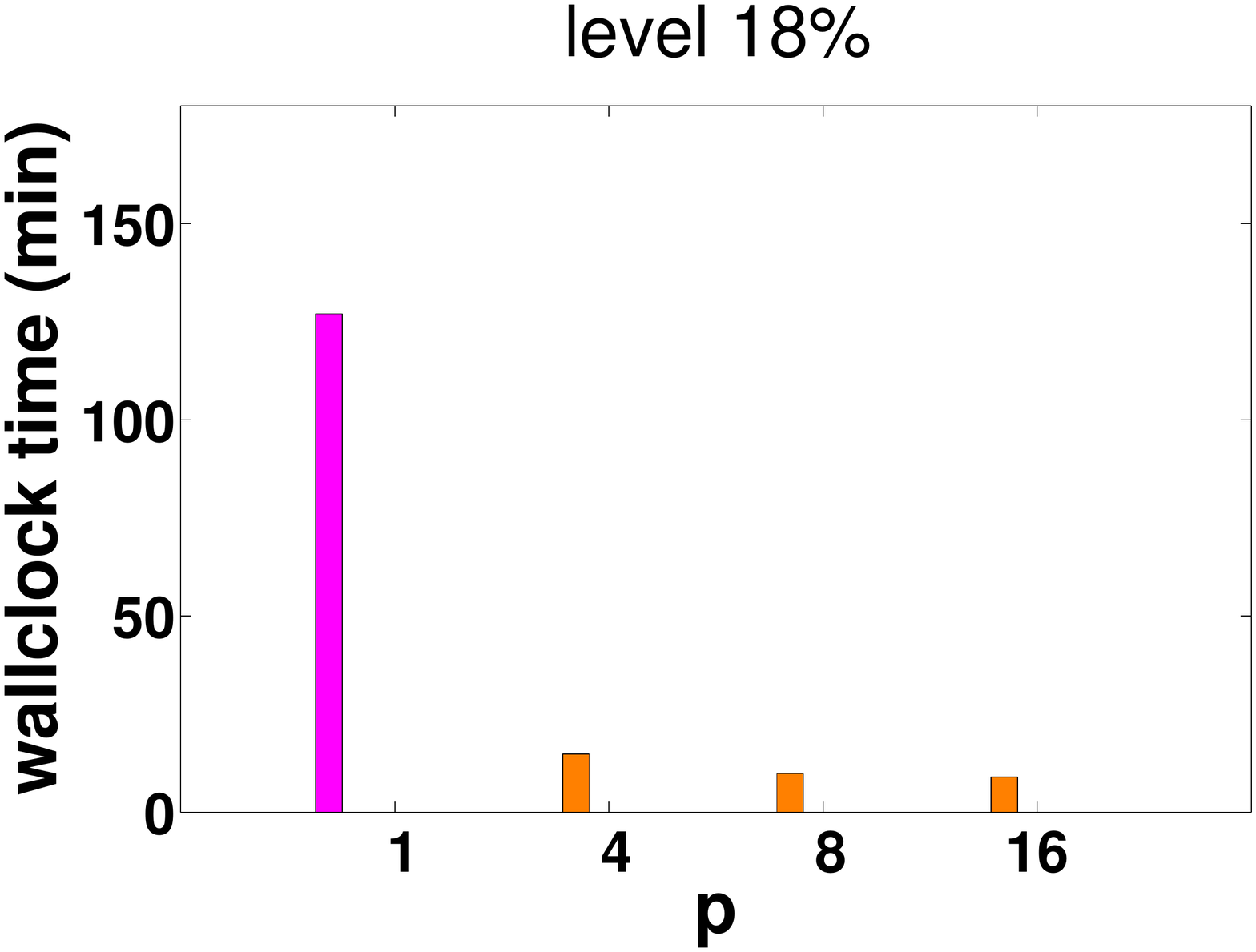}
\caption{The wall clock time needed to achieve the same level of the test error $thr$ as a function of the number of local workers $p$ on the \textit{CIFAR} dataset. From left to right: $thr = \{21\%,20\%,19\%,18\%\}$. Missing bars denote that the method never achieved specified level of test error.}.
\label{fig:CIFARspeedup}
\end{figure}

\begin{figure}[p]
\center
\includegraphics[trim=0cm 0cm 0cm 0cm,clip,width = 3in]{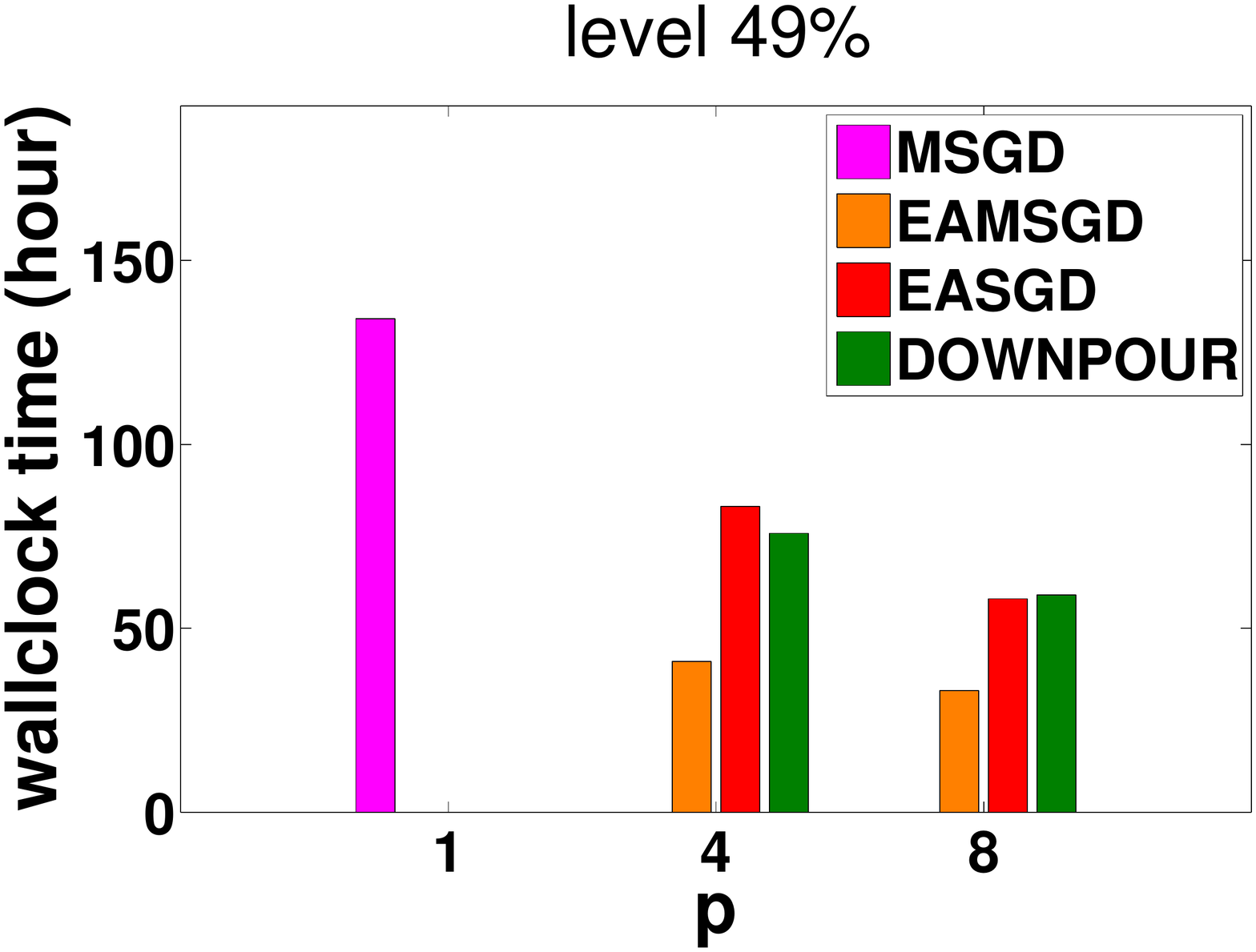}
\hspace{-0.3in}\includegraphics[trim=0cm 0cm 0cm 0cm,clip,angle=0,width = 3in]{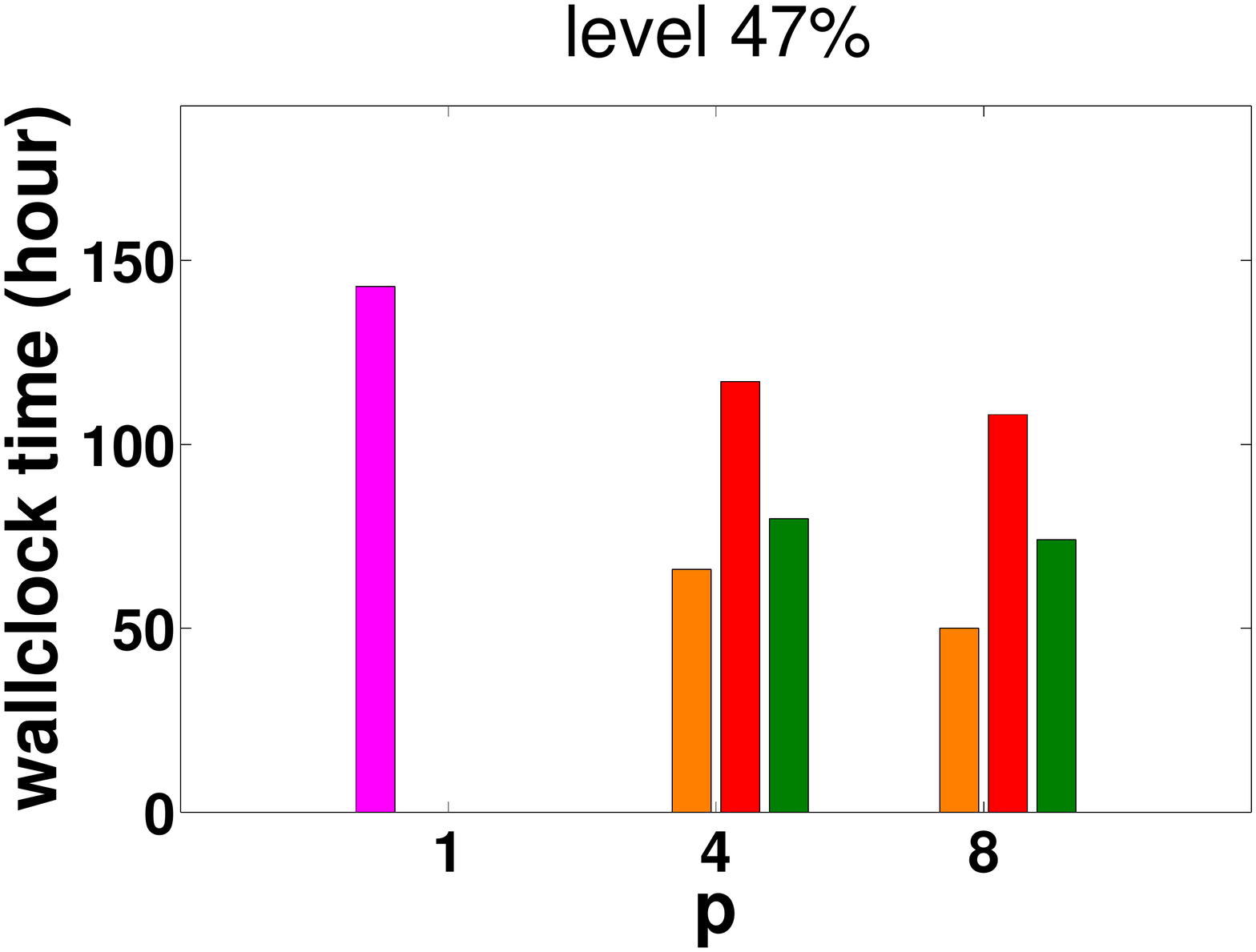} \\
\includegraphics[trim=0cm 0cm 0cm 0cm,clip,angle=0,width = 3in]{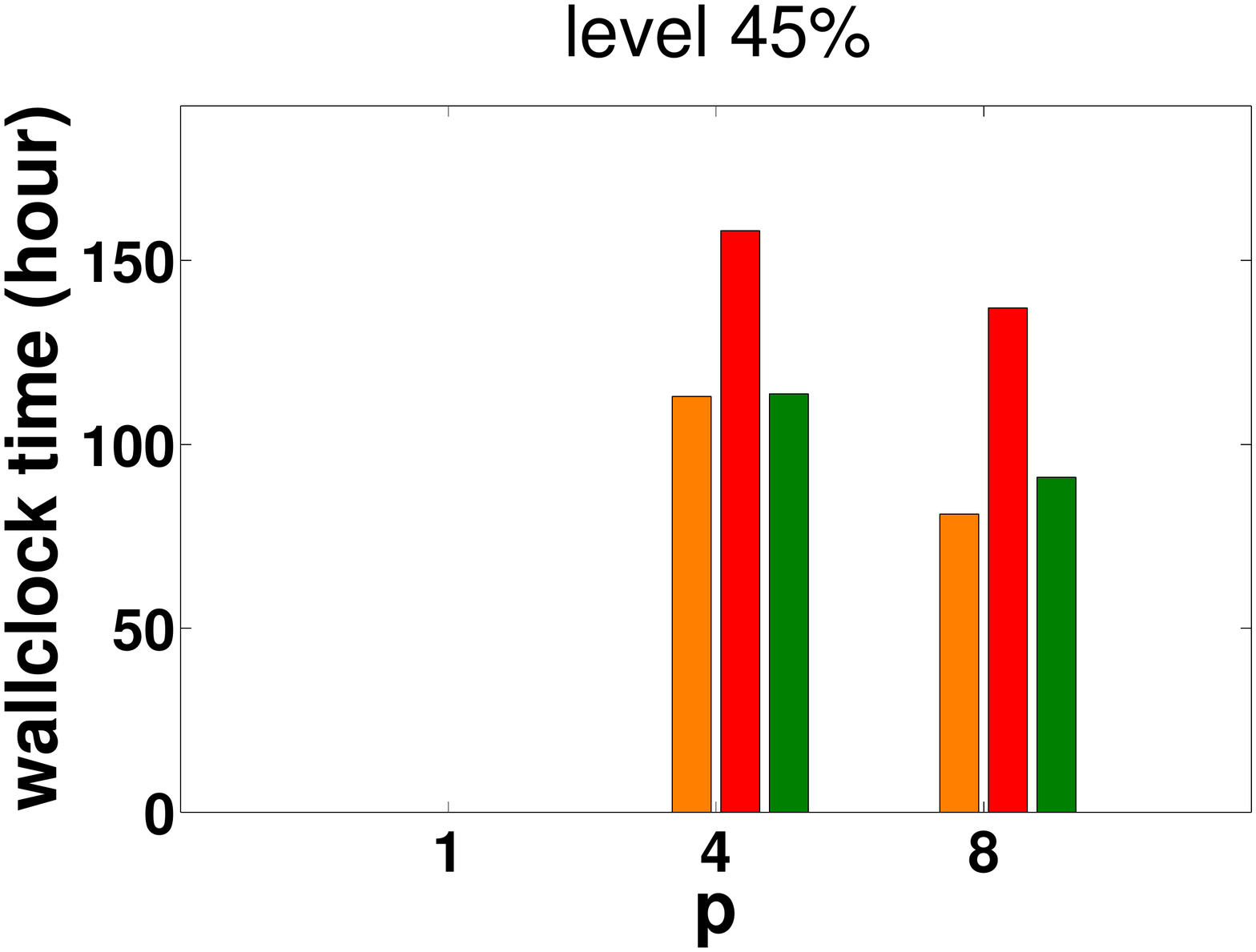}
\hspace{-0.3in}\includegraphics[trim=0cm 0cm 0cm 0cm,clip,angle=0,width = 3in]{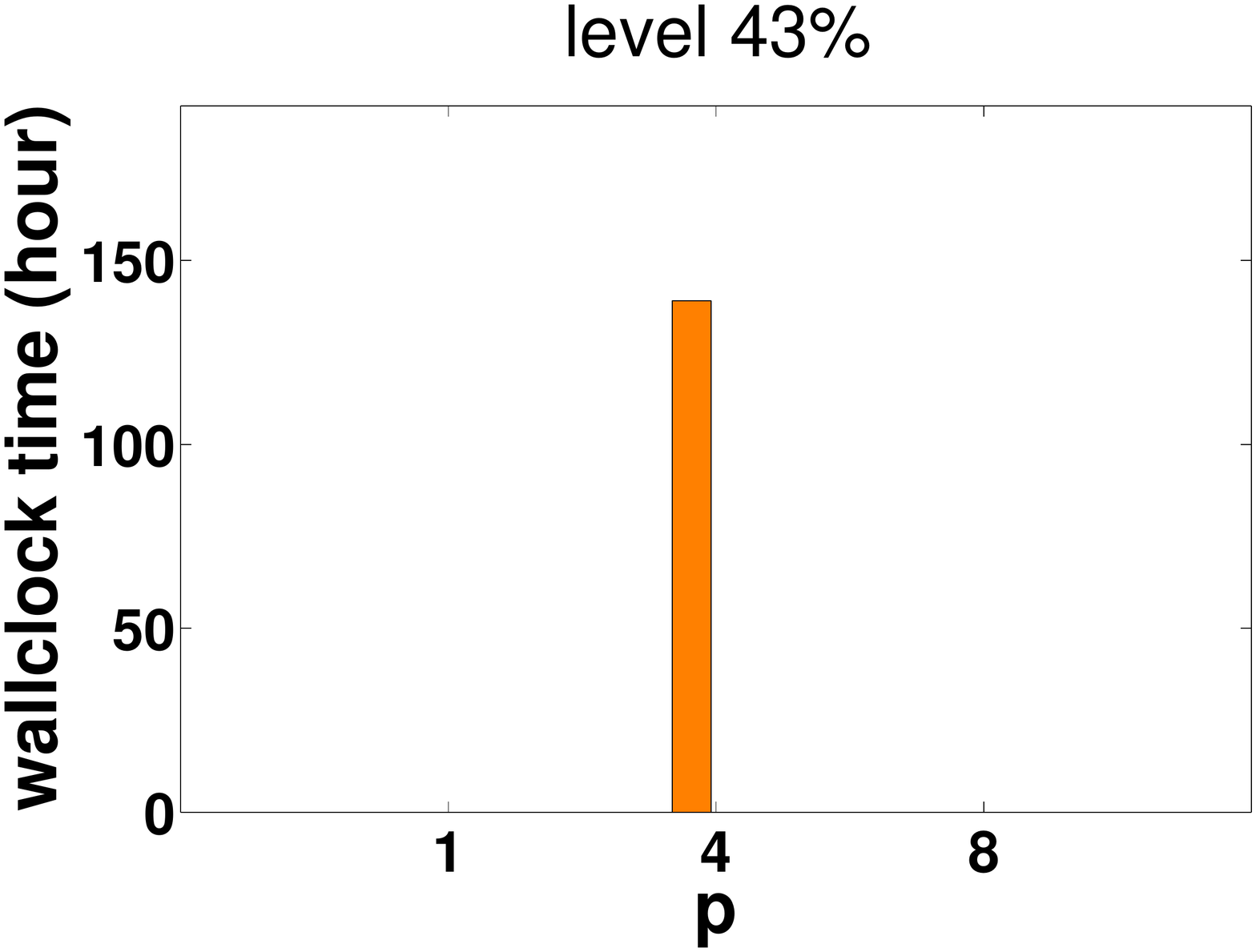}
\caption{The wall clock time needed to achieve the same level of the test error $thr$ as a function of the number of local workers $p$ on the \textit{ImageNet} dataset. From left to right: $thr = \{49\%,47\%,45\%,43\%\}$. Missing bars denote that the method never achieved specified level of test error.}.
\label{fig:ImageNetspeedup}
\end{figure}

In Figure~\ref{fig:CIFARspeedup} and~\ref{fig:ImageNetspeedup}, we summarize the wall clock time needed to achieve the same level of the test error for all the methods in the \textit{CIFAR} and \textit{ImageNet} experiment as a function of the number of local workers $p$. For the \textit{CIFAR} (Figure~\ref{fig:CIFARspeedup}) we examined the following levels: $\{21\%,20\%,19\%,18\%\}$ and for the \textit{ImageNet} (Figure~\ref{fig:ImageNetspeedup}) we examined: $\{49\%,47\%,45\%,43\%\}$. If some method does not appear on the figure for a given test error level, it indicates that this method never achieved this level. For the \textit{CIFAR} experiment we observe that from among \textit{EASGD}, \textit{DOWNPOUR} and \textit{MDOWNPOUR} methods, the \textit{EASGD} method needs less time to achieve a particular level of test error. We observe that with higher $p$ each of these methods does not necessarily need less time to achieve the same level of test error. This seems counter intuitive though recall that the learning rate for the methods is selected based on the smallest achievable test error. For larger $p$ smaller learning rates were selected than for smaller $p$ which explains our results. Meanwhile, the \textit{EAMSGD} method achieves significant speed-up over other methods for all the test error levels. For the \textit{ImageNet} experiment we observe that all methods outperform \textit{MSGD} and furthermore with $p=4$ or $p=8$ each of these methods requires less time to achieve the same level of test error.

\section{Additional pseudo-codes of the algorithms}
\label{sec:addpcode}

\subsubsection{DOWNPOUR pseudo-code}

Algorithm~\ref{alg:asyncD} captures the pseudo-code of the implementation of \textit{DOWNPOUR} used in this paper. Similar to the asynchronous behavior of \textit{EASGD} that we have described in Chapter~\ref{sec:ch_easgd_intro} (Section~\ref{sec:asyncEASGD}), \textit{DOWNPOUR} method also performs several steps of local gradient updates by each worker $i$ before pushing back the accumulated gradients $v^i$ to the center variable. To be more precise, at the beginning of each period, the $i$-th worker reads a new center variable $\tilde{x}$ from the parameter server. Then it performs $\tau$ local \textit{SGD} steps from the new $\tilde{x}$. All the gradients are accumulated (added) and at the end of that period, the total sum $v^i$ is pushed (added) back to the parameter server. The center variable is updated by summing the accumulated gradients from any of the local workers. Notice that we do not use any adaptive learning scheme as having been done in~\cite{2012distbelief}.

\RestyleAlgo{boxruled}
\begin{algorithm}[t]
\caption{DOWNPOUR: Processing by worker $i$ and the master}
\label{alg:asyncD}
\begin{tabular}{l}
\textbf{Input:} learning rate $\eta$, communication period $\tau \in \mathbb{N}$\\
\textbf{Initialize:} $\tilde{x}$ is initialized randomly, $x^i = \tilde{x}$, $v^i = 0$, $t^i = 0\:\:\:\:\:\:\:\:\:\:\:\:\:\:\:\:\:\:\:\:\:\:\:\:\:\:\:\:\:\:\:\:\:\:\:\:\:\:\:\:\:\:\:\:\:\:\:\:\:\:\:\:$\\
\hline
\textbf{Repeat}\\
\:\:\:\:\:\textbf{if} ($\tau $ divides $t^i $) \textbf{then}\\
\:\:\:\:\:\:\:\:\:\:\:$\tilde{x} \:\:\:\!\!\leftarrow \tilde{x} \:\!+ v^i $\\
\:\:\:\:\:\:\:\:\:\:\:$x^i \leftarrow \tilde{x}$\\
\:\:\:\:\:\:\:\:\:\:\:$v^i \leftarrow 0$\\
\:\:\:\:\:\textbf{end}\\
\:\:\:\:\:$ x^i \leftarrow x^i - \eta g^i_{t^i}(x^i)$\\
\:\:\:\:\:$ v^i \leftarrow v^i - \eta g^i_{t^i}(x^i)$\\
\:\:\:\:\:$t^i\:\:\:\:\!\!\!\!\leftarrow \:\!t^i \:\:\!\!+ 1$\\
\textbf{Until forever}
\end{tabular}
\end{algorithm}

\subsubsection{MDOWNPOUR pseudo-code}

Algorithms~\ref{alg:asyncMDworker} and~\ref{alg:asyncMDmaster} capture the pseudo-codes of the implementation of momentum \textit{DOWNPOUR} (\textit{MDOWNPOUR}) used in this paper. Algorithm~\ref{alg:asyncMDworker} describes the behavior of each local worker and Algorithm~\ref{alg:asyncMDmaster} describes the behavior of the master. Note that unlike the~\textit{DOWNPOUR} method, we do not use the communication period $\tau$. This is because the Nesterov's momentum is applied to the center variable. Each worker reads an interpolated variable $\tilde{x}+ \delta v$ from the master, and then sends back the stochastic gradient evaluated at that point.
In case $p=1$, \textit{MDOWNPOUR} is equivalent to the \textit{MSGD} method.

\RestyleAlgo{boxruled}
\begin{algorithm}[h]
\caption{MDOWNPOUR: Processing by worker $i$}
\label{alg:asyncMDworker}
\begin{tabular}{l}
\textbf{Initialize:} $x^i = \tilde{x}\:\:\:\:\:\:\:\:\:\:\:\:\:\:\:\:\:\:\:\:\:\:\:\:\:\:\:\:\:\:\:\:\:\:\:\:\:\:\:\:\:\:\:\:\:\:\:\:\:\:\:\:\:\:\:\:\:\:\:\:\:\:\:\:\:\:\:\:\:\:\:\:\:\:\:\:\:\:\:\:\:\:\:\:\:\:\:\:\:\:\:\:\:\:\:\:\:\:\:\:\:\:\:\:\:\:\:\:\:\:\:\:\:\:\:\:\:\:\:\:\:\:\:\:\:\:\:\:\:\:\:$\\
\hline
\textbf{Repeat}\\
\:\:\:\:\:\textsf{Receive} $\tilde{x}+ \delta v$ \textsf{from the master}: $x^i \leftarrow \tilde{x}+ \delta v$\\
\:\:\:\:\:\textsf{Compute gradient} $g^i = g^i(x^i)$\\
\:\:\:\:\:\textsf{Send} $g^i$ \textsf{to the master}\\
\textbf{Until forever}
\end{tabular}
\end{algorithm}

\RestyleAlgo{boxruled}
\begin{algorithm}[h]
\caption{MDOWNPOUR: Processing by the master}
\label{alg:asyncMDmaster}
\begin{tabular}{l}
\textbf{Input:} learning rate $\eta$, momentum term $\delta$\\
\textbf{Initialize:} $\tilde{x}$ is initialized randomly, $v^i = 0$, $\:\:\:\:\:\:\:\:\:\:\:\:\:\:\:\:\:\:\:\:\:\:\:\:\:\:\:\:\:\:\:\:\:\:\:\:\:\:\:\:\:\:\:\:\:\:\:\:\:\:\:\:\:\:\:\:\:\:\:\:\:\:\:\:\:\:\:\:\:\:\:\:\:\:\:\:\:\:\:$\\
\hline
\textbf{Repeat}\\
\:\:\:\:\:\textsf{Receive} $g^i$\\
\:\:\:\:\:$v \leftarrow \delta v - \eta g^i$\\
\:\:\:\:\:$\tilde{x} \leftarrow \tilde{x} + v$\\
\:\:\:\:\:\textsf{Send} $\tilde{x}+ \delta v$\\
\textbf{Until forever}
\end{tabular}
\end{algorithm}

\newpage
\chapter{The Limit in Speedup}
\label{sec:ch_limit}

This chapter studies the limitation in speedup of several stochastic optimization methods:
the mini-batch \textit{SGD}, the momentum \textit{SGD} and the \textit{EASGD} method.
In Section~\ref{sec:additive_noise}, we study first the asymptotic phase of these methods using an additive noise model. The continuous-time SDE (stochastic differential equation) approximation of its \textit{SGD} update is an Ornstein Uhlenbeck process.
Then we study the initial phase of these methods using a multiplicative noise model in Section~\ref{sec:multiplicative_noise}. The continuous-time SDE approximation of its \textit{SGD} update is a Geometric Brownian motion. In Section~\ref{sec:non-convex}, we study the stability of the critical points of a simple non-convex problem and discuss when the \textit{EASGD} method can get trapped by a saddle point.

\section{Additive noise}
\label{sec:additive_noise}
We (re-)study the simple additive noise model: one-dimensional quadratic objective with Gaussian noise (as in Section~\ref{sec:onedim}). The objective function evaluated at state $x \in \mathbb{R}$ is defined to be the average loss of the quadratic form $(h x-\xi)^2$, i.e.
\begin{equation}
\min_{x \in \mathbb{R}} \mathbb{E} [ (h x - \xi)^2].
\label{eq:pb1}
\end{equation}
Here $h>0$ is a scalar, and the expectation is taken over the random variable $\xi$, which follows a Gaussian distribution. For simplicity, we assume further that $\xi$ is zero mean and has a constant variance $\sigma^2>0$.

\subsection{\textit{SGD} with mini-batch}
The update rule for the \textit{SGD} method for solving the problem in Equation~\ref{eq:pb1} is
\begin{equation}
x_{t+1} = x_{t} - \eta (h x_{t} - \xi_t),
\label{eq:sgd1}
\end{equation}
where $x_0$ is the initial starting point.

Notice that in the continuous-time limit, i.e. for small $\eta$, we can approximate the process in Equation~\ref{eq:sgd1} by an Ornstein Uhlenbeck process~\cite{DBLP:journals/corr/LiTE15} as follows,
\begin{equation*}
dX(t) = - h X(t) dt + \sqrt{\eta} \sigma dB(t).
\end{equation*}

In the discrete-time case,
the bias term (the first-order moment $\mathbb{E} x_t$) in Equation~\ref{eq:sgd1} will decrease at a linear rate $1-\eta h$, i.e.
\begin{equation*}
\mathbb{E} x_{t+1} = (1-\eta h) \mathbb{E} x_{t}.
\end{equation*}
The second-order moment $\mathbb{E} x_t^2$ changes as
\begin{equation}
\mathbb{E} x^2_{t+1} = (1-\eta h)^2 \mathbb{E} x_{t}^2 + \eta^2 \sigma^2.
\label{eq:sgd_l2}
\end{equation}
Thus the variance will increase as
\begin{align*}
\mathbb{V} x_{t+1} &= \mathbb{E} x^2_{t+1} - (\mathbb{E} x_{t+1})^2
= (1-\eta h)^2 \mathbb{E} x_{t}^2 + \eta^2 \sigma^2 - (1-\eta h)^2 (\mathbb{E} x_{t})^2 \\
&= (1-\eta h)^2 \mathbb{V} x_{t} + \eta^2 \sigma^2,
\end{align*}
with $\mathbb{V} x_{0} = 0$. As $t \to \infty$, we get $\mathbb{V} x_{\infty} = \frac{\eta^2}{1-(1-\eta h)^2}\sigma^2$. If we use mini-batch of size $p$, the variance of the noise $\xi_t$ is then reduced by $p$,
and this asymptotic variance becomes $\frac{\eta^2}{1-(1-\eta h)^2} \frac{\sigma^2}{p}$.
The convergence rate of the bias term, $1-\eta h$, is however not improved by increasing the mini-batch size $p$.

\subsection{Momentum \textit{SGD}}
\label{sec:msgd_lin}

We now study the update rule for the \textit{MSGD} (Nesterov's momentum) method for solving Equation~\ref{eq:pb1},
\begin{equation}
\begin{array}{r@{}l}
v_{t+1} &{}= \delta v_t - \eta (h (x_t + \delta v_t) - \xi_t), \\
x_{t+1} &{}= x_{t} + v_{t+1},
\end{array}
\label{eq:msgd1}
\end{equation}
where $x_0$ is the initial starting point and $v_0=0$ is the initial velocity set to zero.

Let $\delta_h = \delta (1-\eta h),\eta_h = \eta h$, then Equation~\ref{eq:msgd1}
is equivalent to
\begin{equation*}
\begin{array}{r@{}l}
v_{t+1} &{}= \delta_h v_t - \eta_h x_t + \eta \xi_t, \\
x_{t+1} &{}= x_{t} + v_{t+1}.
\end{array}
\end{equation*}
In case that $\delta_h$ is chosen independently of $h$, the update is no different to the heavy ball method~\cite{polyak1987introduction}. But the momentum term $\delta_h$ above equals $\delta (1-\eta h)$, thus it implicitly depends on $h$. As we usually choose $\delta$ to be smaller than one, $\delta_h$ is upper bounded by $1-\eta h$. This fact saves us from the variance explosion as $\delta$ tends to one.

More precisely, the second-order moment equation can be computed as follows,
\begin{equation}
\begin{array}{r@{}l}
v_{t+1}^2 &{}= (\delta_h v_t - \eta_h x_t + \eta \xi_t)^2 \\
		 &{}= \delta_h^2 v_t^2 + \eta_h^2 x_t^2 +
				\eta^2 \xi_t^2 - 2 \delta_h \eta_h v_t x_t +
					2 \delta_h \eta v_t \xi_t - 2 \eta_h \eta x_t \xi_t \\
x_{t+1}^2 &{}= (x_{t} + v_{t+1})^2 = x_{t}^2 + 2 v_{t+1}x_{t} + v_{t+1}^2 \\
v_{t+1} x_t &{}= (\delta_h v_t - \eta_h x_t + \eta \xi_t) x_t
				= \delta_h v_t x_t - \eta_h x_t^2 + \eta x_t \xi_t \\
v_{t+1} x_{t+1} &{}= v_{t+1} (x_{t} + v_{t+1}) = v_{t+1} x_t + v_{t+1}^2.
\end{array}
\label{eq:msgd2}
\end{equation}

Now taking expectation on both sides of the Equation~\ref{eq:msgd2}, we obtain the following
recursive relation
\begin{gather}
\begin{pmatrix}
\mathbb{E} v^2_{t+1} \\
\mathbb{E} v_{t+1} x_{t+1} \\
\mathbb{E} x^2_{t+1} \\
\end{pmatrix} =
\underbrace{
\begin{pmatrix}
\delta_h^2 & -2 \delta_h \eta_h & \eta_h^2 \\
\delta_h^2 & \delta_h (1 - 2 \eta_h) & -\eta_h (1-\eta_h) \\
\delta_h^2 & 2 \delta_h (1-\eta_h) & (1-\eta_h)^2
\end{pmatrix}
}_M
\begin{pmatrix}
\mathbb{E} v^2_{t} \\
\mathbb{E} v_{t} x_{t} \\
\mathbb{E} x^2_{t} \\
\end{pmatrix}
+
\begin{pmatrix}
\eta^2 \sigma^2 \\
\eta^2 \sigma^2 \\
\eta^2 \sigma^2 \\
\end{pmatrix}
.
\label{eq:msgd_moment2}
\end{gather}

To see the asymptotic behavior, assume $\underline{v}^2_\infty = \lim_{t \to \infty } \mathbb{E} v^2_{t}$,
$\underline{vx}_\infty = \lim_{t \to \infty} \mathbb{E} v_{t} x_{t}$,
and $\underline{x}^2_\infty = \lim_{t \to \infty } \mathbb{E} x^2_{t} $. Then by solving
\begin{gather*}
\begin{pmatrix}
\underline{v}^2_\infty \\
\underline{vx}_\infty \\
\underline{x}^2_\infty \\
\end{pmatrix} =
\begin{pmatrix}
\delta_h^2 & -2 \delta_h \eta_h & \eta_h^2 \\
\delta_h^2 & \delta_h (1 - 2 \eta_h) & -\eta_h (1-\eta_h) \\
\delta_h^2 & 2 \delta_h (1-\eta_h) & (1-\eta_h)^2
\end{pmatrix}
\begin{pmatrix}
\underline{v}^2_\infty \\
\underline{vx}_\infty \\
\underline{x}^2_\infty \\
\end{pmatrix}
+
\begin{pmatrix}
\eta^2 \sigma^2 \\
\eta^2 \sigma^2 \\
\eta^2 \sigma^2 \\
\end{pmatrix}
,
\end{gather*}
we obtain
\begin{align}
\underline{v}^2_\infty &= \frac{2}{(1-\delta_h)(2(1+\delta_h)-\eta_h)} \eta^2 \sigma^2 \nonumber , \\
\underline{vx}_\infty &= \frac{1}{(1-\delta_h)(2(1+\delta_h)-\eta_h)} \eta^2 \sigma^2 \nonumber , \\
\underline{x}^2_\infty &= \frac{1+\delta_h}{\eta_h (1-\delta_h)(2(1+\delta_h)-\eta_h)} \eta^2 \sigma^2.
\label{eq:msgd_x2inf}
\end{align}

From Equation~\ref{eq:msgd_x2inf}, we see that for the asymptotic variance $\underline{x}^2_\infty$ to be strictly positive, we should assume $-1 < \delta_h<1$ and $0 < \eta_h < 2(1+\delta_h)$. Moreover, compared to
the asymptotic variance of \textit{SGD} (the case that $\delta = 0$), we can check, for example, that in the region $\eta_h \in (0,1)$ and $\delta_h \in (0,1)$, the asymptotic variance of \textit{MSGD} is always larger.

On the other hand, the above condition $-1 < \delta_h<1$ and $0 < \eta_h < 2(1+\delta_h)$ is also the condition for the matrix $M$ in Equation~\ref{eq:msgd_moment2} to remain (strictly) stable, i.e. the
largest absolute eigenvalue is (strictly) smaller than one. In fact, we have three eigenvalues for the matrix $M$ as follows,
\begin{equation}
\begin{array}{r@{}l}
z_1 &{}= \delta_h , \\
z_2 &{}= \frac{ (1-\eta_h)^2 - 2 \eta_h \delta_h + \delta_h^2 - \sqrt{((1-\eta_h)^2 - 2 \eta_h \delta_h + \delta_h^2)^2 - 4 \delta_h^2}}{2}, \\
z_3 &{}= \frac{ (1-\eta_h)^2 - 2 \eta_h \delta_h + \delta_h^2 + \sqrt{((1-\eta_h)^2 - 2 \eta_h \delta_h + \delta_h^2)^2 - 4 \delta_h^2}}{2}.
\end{array}
\label{eq:msgd_z}
\end{equation}
Let $2 b = z_2 + z_3 = (1-\eta_h)^2 - 2 \eta_h \delta_h + \delta_h^2$ and $c = z_2 z_3 = \delta_h^2$.
Then $z_2$ and $z_3$ are the two roots (in $z$) of $z^2 - 2 b z + c = 0$.
In general, if $z_2$, $z_3$ are real-valued, i.e. $b^2 - c > 0$, then we have two cases:
\begin{itemize}
	\item $b> \sqrt{c}$ implies $z_3 = b + \sqrt{b^2 - c} > \sqrt{c} > z_2 = b - \sqrt{b^2 - c} > 0$,
	\item $b < - \sqrt{c}$ implies $z_2 = b - \sqrt{b^2 - c} < - \sqrt{c} < z_3 = b + \sqrt{b^2 - c} < 0 $.
\end{itemize}
On the other hand, if $z_2$, $z_3$ are not real-valued, i.e. $b^2 - c \le 0$, then $|z_3| = |z_1| = | z_2 | = \sqrt{c} $. Recall that $|z_1| = |\delta_h| = \sqrt{c}$.

We see thus for the (strict) stability of the matrix $M$ in Equation~\ref{eq:msgd_moment2}, we need $\sqrt{c} = |\delta_h| < 1$ and $|z_3|< 1$ because the condition $b < - \sqrt{c}$ is never satisfied for any real $\delta_h$. The condition for $|z_3| = b + \sqrt{b^2-c} < 1$
is that $2b-c < 1$, i.e. $0 < \eta_h < 2(1+\delta_h)$.

Moreover, as in~\cite{DBLP:journals/corr/LiTE15}, we can try to minimize $|z_3|$ with respect to $\delta$ such that the rate of convergence of the second order moment is maximized for a given $\eta$. We shall prove that the minimal $|z_3|$ is achieved at $\delta_h = (\sqrt{\eta_h}-1)^2$. In fact, it happens when $b = \sqrt{c}$,
i.e. the optimal rate is at the edge where the eigenvalues transit from real-valued to the complex-valued.
Notice that $b=\sqrt{c}$ gives us two positive solutions: $\delta_h = (\sqrt{\eta_h}-1)^2$ and $\delta_h = (\sqrt{\eta_h}+1)^2$. Since $b =\frac{1}{2} [ (\delta_h - \eta_h)^2 + 1 - 2 \eta_h ]$ is a quadratic function
in $\delta_h$, the fact that $b=\sqrt{c}$ has two positive solutions means that the quadratic function intersects twice with the line $\delta_h$ in the first orthant. If $\eta_h \ge 1/4$, the first intersection point is to the left of the minimum of $b$, and the second intersection point is to the right of the minimum, i.e. $ (\sqrt{\eta_h}-1)^2 \le \eta_h < (\sqrt{\eta_h}+1)^2$. But if $0 < \eta_h < 1/4$, both
intersection points will be both to the right of the minimum of $b$, i.e. $\eta_h < (\sqrt{\eta_h}-1)^2 < (\sqrt{\eta_h}+1)^2$. Nevertheless, in either case, we can show $b > \sqrt{c}$ whenever $\delta_h < (\sqrt{\eta_h}-1)^2$. Thus, in the range $\delta_h < (\sqrt{\eta_h}-1)^2$, we have $z_3 > \sqrt{c} > z_2 > 0$. We thus only need to find the minimum of $z_3$ in the range $\delta_h \in (-1,(\sqrt{\eta_h}-1)^2]$, because $|z_3|=\delta_h$ is monotonically increasing in the range $1 > \delta_h > (\sqrt{\eta_h}-1)^2$. We now show that the minimal value of $z_3$ is $(\sqrt{\eta_h}-1)^2$ in this range. In fact, if $\eta_h \ge 1/4$, $b$ is monotonically decreasing in this range, thus $z_3 \ge b$ reaches its minimum at $\delta_h = (\sqrt{\eta_h}-1)^2$ with $z_3 = b$. If $0 < \eta_h < 1/4$, $b$ firstly decreases for $\delta_h < \eta_h$, then increases for $\eta_h \le \delta_h \le (1-\sqrt{\eta_h})^2$. We show that $z_3$ is still monotonically decreasing in these two ranges of $\delta_h$. We check whether $\frac{ \partial z_3}{\partial \delta_h} = (\delta_h - \eta_h) ( 1 + \frac{b}{\sqrt{b^2- c}} ) - \delta_h \frac{1}{\sqrt{b^2 - c}} < 0$ holds. For $\delta_h < \eta_h$, this is equivalent to $ z_3 > \frac{\delta_h}{\delta_h-\eta_h}$; for $\eta_h < \delta_h \le (1-\sqrt{\eta_h})^2$, this is equivalent to $z_3 < \frac{\delta_h}{\delta_h-\eta_h}$. The former one is true in the range $0 < \delta_h < \eta_h $, because $z_3 $ is always positive. One can check that this is still true in the range $-1 < \delta_h <0$. The latter one is equivalent to $b+ \sqrt{b^2 - c} < \frac{\delta_h}{\delta_h - \eta_h}$. Taking square on both sides, we can check that $b < \frac{\delta_h}{\delta_h - \eta_h}$ and $b^2 -c < (\frac{\delta_h}{\delta_h - \eta_h} - b)^2$, based on $ 0 < \delta_h - \eta_h < 1 - 2 \sqrt{ \eta_h} < 1$.

\begin{figure}[p]
\center
\includegraphics[trim=0in 0in 0in 0in,clip,width = 6in]{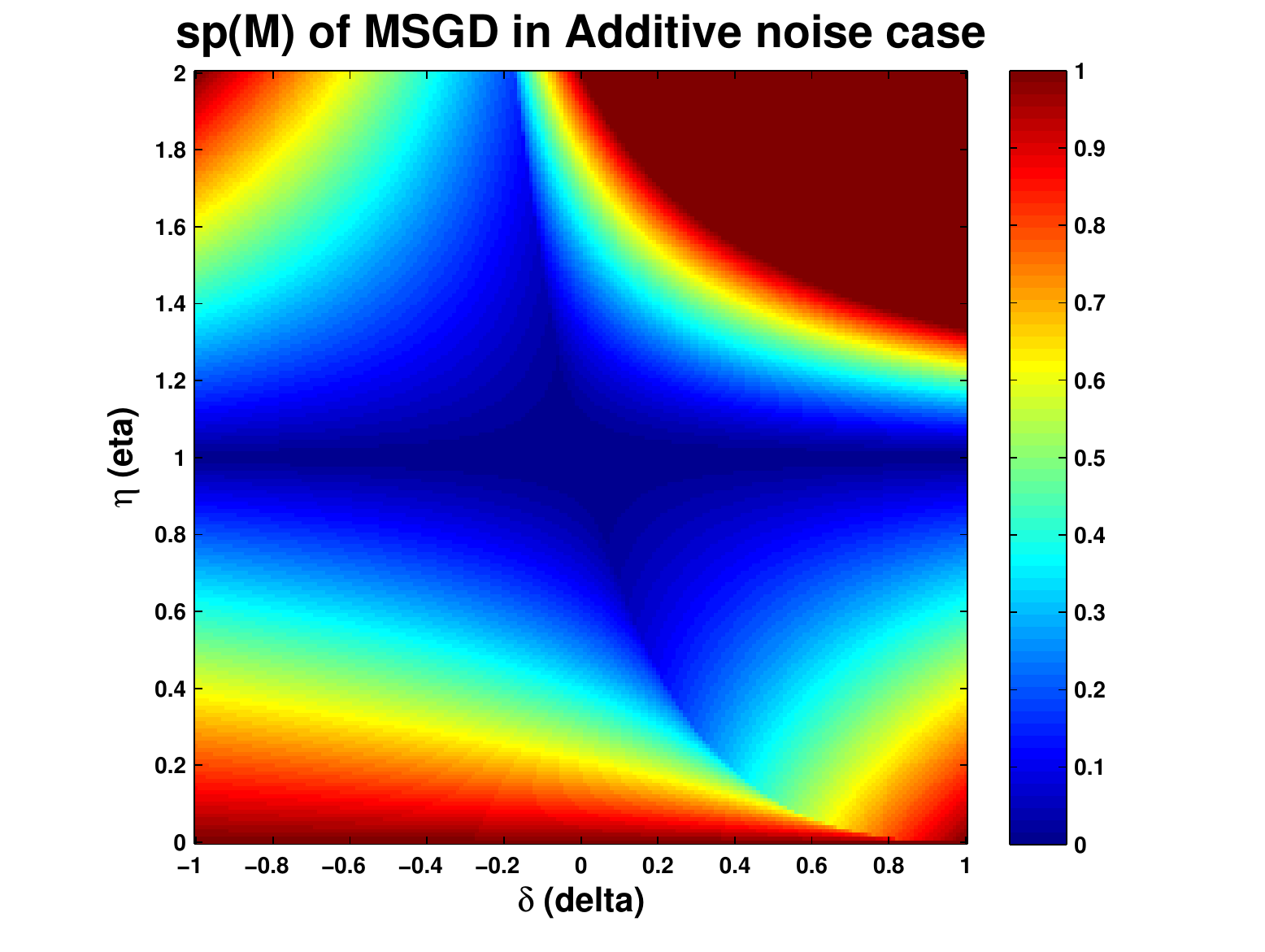}
\caption{The largest absolute eigenvalue of the matrix $M$ (sp($M$)) in Equation~\ref{eq:msgd_moment2} as a function of the learning rate $\eta \in (0,2)$ and the momentum rate $\delta \in (-1,1)$. $h=1$.}
\label{fig:ch5_lin_msgd_sp}
\end{figure}

Thus we conclude that for a fixed $\eta_h$ such that $0 < \eta_h < 2(1+\delta_h)$, the minimal $|z_3|$ over $-1 < \delta_h<1$ is obtained
at \[\delta_h = (\sqrt{\eta_h}-1)^2,\]
with the minimal value $z_3= b + \sqrt{b^2-c}= b = \sqrt{c} = \delta_h = (\sqrt{\eta_h}-1)^2$.
Compared to the rate $(1-\eta_h)^2$ in the \textit{SGD} case (Equation~\ref{eq:sgd_l2}),
\textit{MSGD} can indeed help if $\eta_h$ is small enough (usually $\eta_h = \eta h $ is close to
the inverse of the condition number of the Hessian in higher dimensional case).
To check our above reasoning, we have computed numerically the spectral norm of the matrix $M$, sp($M$),
which is given in Figure~\ref{fig:ch5_lin_msgd_sp}. Note that when $\eta_h>1$, we have the optimal momentum rate being negative, i.e. $\delta = \frac{ (\sqrt{\eta_h}-1)^2}{(1-\eta_h)} < 0$.

Perhaps the most special behavior for \textit{MSGD} is that when the momentum rate $\delta$ gets
close to $1$, the asymptotic variance of $\underline{x}^2_\infty$ in
Equation~\ref{eq:msgd_x2inf} still remains bounded,
i.e. $\underline{x}^2_\infty = \frac{1+1 - \eta_h}{\eta_h^2 (2(1+1 - \eta_h)-\eta_h)} \eta^2 \sigma^2 =
\frac{2- \eta_h}{4 - 3 \eta_h} \frac{\sigma^2}{h^2} $ for $\delta=1$ (or $\delta_h=1-\eta_h$).
This contrasts to the behavior of the heavy ball method, whose asymptotic variance tends to infinity as $\delta \to 1$~\cite{DBLP:journals/corr/LiTE15}.

\subsection{\textit{EASGD} and \textit{EAMSGD}}
We can now study the moment equation for the \textit{EASGD} and \textit{EAMSGD} method. For \textit{EASGD}, we have the following update rules
\begin{equation}
\begin{array}{r@{}l}
x^i_{t+1} &{}= x^i_t - \eta (h x^i_t - \xi_t^i) + \alpha( \tilde{x}_t - x^i_t), \\
\tilde{x}_{t+1} &{}= \tilde{x}_{t} + \beta ( \frac{1}{p} \sum_{i=1}^p x_t^i - \tilde{x}_t ).
\end{array}
\label{eq:easgd1}
\end{equation}
Denote $y_t = \frac{1}{p} \sum_{i=1}^p x_t^i $ and $\xi_t = \frac{1}{p} \sum_{i=1}^p \xi^i_t$, then Equation~\ref{eq:easgd1} can be reduced to
\begin{equation}
\begin{array}{r@{}l}
y_{t+1} &{}= y_t - \eta (h y_t - \xi_t) + \alpha( \tilde{x}_t - y_t), \\
\tilde{x}_{t+1} &{}= \tilde{x}_{t} + \beta ( y_t - \tilde{x}_t ).
\end{array}
\label{eq:easgd2}
\end{equation}

Similar to \textit{MSGD}, the second-order moment equation for Equation~\ref{eq:easgd2} is as follows,
\begin{equation}
\begin{array}{r@{}l}
y_{t+1}^2 &{}= ( (1-\eta h - \alpha) y_t + \alpha \tilde{x}_t + \eta \xi_t )^2 \\
		 &{}= (1-\eta h - \alpha)^2 y_t^2 + \alpha^2 \tilde{x}_t^2 + \eta^2 \xi_t^2 \\
				&{}\quad\quad + 2 \alpha (1-\eta h -\alpha) y_t \tilde{x}_t
				+ 2 \eta (1-\eta h -\alpha) y_t \xi_t + 2\alpha \eta \tilde{x}_t \xi_t , \\
y_{t+1} \tilde{x}_{t+1} &{}=
	((1-\eta h -\alpha) y_t + \alpha \tilde{x}_t + \eta \xi_t)( (1-\beta) \tilde{x}_t + \beta y_t) \\
	&{}=
	(1 - \eta h - \alpha) \beta y_t^2 + (1-\eta h -\alpha)(1-\beta) y_t \tilde{x}_t
		+ \alpha \beta y_t \tilde{x}_t + \alpha (1-\beta) \tilde{x}^2_t \\
	&{}\quad\quad +\eta \xi_t ( (1-\beta) \tilde{x}_t + \beta y_t ), \\
\tilde{x}_{t+1}^2 &{}= (1-\beta)^2 \tilde{x}_t^2 + \beta^2 y_t^2 + 2 \beta (1-\beta) y_t \tilde{x}_t . \\
\end{array}
\label{eq:easgd3}
\end{equation}

Taking expectation on both sides of the Equation~\ref{eq:easgd3}, we get
\begin{gather}
\begin{pmatrix}
\mathbb{E} y^2_{t+1} \\
\mathbb{E} y_{t+1} \tilde{x}_{t+1} \\
\mathbb{E} \tilde{x}^2_{t+1} \\
\end{pmatrix} =
\underbrace{
\begin{pmatrix}
(1-\eta h -\alpha)^2 & 2 \alpha (1 - \eta h - \alpha) & \alpha^2 \\
(1-\eta h -\alpha) \beta & (1 - \eta h - \alpha) (1 - \beta) + \alpha \beta & \alpha (1 - \beta) \\
\beta^2 & 2 \beta (1 - \beta) & (1-\beta)^2
\end{pmatrix}
}_M
\begin{pmatrix}
\mathbb{E} y^2_{t} \\
\mathbb{E} y_{t} \tilde{x}_{t} \\
\mathbb{E} \tilde{x}^2_{t} \\
\end{pmatrix}
+
\begin{pmatrix}
\eta^2 \frac{\sigma^2}{p} \\
0 \\
0 \\
\end{pmatrix}
.
\label{eq:easgd_moment2}
\end{gather}

Similar to the analysis of \textit{MSGD}, we get the following asymptotic variance,
\begin{align}
\underline{y}^2_\infty &= \lim_{t \to \infty } \mathbb{E} y^2_{t} =
\frac{ (2-\beta) (1-\beta) \eta_h + \beta (2 - \alpha - \beta) }
{ \eta_h [(2-\beta)(2-\eta_h)-2\alpha][ \alpha + \beta + \eta_h (1 - \beta) ] }
\frac{\eta^2 \sigma^2}{p} , \label{eq:easgd_asymy2} \\
\underline{y \tilde{x}}_\infty &= \lim_{t \to \infty} \mathbb{E} y_{t} \tilde{x}_{t} =
\frac{\beta ( (2-\beta)(1-\eta_h) - \alpha ) }
{ \eta_h [(2-\beta)(2-\eta_h)-2\alpha][ \alpha + \beta + \eta_h (1 - \beta) ] }
\frac{\eta^2 \sigma^2}{p}, \nonumber \\
\underline{\tilde{x}}^2_\infty &= \lim_{t \to \infty } \mathbb{E} \tilde{x}^2_{t} =
\frac{ - \beta (1-\beta) \eta_h + \beta (2 - \alpha - \beta) }
{ \eta_h [(2-\beta)(2-\eta_h)-2\alpha][ \alpha + \beta + \eta_h (1 - \beta) ] }
\frac{\eta^2 \sigma^2}{p} . \label{eq:easgd_asymx2}
\end{align}

For the above asymptotic variance to be positive, in particular $\underline{x}^2_\infty$, we can assume
\begin{equation}
\begin{array}{r@{}l}
\eta_h > 0 , \\
\beta > 0 , \\
(2-\beta)(2-\eta_h)-2\alpha > 0, \\
\frac{ 2 - \alpha - \beta - \eta_h + \beta \eta_h}{\alpha + \beta + \eta_h (1 - \beta)} > 0. \\
\end{array}
\label{eq:easgd_cond}
\end{equation}

Moreover, if $0< \beta < 1$, the asymptotic variance of the center variable $\underline{\tilde{x}}^2_\infty$ is strictly smaller than that of the spatial average $\underline{y}^2_\infty$ (by comparing Equation~\ref{eq:easgd_asymy2} and~\ref{eq:easgd_asymx2}). Interestingly,
if $\beta > 1$, it becomes strictly bigger.

Note that it's still not very clear whether Equation~\ref{eq:easgd_cond} is the necessary and sufficient condition for the matrix $M$ (in Equation~\ref{eq:easgd_moment2}) to be (strictly) stable. However, if we look back to the earlier result in Section~\ref{sec:onedim} about the condition in Equation~\ref{eq:condition}, they are
nearly the same, except maybe for the last formula. The last formula in Equation~\ref{eq:easgd_cond} reads $ 0 < \alpha + \beta + \eta_h - \beta \eta_h < 2 $. % (or ).
The left inequality implies $ \beta \eta_h < \alpha + \beta + \eta_h $. Based on the third formula in Equation~\ref{eq:easgd_cond}, we have thus $\alpha + \beta + \eta_h < 2 + \frac{\beta \eta_h}{2} < 2 + \frac{\alpha + \beta + \eta_h} {2} $. This gives us the last formula for the condition in Equation~\ref{eq:condition}, which is $\alpha + \beta + \eta_h < 4$. Conversely,
assuming $\alpha > 0$ (as assumed in Section~\ref{sec:onedim} for $\beta = p \alpha$), then
one can check that $\alpha + \beta + \eta_h < 4$ implies the last formula in Equation~\ref{eq:easgd_cond}, which
also reads $\alpha - 1 < (\eta_h -1)(\beta-1) < \alpha + 1$.

Perhaps the most striking observation is that the optimal $\alpha$ such that
the convergence rate (of the moment Equation~\ref{eq:easgd_moment2}) is maximized turns out to be negative, given $\eta_h>0$ and $\beta>0$ fixed.

The three eigenvalues of the matrix $M$ in Equation~\ref{eq:easgd_moment2} are
\begin{equation}
\begin{array}{r@{}l}
z_1 &= -\alpha + (1-\eta_h)(1-\beta), \\
z_2 &= b - \sqrt{b^2-c}, \\
z_3 &= b + \sqrt{b^2-c},
\end{array}
\label{eq:easgd_eigs}
\end{equation}
where $b = \frac{1}{2} [(\alpha - (1- \eta_h - \beta) )^2 + 1 - 2 \beta \eta_h ]$,
$c = [ \alpha - (1-\eta_h)(1- \beta) ]^2 = z_1^2$. Denote $\alpha' = z_1 = -\alpha + (1-\eta_h)(1-\beta)$,
then $b =\frac{1}{2} [(\alpha' - \eta_h \beta )^2 + 1 - 2 \beta \eta_h ]$ and $c = (\alpha')^2$.
Using exactly the same analysis and the result from the \textit{MSGD} case, we get that the
minimal $|z_3|$ over $-1 < \alpha' <1$ is obtained at
\[\alpha' = (\sqrt{\beta \eta_h}-1)^2,\]
which is equivalent to
\begin{equation}
\alpha = - ( \sqrt{\beta} - \sqrt{\eta_h})^2.
\label{eq:easgd_danger}
\end{equation}

\begin{figure}[p]
\center
\includegraphics[trim=0in 0in 0in 0in,clip,width = 6in]{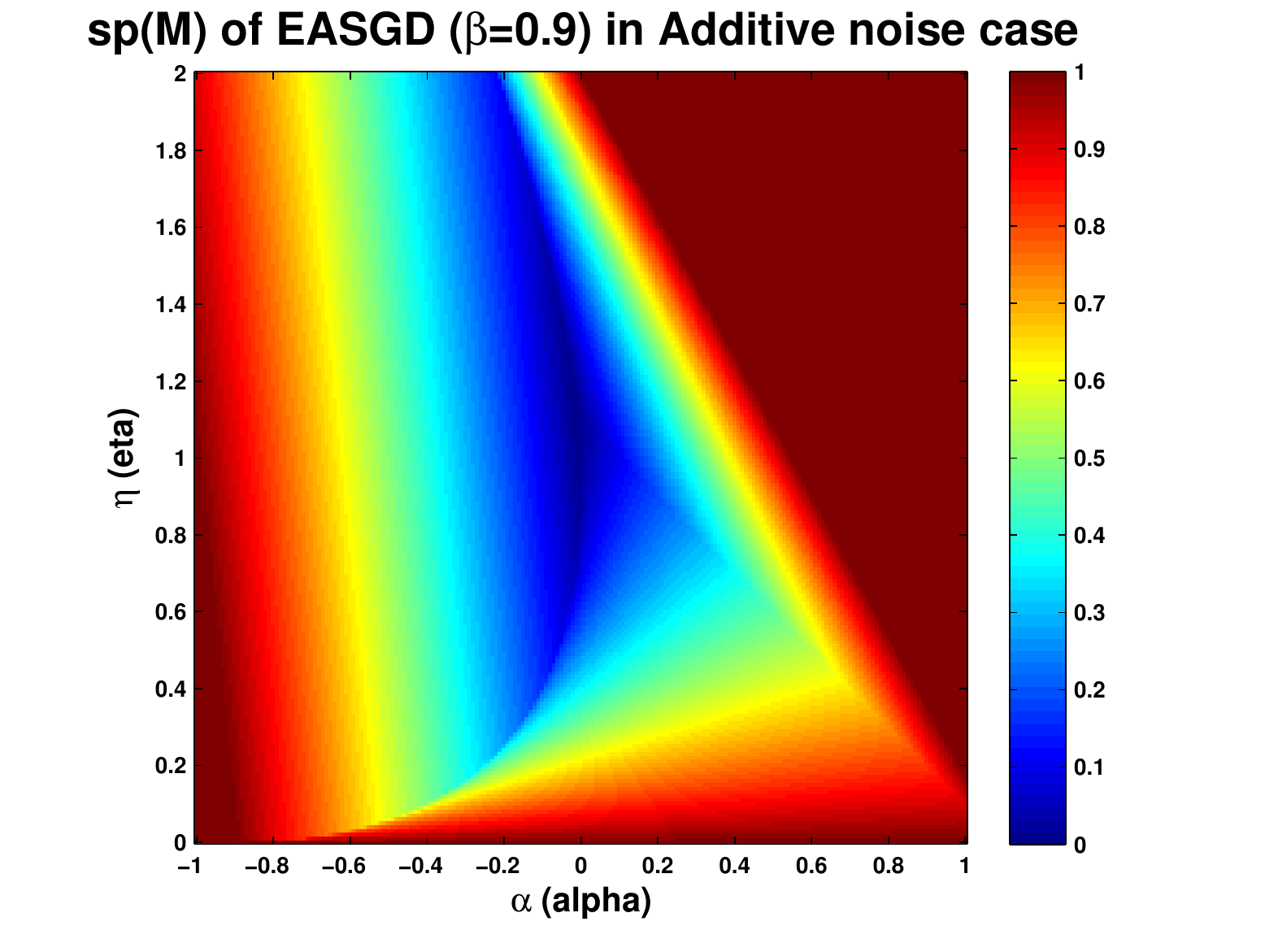}
\caption{The largest absolute eigenvalue of the matrix $M$ in Equation~\ref{eq:easgd_moment2} as a function of the learning rate $\eta \in (0,2)$ and the moving rate $\alpha \in (-1,1)$. $h=1$ and $\beta=0.9$.}
\label{fig:ch5_lin_easgd_sp}
\end{figure}

\begin{figure}[p]
\center
\includegraphics[trim=0in 0in 0in 0in,clip,width = 6in]{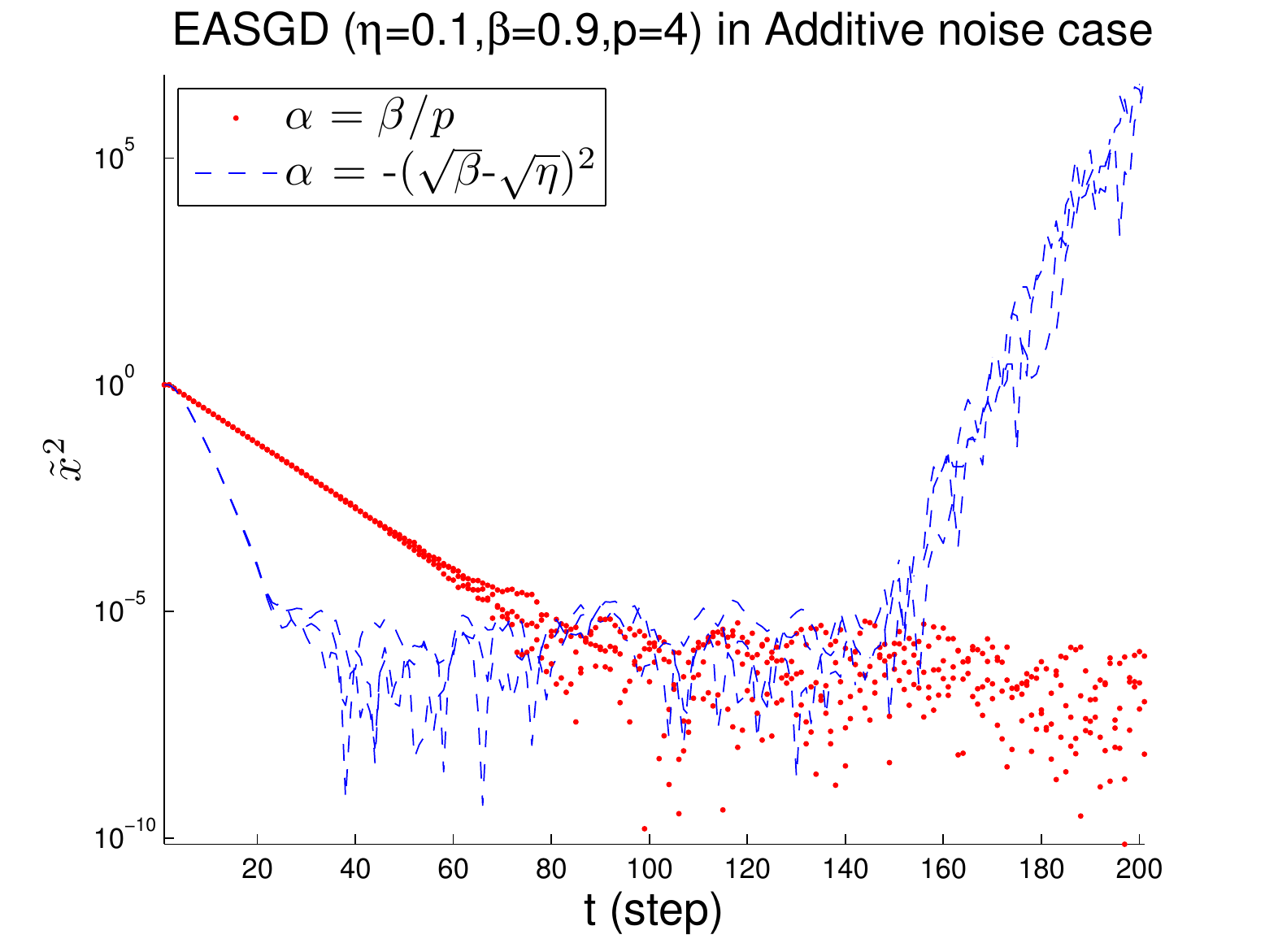}
\caption{Three independent simulations of \textit{EASGD} using the elastic averaging $\alpha=\beta/p$ and the optimal $\alpha$ given in Equation~\ref{eq:easgd_danger}. The x-axis is the time step $t$ in the \textit{EASGD} updates of Equation~\ref{eq:easgd1}. The y-axis is the squared distance of the center variable to the optimum zero, i.e. $\tilde{x}_t^2$. We have chosen $h=1$, $\sigma=10^{-2}$, $p=4$, $\eta=0.1$ and $\beta=0.9$.}
\label{fig:ch5_lin_easgd_beta09}
\end{figure}

The situation that the coupling constant $\alpha$ being negative, while $\beta$ being positive suggests
a very different perspective to understand \textit{EASGD}. It seems that our earlier condition $\beta = p \alpha$ is unnecessary and is sub-optimal in this case. To check our above results, we have computed numerically the spectral norm of the matrix $M$ for a fixed $\beta=0.9$, which is given in Figure~\ref{fig:ch5_lin_easgd_sp}.
However, we are in danger this time if we were to simulate \textit{EASGD} using the optimal $\alpha$ given in Equation~\ref{eq:easgd_danger}. In Figure~\ref{fig:ch5_lin_easgd_beta09}, we illustrate
an unstable behavior in such optimal case. The reason is that our above analysis is based on the reduced Equation~\ref{eq:easgd2}, rather than the original Equation~\ref{eq:easgd1}.

In the original Equation~\ref{eq:easgd1}, we have the following form of the drift matrix
\begin{equation}
M_p=
\begin{bmatrix}
1-\alpha-\eta_h & 0 & ... & 0 & \alpha \\
0 & 1-\alpha-\eta_h& 0 & ... & \alpha \\
... & 0 & ... &0 & ... \\
0 & ... &0 & 1-\alpha-\eta_h & \alpha \\
\beta' & \beta' & ... & \beta' & 1-\beta
\end{bmatrix},
\label{eq:easgd_mp}
\end{equation}
whose first $p$ rows correspond to the local workers' updates, and the last row correspond to
the master's update. The $p+1$ eigenvalues $M_p$ can be computed recursively as follows: let $I_{p+1}(z) = \mbox{det}(M_p- z)$, then we have $I_{p+1}(z) = (1-\alpha -\eta_h -z ) I_p(z) - \alpha \beta' (1-\alpha -\eta_h -z)^{p-1} = \cdots = (1-\alpha -\eta_h -z)^{p-1} [ (1-\beta-z)(1-\alpha -\eta_h-z)-p \alpha \beta' ]$. Notice that $\beta'= \beta/p$, thus we have two eigenvalues which do not depend on $p$, i.e. $(1-\beta-z)(1-\alpha -\eta h-z)- \alpha \beta = 0$, and an extra eigenvalue which only shows up for $p>1$, i.e. $z = 1-\alpha -\eta_h$.
This extra eigenvalue $z = 1-\alpha -\eta_h$ is completely ignored in our reduced Equation~\ref{eq:easgd2}.
Then what is the optimal $\alpha$ for the matrix $M_p$ instead? The three eigenvalues of the matrix $M_p$ in Equation~\ref{eq:easgd_mp} are
\begin{equation}
\begin{array}{r@{}l}
z_1 &= 1-\alpha -\eta_h , \\
z_2 &= b - \sqrt{b^2-c}, \\
z_3 &= b + \sqrt{b^2-c},
\end{array}
\label{eq:easgd_mp_eigs}
\end{equation}
where $b = \frac{1}{2} (2 - \beta -\eta_h - \alpha )$,
$c = (1-\eta_h)(1- \beta) - \alpha $.

Given $\eta_h >0$ and $\beta>0$ fixed, we shall prove that
\begin{itemize}
	\item if $\beta > \eta_h$: the optimal $\alpha = 0$.
	\item if $\beta < \eta_h$: the optimal $\alpha = - ( \sqrt{\beta} - \sqrt{\eta_h})^2$.
\end{itemize}

\begin{figure}[p]
\center
\includegraphics[trim=0in 0in 0in 0in,clip,width = 6in]{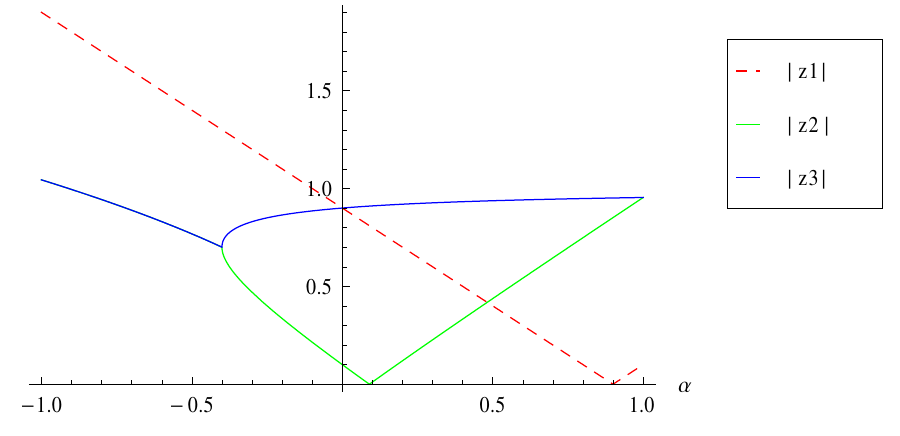}
\caption{The absolute value of the eigenvalues of $z_1$, $z_2$ and $z_3$ in Equation~\ref{eq:easgd_mp_eigs}. $\eta_h=0.1$ and $\beta=0.9$.}
\label{fig:ch5_lin_easgd_original_mp1}
\end{figure}

\begin{figure}[p]
\center
\includegraphics[trim=0in 0in 0in 0in,clip,width = 6in]{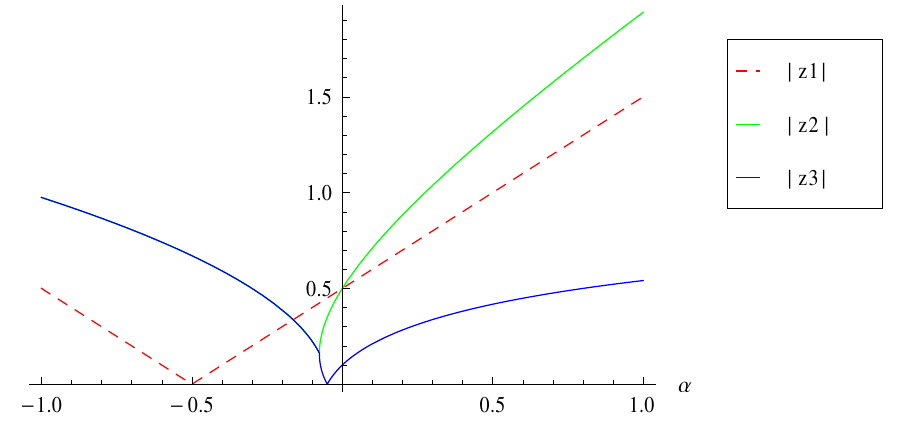}
\caption{The absolute value of the eigenvalues of $z_1$, $z_2$ and $z_3$ in Equation~\ref{eq:easgd_mp_eigs}. $\eta_h=1.5$ and $\beta=0.9$.}
\label{fig:ch5_lin_easgd_original_mp2}
\end{figure}

\begin{figure}[p]
\center
\includegraphics[trim=0in 0in 0in 0in,clip,width = 6in]{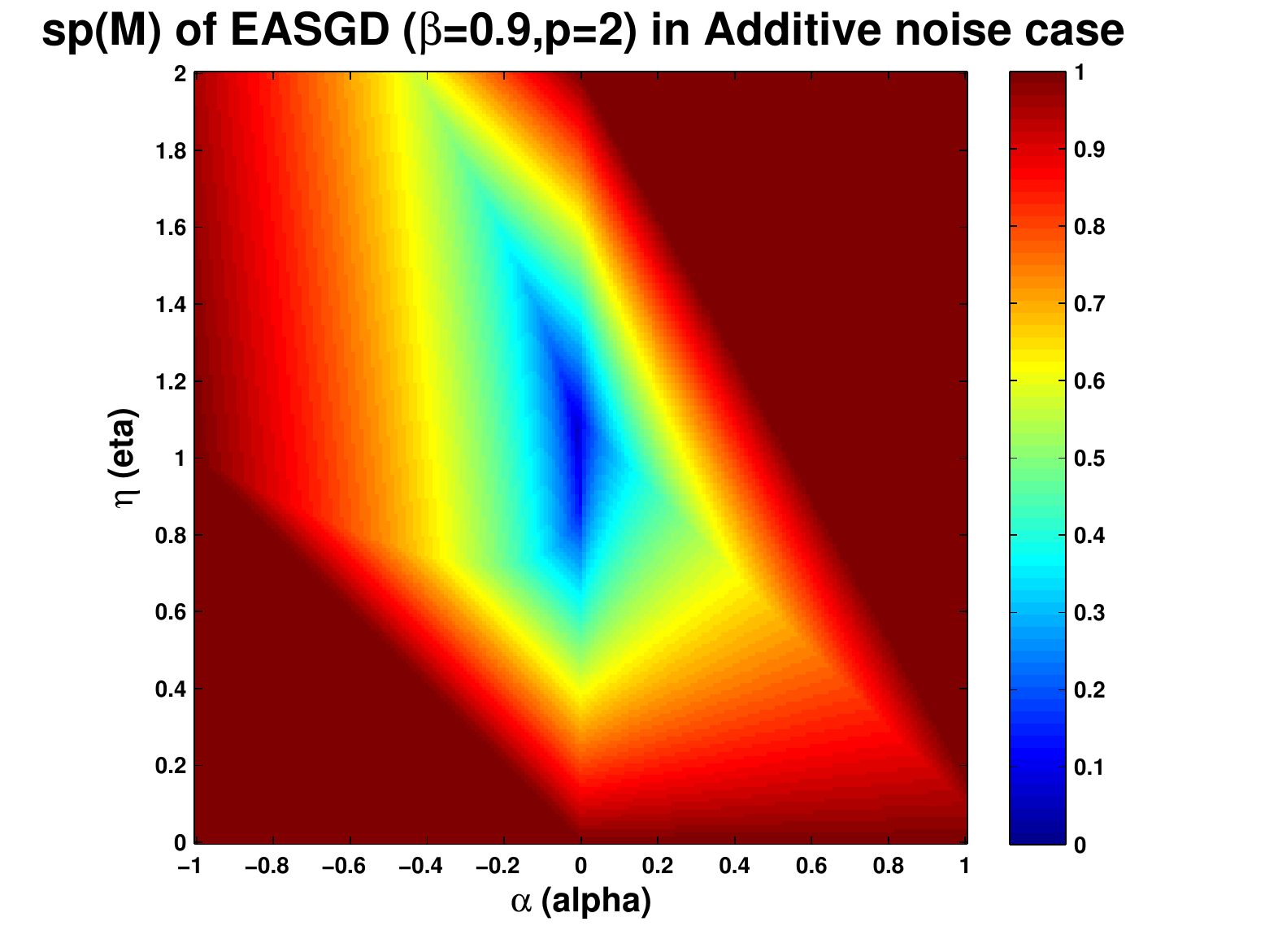}
\caption{The largest absolute eigenvalue of the matrix $M_p$ in Equation~\ref{eq:easgd_mp} as a function of the learning rate $\eta \in (0,2)$ and the moving rate $\alpha \in (-1,1)$. $h=1$ and $\beta=0.9$. Note that we computed this spectrum using $p=2$, as we have discussed in the text, it is independent of the choice of $p$ for $p>1$.}
\label{fig:ch5_lin_easgd_sp_original}
\end{figure}

\begin{figure}[p]
\center
\includegraphics[trim=0in 0in 0in 0in,clip,width = 6in]{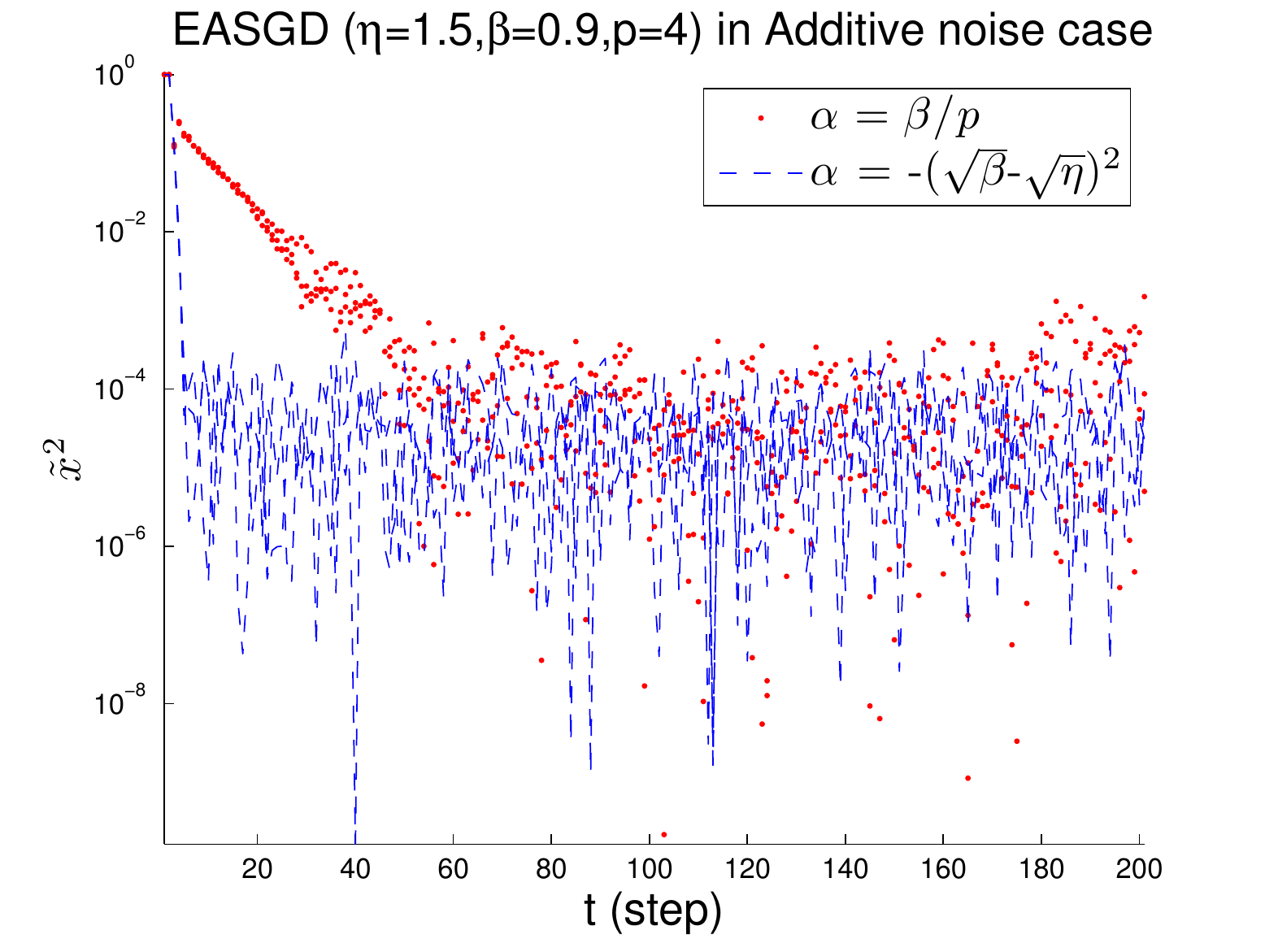}
\caption{Three independent simulations of \textit{EASGD} using the elastic averaging $\alpha=\beta/p$ and the optimal $\alpha$ given in Equation~\ref{eq:easgd_danger}. The x-axis is the time step $t$ in the \textit{EASGD} updates of Equation~\ref{eq:easgd1}. The y-axis is the squared distance of the center variable to the optimum zero, i.e. $\tilde{x}_t^2$. We have chosen $h=1$, $\sigma=10^{-2}$, $p=4$, $\eta=1.5$ and $\beta=0.9$.}
\label{fig:ch5_lin_easgd_beta09_eta15}
\end{figure}

The proof idea is similar to the \textit{MSGD} case, so we shall be more brief.
Let's focus on the variable $c = (1-\eta_h)(1- \beta) - \alpha$ instead of the variable $\alpha$ itself.
Rewrite Equation~\ref{eq:easgd_mp_eigs} with $z_1=c + \beta(1-\eta_h)$, $b = \frac{1}{2}(c-\beta \eta_h +1)$. We find that for $b^2>c$, we only need $c>c_2$ or $c<c_1$, where $c_1 = (\sqrt{\beta \eta_h}-1)^2$, and $c_2 = (\sqrt{\beta \eta_h}+1)^2$. We also observe that the line $z_1$ as a function of $c$ intersects with the quadratic curve $z_2$ or $z_3$ at $c_0 = (1-\eta_h)(1-\beta)$ (i.e. at $\alpha = 0$). At $c=c_0$, $z_1=1-\eta_h$, $z_2 = 1 - \frac{1}{2}(\beta + \eta_h) - \frac{1}{2}|\beta - \eta_h|$, and $z_3 = 1 - \frac{1}{2}(\beta + \eta_h) + \frac{1}{2}|\beta - \eta_h|$. If $\beta > \eta_h$, then $z_3 = z_1$ at $c=c_0$. Otherwise $z_2 = z_1$ at $c=c_0$. We can check that the negative optimal $\alpha= - ( \sqrt{\beta} - \sqrt{\eta_h})^2$ is given at $c=c_1$, where $z_2$ and $z_3$ meet and transit from real-valued to complex-valued. Note that $z_1$ is an increasing function of $c$, and if it intersects with $z_3$, the optimal $\alpha$ becomes $0$ (rather than being negative). They are illustrated in Figure~\ref{fig:ch5_lin_easgd_original_mp1} and~\ref{fig:ch5_lin_easgd_original_mp2} as a function of $\alpha$ under the two conditions, i.e. $\beta > \eta_h$ and $\beta < \eta_h$. We also computed numerically the spectral norm of the matrix $M_p$ for a fixed $\beta=0.9$, which is given in Figure~\ref{fig:ch5_lin_easgd_sp_original}. Finally, we show in Figure~\ref{fig:ch5_lin_easgd_beta09_eta15} the optimal case of \textit{EASGD} under the condition $\beta < \eta_h$.

We have studied the the second-order moment equation~\ref{eq:easgd_moment2} of the reduced system (Equation~\ref{eq:easgd2}), and the first-order moment equation~\ref{eq:easgd_mp} of the original system (Equation~\ref{eq:easgd1}) for the \textit{EASGD} method. We found that the reduced system can lose critical information about the stability of the original system. However, the eigenvalues are still closely related in the first-order moment matrix $M_p$ and the second-order moment $M$, i.e. the $z_2$ and $z_3$ in Equation~\ref{eq:easgd_eigs} and~\ref{eq:easgd_mp_eigs}. Thus for the \textit{EAMSGD} method, we shall only focus on the first-order moment equation. Its asymptotic variance, which can obtained from the second-order moment equation, is rather complicated and will not be discussed.

For \textit{EAMSGD}, we have the following update rules
\begin{equation*}
\begin{array}{r@{}l}
v^i_{t+1} &{}= \delta v^i_t - \eta (h (x^i_t + \delta v^i_t) - \xi_t^i), \\
x^i_{t+1} &{}= x^i_t +v_{t+1}^i + \alpha( \tilde{x}_t - x^i_t), \\
\tilde{x}_{t+1} &{}= \tilde{x}_{t} + \beta ( \frac{1}{p} \sum_{i=1}^p x_t^i - \tilde{x}_t ).
\end{array}
\end{equation*}

We have the following form of the drift matrix for the first-order moment equation,
\begin{equation}
M_p=
\begin{bmatrix}
\delta_h & -\eta_h & 0 & 0 & ... & ... & 0 \\
\delta_h & 1-\eta_h-\alpha & 0 & 0 & ... & ... & \alpha \\
0 & 0 & \delta_h & -\eta_h & ... & ... & 0 \\
0 & 0 & \delta_h & 1-\eta_h-\alpha & ... & ...& \alpha \\
... & ... & ... & ... & ... & ...& ... \\
... & ... & ... & ... &... & ...& ...\\
0 & \beta' & 0 & \beta' & ... & ...& 1-\beta
\end{bmatrix},
\label{eq:eamsgd1_mp}
\end{equation}
such that $[v^1_{t+1},x^1_{t+1},v^2_{t+1},x^2_{t+1},\ldots,\tilde{x}_{t+1}]^{T}=M_p [v^1_{t},x^1_{t},v^2_{t},x^2_{t},\ldots,\tilde{x}_{t}]^{T}$.
Recall that $\delta_h = \delta(1-\eta h)$, $\eta_h = \eta h$, and $\beta' = \beta / p$.
We can compute the eigenvalues of the above drift matrix $M_p$ recursively,
and one can check that they are again independent of the choice of $p$ for $p>1$,
as in the \textit{EASGD} case. More precisely, we have
\begin{equation}
\mbox{det} | M_p - z | = (u(z))^{p-1} v(z) = 0,
\label{eq:eamsgd1_eigs}
\end{equation}
where $u(z) = z^2 - (1-\eta-\alpha-\delta)z + \delta(1-\alpha)$ and
$v(z) = (\delta - z)(1-\eta-\alpha-z)(1-\beta-z)+\eta \delta(1-\beta-z) - \alpha \beta (\delta -z)$.

\begin{figure}[p]
\center
\includegraphics[trim=0in 0in 0in 0in,clip,width = 6in]{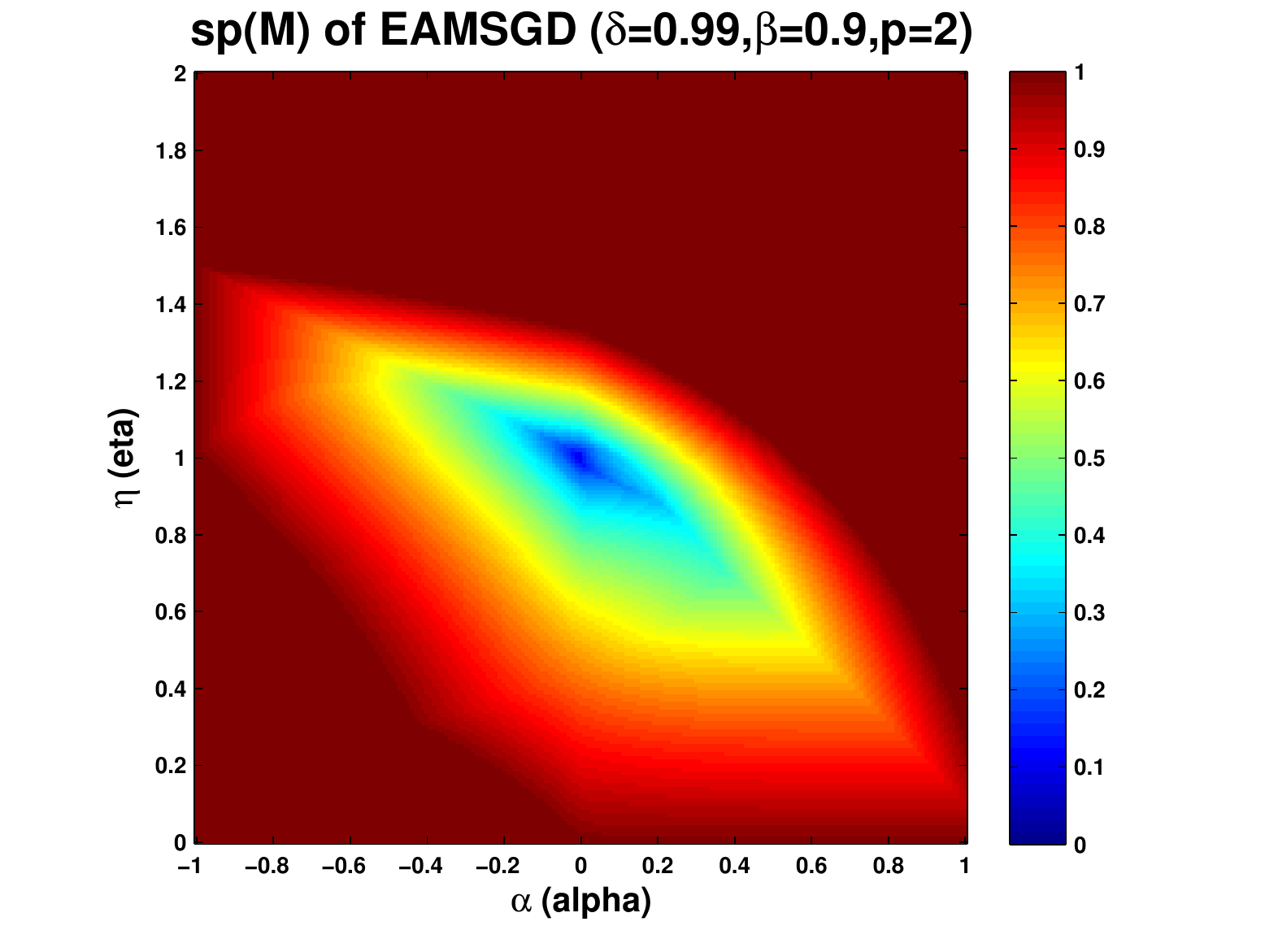}
\caption{The largest absolute eigenvalue of the matrix $M_p$ in Equation~\ref{eq:eamsgd1_mp} as a function of the learning rate $\eta \in (0,2)$ and the moving rate $\alpha \in (-1,1)$. $h=1$, $\beta=0.9$ and $\delta=0.99$. Note that we computed this spectrum again using $p=2$, as we have discussed in the text, it is independent of the choice of $p$ for $p>1$.}
\label{fig:ch5_lin_eamsgd_sp_p2}
\end{figure}

The solution of the Equation~\ref{eq:eamsgd1_eigs} is quite complicated as it involves a third-order
polynomial which is irreducible. It would be difficult to obtain the optimal $\delta$ or $\alpha$ as before.
So we computed numerically the spectral norm of the matrix $M_p$ in
Equation~\ref{eq:eamsgd1_mp} for a fixed $\beta=0.9$ and $\delta=0.99$ (which we used in the experiment in Chapter~\ref{sec:ch_exp}).
It is given in Figure~\ref{fig:ch5_lin_eamsgd_sp_p2}. This result suggests that the optimal $\alpha$ increases
as the $\eta$ (or $\eta_h$ as $h=1$) decreases. Recall our result of \textit{MSGD} in Figure~\ref{fig:ch5_lin_msgd_sp}: the optimal $\delta$ increases as $\eta$ decreases.
This is consistent with the choice of the learning rate and the momentum rate scheduling in the Nesterov's optimal methods in literature~\cite{lan2012optimal}. We have seen that there's an intimate connection
between \textit{MSGD} and \textit{EASGD} when we were studying the optimal momentum rate $\delta$ and the optimal moving rate $\alpha$. It suggests that we may also find interesting rate scheduling for \textit{EASGD}, as well as \textit{EAMSGD} based on the convex analysis. However, we should be careful as the optimal $\alpha$ for \textit{EASGD} is either zero or negative, while for \textit{EAMSGD} it can be positive as well.

\section{Multiplicative noise}
\label{sec:multiplicative_noise}
In this section, we study a multiplicative noise model, which attempts to capture the
initial behavior of the stochastic optimization method. It is complementary to the additive noise model, which captures the asymptotic behavior.

Our starting point is the following linear regression problem
\begin{equation*}
\min_{a \in \mathbb{R}} \mathbb{E} [(v-au)^2],
\end{equation*}
where the expectation $ \mathbb{E}$ is taken over the joint distribution of $(u,v)$. If the input data $(u,v)$ satisfies $v=a^\ast u$, we may reduce the above problem to the following one-dimensional case,
\begin{equation}
\min_{x \in \mathbb{R}} \mathbb{E} [(x u)^2].
\label{eq:mulpb}
\end{equation}
An interesting perspective of this problem is that we can assume that $u^2$ of the input data follows a Gamma distribution $\Gamma(\lambda,\omega)$, with mean $\lambda / \omega$ and variance $\lambda/\omega^2$. Note that if $u$ follows a Gaussian distribution with mean zero and variance $\sigma^2$, then $\lambda = 1/2$ and $\omega = 1/(2 \sigma^2)$.

\subsection{\textit{SGD} with mini-batch}
Given $x_0 \in \mathbb{R}$, the mini-batch \textit{SGD} method for solving the multiplicative noise problem in Equation \ref{eq:mulpb} is as follows,
\begin{equation}
	x_{t+1} = x_{t} - \eta u_t^2 x_t ,
\label{eq:mulsgd}
\end{equation}
where $\eta > 0$, and $u_t$ is an i.i.d input process.

As we assumed that $u_t^2$ follow a Gamma distribution $\Gamma(\lambda,\omega)$, it's easy to verify
that if we use mini-batch of size $p$, Equation~\ref{eq:mulsgd} becomes
\begin{equation}
	x_{t+1} = x_{t} - \eta \frac{1}{p} \sum_{i=1}^p (u^i_t)^2 x_t,
\label{eq:mulmbsgd}
\end{equation}
and this mini-batch $\frac{1}{p} \sum_{i=1}^p (u^i_t)^2$ also follows a Gamma distribution $\Gamma(p \lambda, p \omega)$, since the mean of the mini-batch is not changed and the variance is divided by $p$.

We also notice that in the continuous-time limit, i.e. for small $\eta$, we can approximate the process in Equation~\ref{eq:mulmbsgd} by a Geometric Brownian motion~\cite{DBLP:journals/corr/LiTE15} as follows,
\begin{equation*}
dX(t) = - \frac{\lambda}{\omega} X(t) dt + \sqrt{\eta} \sqrt{\frac{\lambda}{p \omega^2}} X(t) dB(t).
\end{equation*}

An interesting behavior of the geometric Brownian Motion is that its variance can explode (go to infinity), yet the path of the stochastic process still converges almost surely towards zero. In such case, we can observe along the path a few extreme large values. These extreme values may however cause trouble to the stability of the nonlinear dynamics in the training of deep learning models. So our goal here is to control this variance by studying again the second-order moment equation.

Going back to the discrete process defined by Equation~\ref{eq:mulmbsgd}. Denote $\xi_t = \frac{1}{p} \sum_{i=1}^p (u^i_t)^2$, we have
\begin{equation}
	x_{t+1}^2 = (1-\eta \xi_t)^2 x_{t}^2.
\label{eq:mulmbsgd2}
\end{equation}

Taking expectation on both sides of the Equation~\ref{eq:mulmbsgd2}, we obtain
\begin{equation}
	\mathbb{E} [ x_{t+1}^2 ] =
(1-2 \eta \frac{p \lambda}{p \omega} + \eta^2 \frac{p \lambda (p \lambda+1)}{(p\omega)^2} )
\mathbb{E} [ x_{t}^2 ]
		= (1-2 \eta \frac{\lambda}{\omega} + \eta^2 \frac{\lambda (p \lambda+1)}{p \omega^2} )
\mathbb{E} [ x_{t}^2 ].
\label{eq:mulmbsgdm2}
\end{equation}

The rate of convergence defined by the above second-moment equation~\ref{eq:mulmbsgdm2} is
$1-2 \eta \frac{\lambda}{\omega} + \eta^2 \frac{\lambda (p \lambda+1)}{p \omega^2}$. It is monotone decreasing as $p$ increases and will saturate at a limit $1-2 \eta \frac{\lambda}{\omega} + \eta^2 \frac{\lambda^2}{\omega^2} = (1- \eta \frac{\lambda}{\omega})^2$. Given $p$, we can optimize this convergence rate with respect to the choice of the learning rate $\eta$. This optimal learning rate is given by
\begin{equation}
	\eta_p = \frac{p \omega}{p \lambda+1} = \frac{\omega}{\lambda + 1/p} .
\label{eq:muloptlr}
\end{equation}

\begin{figure}[p]
\center
\includegraphics[trim=0in 0in 0in 0in,clip,width = 6in]{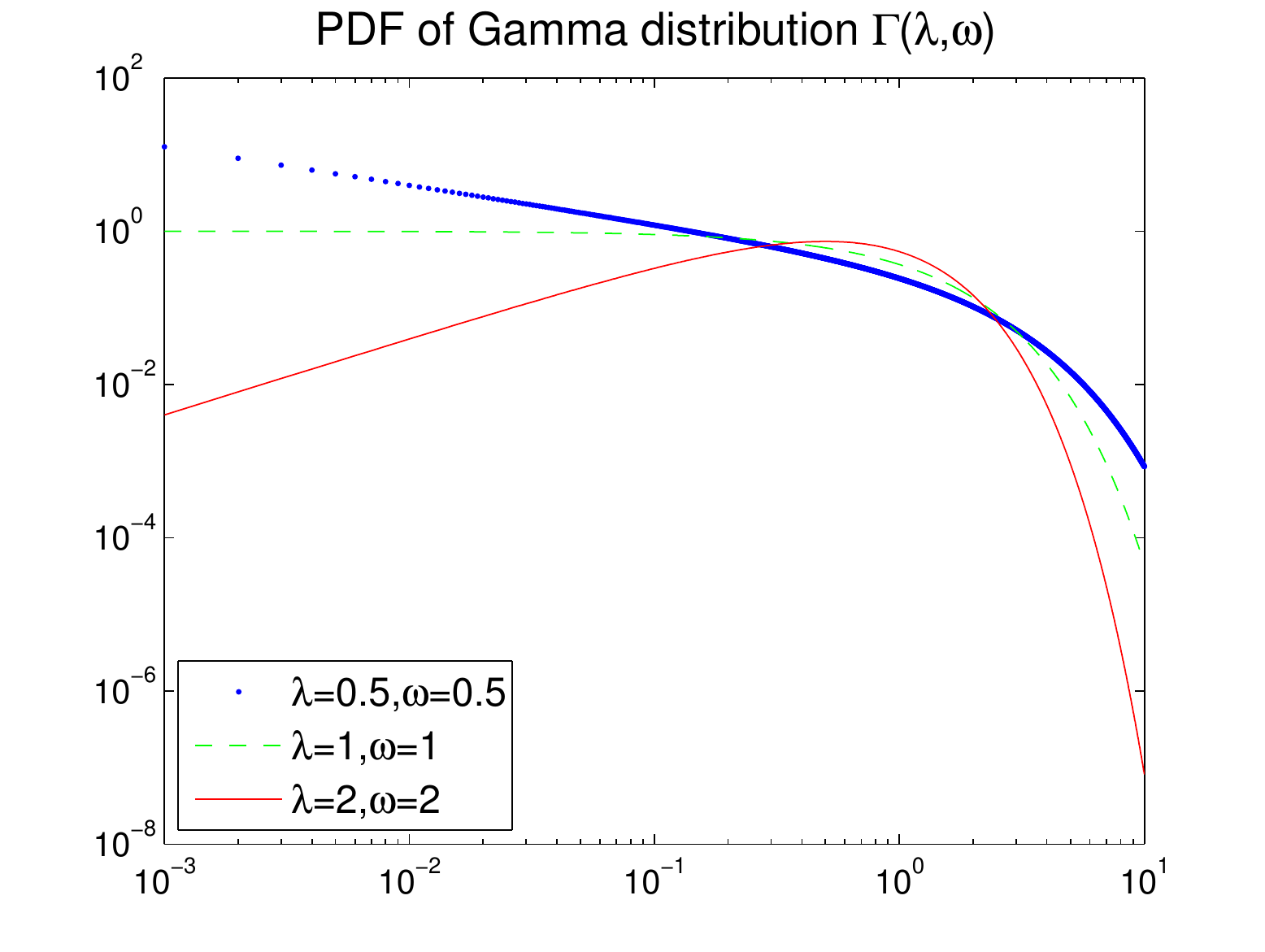}
\caption{The probability density function of the Gamma distribution $\Gamma(\lambda,\omega)$. The x-axis and y-axis are both in log-scale to enlarge the singularity at zero and the decay of the tail toward infinity.}
\label{fig:ch5_mul_gamma_pdf}
\end{figure}

From this formula, we can see that as $p$ evolves, $\eta_p$ will change a lot if $\lambda$ is much smaller than $1/p$. In other words, if $\lambda$ is very big compared to $1/p$, then the change in $\eta_p$ is not that much as we increases the mini-batch size $p$. Thus we can expect that the speedup we would gain depends heavily on the distribution of the input data, in particular the shape parameter $\lambda$ of the Gamma distribution. Suppose for $p=1$, the $\xi_t$ follows the Gamma distribution $\Gamma(\lambda, \omega)$. Its probability density function is $\frac{\omega^\lambda}{\Gamma(\omega)} \xi^{ \lambda-1} e^{- \omega \xi}$. We illustrate this probability density function in Figure~\ref{fig:ch5_mul_gamma_pdf}
for the standard Gaussian case ($\lambda=1/2$, $\omega=1/2$) with $p=1$, $p=2$ and $p=4$.
We see that when $\lambda$ is smaller than one, the probability density function has a pole at zero and has also a slower decay (heavier tail) toward infinity. By increasing the mini-batch size $p$, the probability density function becomes more and more concentrated near its mean. Thus the mini-batch is more effective for such input distribution with very large spread (i.e. small $\lambda$).

\subsection{Momentum \textit{SGD}}
\label{sec:mom_tradeoff}
Given $x_0 \in \mathbb{R}$ and $v_0 = 0$, the \textit{MSGD} method for solving the multiplicative noise problem in Equation \ref{eq:mulpb} is as follows,
\begin{equation}
\begin{array}{r@{}l}
v_{t+1} &{}= \delta v_t - \eta \xi_t ( x_t + \delta v_t), \\
x_{t+1} &{}= x_t +v_{t+1},
\end{array}
\label{eq:mulmsgd1}
\end{equation}
where $\xi_t $ follows a Gamma distribution $\Gamma( \lambda, \omega)$.

Denote $\delta_t = \delta (1-\eta \xi_t)$ and $\eta_t = \eta \xi_t$, the second-order moment equation can be computed as follows,
\begin{equation}
\begin{array}{r@{}l}
v_{t+1}^2 &{}= (\delta ( 1 - \eta \xi_t) v_t - \eta \xi_t x_t )^2 = (\delta_t v_t - \eta_t x_t)^2 \\
		 &{}= \delta_t^2 v_t^2 -2 \eta_t \delta_t v_t x_t + \eta_t^2 x_t^2, \\
v_{t+1} x_t &{}= (\delta_t v_t - \eta_t x_t ) x_t
			 = \delta_t v_t x_t - \eta_t x_t^2, \\
x_{t+1}^2 &{}= (x_{t} + v_{t+1})^2 = x_{t}^2 + 2 v_{t+1}x_{t} + v_{t+1}^2 \\
		 &{}= x_{t}^2 + 2(\delta_t v_t x_t - \eta_t x_t^2) + \delta_t^2 v_t^2 -2 \eta_t \delta_t v_t x_t + \eta_t^2 x_t^2
, \\
		 &{}= \delta_t^2 v_t^2 + (1-2 \eta_t + \eta^2_t) x_t^2 + (2 \delta_t-2 \eta_t \delta_t) v_t x_t, \\
v_{t+1} x_{t+1} &{}= v_{t+1} (x_{t} + v_{t+1}) = v_{t+1} x_t + v_{t+1}^2 \\
		 &{}= \delta_t v_t x_t - \eta_t x_t^2 + \delta_t^2 v_t^2 -2 \eta_t \delta_t v_t x_t + \eta_t^2 x_t^2 \\
		 &{}= \delta_t^2 v_t^2 + ( - \eta_t + \eta_t^2) x_t^2 + \delta_t(1-2\eta_t) v_t x_t.
\end{array}
\label{eq:mulmsgd2}
\end{equation}

Taking expectation on both sides of the Equation~\ref{eq:mulmsgd2}, we obtain the following recursive relation $(\mathbb{E} v^2_{t+1},\mathbb{E} x^2_{t+1}, \mathbb{E} v_{t+1} x_{t+1})^T = M
(\mathbb{E} v^2_{t},\mathbb{E} x^2_{t}, \mathbb{E} v_{t} x_{t})^T$, where
\begin{gather}
M =
\begin{pmatrix}
\delta^2 ( 1 - 2 \eta u_1 + \eta^2 u_2) &
\eta^2 u_2 & -2 \delta \eta (u_1 - \eta u_2) \\
\delta^2 ( 1 - 2 \eta u_1 + \eta^2 u_2) &
	1 - 2 \eta u_1 + \eta^2 u_2 &
	2 \delta ( 1- \eta u_1) - 2 \delta \eta ( u_1 - \eta u_2 ) \\
\delta^2 ( 1 - 2 \eta u_1 + \eta^2 u_2) &
	- \eta u_1 + \eta^2 u_2 & \delta (1 - \eta u_1) - 2 \delta \eta (u_1 - \eta u_2)
\end{pmatrix}
,
\label{eq:mulmsgd_moment2}
\end{gather}
$u_1 = \frac{\lambda}{\omega} $ and $u_2 = \frac{\lambda(\lambda+1)}{\omega^2}$.

\begin{figure}[p]
\center
\includegraphics[trim=0in 0in 0in 0in,clip,width = 6in]{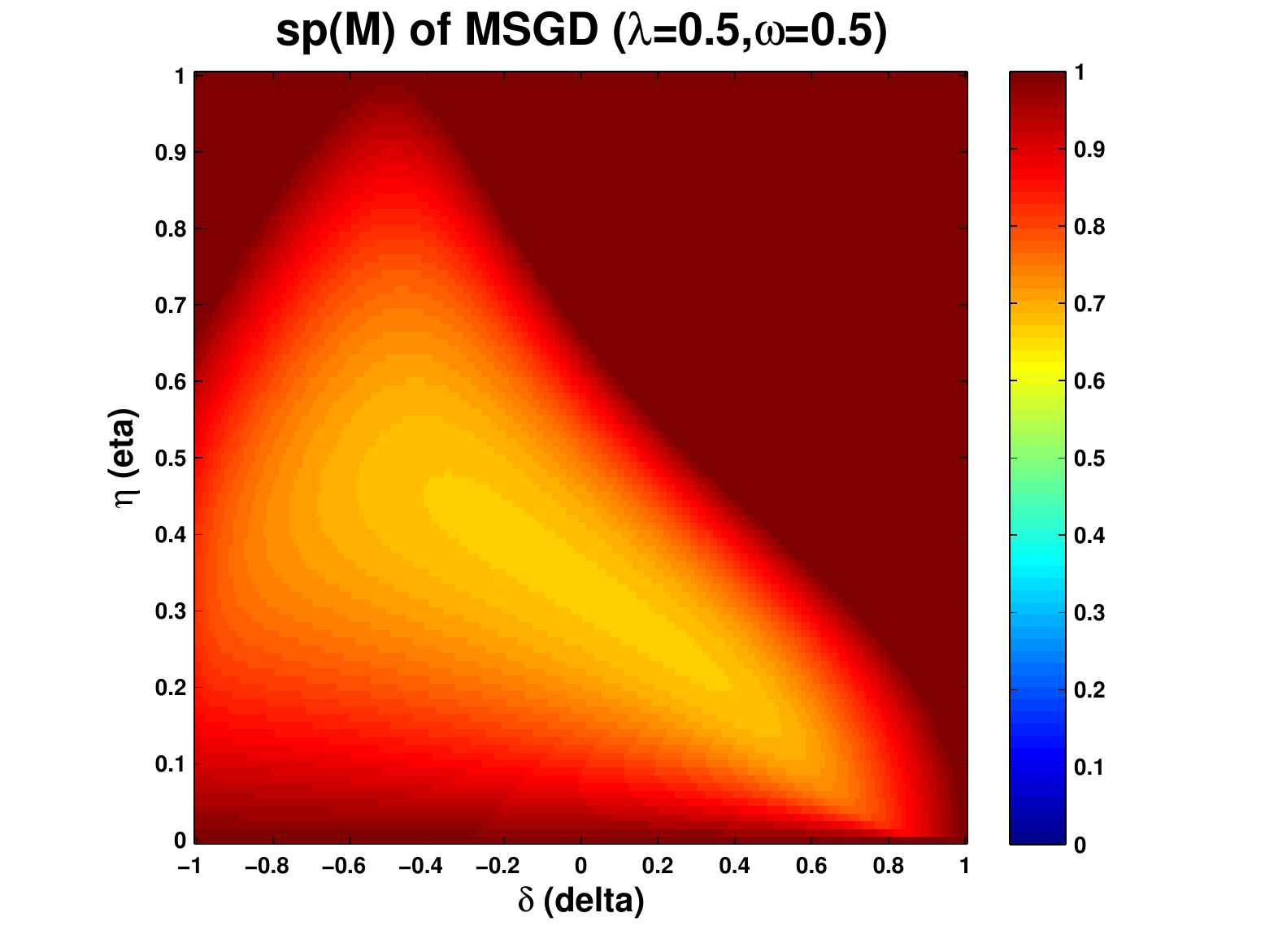}
\caption{The largest absolute eigenvalue of the matrix $M$ in Equation~\ref{eq:mulmsgd_moment2} as a function of the learning rate $\eta \in (0,1)$ and the momentum rate $\delta \in (-1,1)$. $\lambda=0.5$, $\omega=0.5$.}
\label{fig:ch5_mul_msgd_sps_lambda05_omega05}
\end{figure}

\begin{figure}[p]
\center
\includegraphics[trim=0in 0in 0in 0in,clip,width = 6in]{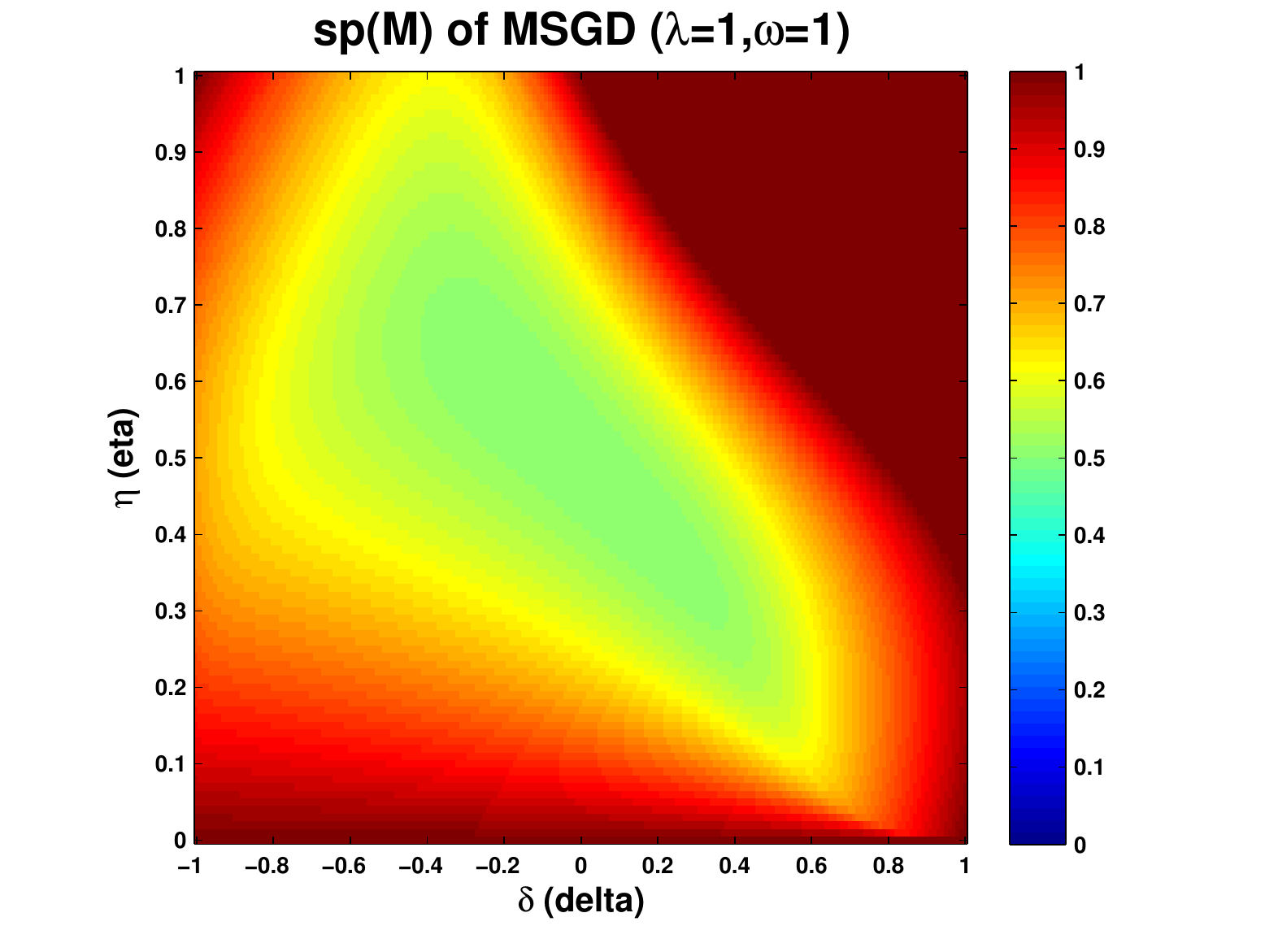}
\caption{The largest absolute eigenvalue of the matrix $M$ in Equation~\ref{eq:mulmsgd_moment2} as a function of the learning rate $\eta \in (0,1)$ and the momentum rate $\delta \in (-1,1)$. $\lambda=1$, $\omega=1$.}
\label{fig:ch5_mul_msgd_sps_lambda1_omega1}
\end{figure}

\begin{figure}[p]
\center
\includegraphics[trim=0in 0in 0in 0in,clip,width = 6in]{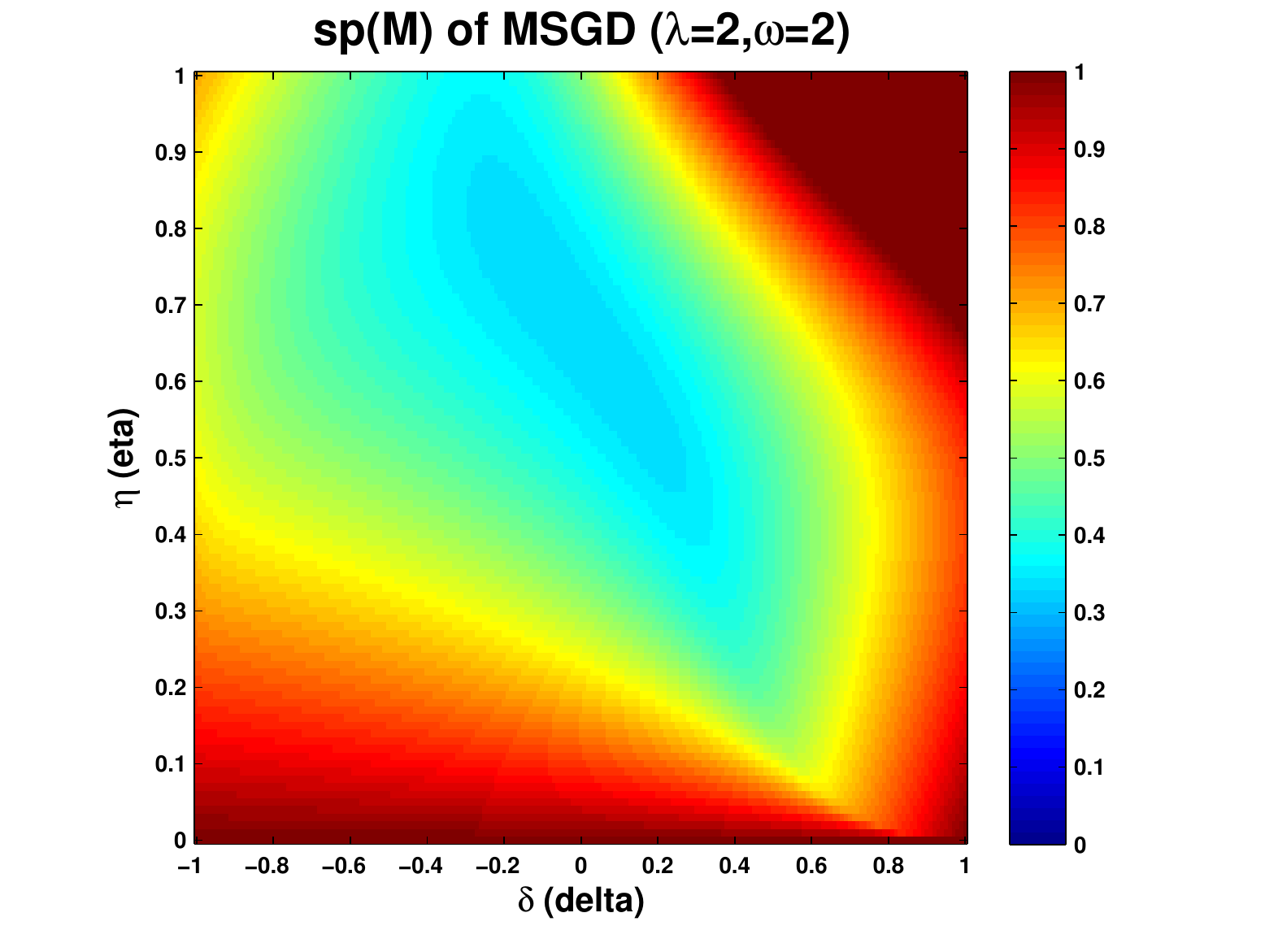}
\caption{The largest absolute eigenvalue of the matrix $M$ in Equation~\ref{eq:mulmsgd_moment2} as a function of the learning rate $\eta \in (0,1)$ and the momentum rate $\delta \in (-1,1)$. $\lambda=2$, $\omega=2$.}
\label{fig:ch5_mul_msgd_sps_lambda2_omega2}
\end{figure}

\begin{figure}[p]
\center
\includegraphics[trim=0in 0in 0in 0in,clip,width = 6in]{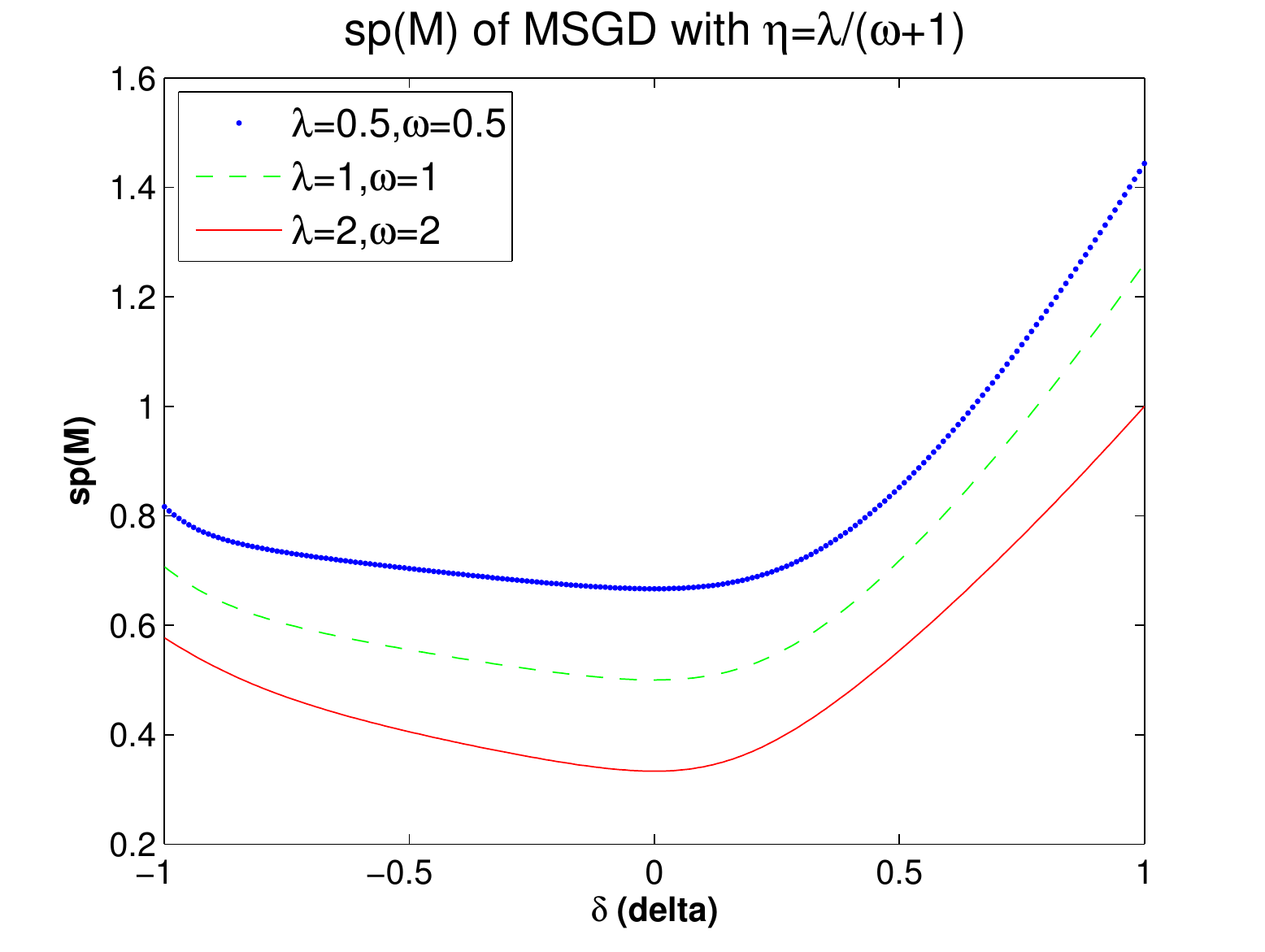}
\caption{The largest absolute eigenvalue of the matrix $M$ in Equation~\ref{eq:mulmsgd_moment2} as a function of the momentum rate $\delta \in (-1,1)$ at $\eta = \frac{\lambda}{\omega+1}$. $(\lambda,\omega) \in \{(0.5,0.5),(1,1),(2,2)\}$.}
\label{fig:ch5_mul_msgd_opt}
\end{figure}

\begin{figure}[p]
\center % 3.5
\includegraphics[trim=0in 0in 0in 0in,clip,width = 2.8in]{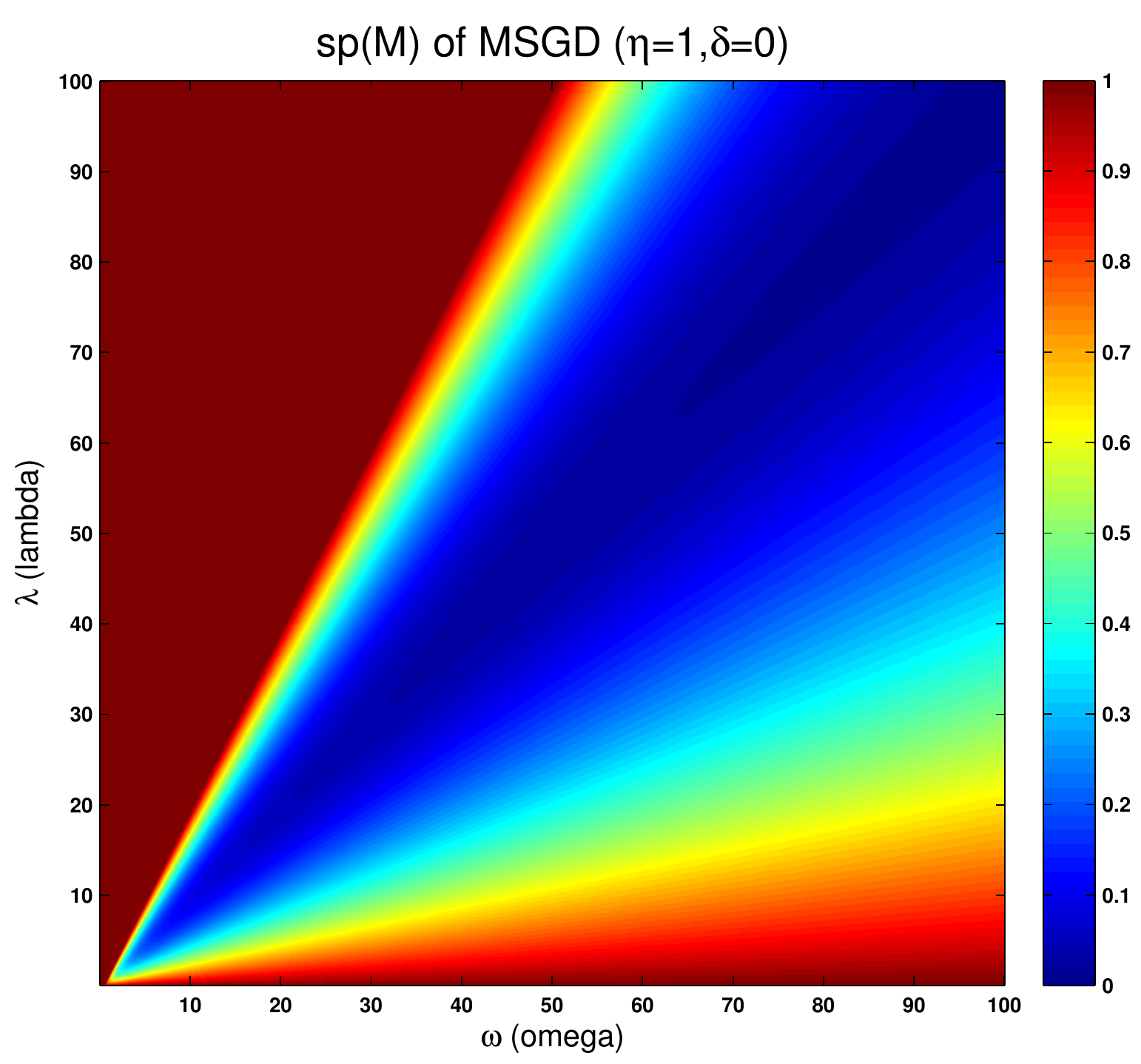} \\
\includegraphics[trim=0in 0in 0in 0in,clip,width = 2.8in]{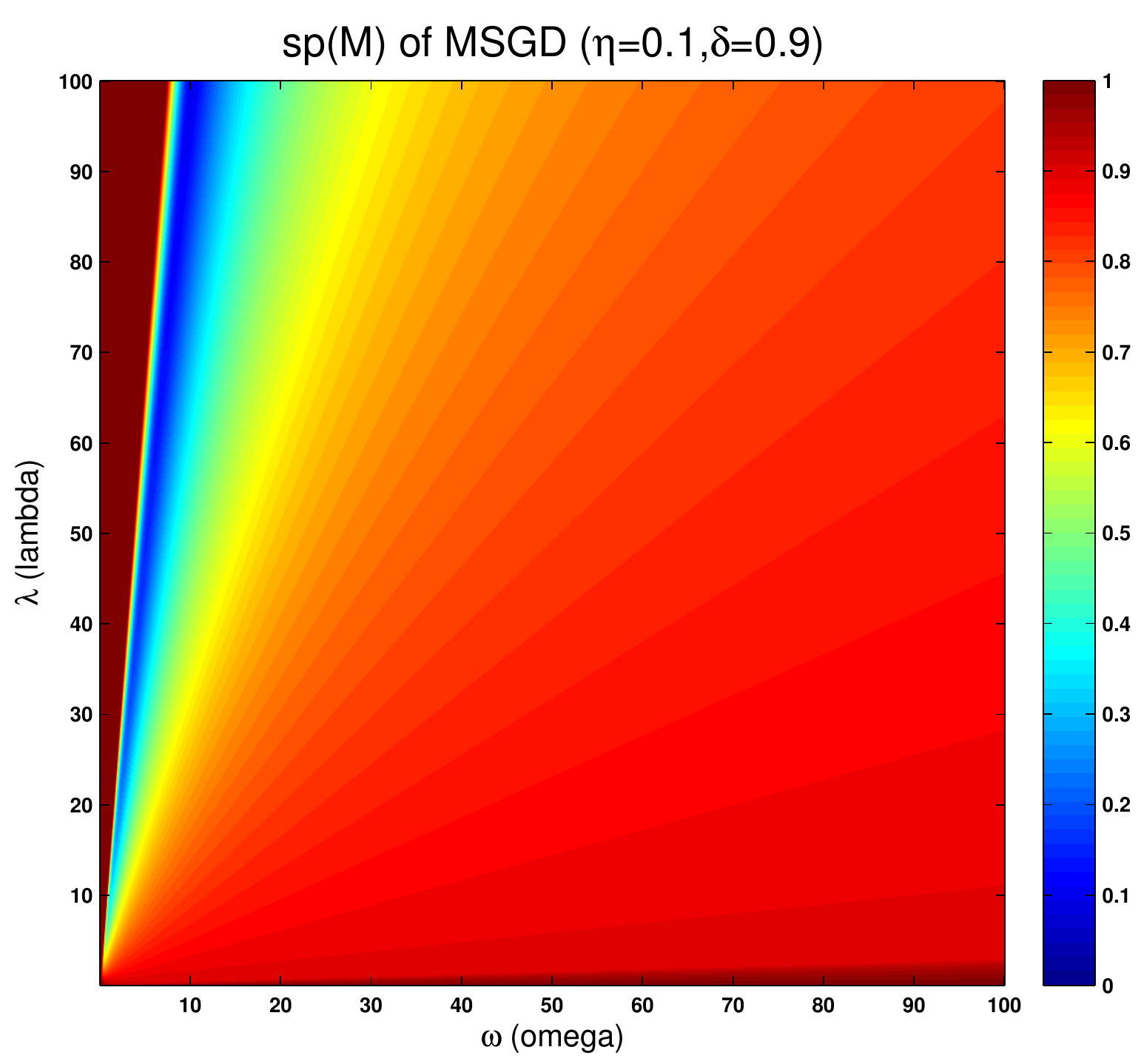}
\includegraphics[trim=0in 0in 0in 0in,clip,width = 2.8in]{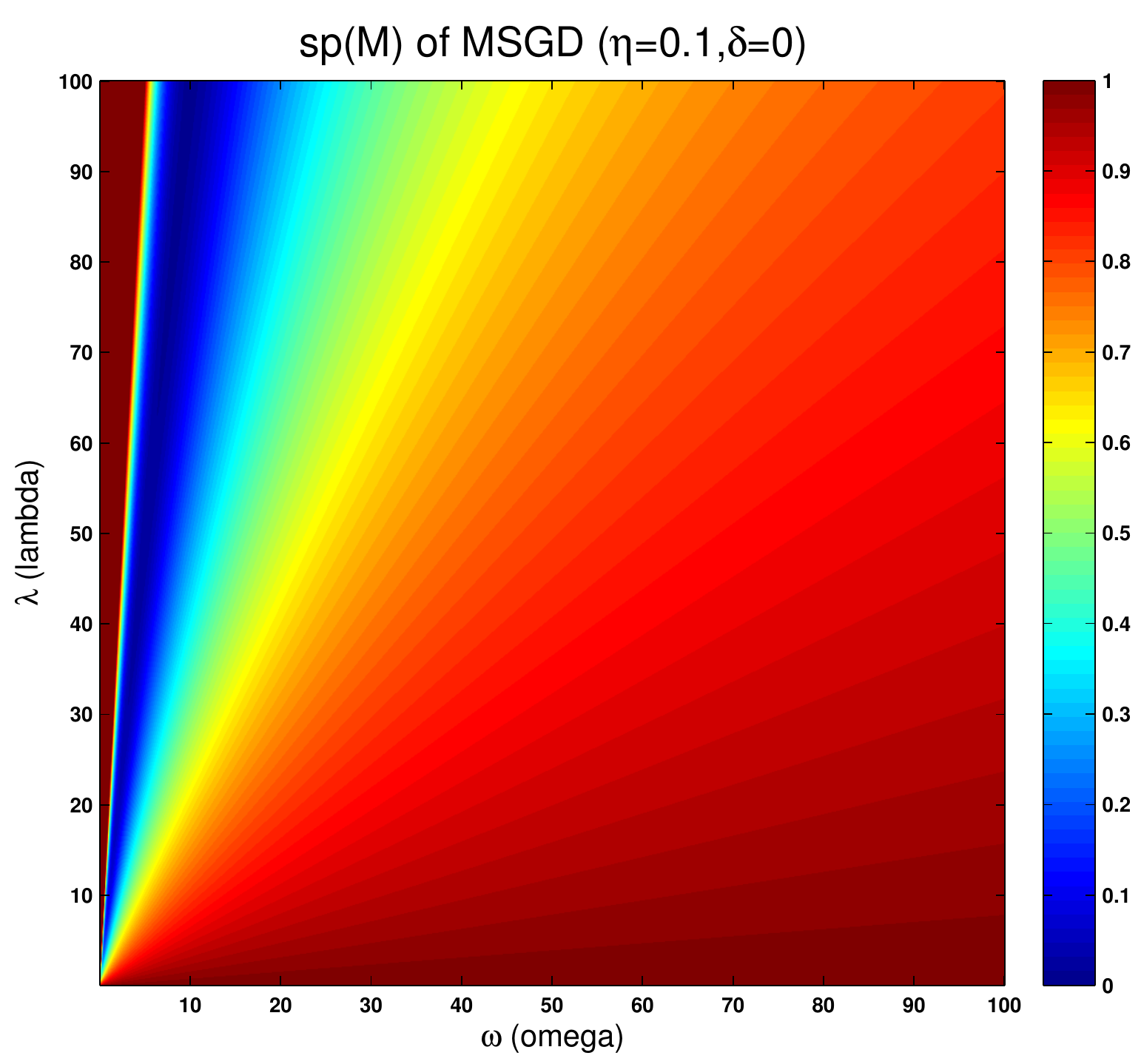} \\
\includegraphics[trim=0in 0in 0in 0in,clip,width = 2.8in]{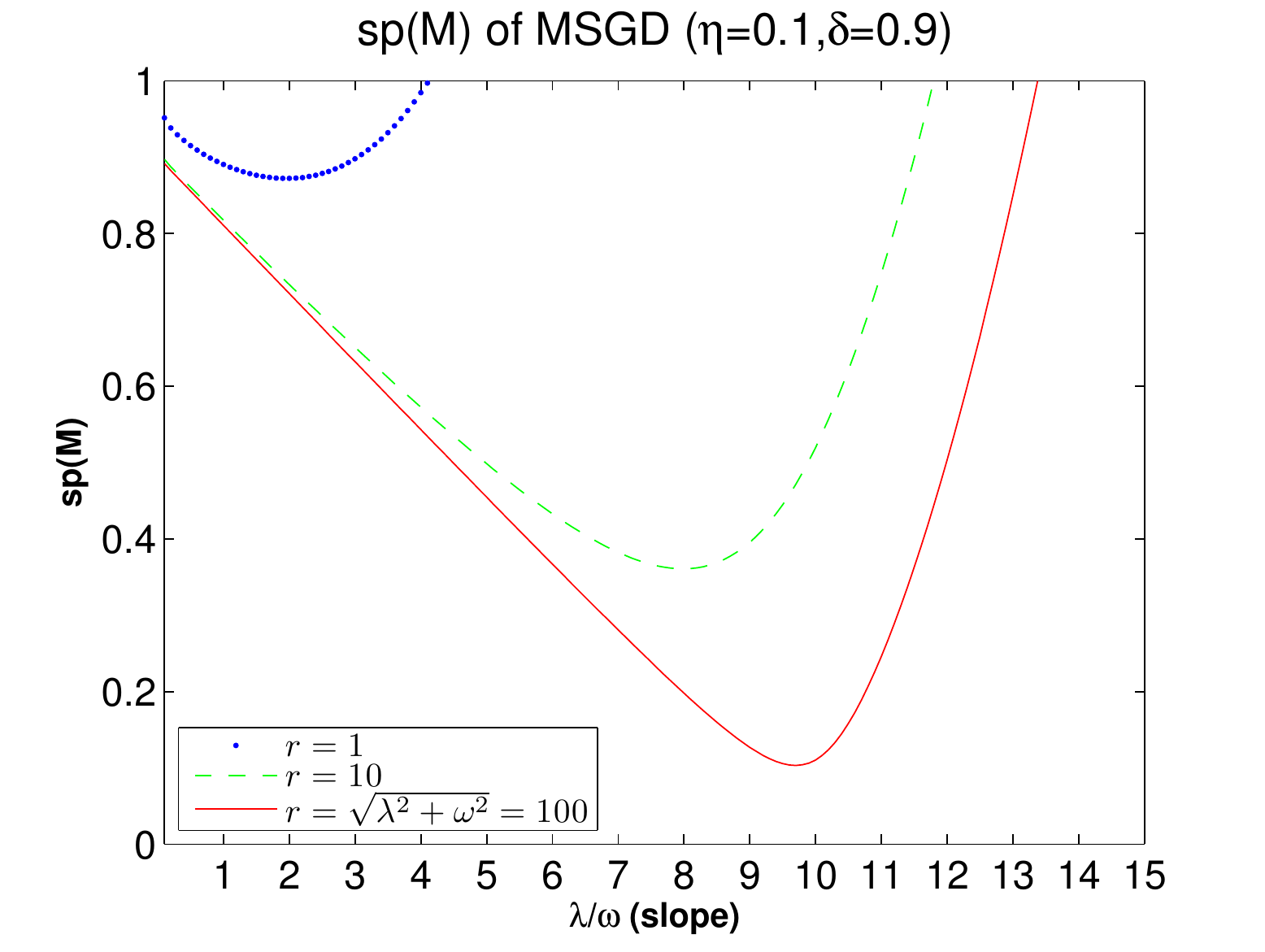}
\includegraphics[trim=0in 0in 0in 0in,clip,width = 2.8in]{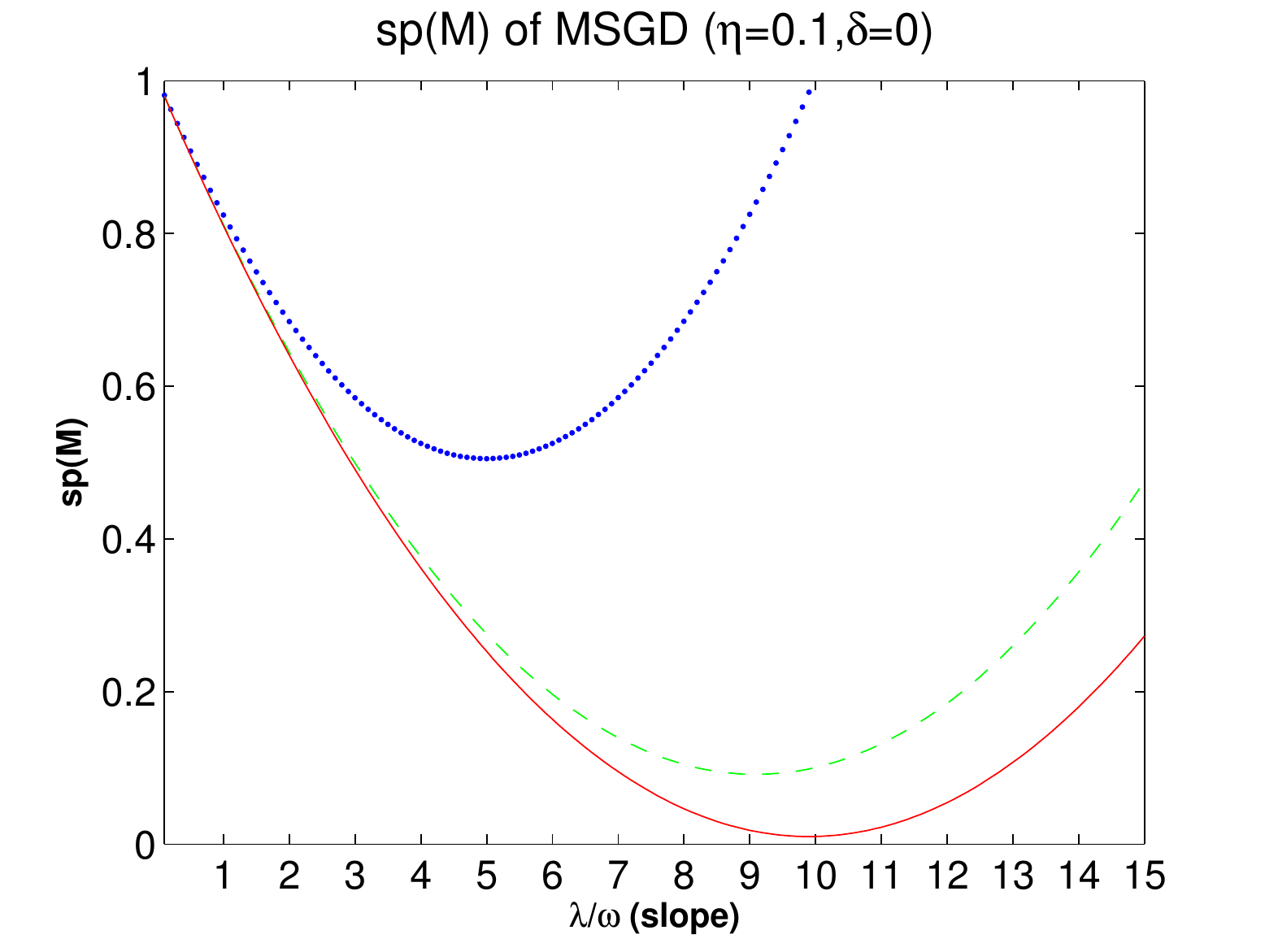}
\caption{The largest absolute eigenvalue of the matrix $M$ in Equation~\ref{eq:mulmsgd_moment2} as a function of the input Gamma distribution $\Gamma( \lambda, \omega)$ in the range $\omega \in (0,100)$ and the $\lambda \in (0,100)$. $(\eta,\delta) \in \{(1,0),(0.1,0),(0.1,0.9)\}$.}
\label{fig:ch5_mul_msgd_sp2}
\end{figure}

Now we can compare the numerical spectral norm of the matrix $M$ for various choices of
the input data distribution parameterized by $\lambda$ and $\omega$. In Figure~\ref{fig:ch5_mul_msgd_sps_lambda05_omega05}, we illustrate the standard Gaussian case, i.e. $u_t^i$ follows i.i.d. $\mathcal{N}(0,1)$ in Equation~\ref{eq:mulmbsgd}. In Figure~\ref{fig:ch5_mul_msgd_sps_lambda1_omega1} and~\ref{fig:ch5_mul_msgd_sps_lambda2_omega2} we illustrate the resulting effect if we use mini-batch of size $p=2$ and $p=4$. Recall that for $\delta=0$, we have the optimal learning rate in Equation~\ref{eq:muloptlr}. It achieves nearly the optimal convergence rate (smallest sp($M$)) in these Figures. We also observe that the optimal momentum rate is actually $0$ at these
optimal learning rates. One can check such observation in Figure~\ref{fig:ch5_mul_msgd_opt}.
Thus contrary to our earlier \textit{MSGD} results in the additive noise case (Figure~\ref{fig:ch5_lin_msgd_sp}),
the momentum can slow down the optimal convergence rate. However, we can yet ask, given $\eta$ and $\delta$ fixed, what is the region of input data distribution $\lambda$ and $\omega$ such that the momentum helps. We find that the answer is for relatively small slope $\frac{\lambda}{\omega}$ (c.f. Figure~\ref{fig:ch5_mul_msgd_sp2}). This phenomenon of acceleration
is consistent with our earlier analysis of \textit{MSGD} in the additive noise case (c.f. Figure~\ref{fig:ch5_lin_msgd_sp}).

\subsection{\textit{EASGD} and \textit{EAMSGD}}

We now focus on the \textit{EASGD} updates for solving the problem in Equation~\ref{eq:mulpb},
\begin{equation}
\begin{array}{r@{}l}
	x^i_{t+1} &{}= x^i_{t} - \eta \xi_t^i x^i_t + \alpha (\tilde{x}_t - x^i_t), \\
	\tilde{x}_{t+1} &{}= \tilde{x}_{t} - \frac{\beta}{p} \sum_{i=1}^{p} (\tilde{x}_t - x^i_t),
\end{array}
\label{eq:mul_easgd1}
\end{equation}
where $\xi_t^i$ follows a Gamma distribution $\Gamma( \lambda, \omega)$.

The second-order moment equation can be computed as follows,
\begin{equation}
\begin{array}{r@{}l}
(x^i_{t+1})^2 &{}= (1 - \alpha - \eta \xi_t^i)^2 (x^i_t)^2 + 2 \alpha (1 - \alpha - \eta \xi_t^i) \tilde{x}_t x^i_t + 		\alpha^2 (\tilde{x}_t)^2, \\
(\tilde{x}_{t+1})^2 &{}= (1 - \beta)^2 (\tilde{x}_{t})^2 + 2 \beta (1 - \beta) (\frac{1}{p} \sum_{i=1}^p 		\tilde{x}_{t} x_t^i) + \frac{\beta^2}{p^2} \sum_{i=1,j=1}^p x_t^i x_t^j, \\
\tilde{x}_{t+1} x_{t+1}^i &{}= (1-\beta)(1-\alpha-\eta \xi_t^i) \tilde{x}_t x_t^i +
	\alpha (1-\beta) (\tilde{x}_t)^2 \\
	&{} \quad + (1-\alpha-\eta \xi_t^i) \frac{\beta}{p} \sum_{j=1}^p x_t^i x_t^j, \\
x_{t+1}^i x_{t+1}^j &{}= (1 - \alpha -\eta \xi_t^i)(1 - \alpha -\eta \xi_t^j) x_t^i x_t^j + \alpha^2 (\tilde{x}_t)^2 \\
&{} \quad + \alpha (1-\alpha -\eta \xi_t^i) \tilde{x}_t x_t^i + \alpha (1-\alpha -\eta \xi_t^j) \tilde{x}_t x_t^j , \quad i \ne j.
\end{array}
\label{eq:mul_easgd2}
\end{equation}

Taking expectation on both sides of the Equation~\ref{eq:mul_easgd2}, we get
\begin{equation}
\begin{array}{r@{}l}
\mathbb{E} (x^i_{t+1})^2 &{}= [ (1 - \alpha - \eta \frac{\lambda}{\omega})^2 + \eta^2 \frac{\lambda}			{\omega^2}] \mathbb{E} (x^i_t)^2 + 2 \alpha (1 - \alpha - \eta \frac{\lambda}{\omega}) \mathbb{E} 		( \tilde{x}_t x^i_t) + 	\alpha^2 \mathbb{E} (\tilde{x}_t)^2, \\
\mathbb{E}(\tilde{x}_{t+1})^2 &{}= (1 - \beta)^2 \mathbb{E} (\tilde{x}_{t})^2 +
	2 \beta (1 - \beta) \frac{1}{p} \sum_{i=1}^p \mathbb{E} ( \tilde{x}_{t} x_t^i) +
	\frac{\beta^2}{p^2} \sum_{i=1,j=1}^p \mathbb{E} (x_t^i x_t^j), \\
\mathbb{E} ( \tilde{x}_{t+1} x_{t+1}^i) &{}=
	(1-\beta)(1-\alpha-\eta \frac{\lambda}{\omega}) \mathbb{E} (\tilde{x}_t x_t^i) +
	\alpha (1-\beta) \mathbb{E} (\tilde{x}_t)^2 \\
	&{} \quad + (1-\alpha-\eta \frac{\lambda}{\omega}) \frac{\beta}{p} \sum_{j=1}^p \mathbb{E} (x_t^i x_t^j)
+ \alpha \beta ( \frac{1}{p} \sum_{j=1}^p \mathbb{E} ( \tilde{x}_t x_t^j )), \\
\mathbb{E} (x_{t+1}^i x_{t+1}^j) &{}= (1 - \alpha -\eta \frac{\lambda}{\omega})(1 - \alpha -\eta \frac{\lambda}{\omega}) \mathbb{E}( x_t^i x_t^j) + \alpha^2 \mathbb{E} (\tilde{x}_t)^2 \\
&{} \quad + \alpha (1-\alpha -\eta \frac{\lambda}{\omega}) \mathbb{E} (\tilde{x}_t x_t^i) + \alpha (1-\alpha -\eta \frac{\lambda}{\omega}) \mathbb{E} ( \tilde{x}_t x_t^j ), \quad i \ne j.
\end{array}
\label{eq:mul_easgd2e}
\end{equation}

We can reduce the above system (Equation~\ref{eq:mul_easgd2e}) into a closed form by introducing
\begin{equation*}
	a_t = \mathbb{E} [ (\tilde{x}_{t})^2], \quad b_t = \frac{1}{p} \sum_{i=1}^{p} \mathbb{E} [ (x^i_{t})^2], \quad
	c_t = \frac{1}{p} \sum_{i=1}^{p} \mathbb{E} [ \tilde{x}_t x^i_{t} ], \quad d_t = \frac{1}{p^2} \sum_{i=1}^{p} \sum_{j=1}^{p} \mathbb{E} [ x^i_{t} x^j_{t} ].
\end{equation*}
We get $(a_{t+1}, b_{t+1}, c_{t+1} , d_{t+1})^T = M (a_{t}, b_{t}, c_{t} , d_{t})^T$, where
\begin{equation}
M = \left(
\begin{array}{cccc}
(1-\beta)^2 & 0 & 2\beta(1-\beta) & \beta^2 \\
\alpha^2 & (1-\alpha-\eta \frac{\lambda}{\omega})^2+ \eta^2 \frac{\lambda}{\omega^2} & 2 \alpha (1-\alpha-\eta \frac{\lambda}{\omega}) & 0 \\
\alpha (1-\beta) & 0 & (1-\beta)(1-\alpha-\eta \frac{\lambda}{\omega}) + \alpha \beta & (1-\alpha-\eta \frac{\lambda}{\omega})\beta \\
\alpha^2 & \eta^2 \frac{\lambda}{p \omega^2} & 2 \alpha (1-\alpha - \eta \frac{\lambda}{\omega}) & (1-\alpha-\eta \frac{\lambda}{\omega})^2
\end{array} \right).
\label{eq:mul_easgd2m}
\end{equation}

We will analyze the spectral norm of the matrix $M$ in the limit $p$ goes to infinity.
We focus on the question that whether the stability region (i.e. the spectral norm be smaller than one) with respect to the learning rate $\eta$ will be enlarged for large $p$.

\subsubsection{Case I: $\alpha = \beta / p$}
\begin{figure}[p]
\center
\includegraphics[trim=0in 0in 0in 0in,clip,width = 6in]{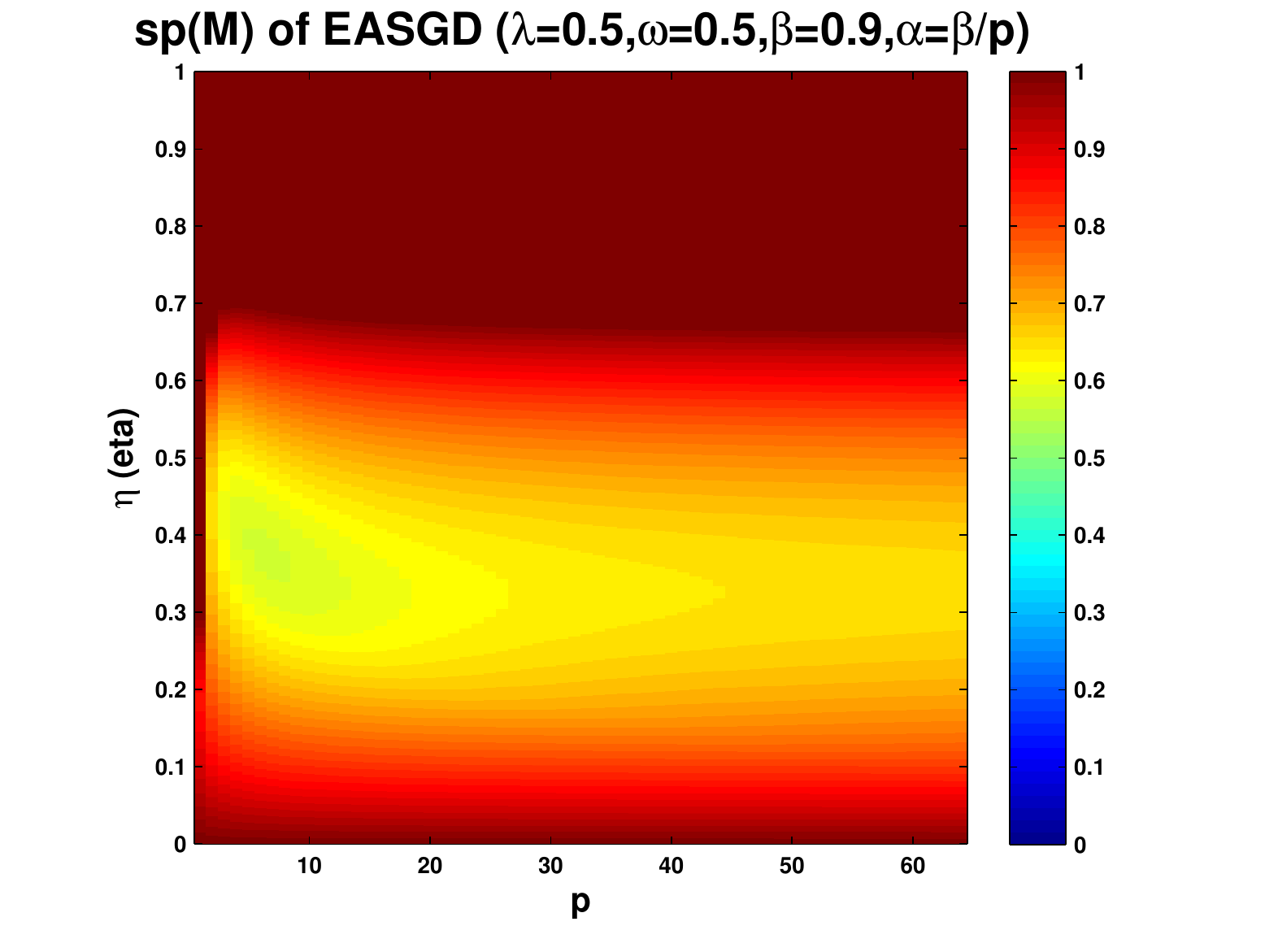}
\caption{The largest absolute eigenvalue of the matrix $M$ in Equation~\ref{eq:mul_easgd2m} as a function of the learning rate $\eta \in (0,1)$ and the number of workers $p \in [1,64]$. $\lambda=0.5$, $\omega=0.5$, $\beta=0.9$, $\alpha=\beta/p$.}
\label{fig:ch5_mul_easgd_stability1_05}
\end{figure}

\begin{figure}[p]
\center
\includegraphics[trim=0in 0in 0in 0in,clip,width = 6in]{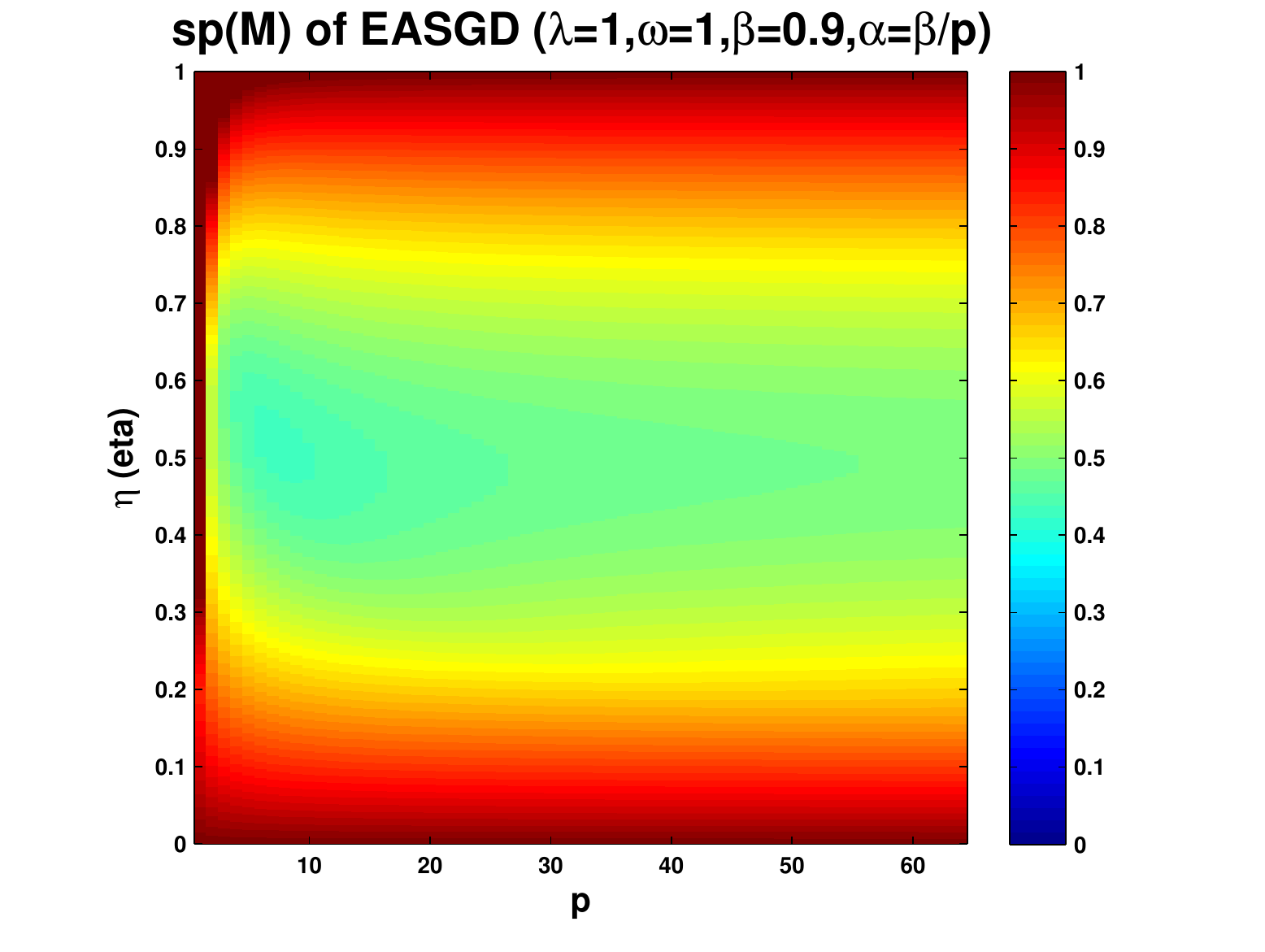}
\caption{The largest absolute eigenvalue of the matrix $M$ in Equation~\ref{eq:mul_easgd2m} as a function of the learning rate $\eta \in (0,1)$ and the number of workers $p \in [1,64]$. $\lambda=1$, $\omega=1$, $\beta=0.9$, $\alpha=\beta/p$.}
\label{fig:ch5_mul_easgd_stability1_1}
\end{figure}

\begin{figure}[p]
\center
\includegraphics[trim=0in 0in 0in 0in,clip,width = 6in]{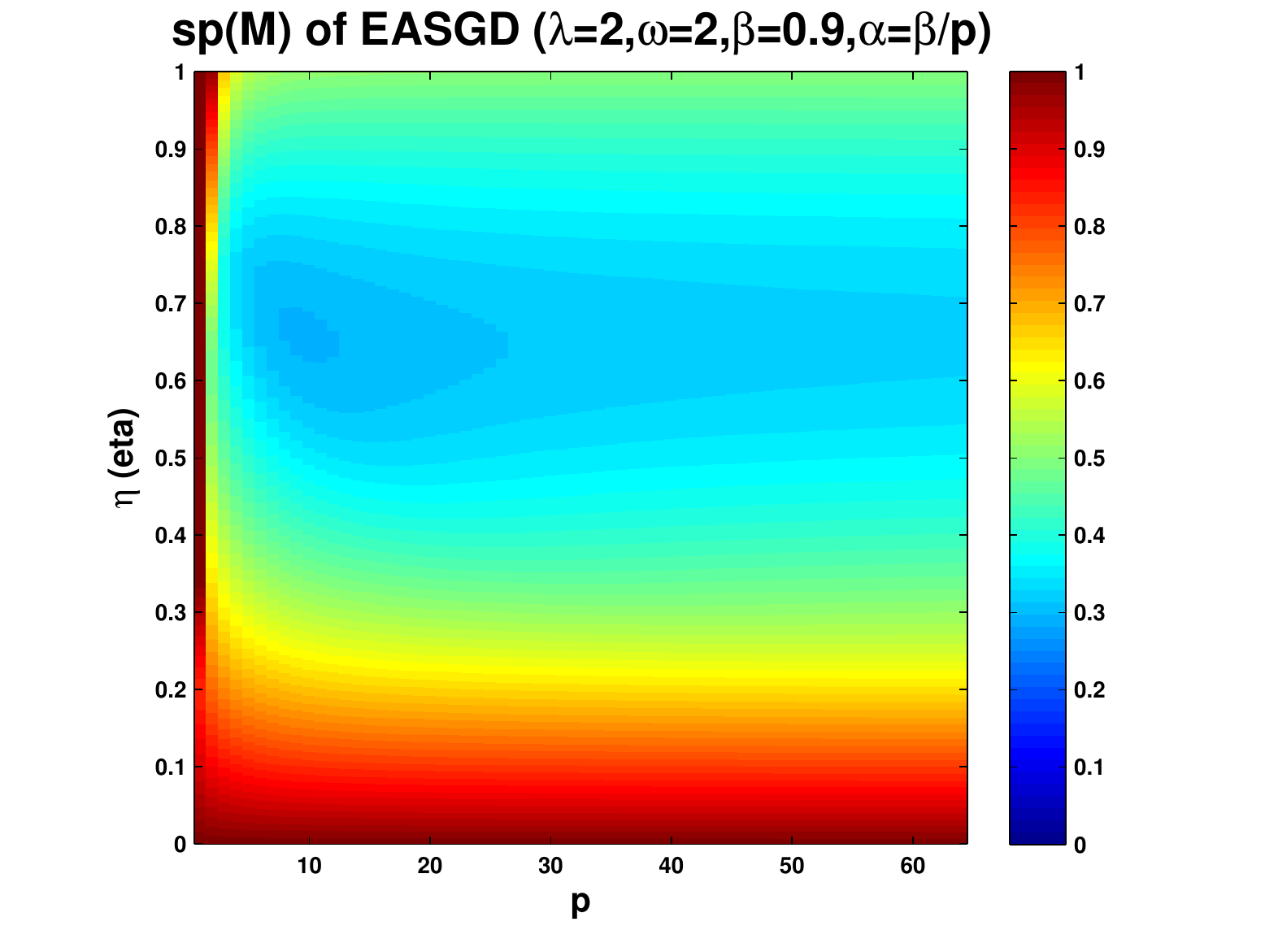}
\caption{The largest absolute eigenvalue of the matrix $M$ in Equation~\ref{eq:mul_easgd2m} as a function of the learning rate $\eta \in (0,1)$ and the number of workers $p \in [1,64]$. $\lambda=2$, $\omega=2$, $\beta=0.9$, $\alpha=\beta/p$.}
\label{fig:ch5_mul_easgd_stability1_2}
\end{figure}

\begin{figure}[p]
\center
\includegraphics[trim=0in 0in 0in 0in,clip,width = 6in]{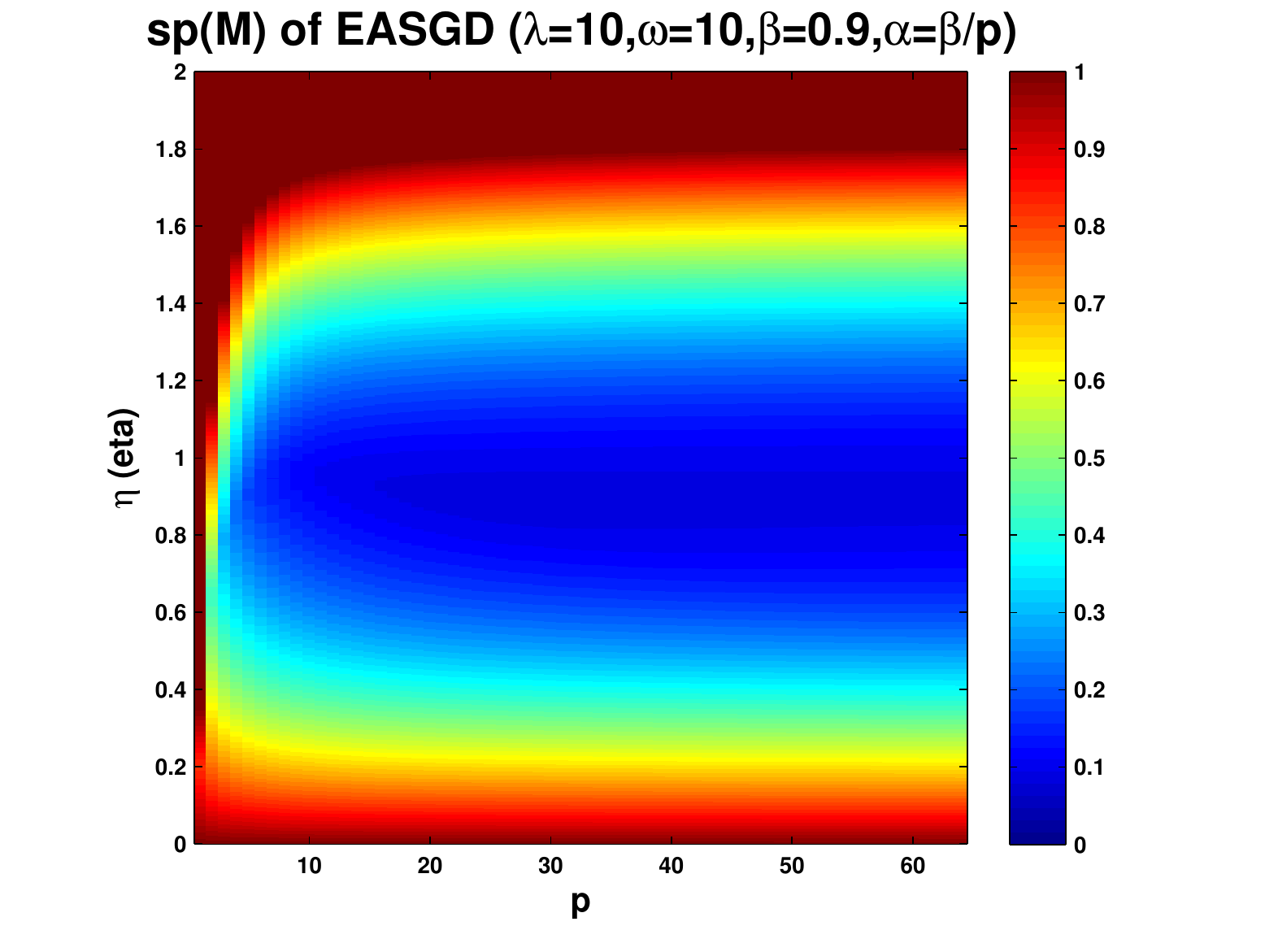}
\caption{The largest absolute eigenvalue of the matrix $M$ in Equation~\ref{eq:mul_easgd2m} as a function of the learning rate $\eta \in (0,2)$ and the number of workers $p \in [1,64]$. $\lambda=10$, $\omega=10$, $\beta=0.9$, $\alpha=\beta/p$. The minimal $\mbox{sp}(M)=0.0868$ is achieved at $p=29$ and $\eta = 0.8929$.}
\label{fig:ch5_mul_easgd_stability1_10}
\end{figure}

The characteristic polynomial of $M$ in Equation~\ref{eq:mul_easgd2m} can be written as
\begin{equation*}
P(z)= [z - (1 - \beta)^2]
	[ z - (1 - \beta) (1-\eta \frac{\lambda}{\omega})]
	[z - (1 - 2 \eta \frac{\lambda}{\omega} + \eta^2 \frac{ \lambda (\lambda+1) }{\omega^2} )]
	[z - (1 - \eta \frac{\lambda}{\omega})^2] + O_z(\frac{1}{p}),
\end{equation*}
where $O_z(\frac{1}{p})$ is the remaining polynomial in $z$ of order $\frac{1}{p}$ and higher.
The four leading order eigenvalues of $P(z)$ are $z_1 = (1-\beta)^2$,
$z_2 = (1 - \beta) (1-\eta \frac{\lambda}{\omega})$,
$z_3 = (1 - 2 \eta \frac{\lambda}{\omega} + \eta^2 \frac{ \lambda (\lambda+1) }{\omega^2} )$,
$z_4 = (1 - \eta \frac{\lambda}{\omega})^2$. We need necessarily $\beta \in (0,2)$, and in such case
the stability condition for $\eta$ is the same as the \textit{SGD} case (c.f. Equation~\ref{eq:mulmbsgdm2}),
i.e. $| 1 - 2 \eta \frac{\lambda}{\omega} + \eta^2 \frac{ \lambda (\lambda+1) }{\omega^2} | < 1$.
Thus the stability region of \textit{EASGD} remains the same as \textit{SGD}
in the limit $p$ goes to infinity, i.e.
\begin{equation*}
	0< \eta < \frac{2 \omega}{\lambda + 1}.
\end{equation*}
In the Figure~\ref{fig:ch5_mul_easgd_stability1_05},~\ref{fig:ch5_mul_easgd_stability1_1},~\ref{fig:ch5_mul_easgd_stability1_2} and~\ref{fig:ch5_mul_easgd_stability1_10}, we illustrate the stability region for different $\lambda$, $\omega$ and $p$. Compared with the \textit{MSGD} method in the Figure~\ref{fig:ch5_mul_msgd_opt},
\textit{EASGD} improves its optimal convergence rate:
\begin{itemize}
\item $\lambda=0.5, \omega=0.5$: $\mbox{sp}(M)=0.5742$ at $p=6$ and $\eta=0.3814$ for \textit{EASGD} vs. $\mbox{sp}(M) = 0.6667$ for \textit{MSGD} with $\eta=\frac{\lambda}{\omega+1}=\frac{1}{3}$ and $\delta=0$.
\item $\lambda=1, \omega=1$: $\mbox{sp}(M)=0.4317$ at $p=7$ and $\eta=0.5225$ vs. $\mbox{sp}(M) = 0.5$ for \textit{MSGD} with $\eta=\frac{\lambda}{\omega+1}=\frac{1}{2}$ and $\delta=0$.
\item $\lambda=2, \omega=2$: $\mbox{sp}(M)=0.2945$ at $p=9$ and $\eta=0.6647$ vs. $\mbox{sp}(M) = 0.3333$ for \textit{MSGD} with $\eta=\frac{\lambda}{\omega+1}=\frac{2}{3}$ and $\delta=0$.
\end{itemize}

Note that \textit{EASGD}'s optimal convergence rate is achieved for some finite $p$.
This is in contrast to the mini-batch \textit{SGD} method (c.f. Equation~\ref{eq:muloptlr}).

\subsubsection{Case II: $\alpha$ is independent of $\beta$ and $p$}
The characteristic polynomial of $M$ of Equation~\ref{eq:mul_easgd2m} can be written as
\begin{equation*}
	P(z)= (z - z_1)(z-z_2)(z-z_3)(z-z_4) + O_z(\frac{1}{p}),
\end{equation*}
where $z_1 = (1-\beta)(1-\eta \frac{\lambda}{\omega})-\alpha$,
$z_2 = (1-\alpha)^2 - 2(1-\alpha) \eta \frac{\lambda}{\omega} + \eta^2 \frac{\lambda (\lambda+1)}{\omega^2} $,
$z_3+z_4 = 2 + \alpha^2 - (2-\beta)\beta - (2- \eta \frac{\lambda}{\omega}) \eta \frac{\lambda}{\omega}
- 2 \alpha (1- \beta - \eta \frac{\lambda}{\omega})$, $z_3 z_4 = z_1^2$.

\begin{figure}[p]
\center
\includegraphics[trim=0in 0in 0in 0in,clip,width = 6in]{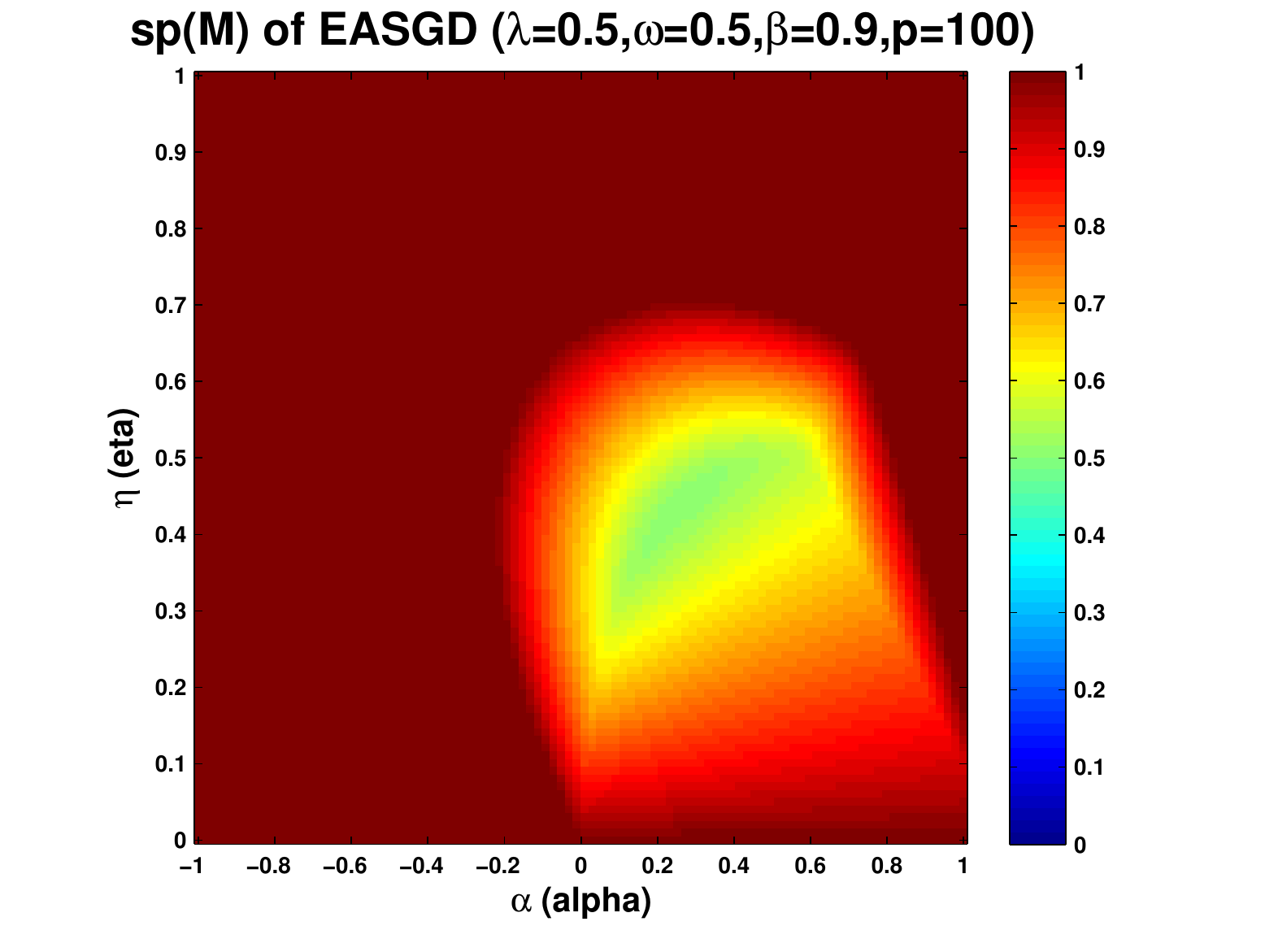}
\caption{The largest absolute eigenvalue of the matrix $M$ in Equation~\ref{eq:mul_easgd2m} as a function of the learning rate $\eta \in (0,1)$ and the moving rate $\alpha \in (-1,1)$. $\lambda=0.5$, $\omega=0.5$, $\beta=0.9$, $p=100$. The minimal $\mbox{sp}(M)=0.5024$ is achieved at $\eta=0.4343$ and $\alpha=0.2525$.}
\label{fig:ch5_mul_easgd_stability2_05_p100}
\end{figure}

These four eigenvalues depend on $\lambda$, $\omega$, $\beta$, $\alpha$, $p$ and $\eta$. It's not as clear here on how to choose an optimal $\alpha$. We can however still find the stability region with respect of $\eta$ in the limit $p \to \infty$. We ask given $\lambda$, $\omega$, $\beta$ and $\alpha$, what is the range of $\eta$ such that the four eigenvalues lie on the unit circle of the complex plane. We also observe that the optimal $\alpha$ can be positive for large $p$, in contrast to the additive noise case (c.f. Figure~\ref{fig:ch5_mul_easgd_stability2_05_p100} and~\ref{fig:ch5_lin_easgd_sp_original}).
Thus to simplify our analysis, we only consider $\eta>0$, $\alpha \in (0,1)$ and $\beta \in (0,1)$.
It turns out that in this case,
\begin{itemize}
\item $|z_1| < 1$: $0 < \eta < \frac{2 - \alpha -\beta} { 1-\beta } \frac{\omega}{\lambda}$.
\item $|z_2| < 1$: $0 < \eta < \frac{\omega(1-\alpha)}{\lambda+1} + \frac{\omega^2}{\lambda(\lambda+1)}
\sqrt{ \frac{\lambda^2}{\omega^2} + \frac{\lambda}{\omega^2} ( 2 \alpha - \alpha^2)}$.
\item $|z_3| < 1$ and $|z_4| < 1$: $ 0 < \eta < \frac{ 4 - 2 \alpha - 2 \beta}{ 2 - \beta } \frac{\omega}{\lambda}$.
\end{itemize}
We can also check that $ \frac{2 - \alpha -\beta} { 1-\beta } \frac{\omega}{\lambda} > \frac{ 4 - 2 \alpha - 2 \beta}{ 2 - \beta } \frac{\omega}{\lambda}$, thus we only need to maximize the $\min_{0<\alpha<1} \{\frac{ 4 - 2 \alpha - 2 \beta}{ 2 - \beta } \frac{\omega}{\lambda} , \frac{\omega (1-\alpha)}{\lambda+1} + \frac{\omega}{\lambda(\lambda+1)} \sqrt{ \lambda^2 + \lambda ( 2 \alpha - \alpha^2)} \}$. We now prove that the maximum is achieved at $\alpha = 1 - \sqrt{\lambda}$. In fact, this is the maximum of the second term, and at that $\alpha$, the first term equals $\frac{ 4 - 2 \alpha - 2 \beta}{ 2 - \beta } \frac{\omega}{\lambda} = \frac{ 2 + 2 \sqrt{ \lambda } -2 \beta}{2 - \beta} \frac{\omega}{\lambda}$, and
the second term equals $\frac{\sqrt{\lambda}}{\lambda+1}(\omega+\frac{\omega}{\lambda}) =
\frac{\omega}{\sqrt{\lambda}}$. The first term being larger than the second term, thus we conclude that
the maximal value $\frac{\omega}{\sqrt{\lambda}}$ is indeed achieved at $\alpha = 1 - \sqrt{\lambda}$.

Based on the Equation~\ref{eq:muloptlr}, the stability region for \textit{SGD} is $0 < \eta < \frac{2 \omega}{\lambda+1}$; for mini-batch \textit{SGD} with $p \to \infty$, it is $ 0 < \eta < \frac{2 \omega}{\lambda}$.
For \textit{EASGD}, it is $0 < \eta < \frac{\omega}{\sqrt{\lambda}}$, with $p \to \infty$ and $\alpha = 1 - \sqrt{\lambda}$. In case $\lambda=0.5$, $\omega=0.5$, we can check from Figure~\ref{fig:ch5_mul_easgd_stability2_05_p100} that $\alpha = 1 - \sqrt{0.5} = 0.2929$ achieves the widest range of $\eta \in (0,\sqrt{0.5})$ for $M$ to be stable.

For the \textit{EAMSGD} method, it's much more difficult to analyze the second-order moment equation.
It involves a linear system of nine variables: $a_t = \mathbb{E}(\tilde{x}_t^2)$,
$b_t = \frac{1}{p} \sum_i \mathbb{E} (x_t^i)^2$,
$c_t = \frac{1}{p} \sum_{i} \mathbb{E} (\tilde{x}_t x^i_t)$,
$d_t = \frac{1}{p^2} \sum_{i,j} \mathbb{E} (x_t^i x_t^j)$,
$e_t = \frac{1}{p} \mathbb{E} (\tilde{x}_t v_t^i)$,
$f_t = \frac{1}{p^2} \sum_{i,j} \mathbb{E} (v_t^i v_t^j)$,
$g_t =\frac{1}{p} \mathbb{E} (v_t^i)^2$,
$h_t = \frac{1}{p} \sum_i \mathbb{E} (x_t^i v_t^i)$,
$k_t = \frac{1}{p^2} \sum_{i,j} \mathbb{E} (x_t^i v_t^j)$.
The linear matrix is quite complicated, so we will not present it here.

\section{A non-convex case}
\label{sec:non-convex}

Here's an amusing non-convex case which sheds some light on
when \textit{EASGD} will work and when it will not work.
Recall our formalization of the problem in Chapter~\ref{sec:ch_intro}, Equation~\ref{eq:penalty}:
\begin{equation*}
\min_{x^1,\ldots,x^p,\tilde{x}} \sum_{i=1}^p \mathbb{E} [f(x^i,\xi^i)] + \frac{\rho}{2}\|x^i - \tilde{x}\|^2,
\end{equation*}
where we refer to $x^i$'s as local variables and we refer to $\tilde{x}$ as a center variable.

If $f(x^i,\xi^i) = \frac{1}{4} (1-(x^i)^2)^2$, which is deterministic (independent of $\xi^i$) and one-dimensional. We see that $f$ has two minimum $x^i = 1$ and $x^i = -1$. It also has a saddle point at $x^i = 0$. We ask how large $\rho$ should be such that the $x^i$'s will have a common minimum. Let's assume that $p=2$, we see that if $\rho=0$, then $x^1=-1$ and $x^2=1$ can be a stable solution. This is the scenario that the 'elasticity' of \textit{EASGD} is broken. We have observed a related phenomenon in \textit{EAMSGD} when the communication period is too large ($\tau$ big) in the Figure~\ref{fig:CIFAR3_supp} of Chapter $4$. More formally, we have the following objective for $p=2$ workers,
\begin{equation}
\min_{x ,y,z \in \mathbb{R}} \frac{1}{4} (1-x^2)^2 + \frac{1}{4} (1-y^2)^2 +
\frac{\rho}{2}\|x - z \|^2 + \frac{\rho}{2}\|y - z \|^2.
\label{eq:egnonconvex}
\end{equation}

The partial derivative of our objective in Equation~\ref{eq:egnonconvex} with respect to
$x$, $y$ and $z$ is:
\begin{equation}
\begin{array}{r@{}l}
\frac{\partial}{\partial x} &{}= (x^2-1) x + \rho (x-z) \\
\frac{\partial}{\partial y} &{}= (y^2-1) y + \rho (y-z) \\
\frac{\partial}{\partial z} &{}= \rho (z - x) + \rho (z-y).
\end{array}
\label{eq:egnonconvexder}
\end{equation}

The question is that what are all the critical points of the objective in Equation~\ref{eq:egnonconvex}. By setting the
derivative to zero in Equation~\ref{eq:egnonconvexder}, they should satisfy $z = \frac{x+y}{2}$ and
\begin{equation}
\begin{array}{r@{}l}
(x^2-1) x + \rho (x- \frac{x+y}{2}) &{}= 0 \\
(y^2-1) y + \rho (y-\frac{x+y}{2}) &{}= 0.
\end{array}
\label{eq:nonconvreduced}
\end{equation}

Observing three special cases: $x=y=z=1$, $x=y=z=-1$ and $x=y=z=0$. We can guess that $x=-y$ is also a solution.
In fact, they should satisfy $z=0$, $(x^2-1) x + \rho ( x - 0) = 0$ and $(y^2-1) y + \rho ( y - 0) = 0$. We have thus either $x=0$ or $x^2 = 1 - \rho$ (resp. $y=0$ or $y^2 = 1 - \rho$). If $\rho < 1$, then we indeed find a real critical point $x=\sqrt{1-\rho}$, $y=-\sqrt{1-\rho}$, $z=0$. Is this critical point stable? To answer this, we compute the Hessian of our objective in Equation~\ref{eq:egnonconvex} with respect to
$x$, $y$ and $z$:
\begin{equation}
H = \left [
\begin{array}{ccc}
3x^2 -1 + \rho & 0 & -\rho \\
0 & 3y^2 -1 + \rho & -\rho \\
-\rho & -\rho & 2 \rho.
\end{array} \right].
\label{eq:hessnonconv}
\end{equation}

\begin{figure}[h]
\center
\includegraphics[trim=0in 0in 0in 0in,clip,width = 6in]{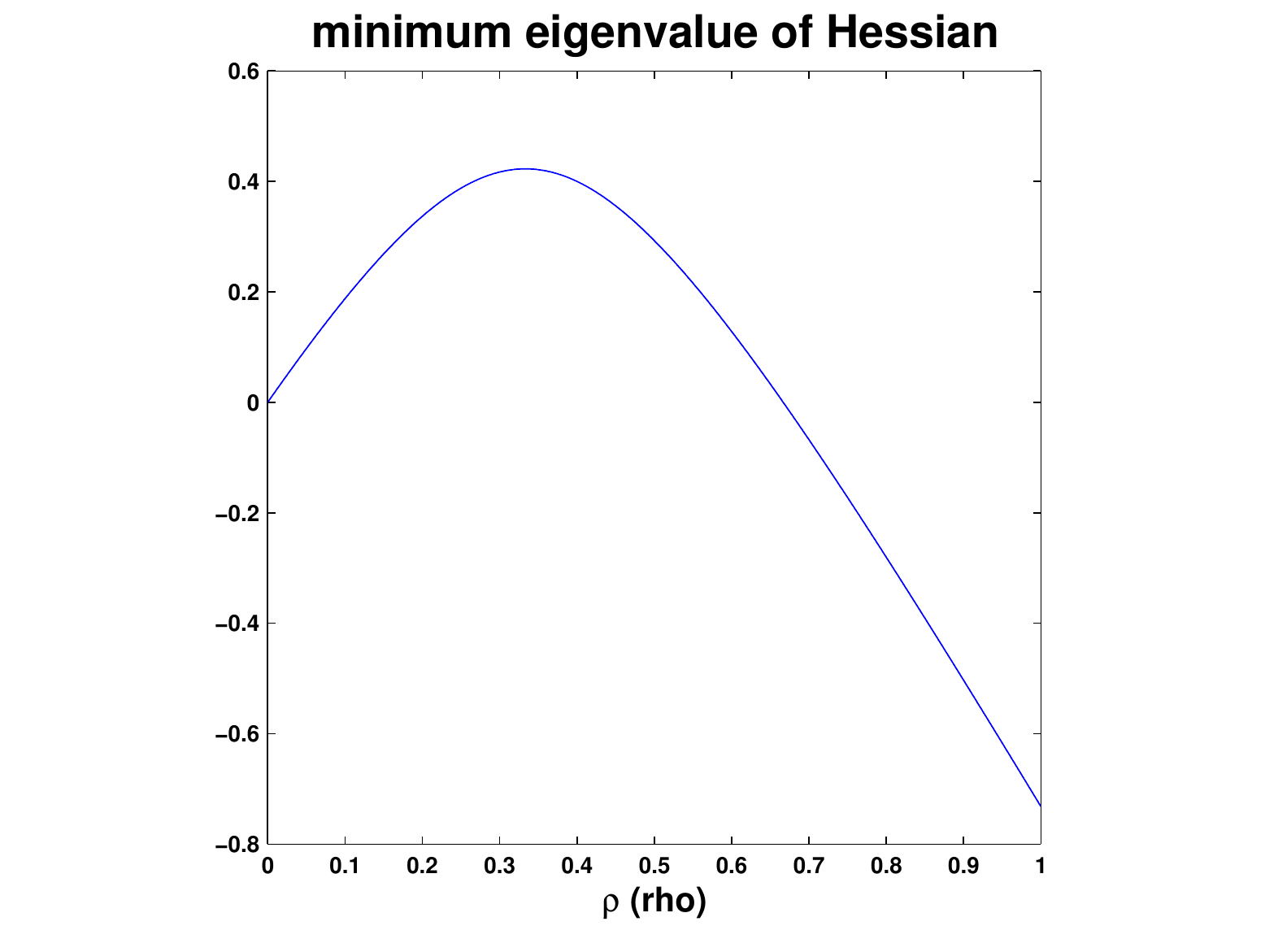}
\caption{The smallest eigenvalue of the Hessian matrix $H$ in Equation~\ref{eq:hessnonconv} as a function of the penalty term $\rho$, evaluated at the critical point $x=\sqrt{1-\rho}$, $y=-\sqrt{1-\rho}$, $z=0$.}
\label{fig:ch5_nonconvex_hessian}
\end{figure}

Evaluating this Hessian at the critical point $x=\sqrt{1-\rho}$, $y=-\sqrt{1-\rho}$, $z=0$, we can compute its
smallest eigenvalue and see when it is positive definite. Figure~\ref{fig:ch5_nonconvex_hessian} shows that the smallest eigenvalue is always positive in the range $\rho \in (0,2/3)$. This suggests that this critical point can indeed be stable, i.e. it is a local optimum introduced by \textit{EASGD} when the penalty term $\rho$ is small enough.

For completeness, we now prove that all the critical points are either of the form $x=y$ or of the form $x=-y$. In fact, adding the two Equations in~\ref{eq:nonconvreduced} gives us $(x^2-1)x + (y^2-1)y = 0$, i.e. $x^3 + y^3 = x + y = (x^2 - xy + y^2) ( x+y)$. Subtracting these two Equations gives us $x^3 - x - y^3 + y + \rho (x-y) = 0$, i.e. $x^3 - y^3 = (1-\rho) (x-y) = (x^2 + xy + y^2) (x-y)$. If $x \neq y$ and $x \neq -y$, then
\begin{equation}
\begin{array}{r@{}l}
x^2 - xy + y^2 &{}= 1 \\
x^2 + xy + y^2 &{} = 1 - \rho.
\end{array}
\label{eq:nonconvreduced2}
\end{equation}

Adding and subtracting again the two Equations in~\ref{eq:nonconvreduced2}, we obtain $x^2 + y^2 = 1 - \frac{\rho}{2}$, and $ x y = -\frac{\rho}{2}$. There's no real solution satisfying these two conditions for any $\rho > 0$.

\newpage
\chapter{Scaling up Elastic Averaging SGD}
\label{sec:ch_tree}

This chapter discusses how to scale up the \textit{EASGD} method to hundreds and thousands of processors. In Section~\ref{sec:easgd_tree}, we first propose a tree-structured extension of the \textit{EASGD} method called \textit{EASGD Tree}. The basic idea and the design principle are discussed in Section~\ref{sec:easgd_tree_algo}, and the numerical results are presented in Section~\ref{sec:easgd_tree_result}.
We present two different communication schemes for the \textit{EASGD Tree} method.
As we had seen the advantage of \textit{EAMSGD}, we also accelerate \textit{EASGD Tree} with \textit{Nesterov}'s momentum method.
In Section~\ref{sec:unify}, we unify the \textit{DONWPOUR} method and the \textit{EASGD} method by considering a \textit{Gauss-Seidel} reformulation of the \textit{EASGD} update rules in the synchronous scenario. This unification suggests the possibility of using both \textit{DONWPOUR} and \textit{EASGD} under the \textit{EASGD Tree}.
It also suggests that in-between the \textit{DONWPOUR} and the \textit{EASGD} method there may be some even better method.

\section{\textit{EASGD Tree}}
\label{sec:easgd_tree}

The original motivation of \textit{EASGD Tree} is to run \textit{SGD} at multiple time scales, where each scale corresponds to the use of a different learning rate. It naturally gives rise to a hierarchical tree structured organization of the processors. In literature, the tree idea has shown up in various contexts. For example, the tree was used to scale up the asynchronous \textit{SGD} method by aggregating the delayed gradients computed by the intermediate nodes and the leaf nodes~\cite{agarwal2011distributed}. It can also be used to efficiently implement the Broadcast/AllReduce operation as in MPI~\cite{agarwal2014reliable,zhang2015staleness}.
The benefit of using a tree is that the number of links to connect a very large number of nodes is minimal. The trade-off is that the connectivity of the whole tree is not robust to the link failure. However, the tree structure has its own charm for its simplicity. The main theoretical challenge is to understand its convergence property in terms of the root of the tree. This still remains open.

\subsection{The algorithm}
\label{sec:easgd_tree_algo}
We start from introducing the \textit{EASGD Tree} algorithm.
There are two important aspects that can guide us to design such an algorithm, one is the computation, the other is the communication.

\begin{itemize}
\item Do we need intermediate node to perform the gradient computation?
We had hoped that each tree node runs \textit{SGD} in parallel with a smaller and smaller learning rates as the depth of the tree node decreases (from bottom to top). Each intermediate node (i.e. except the leaf node) is expected to average out the noise of its children. The root can thus achieve the smallest variance. One benefit that each node performs its own gradient computation is that it can reduce the bias of the spatial averaging of its children away from the local optimum. The disadvantage is that each node will have the overhead for both computation and communication. In case each tree node uses only one CPU core, it needs to tradeoff the computation speed for the communication throughput. In addition, in the HPC environment (where we run our experiments), we have not observed any visible performance improvement for the intermediate node to perform the gradient computation. It's also complicated to choose a proper learning rate decay for different levels of the tree nodes. On the other hand, if we were to remove the intermediate node's gradient computation, we can still preserve the multi-scale variance reduction property using the elastic averaging. Moreover, the root node of the tree behaves as the center variable of the \textit{EASGD} method. Thus in the following we only discuss the \textit{EASGD Tree} method without intermediate node's gradient computation.

\item There are at least two different ways to design the communication protocol. One is locally synchronous, the other is fully asynchronous. We observe that the local synchronization protocol (as in our asynchronous \textit{EASGD} in Section~\ref{sec:asyncEASGD}) may cause extra waiting time for the parent and the children nodes, but it can give better convergence property (due to a smaller parameter staleness).
The fully asynchronous protocol can in theory hide completely this waiting time by allowing each node asynchronously broadcasts its parameter to its neighbours without blocking any computation. In the scale that we simulate the \textit{EASGD Tree}, we have to use the fully asynchronous protocol as it saves significantly the time which is spent on the chain of waiting. In our HPC environment, the fully asynchronous protocol works well if the network link is stable over time.

\item The communication cost is different within a machine and across a machine. We would thus structure the tree nodes from bottom-up, i.e. the leaf nodes are allocated on one machine so that their variance can be averaged out the quickest possible. The intermediate nodes including the root of the tree will then have to communicate across the machine boundary (in our HPC environment, there are 20 physical CPU cores per machine). These communication should happen less frequently. We will thus introduce two levels of the communication periods. The first communication period is between the leaf nodes and their parents. The second (smaller) one is between the intermediate nodes. We will also distinguish the push up (toward parent) period and the push down (toward children) period. With this, we can trade-off more push up communication for less push down communication.
\end{itemize}

\begin{figure}[h]
\center
\includegraphics[trim=0cm 0cm 0cm 0cm,clip,width = 3in]{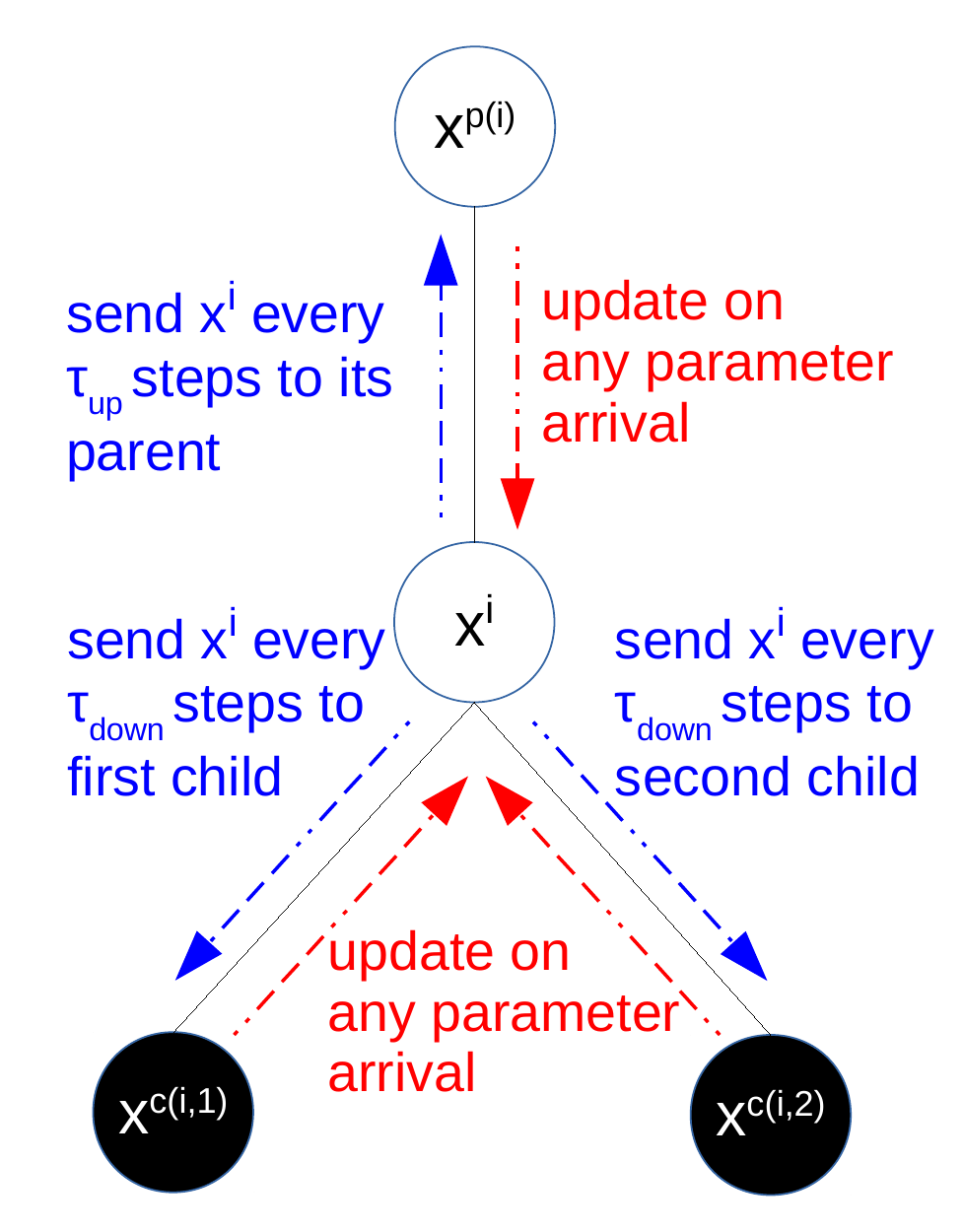}
\caption{The behavior of \textit{EASGD Tree} in Algorithm~\ref{alg:easgdtree}.}
\label{fig:EASGDTree4}
\end{figure}

\begin{figure}[h]
\center
\includegraphics[trim=0cm 0cm 0cm 0cm,clip,width = 6in]{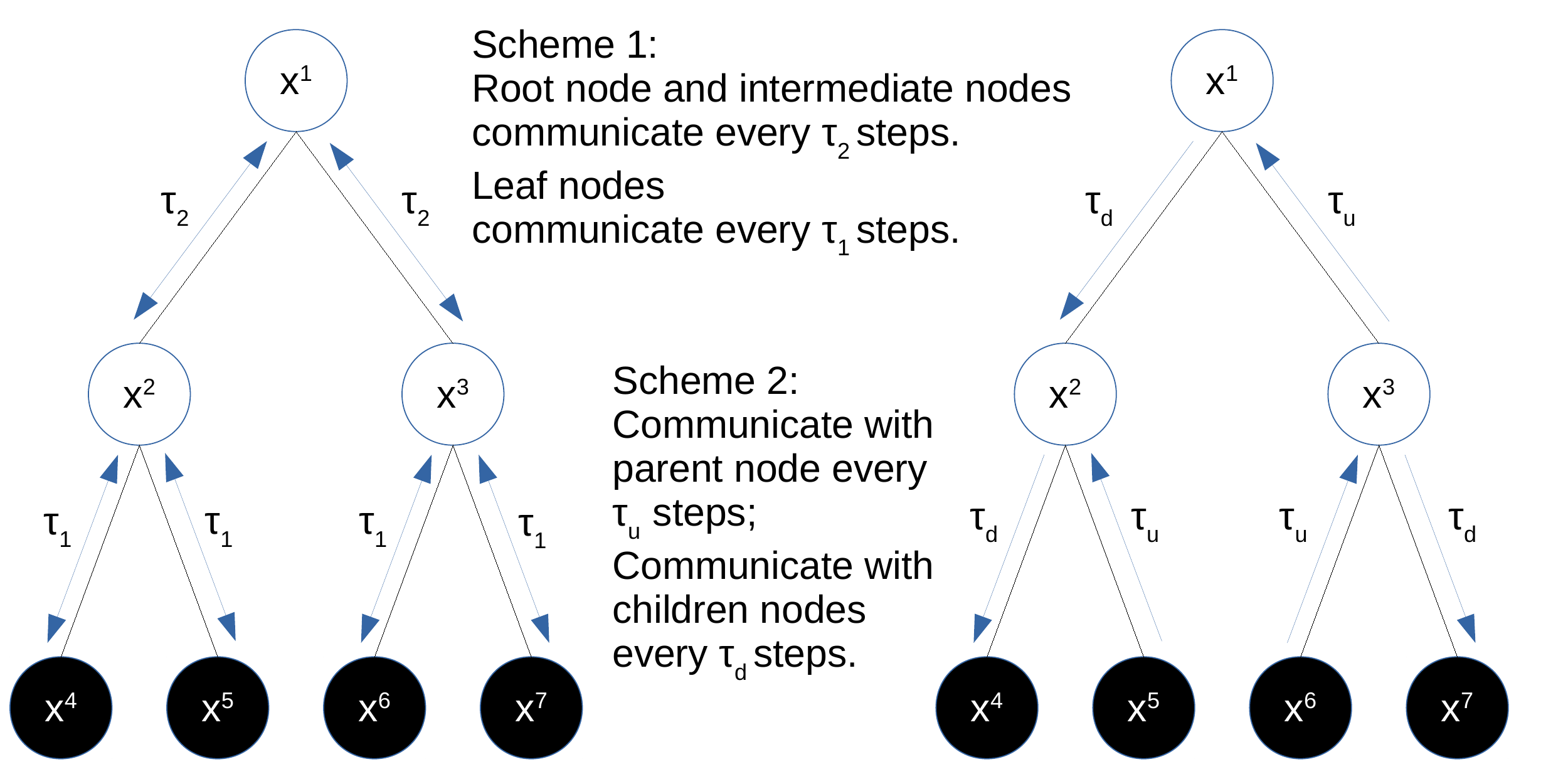}
\caption{The two communication schemes of the \textit{EASGD Tree}.}
\label{fig:easgdTree3}
\end{figure}

Based on the above discussion, we now present the \textit{EASGD Tree} in Algorithm~\ref{alg:easgdtree}.
Figure~\ref{fig:EASGDTree4} illustrates this. This generic pseudocode does not distinguish the leaf node, the intermediate node and the root node, so let's describe their difference first.
As in \textit{EASGD}, each node starts with
the same initial parameter $x_0$ and sets its local clock $t^i$ to $0$. It performs non-blocking read (Irecv) and non-blocking send (Isend) with its parent and children (if any). Every $\tau_{up}$ local steps, it sends its parameter to its parent. Every $\tau_{down}$ local steps (number of local gradient updates), it sends its parameter to the $d$ children. We denote $d$ to be the degree of the $d$-ary tree we consider here. Each node also needs to check if there's any new arrival of the parameter from its parent or its $d$ children. This needs to be interleaved with the gradient computation to achieve high I/O throughput. On receiving a new parameter, a moving average with the moving rate $\alpha$ is performed. If it is the leaf node, then a stochastic gradient descent step with learning rate $\eta$ is performed per local step; otherwise it replaces the gradient descent step with a while loop, looping for roughly the same amount of time as a gradient step, just to apply the moving average on any new parameter arrival. Note that for each node, they can use different communication period $\tau_{up}$ and $\tau_{down}$. We discuss two such possible schemes below. They are illustrated in Figure~\ref{fig:easgdTree3}.

The first scheme is based on the idea of multi-scale averaging. We would like to have a fast averaging process at the bottom level, and a slower averaging process at the top and intermediate level. Let $\tau_1$ be the
fast communication period between the leaf nodes and their parents. Let $\tau_2$ be the slow
communication period between the intermediate nodes.
For the leaf nodes, i.e. without child, we set $\tau_{up} = \tau_1$, $\tau_{down} = NaN$.
For their parents (which should sit on the same machine as their children), we set $\tau_{up} = \tau_2$, $\tau_{down} = \tau_1$. For the rest of the intermediate nodes, except for the root node, we set $\tau_{up} = \tau_{down} = \tau_2$. Lastly, for the root node, we set $\tau_{up} = NaN$, $\tau_{down} = \tau_2$.

The second scheme is to mimic the behavior of \textit{EASGD} for $\beta$ being large and $\alpha$ being small (c.f. Equation~\ref{eq:movingaverage1} and~\ref{eq:movingaverage2}). We use a large $\tau_{up}$ to represent large $\beta$, and a small $\tau_{down}$ to represent small $\alpha$. For simplicity, we set each tree node with the same $\tau_{up}=\tau_{u}$ and $\tau_{down}=\tau_{d}$, except for the leaf node with $\tau_{down}=NaN$ and the root node with $\tau_{up}=NaN$.

There's a subtle difference between \textit{EASGD} and \textit{EASGD Tree}. As we discussed in Section~\ref{sec:asyncEASGD}, the design of \textit{EASGD} follows the \textit{Jacobi} form. In
\textit{EASGD Tree}, we use instead the \textit{Gauss-Seidel} form to perform the moving average. This is because the time when the parameter will arrive is not predictable in general, so we'd rather perform the update just-in-time. Nevertheless, we should be careful to avoid performing the moving average during the gradient update. Moreover, there's a deeper connection between the \textit{Jacobi/Gauss-Seidel} method and \textit{EASGD}/\textit{DOWNPOUR} method that we shall discuss in the next Section~\ref{sec:unify}.
Thus \textit{EASGD Tree} is somewhere in-between \textit{EASGD} and \textit{DOWNPOUR}.

\begin{algorithm}
\caption{\textit{EASGD Tree} \newline Processing by worker $i$, with its parent $p(i)$ and children $\{c(i,j)| j=1,\ldots,d\}$.}
\label{alg:easgdtree}
\begin{tabular}{l}
\textbf{Input:} learning rate $\eta$, moving rate $\alpha$, communication period $\tau_{up}$ and $\tau_{down} \in \mathbb{N}$ \\		
\textbf{Initialize:} $x^i = x_0$, $t^i = 0$\\
\hline
\textbf{Irecv} $x^{p(i)}$ from parent $p(i)$ \\
\textbf{Irecv} $x^{c(i,j)}$ from children $\{c(i,j)| j=1,\ldots,d\}$ \\
\textbf{Repeat}\\
\quad \textbf{if} ($\tau_{up} $ divides $t^i $) \textbf{then}\\
\quad \quad \textbf{Isend} $x^i$ to parent $p(i)$ \\
\quad \textbf{end} \\
\quad \textbf{if} ($\tau_{down}$ divides $t^i$) \textbf{then}\\
\quad \quad \textbf{Isend} $x^i$ to children $\{c(i,j)| j=1,\ldots,d\}$ \\
\quad \textbf{end} \\
\quad \textbf{if} there is new arrival of $x^{p(i)}$ \textbf{then}\\
\quad \quad $x^i \leftarrow x^i + \alpha (x^{p(i)}-x^i)$ \\
\quad \textbf{end} \\
\quad \textbf{if} there is new arrival of $x^{c(i,j)}$ \textbf{then}
\\
\quad \quad $x^i \leftarrow x^i + \alpha (x^{c(i,j)} - x^i )$ \\
\quad \textbf{end} \\
\quad $x^i \leftarrow x^i - \eta \nabla{f}(x^i,\xi^i_{t^i})$ \\
\quad $t^i \leftarrow t^i + 1$ \\
\textbf{Until forever}
\end{tabular}
\end{algorithm}

\subsection{The result}
\label{sec:easgd_tree_result}

In this section, we show empirically the advantage and the challenge to scale up the \textit{EASGD Tree} to a few hundreds of CPU cores. Since we are limited by the number GPUs available, all the results are run on CPUs. The experiment setup is thus different to our earlier results in Section~\ref{sec:exp_setup1}.
We report the results based on CIFAR-10 dataset using a low-rank convolutional neural-network model~\cite{DBLP:journals/corr/TaiXWE15}. The low-rank convolution approximation saves us a lot of CPU computation time, so we can simulate the long term behavior of the algorithm within 12 hours. We do not use any data augmentation as we did in Chapter~\ref{sec:ch_exp}, because it requires extra CPUs to pre-process the input data. Each data point is sampled uniformly random with replacement (different to what we did in Section~\ref{sec:exp_setup1}). As in~\cite{DBLP:journals/corr/TaiXWE15}, we center the input pixel values by the mean and the standard deviation for each of the three channels using the training data.
We also add $l_2$-regularization $\frac{\lambda}{2}\norm{}{x}^2$ to the loss function with $\lambda = 0.0001$.

For consistency, we describe the detailed model\footnote{see also in \url{https://github.com/chengtaipu/lowrankcnn/blob/master/cifar/models/baseline_lowrank_cudnn.lua}} using the notation in Section~\ref{sec:exp_setup1}:
$(3,32,32)
\xrightarrow[(1,5,1,1,0,2)]{C}
(10,32,32)
\xrightarrow[(5,1,1,1,2,0)]{C,R}
(96,32,32)
\xrightarrow[(2,2,2,2)]{P}
(96,16,16)
\xrightarrow[(1,5,1,1,0,2)]{C}
(51,16,16)
\xrightarrow[(5,1,1,1,2,0)]{C,R}
(128,16,16)
\xrightarrow[(2,2,2,2)]{P}
(128,8,8)
\xrightarrow[(1,5,1,1,0,2)]{C}
(51,8,8)
\xrightarrow[(5,1,1,1,2,0)]{C,R}
(256,8,8)
\xrightarrow[(2,2,2,2)]{P}
(256,4,4)
\xrightarrow[(1,1,1,1,0,0)]{C,R}
(64,4,4)
\xrightarrow[0.5]{D,L,R}
(256,1,1)
\xrightarrow[]{D,L,S}(10,1,1)$.

\begin{figure}[p]
\center
\includegraphics[trim=0cm 0cm 0cm 0cm,clip,width = 3in]{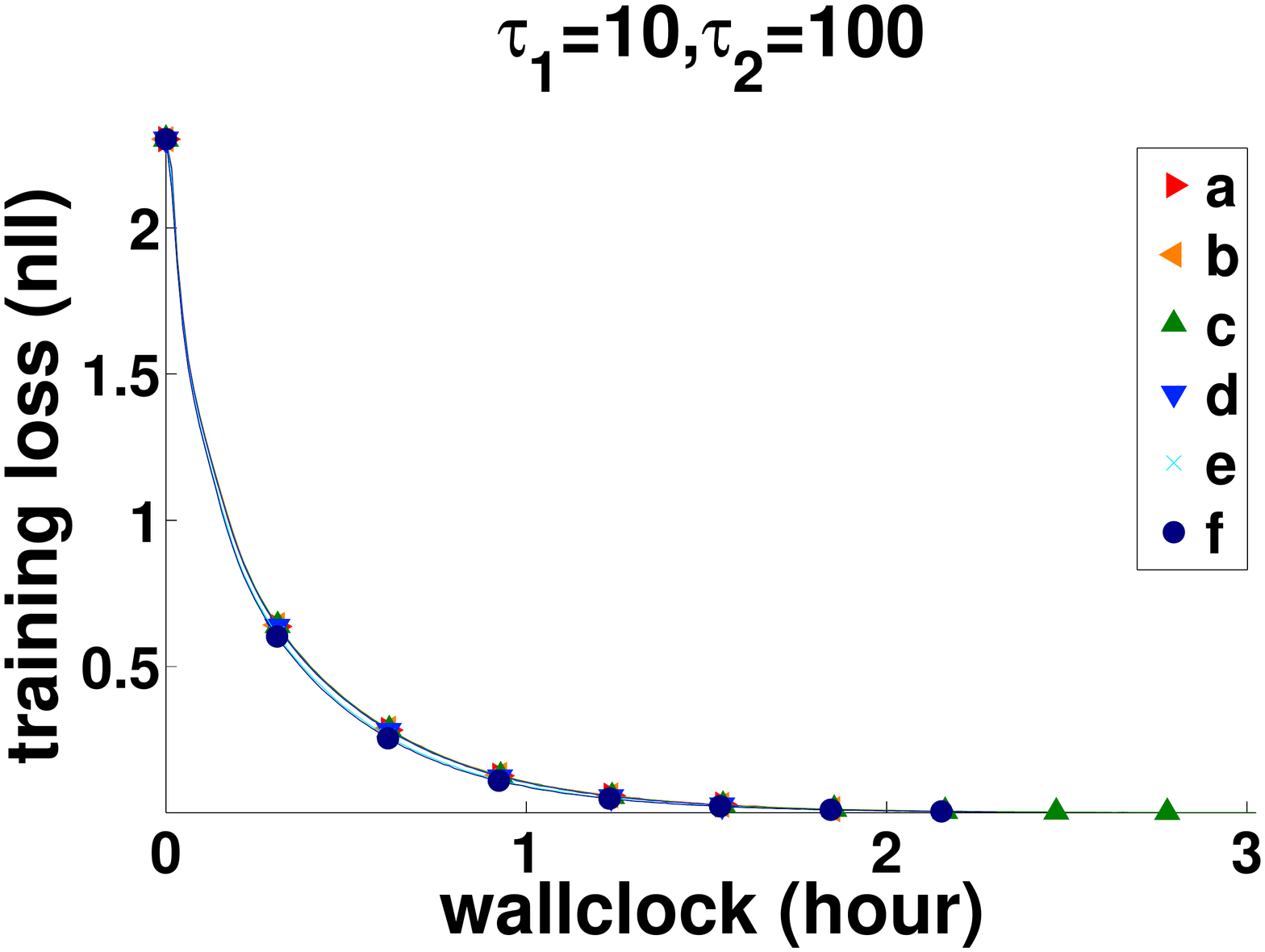}
\hspace{-0.3in}\includegraphics[trim=0cm 0cm 0cm 0cm,clip,width = 3in]{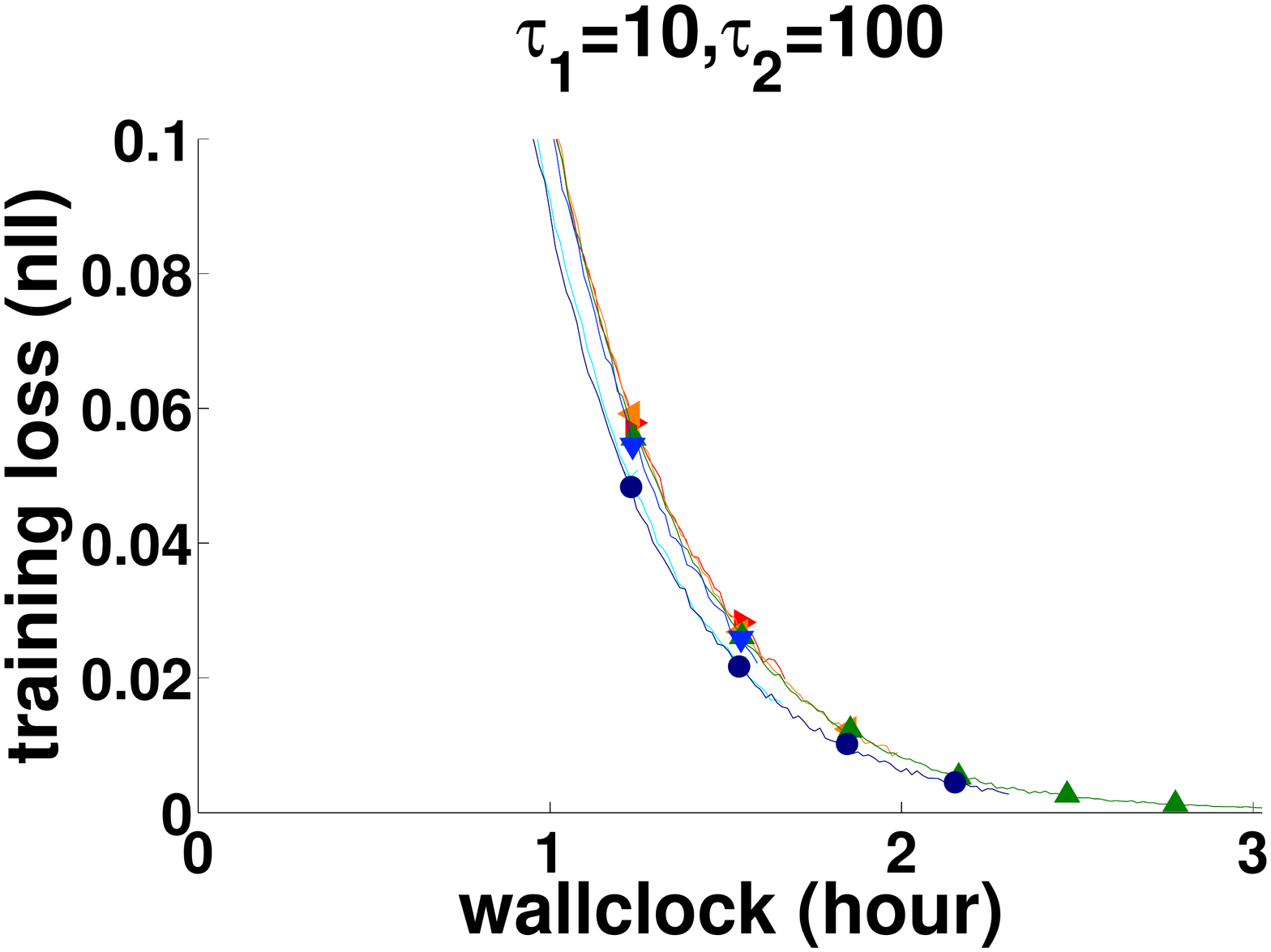} \\
\includegraphics[trim=0cm 0cm 0cm 0cm,clip,width = 3in]{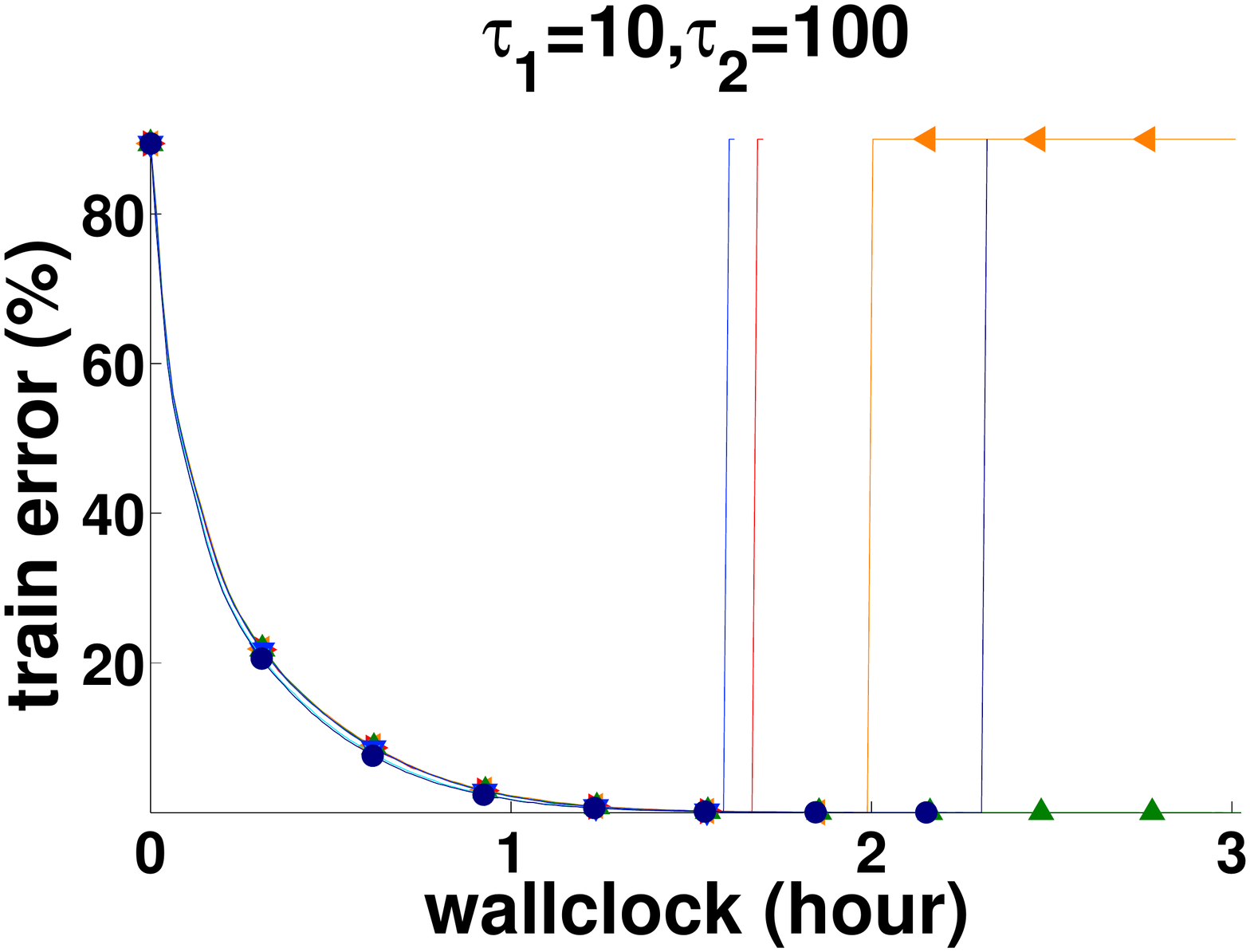}
\hspace{-0.3in}\includegraphics[trim=0cm 0cm 0cm 0cm,clip,width = 3in]{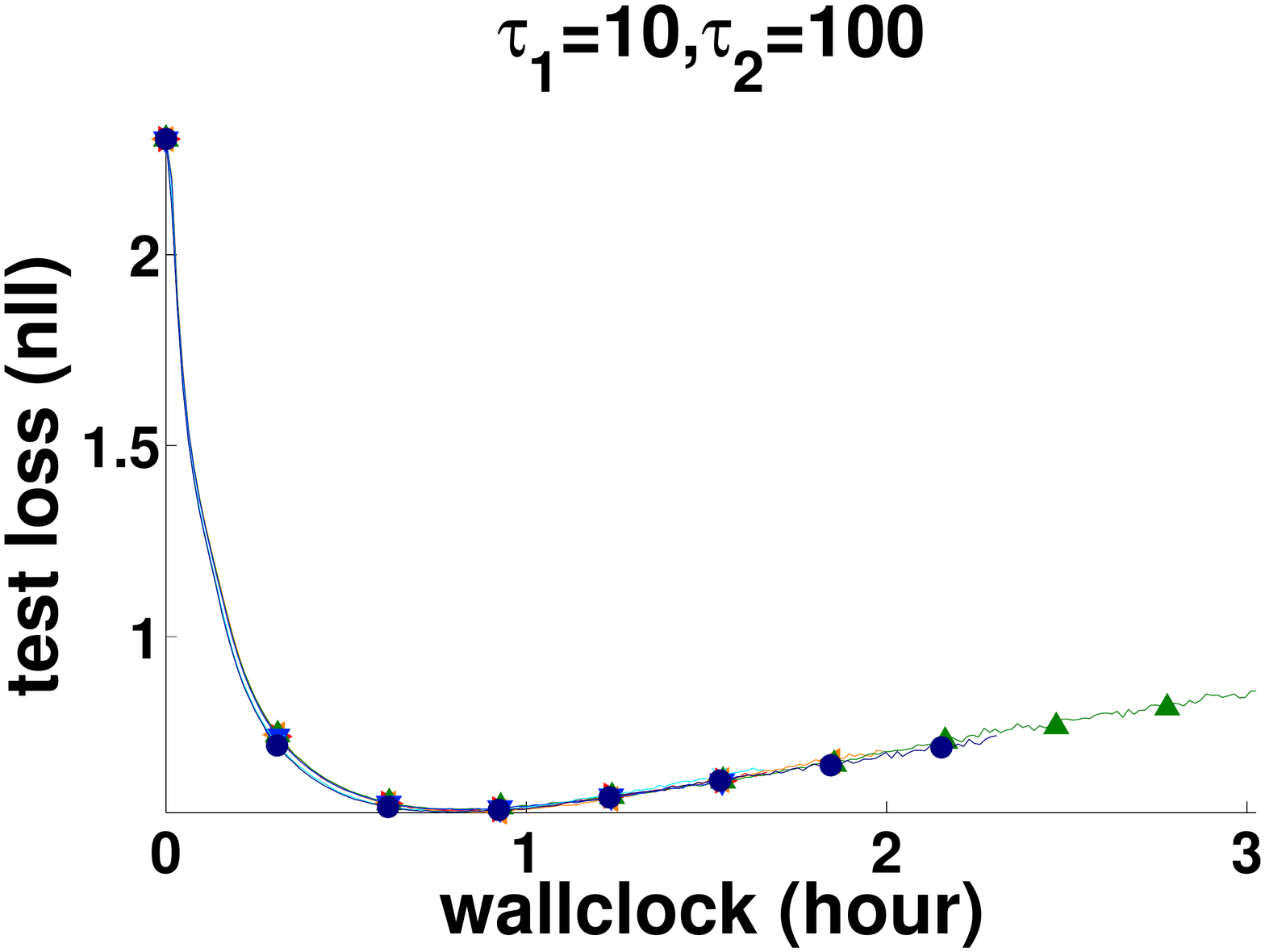} \\
\includegraphics[trim=0cm 0cm 0cm 0cm,clip,width = 3in]{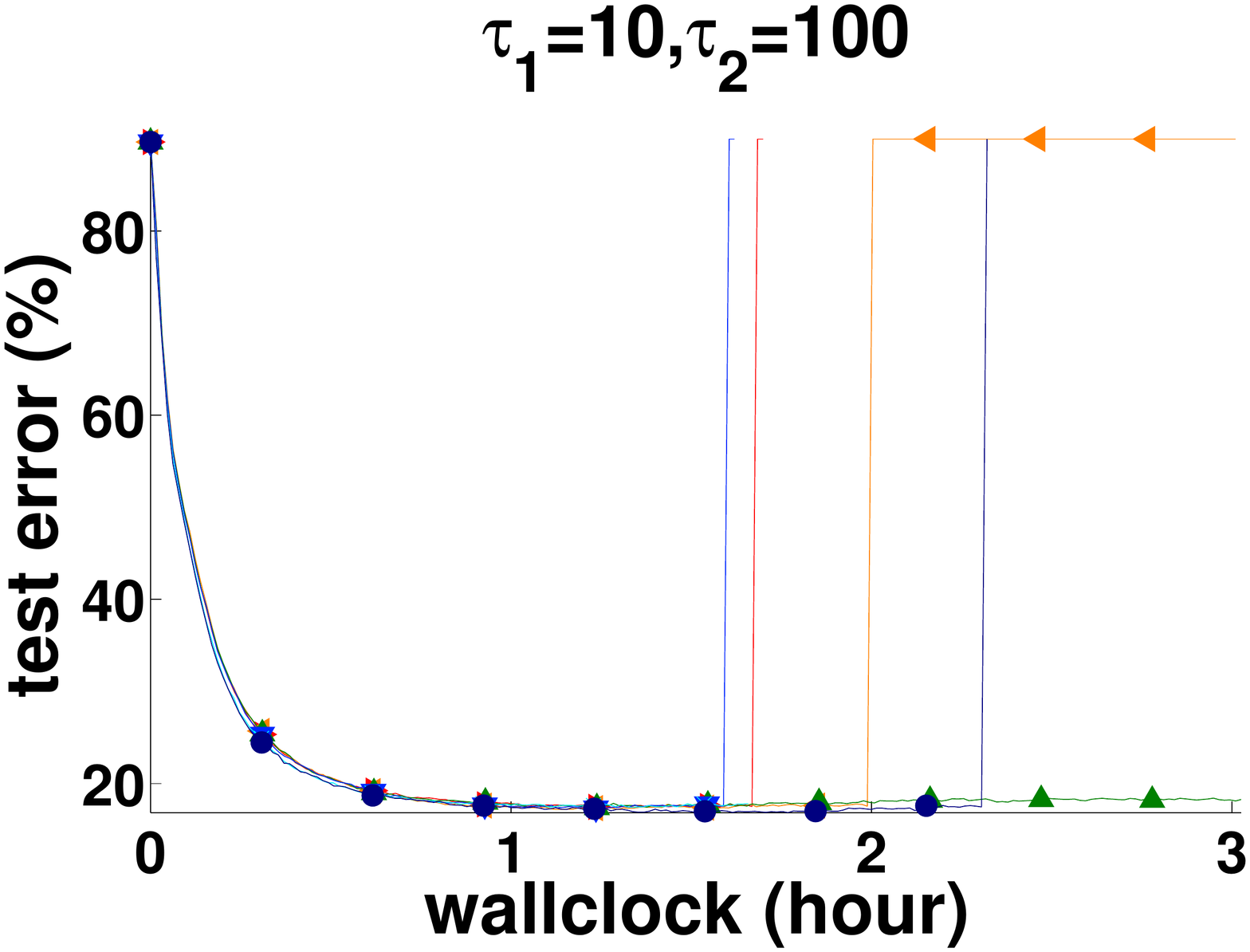}
\hspace{-0.3in}\includegraphics[trim=0cm 0cm 0cm 0cm,clip,width = 3in]{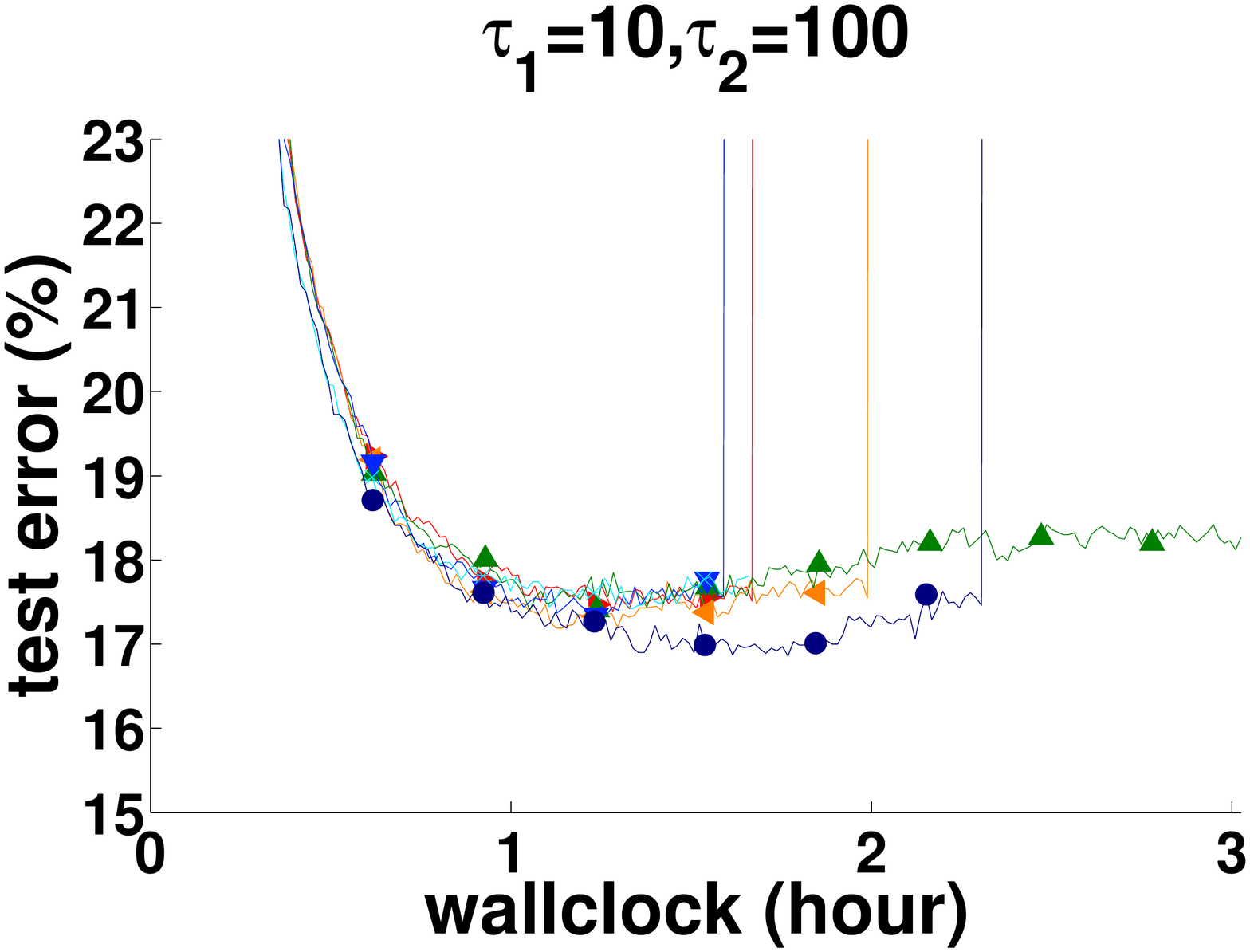}\\
\caption{\textit{EASGD Tree} on \textit{CIFAR-lowrank} using the first communication scheme with $\tau_1=10$ and $\tau_2=100$. Training loss and error, test loss and error of the root node versus a wallclock time. We run this experiment six times independently (labeled with a,b,c,d,e,f) with a same random initialization. $p=256,d=16,\eta=5e-3,\alpha=0.9/(d+1)$.
}
\label{fig:CIFARlr_mix111}
\end{figure}

\begin{figure}[p]
\center
\includegraphics[trim=0cm 0cm 0cm 0cm,clip,width = 3in]{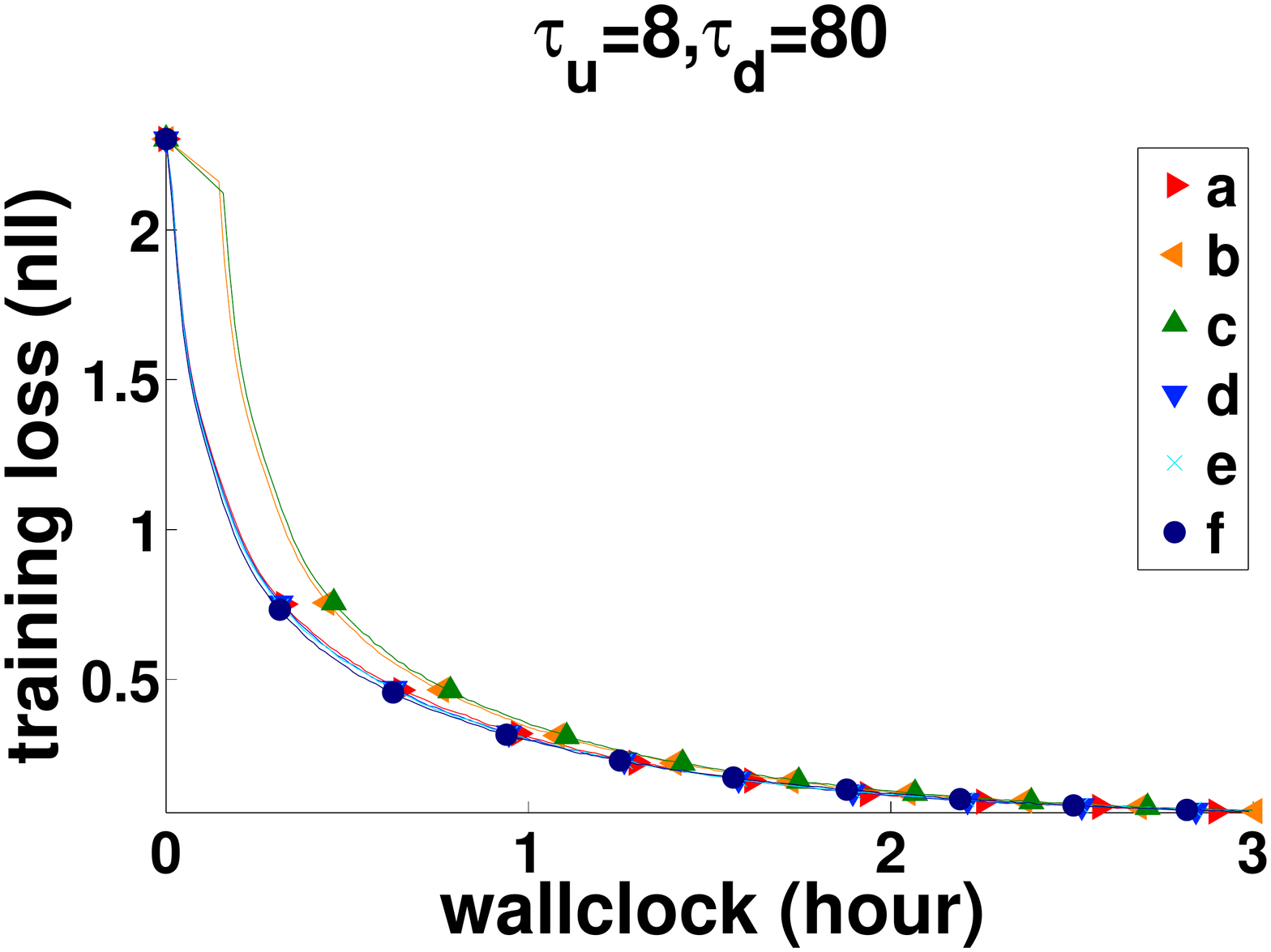}
\hspace{-0.3in}\includegraphics[trim=0cm 0cm 0cm 0cm,clip,width = 3in]{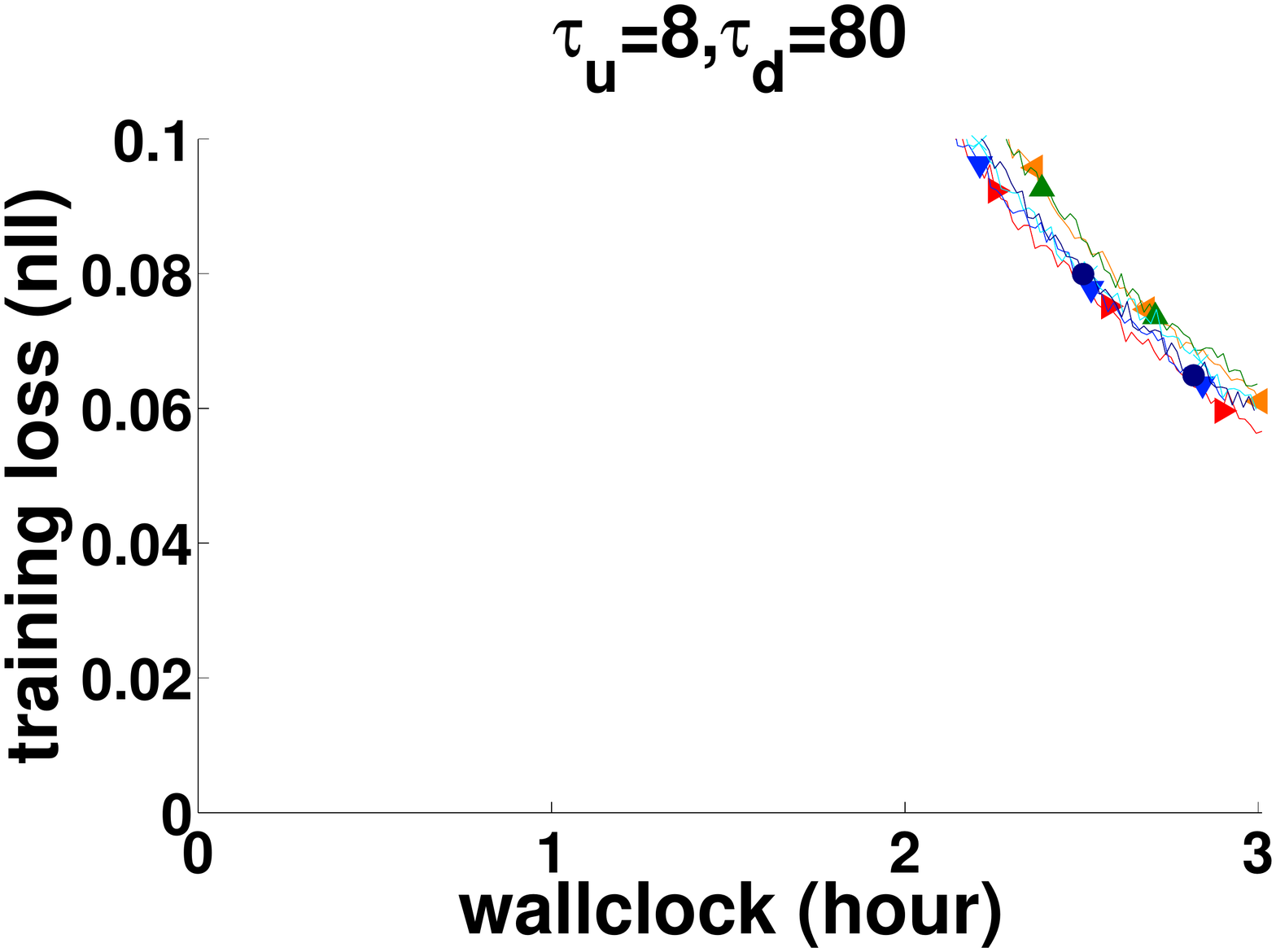} \\
\includegraphics[trim=0cm 0cm 0cm 0cm,clip,width = 3in]{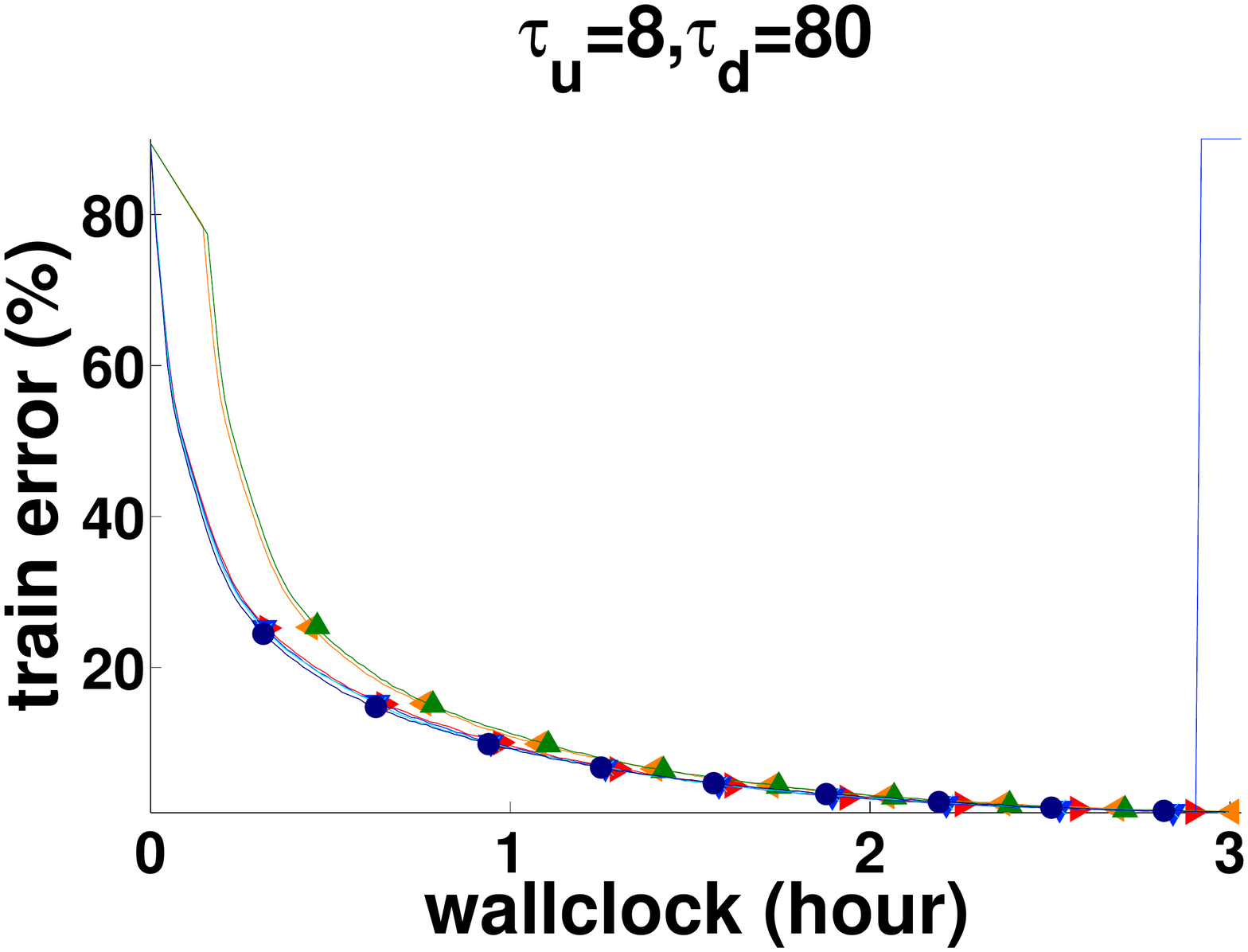}
\hspace{-0.3in}\includegraphics[trim=0cm 0cm 0cm 0cm,clip,width = 3in]{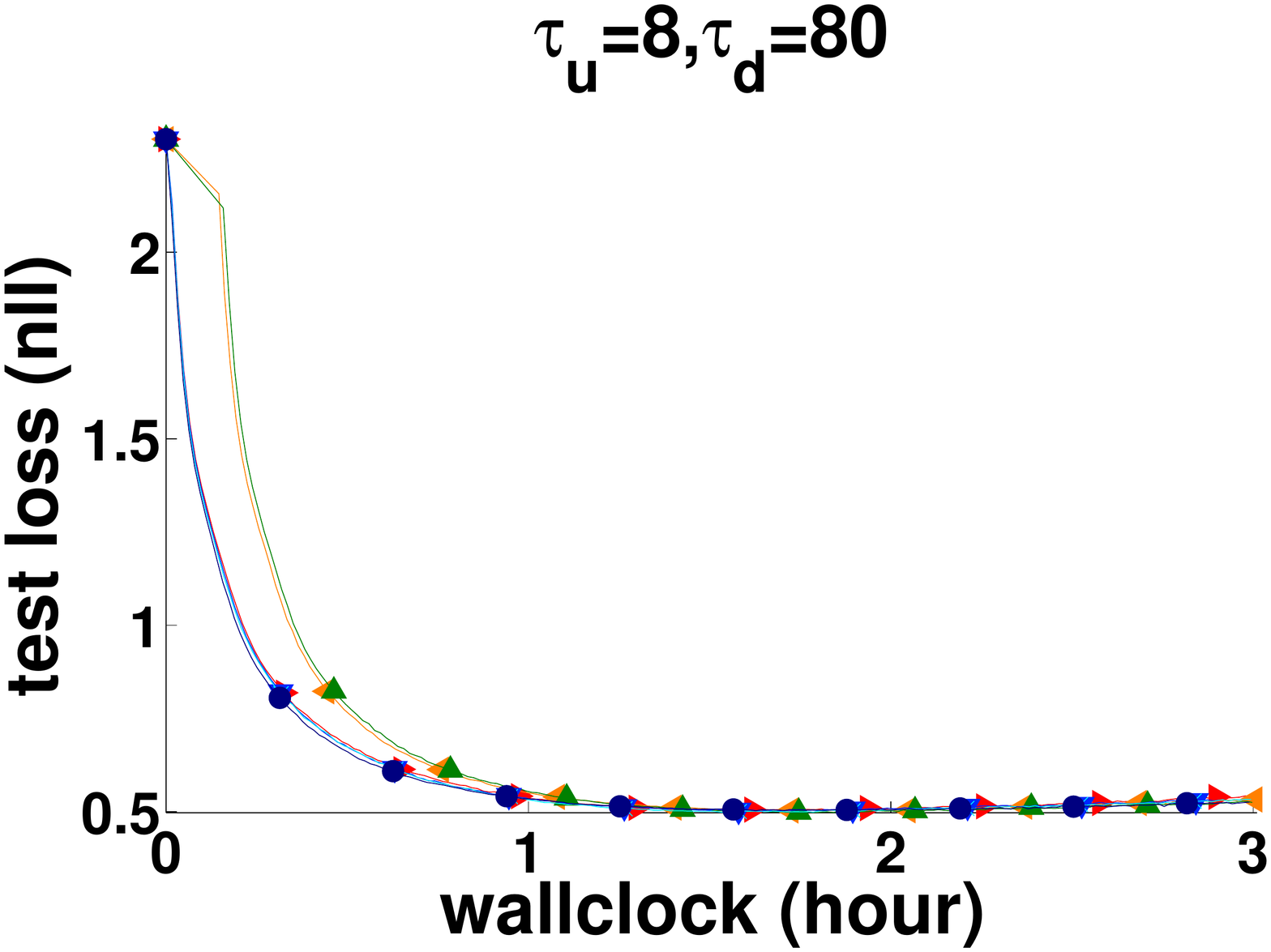} \\
\includegraphics[trim=0cm 0cm 0cm 0cm,clip,width = 3in]{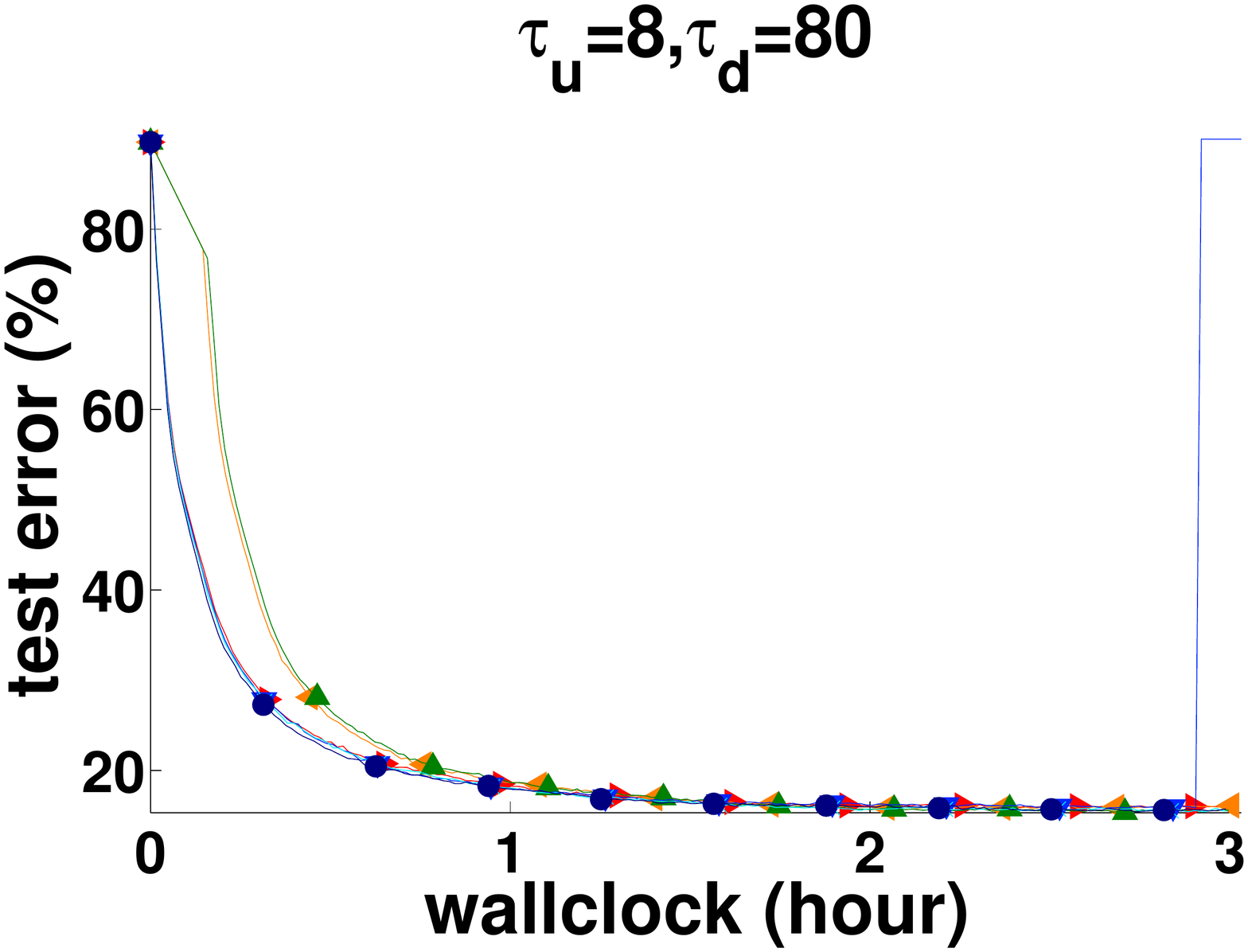}
\hspace{-0.3in}\includegraphics[trim=0cm 0cm 0cm 0cm,clip,width = 3in]{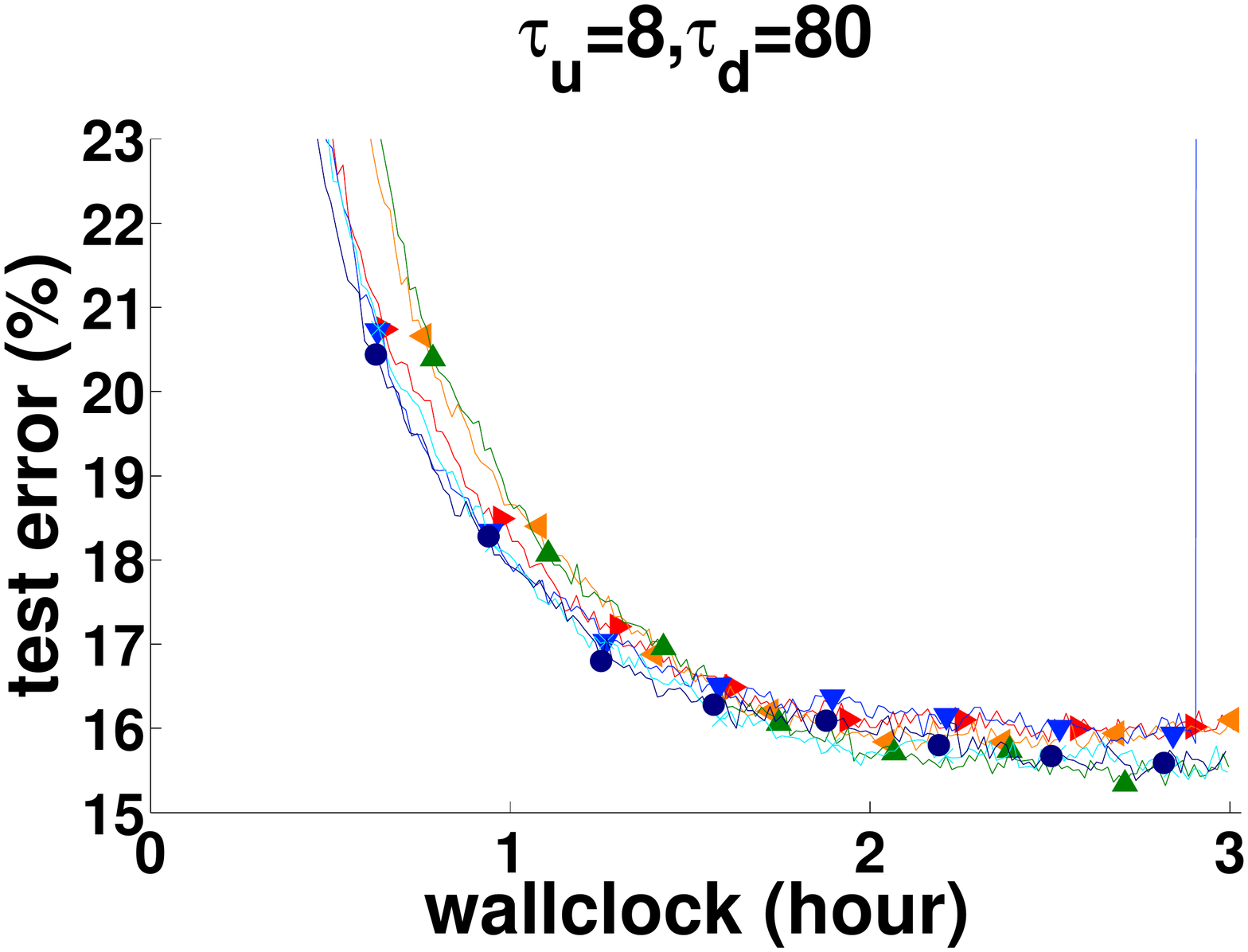}\\
\caption{\textit{EASGD Tree} on \textit{CIFAR-lowrank} using the second communication scheme with $\tau_{u}=8$ and $\tau_{d}=80$. Training loss and error, test loss and error of the root node versus a wallclock time. We run this experiment six times independently (labeled with a,b,c,d,e,f) with a same random initialization. $p=256,d=16,\eta=5e-3,\alpha=0.9/(d+1)$.
}
\label{fig:CIFARlr_mix112}
\end{figure}

In the following experiments, we call \textit{CIFAR-lowrank} the experiment that is just described. We use a $d$-ary tree with degree $d=16$ and $p=256$ leaf nodes running on 16 machines.
For simplicity, we set the moving rate $\alpha$ at each node to be a constant $0.9/(16+1)$.
First we examine the two communication schemes that we have discussed above. Figure~\ref{fig:CIFARlr_mix111} shows the performance of the first communication scheme using period $\tau_1=10$ and $\tau_2=100$. Figure~\ref{fig:CIFARlr_mix112} shows the performance of the second communication scheme using $\tau_{u}=8$ and $\tau_{d}=80$. We see that the first communication scheme achieves a faster convergence in terms of the training loss. All the six runs are nearly the same at the beginning. However, five runs out of the six have diverged in the middle of the training. We have chosen a quite large learning rate which can trigger this instability. The second communication scheme is more stable. Only one out of the six runs has diverged. It has a slower convergence in terms of the training loss, but it consistently achieves a lower test error (the smallest one is 15.32\%, compared to the smallest test error of the first scheme which is 16.86\%). Also notice that there are two runs with a slower initial convergence rate. This may be related to the network traffic.

\begin{figure}[p]
\center
\includegraphics[trim=0cm 0cm 0cm 0cm,clip,width = 3in]{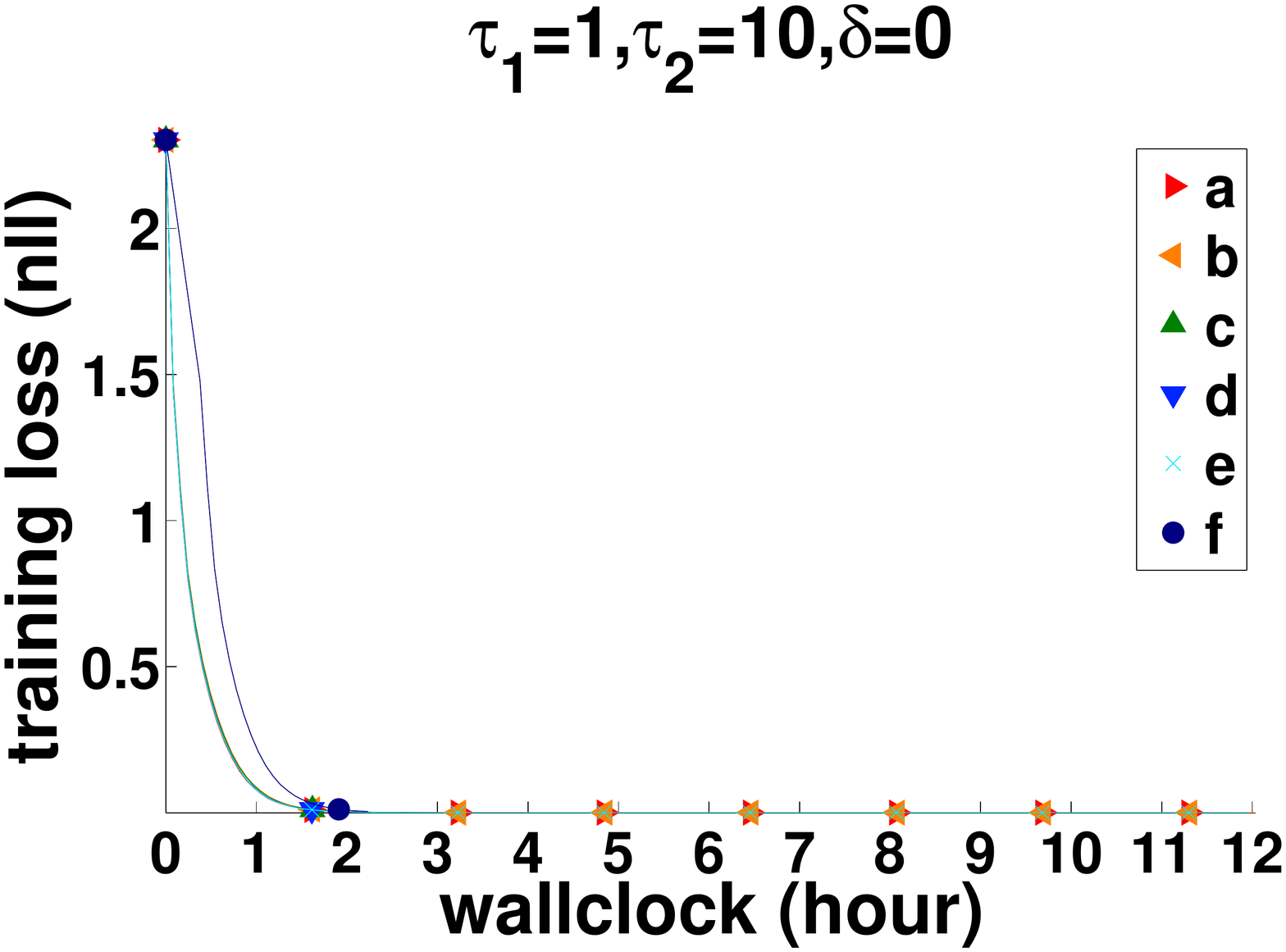}
\hspace{-0.3in}\includegraphics[trim=0cm 0cm 0cm 0cm,clip,width = 3in]{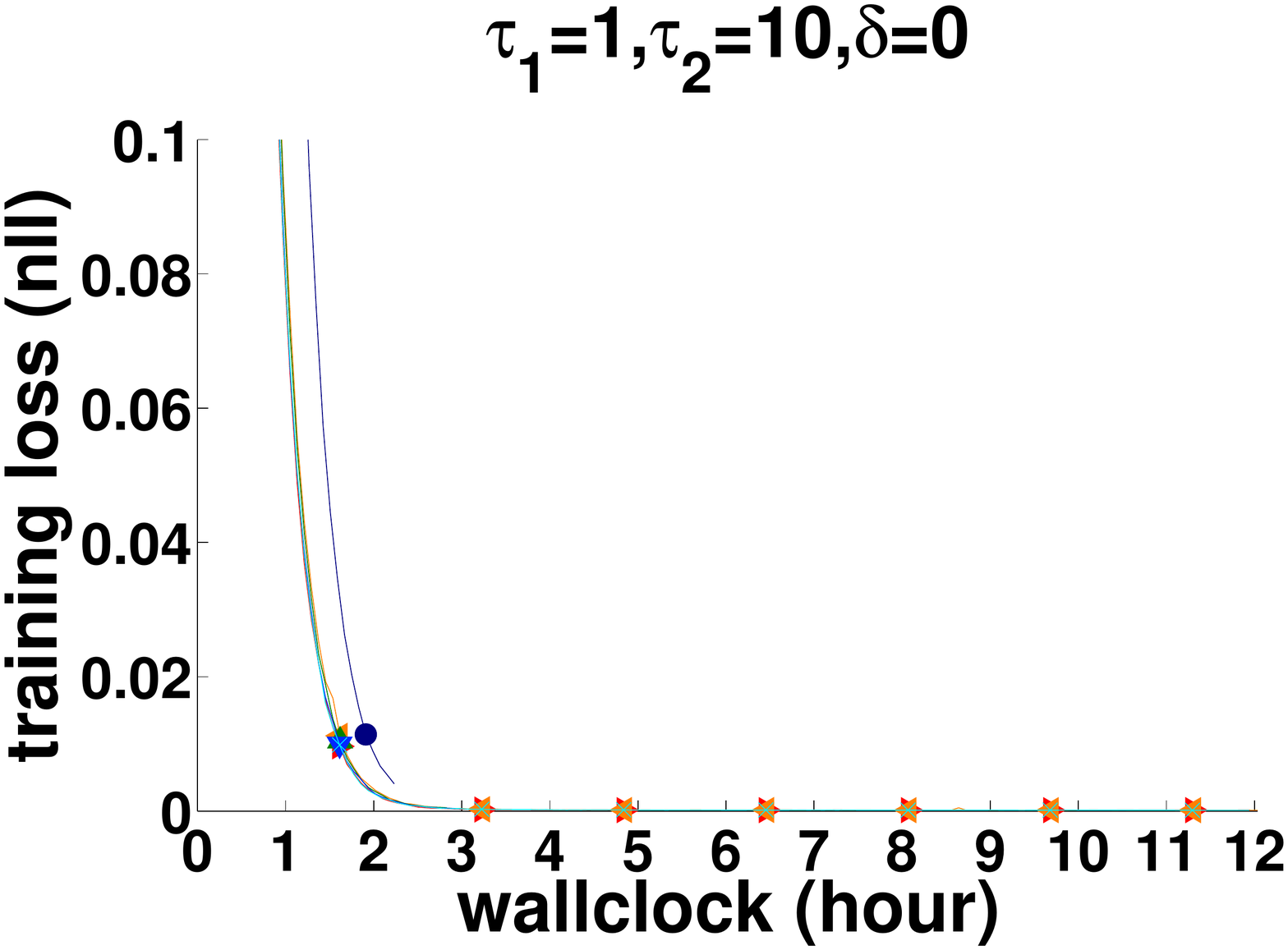} \\
\includegraphics[trim=0cm 0cm 0cm 0cm,clip,width = 3in]{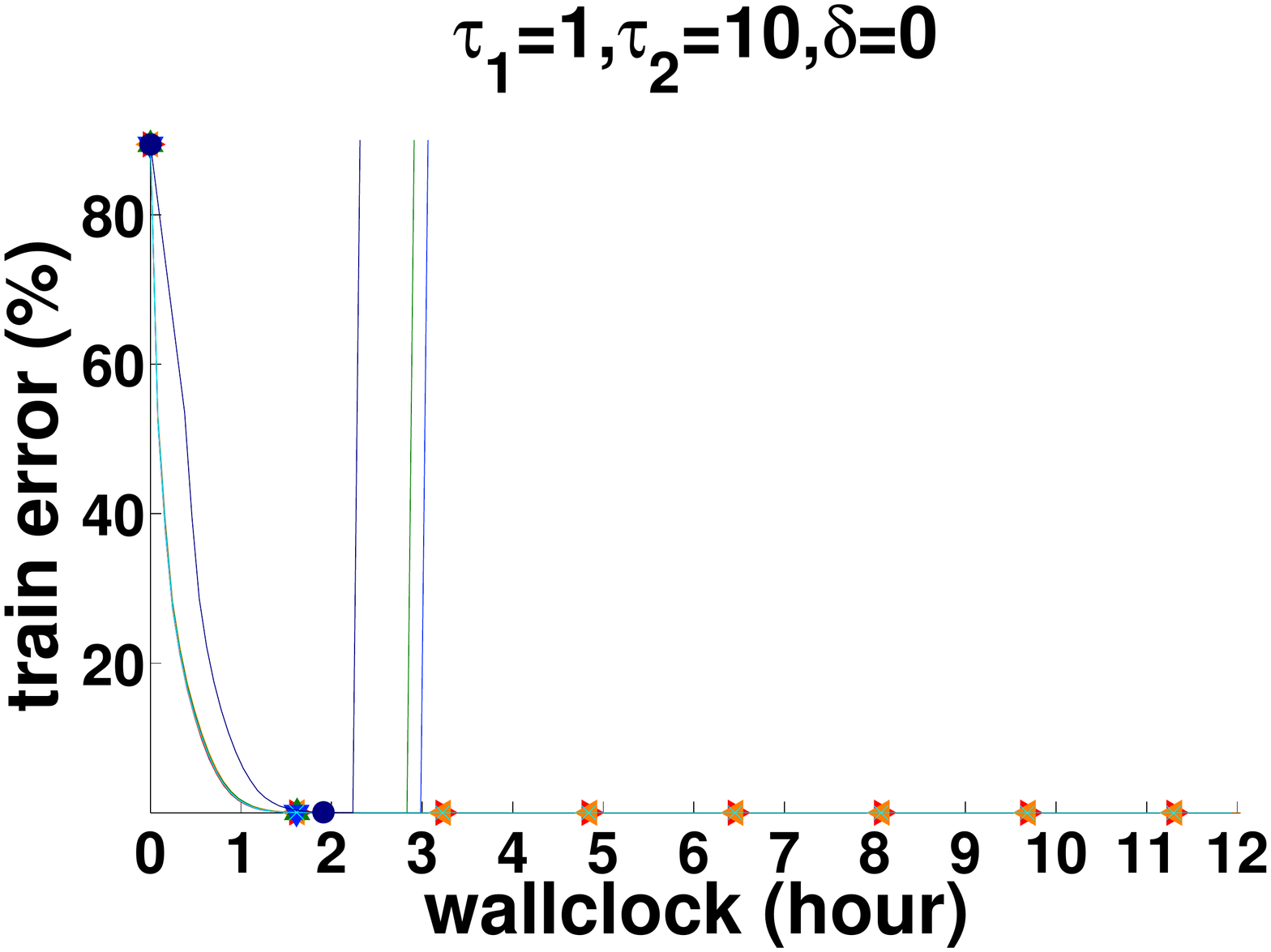}
\hspace{-0.3in}\includegraphics[trim=0cm 0cm 0cm 0cm,clip,width = 3in]{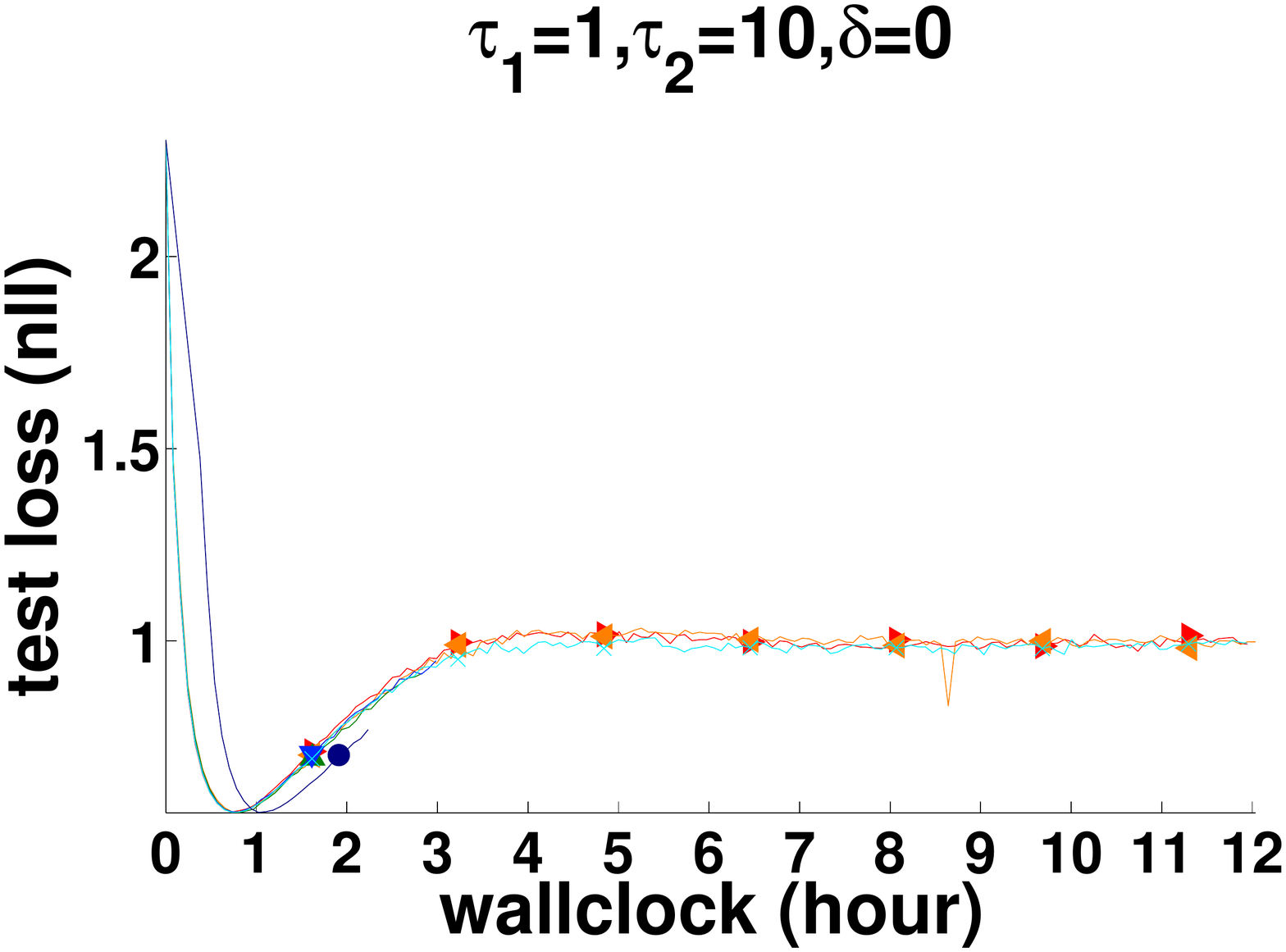} \\
\includegraphics[trim=0cm 0cm 0cm 0cm,clip,width = 3in]{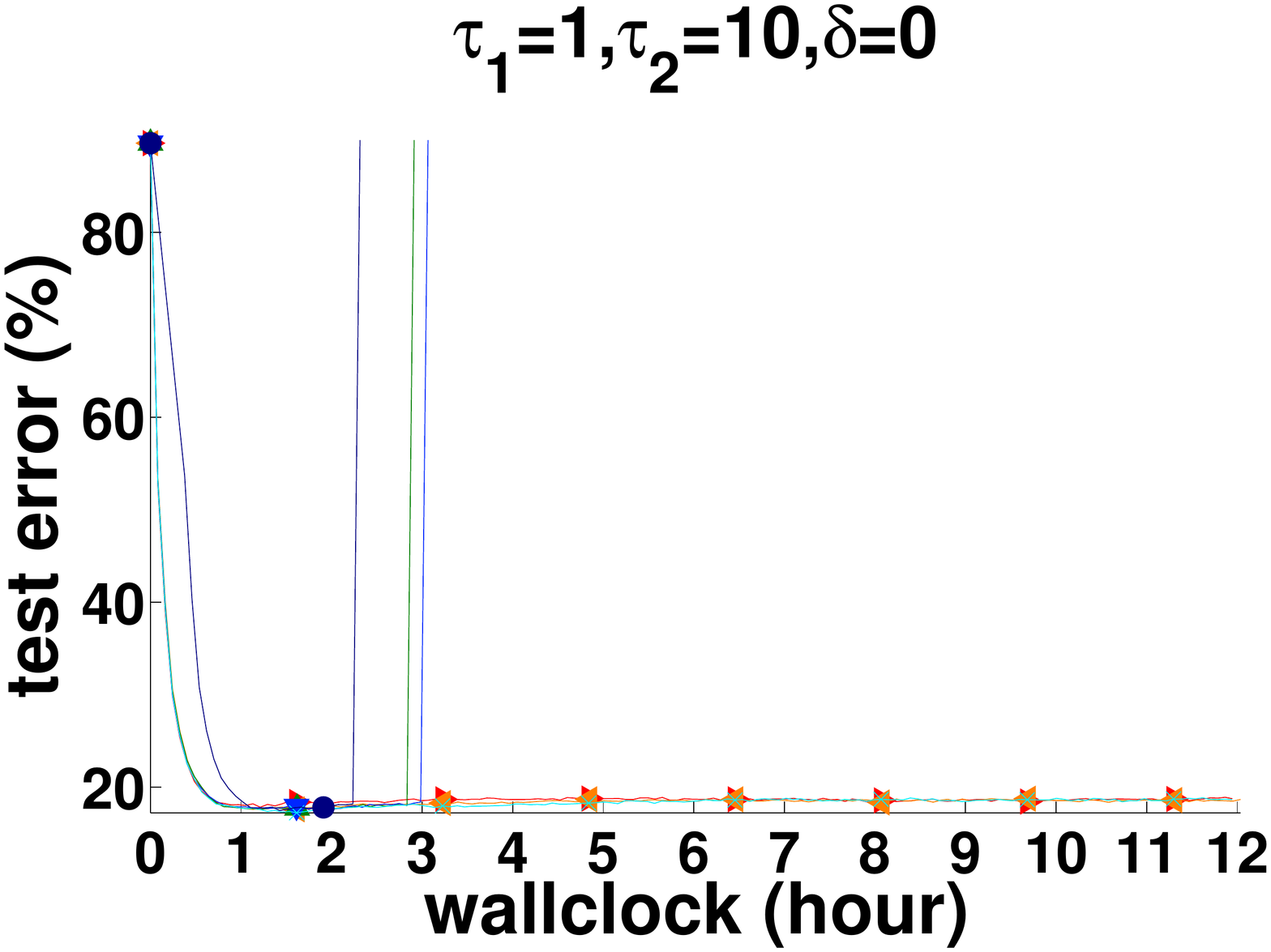}
\hspace{-0.3in}\includegraphics[trim=0cm 0cm 0cm 0cm,clip,width = 3in]{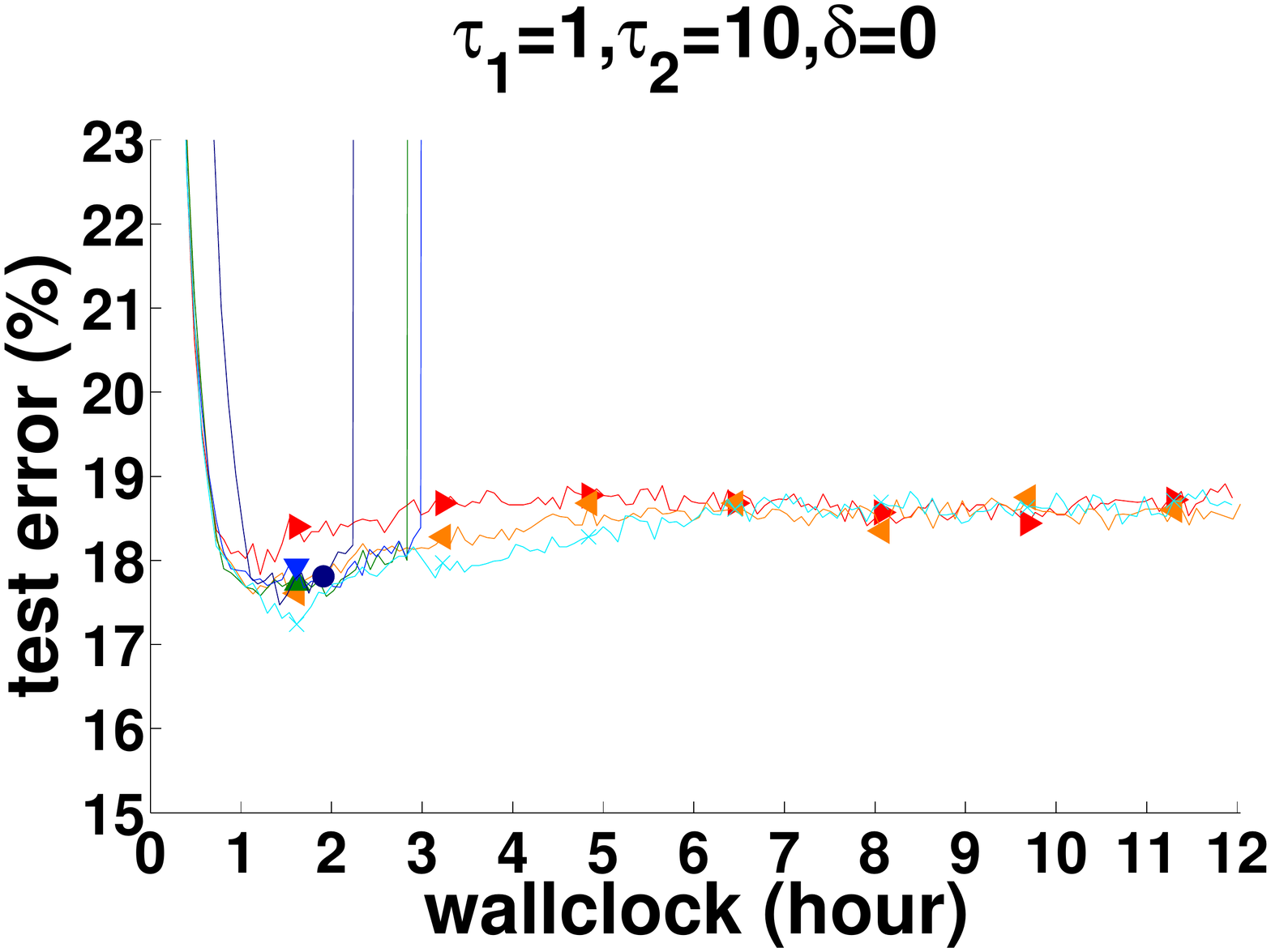}\\
\caption{\textit{EASGD Tree} on \textit{CIFAR-lowrank} using the first communication scheme with $\tau_1=1$ and $\tau_2=10$. Training loss and error, test loss and error of the root node versus a wallclock time. We run this experiment six times independently (labeled with a,b,c,d,e,f) with a same random initialization. $p=256,d=16,\eta=5e-2,\alpha=0.9/(d+1)$. The momentum rate $\delta=0$.
}
\label{fig:CIFARlr_mix121}
\end{figure}

\begin{figure}[p]
\center
\includegraphics[trim=0cm 0cm 0cm 0cm,clip,width = 3in]{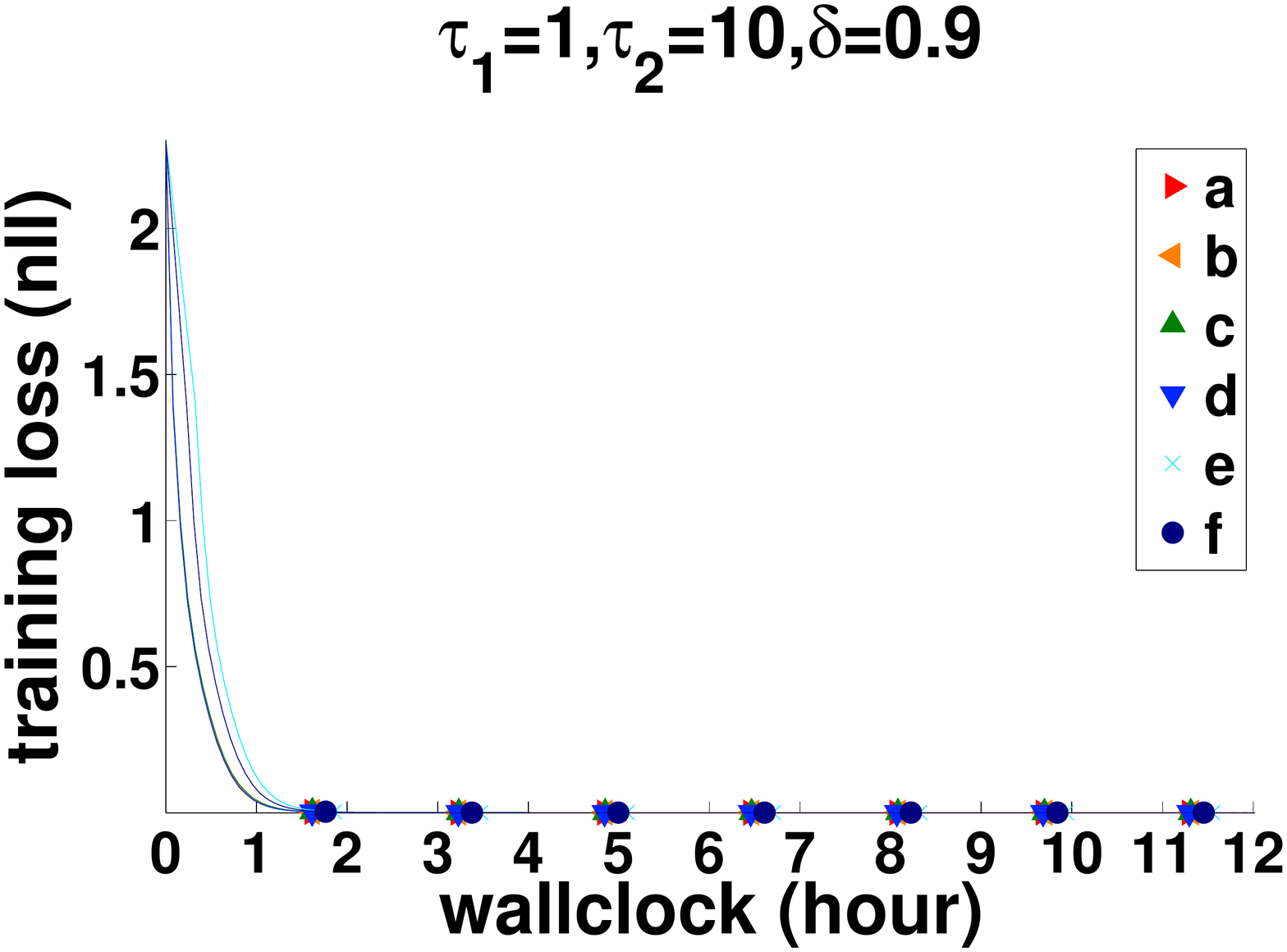}
\hspace{-0.3in}\includegraphics[trim=0cm 0cm 0cm 0cm,clip,width = 3in]{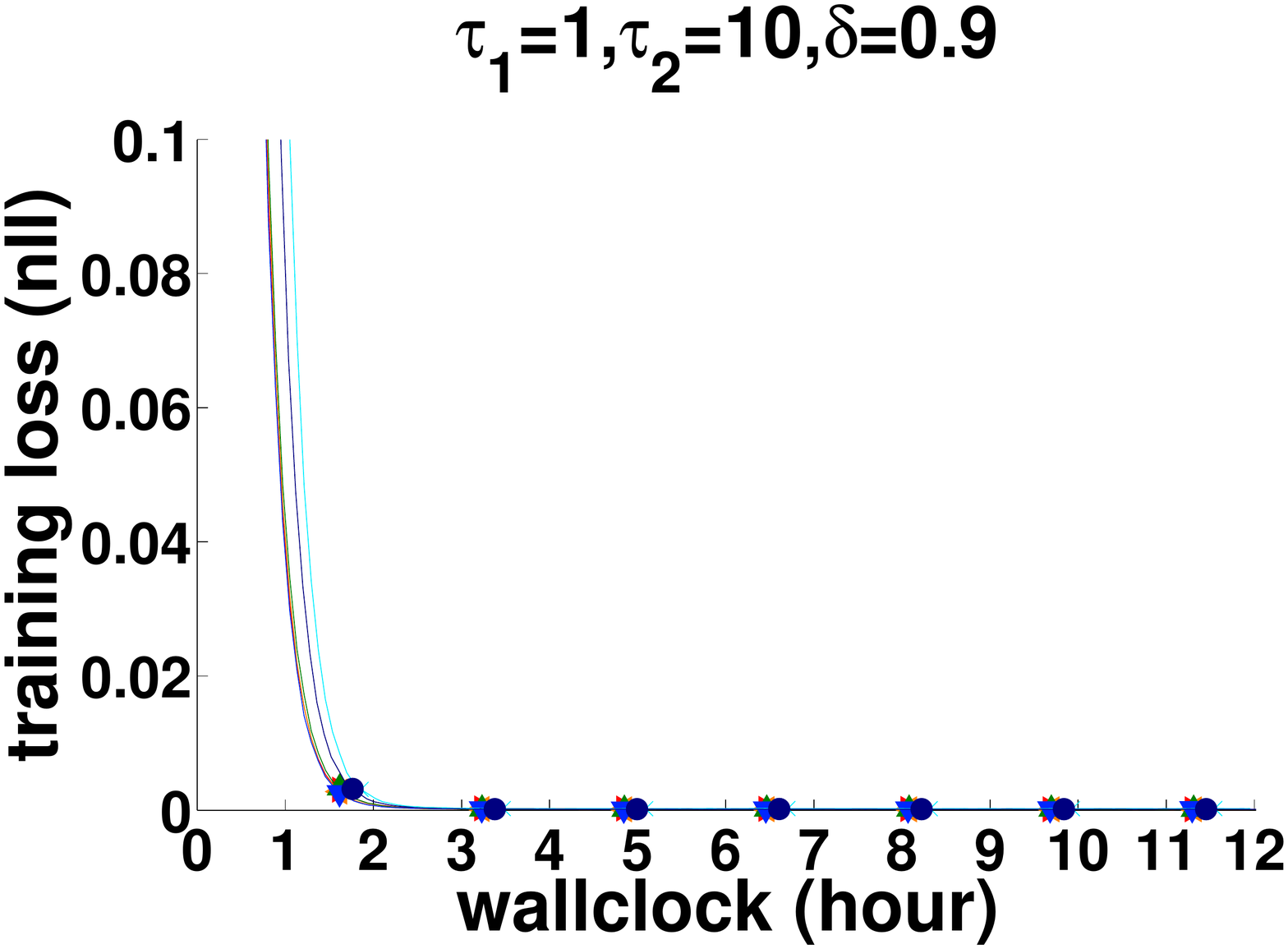} \\
\includegraphics[trim=0cm 0cm 0cm 0cm,clip,width = 3in]{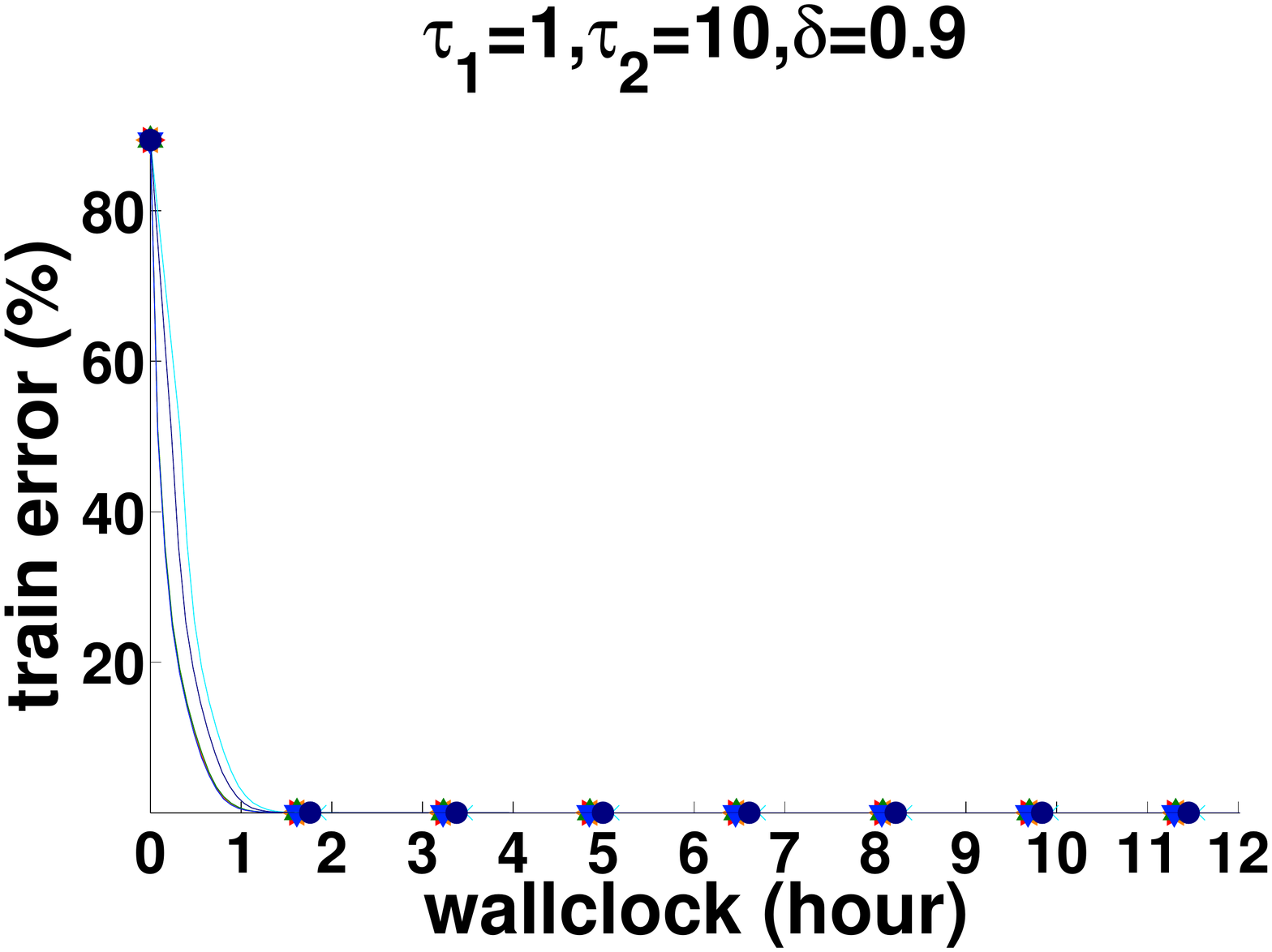}
\hspace{-0.3in}\includegraphics[trim=0cm 0cm 0cm 0cm,clip,width = 3in]{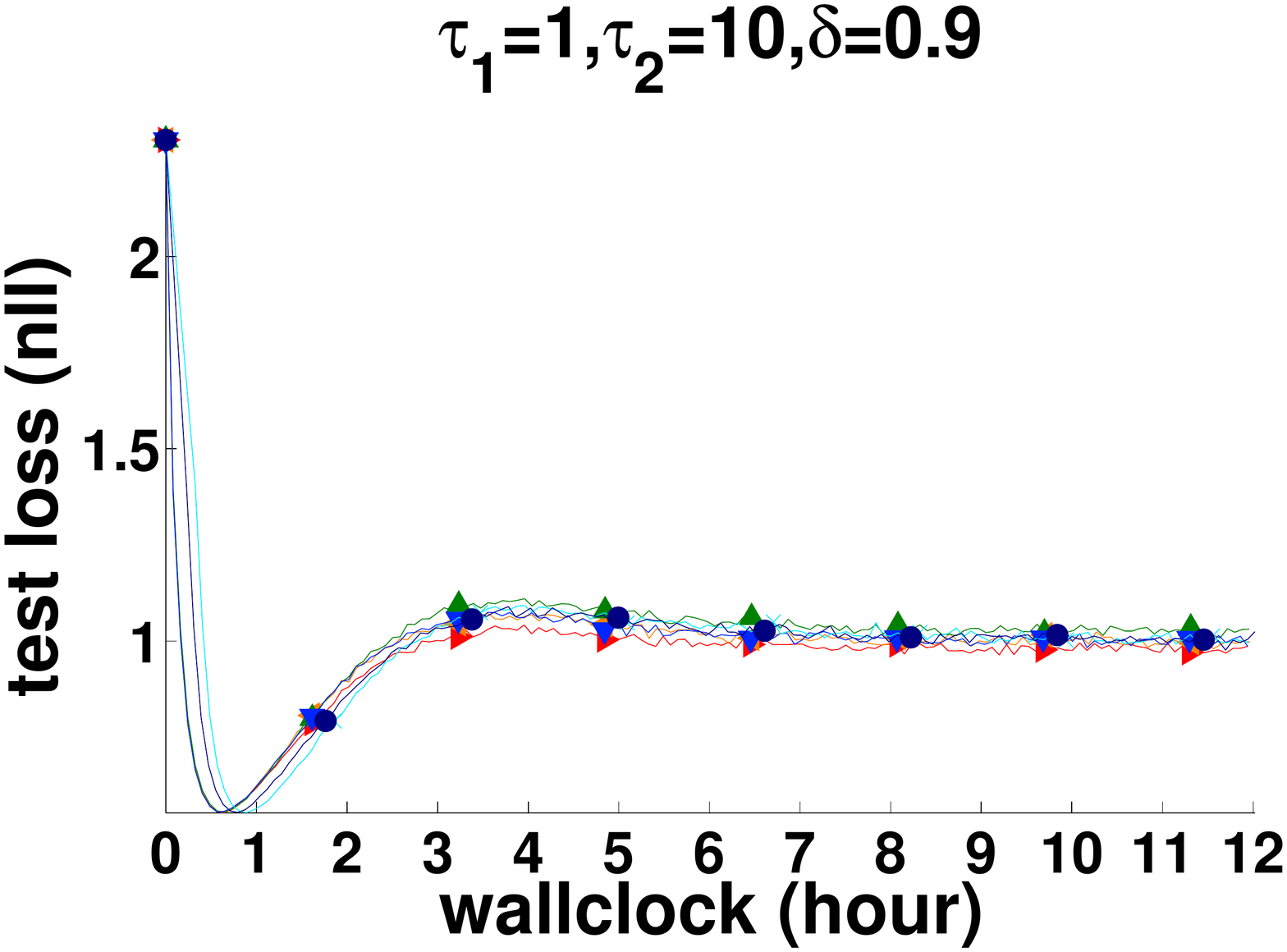} \\
\includegraphics[trim=0cm 0cm 0cm 0cm,clip,width = 3in]{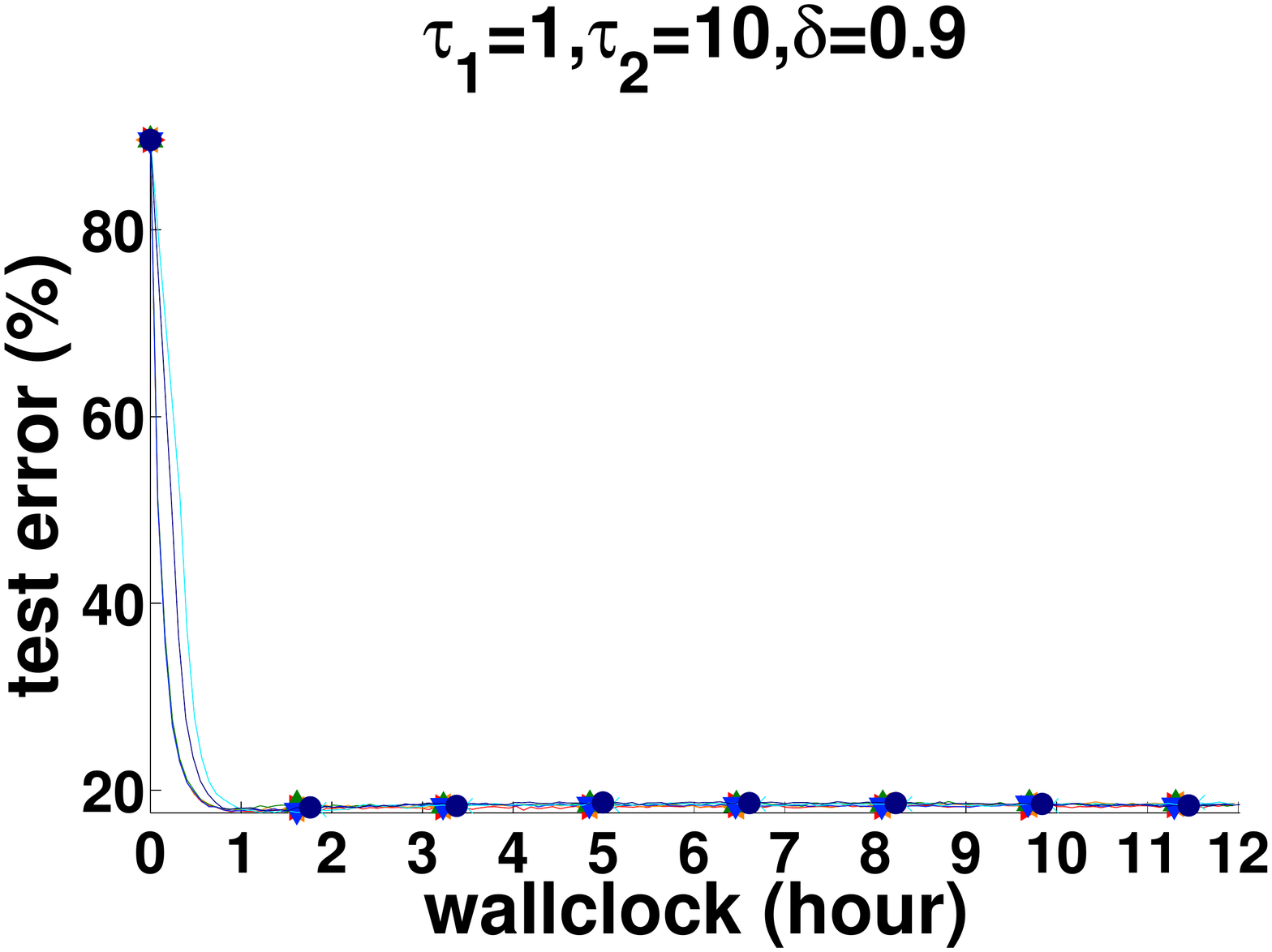}
\hspace{-0.3in}\includegraphics[trim=0cm 0cm 0cm 0cm,clip,width = 3in]{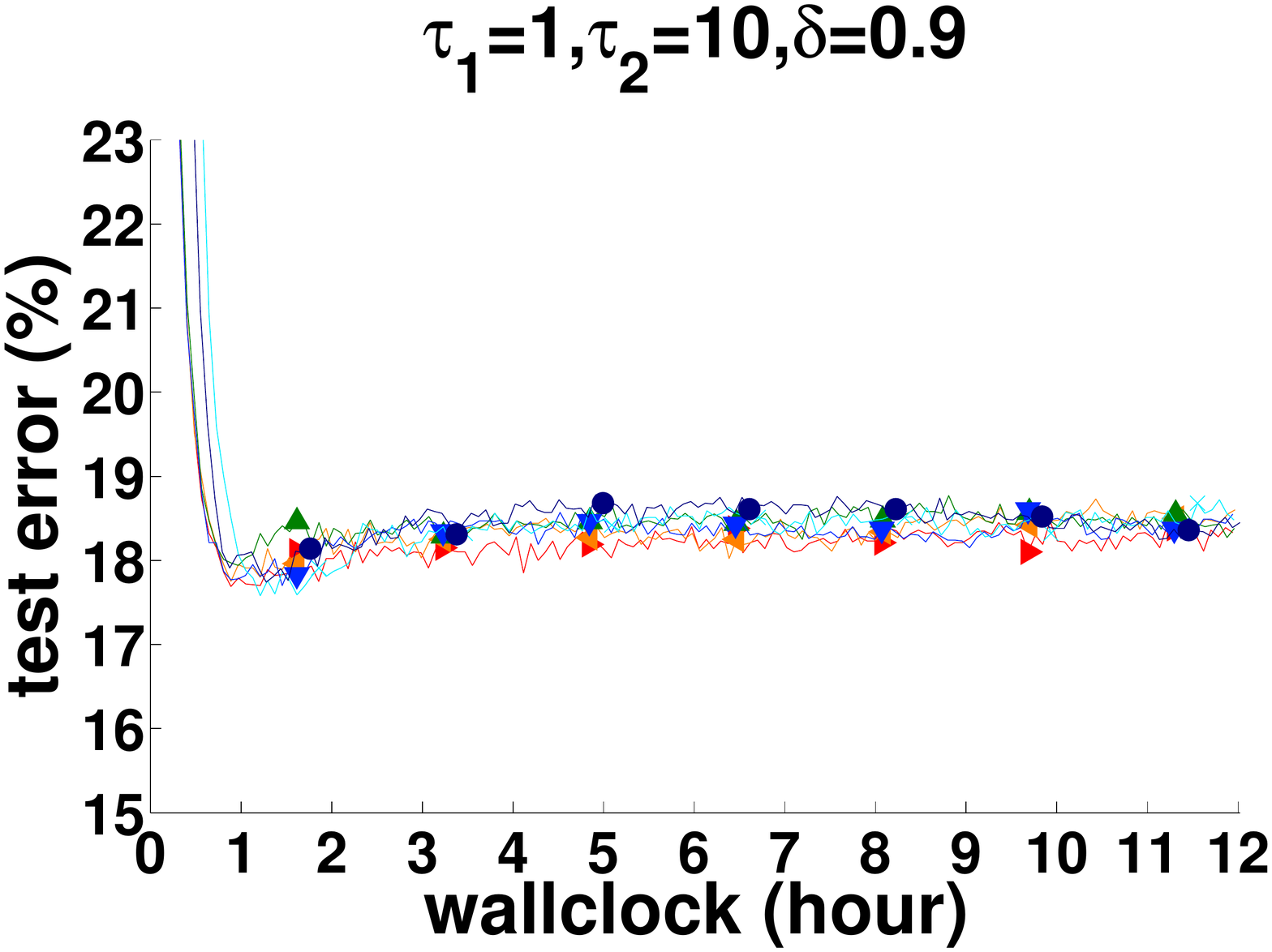}\\
\caption{\textit{EASGD Tree} on \textit{CIFAR-lowrank} using the first communication scheme with $\tau_1=1$ and $\tau_2=10$. Training loss and error, test loss and error of the root node versus a wallclock time. We run this experiment six times independently (labeled with a,b,c,d,e,f) with a same random initialization. $p=256,d=16,\eta=5e-3,\alpha=0.9/(d+1)$. The momentum rate $\delta=0.9$.
}
\label{fig:CIFARlr_mix122}
\end{figure}

\begin{figure}[p]
\center
\includegraphics[trim=0cm 0cm 0cm 0cm,clip,width = 3in]{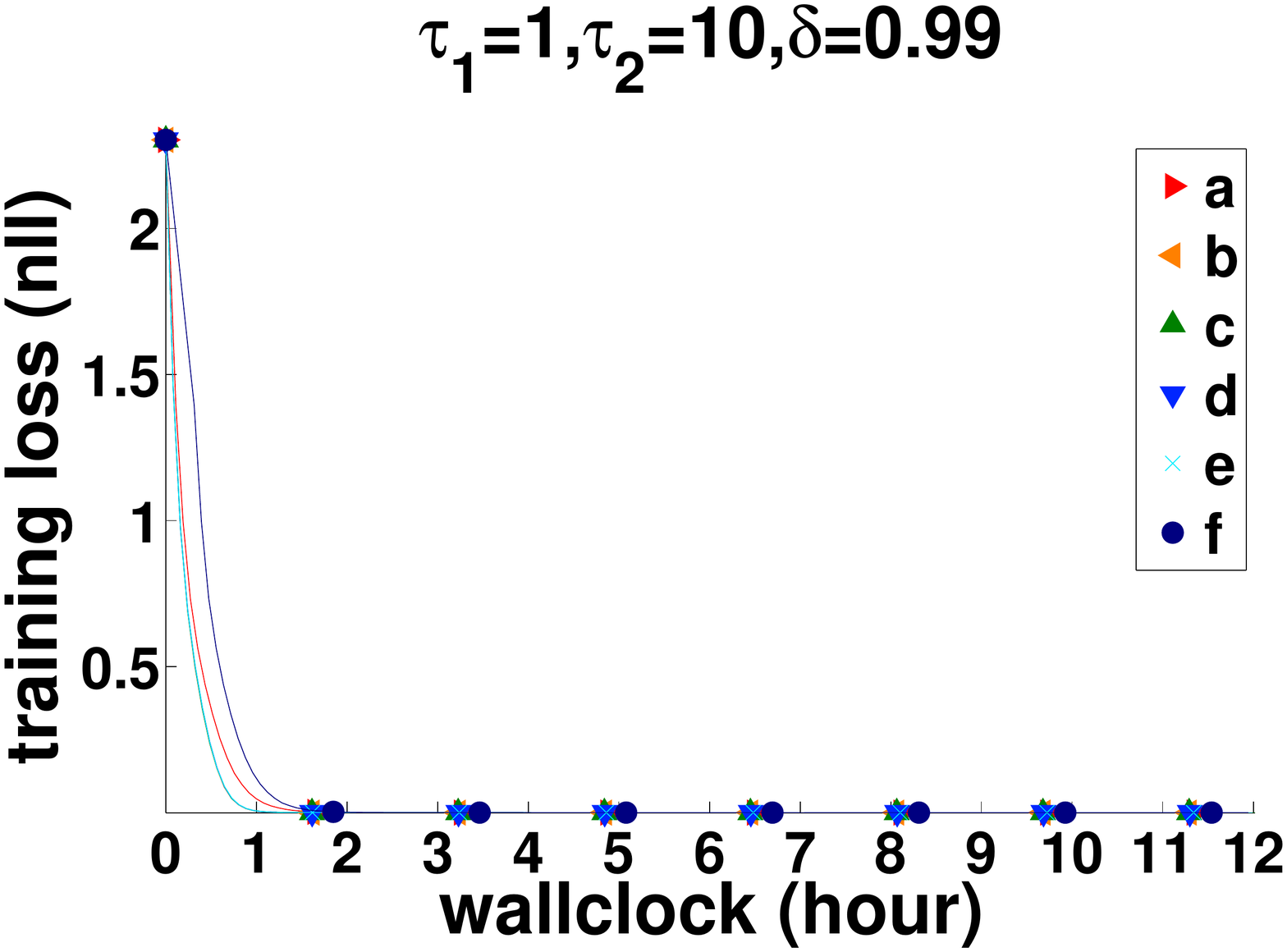}
\hspace{-0.3in}\includegraphics[trim=0cm 0cm 0cm 0cm,clip,width = 3in]{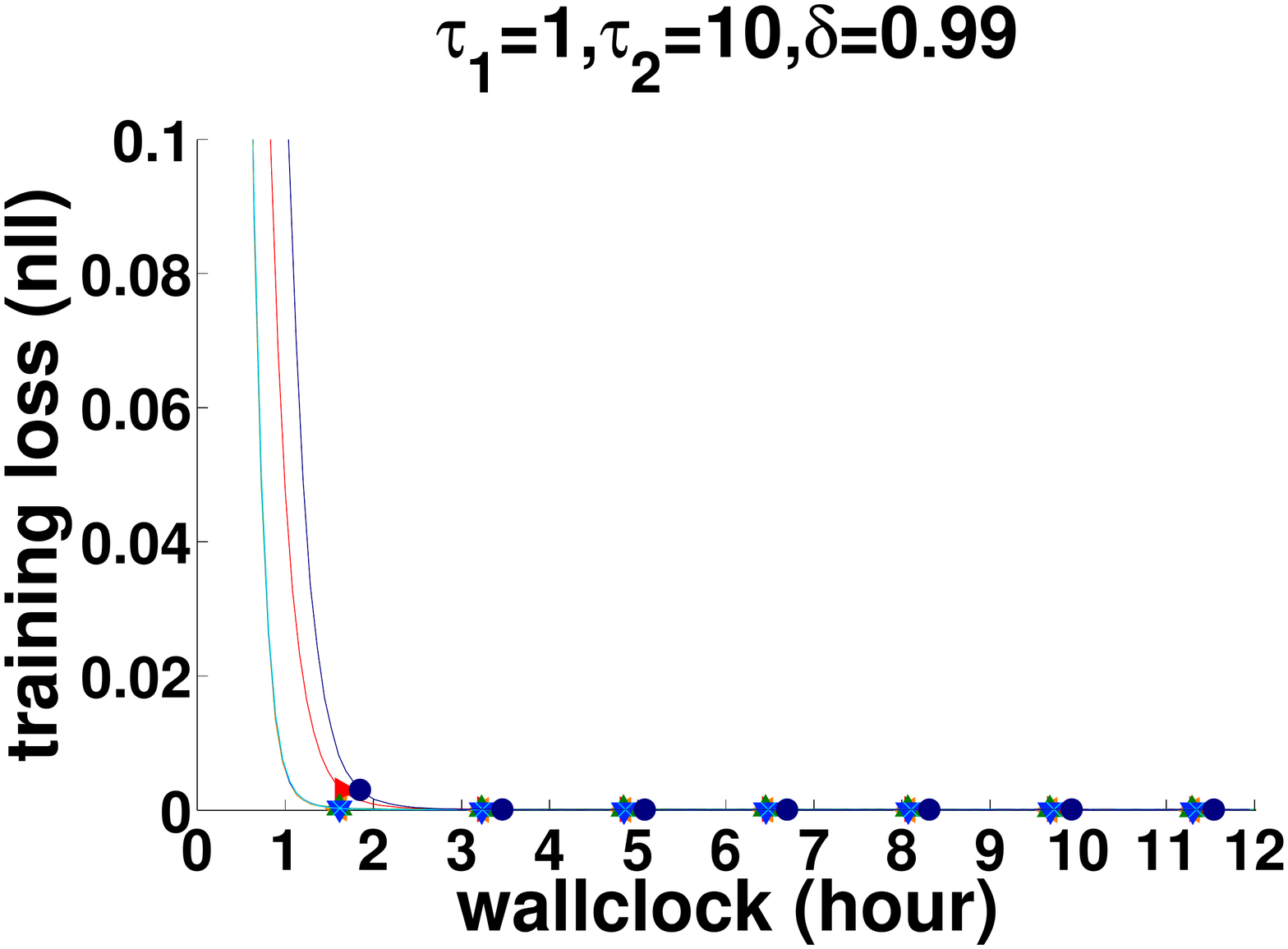} \\
\includegraphics[trim=0cm 0cm 0cm 0cm,clip,width = 3in]{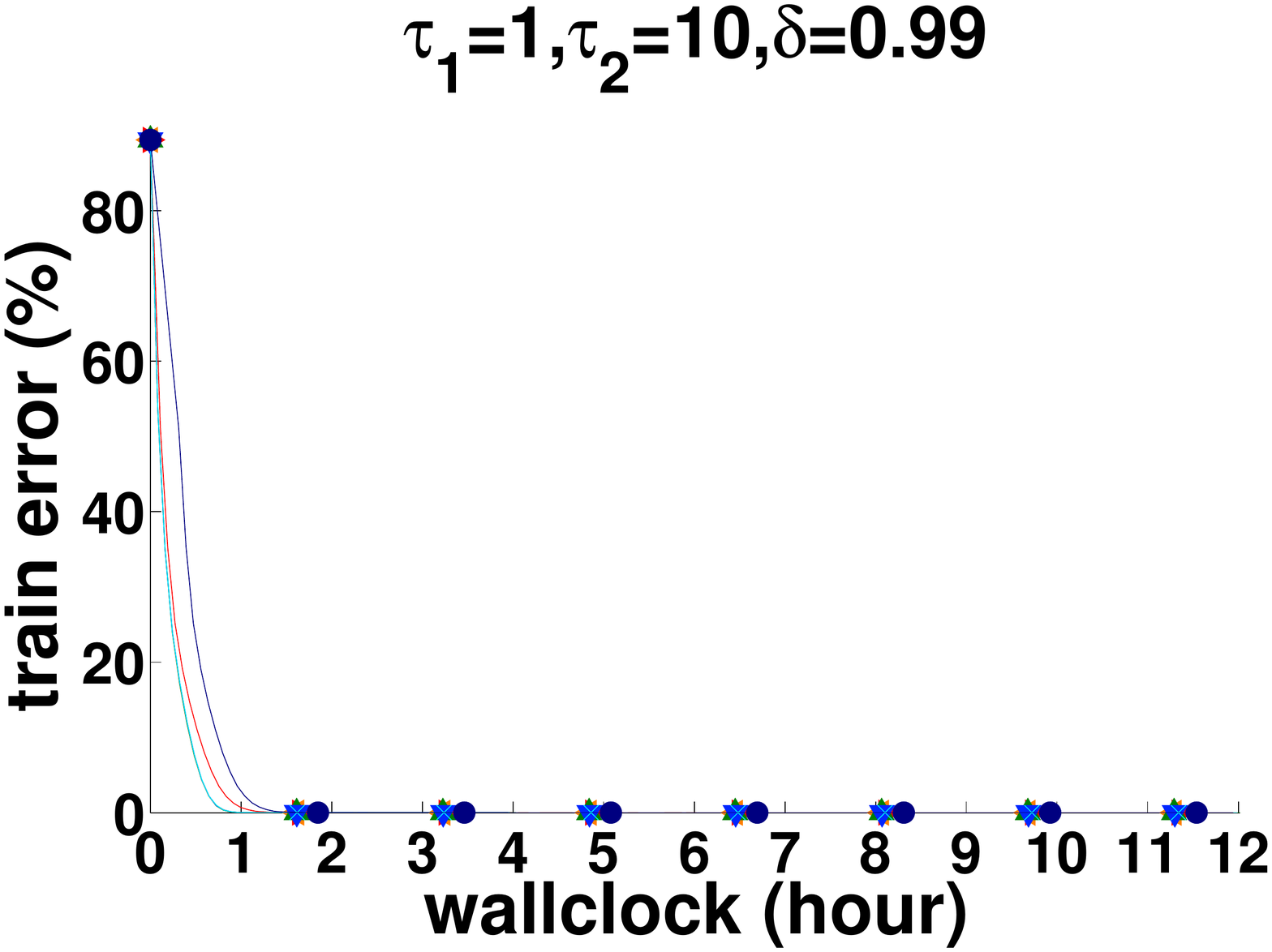}
\hspace{-0.3in}\includegraphics[trim=0cm 0cm 0cm 0cm,clip,width = 3in]{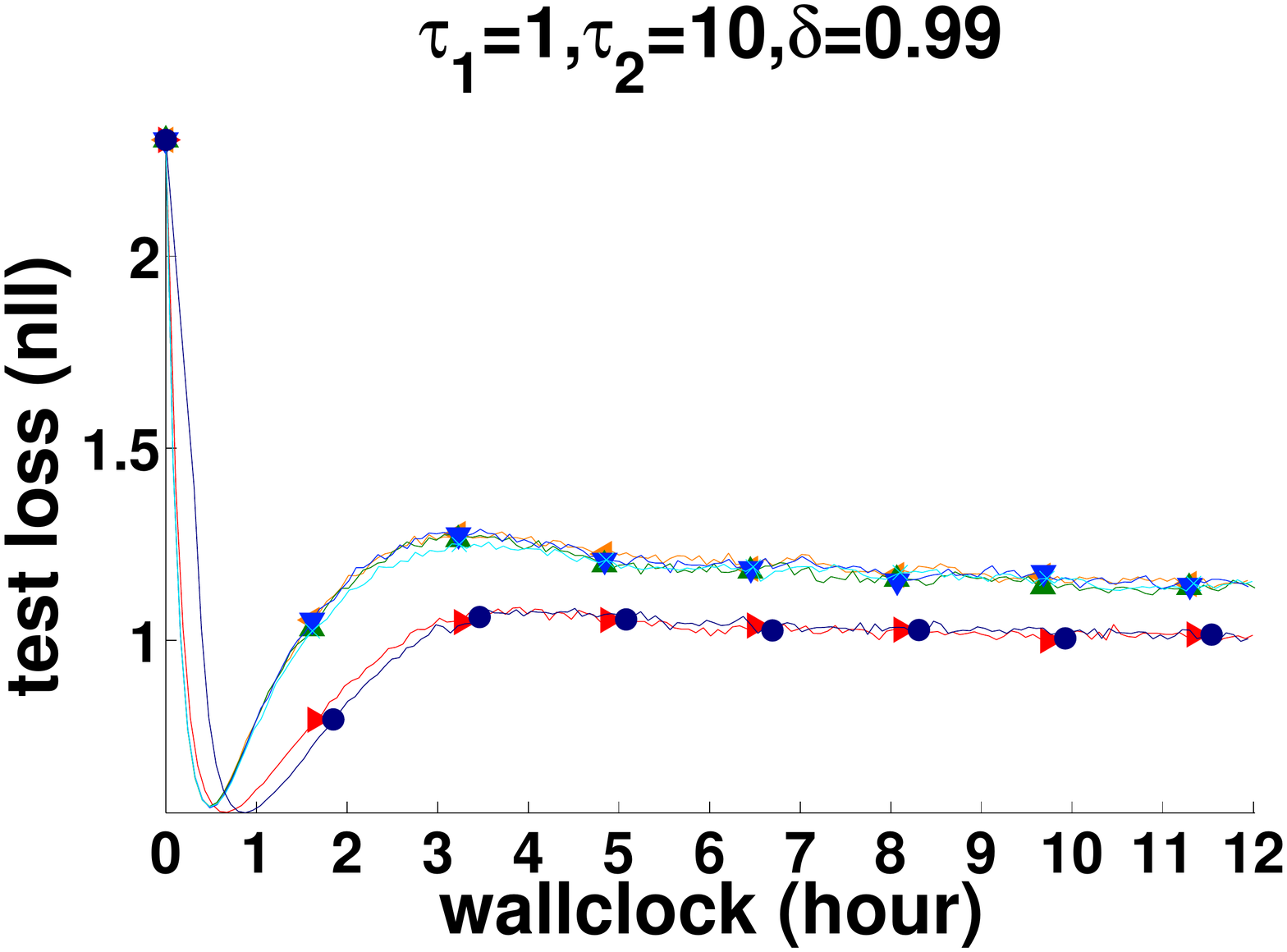} \\
\includegraphics[trim=0cm 0cm 0cm 0cm,clip,width = 3in]{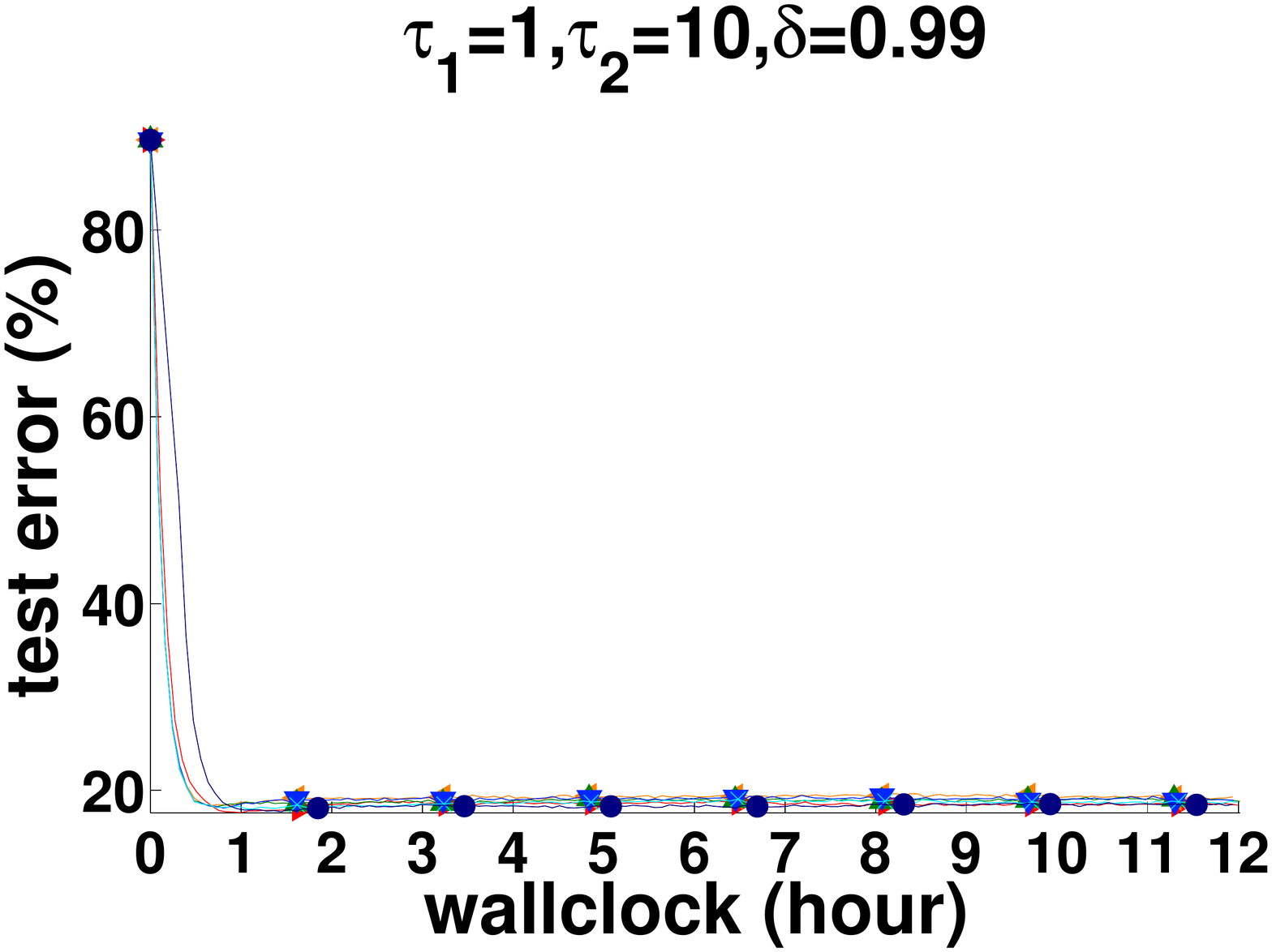}
\hspace{-0.3in}\includegraphics[trim=0cm 0cm 0cm 0cm,clip,width = 3in]{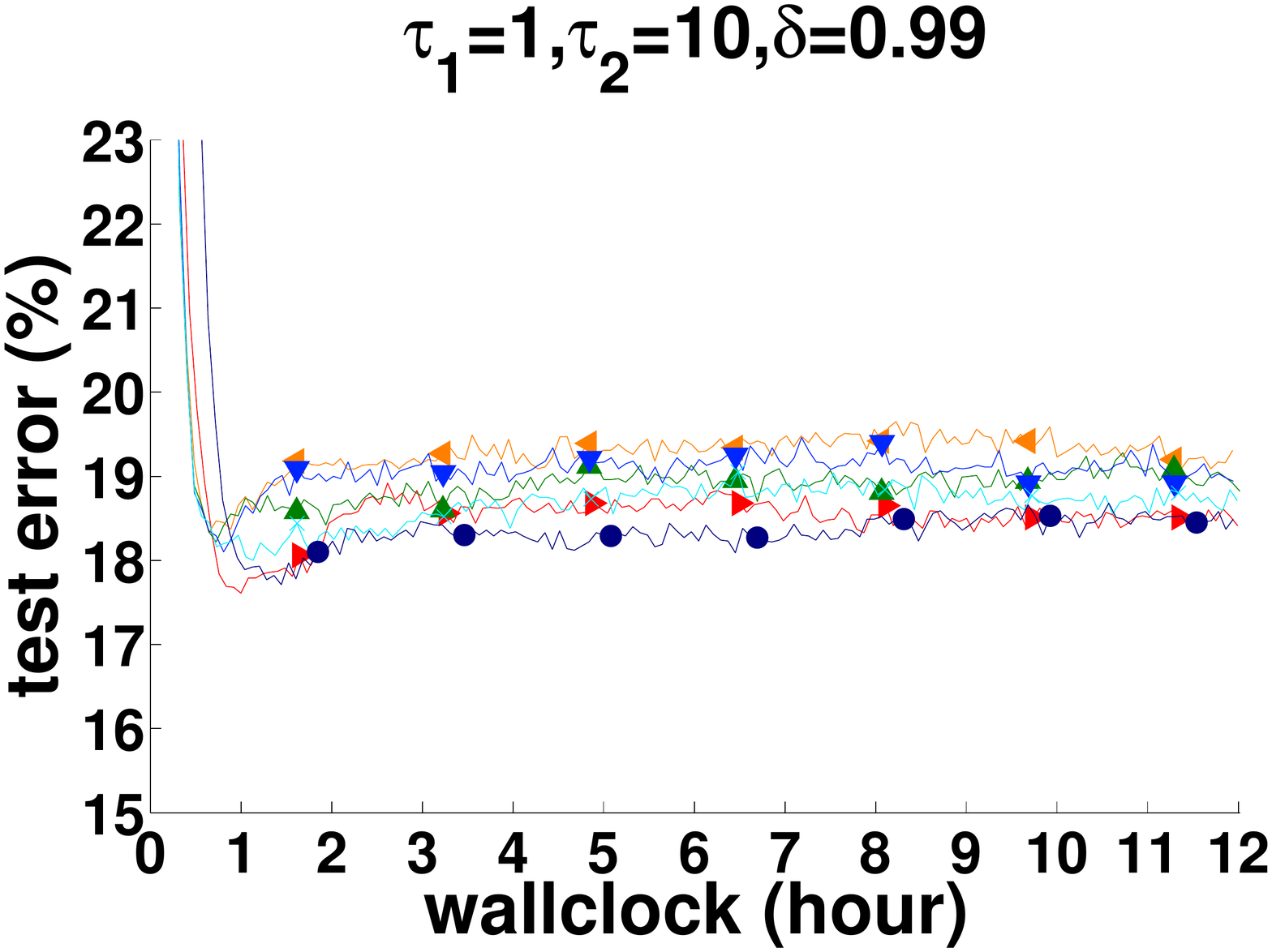}\\
\caption{\textit{EASGD Tree} on \textit{CIFAR-lowrank} using the first communication scheme with $\tau_1=1$ and $\tau_2=10$. Training loss and error, test loss and error of the root node versus a wallclock time. We run this experiment six times independently (labeled with a,b,c,d,e,f) with a same random initialization. $p=256,d=16,\eta=5e-4,\alpha=0.9/(d+1)$. The momentum rate $\delta=0.99$.
}
\label{fig:CIFARlr_mix123}
\end{figure}

\begin{figure}[p]
\center
\includegraphics[trim=0cm 0cm 0cm 0cm,clip,width = 3in]{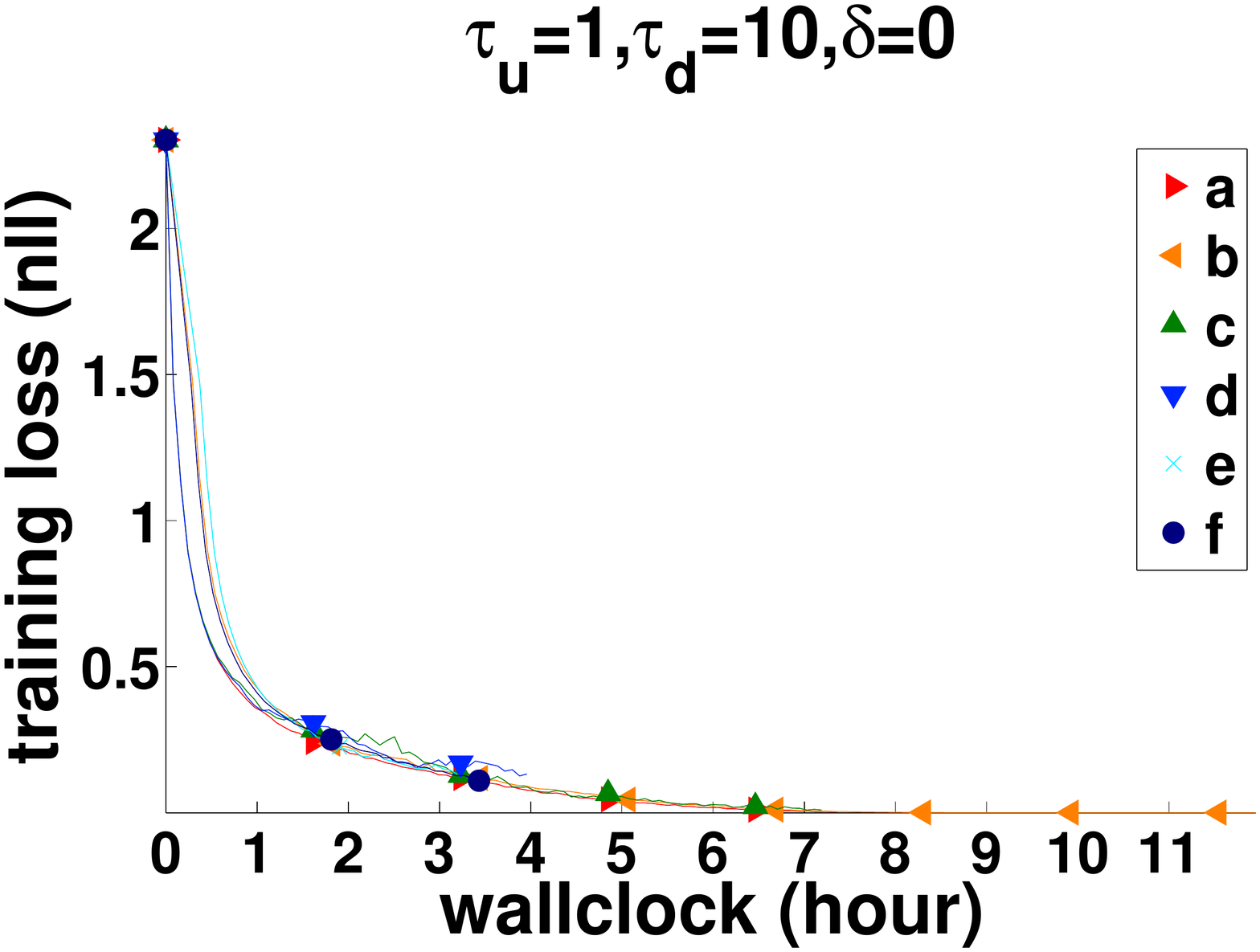}
\hspace{-0.3in}\includegraphics[trim=0cm 0cm 0cm 0cm,clip,width = 3in]{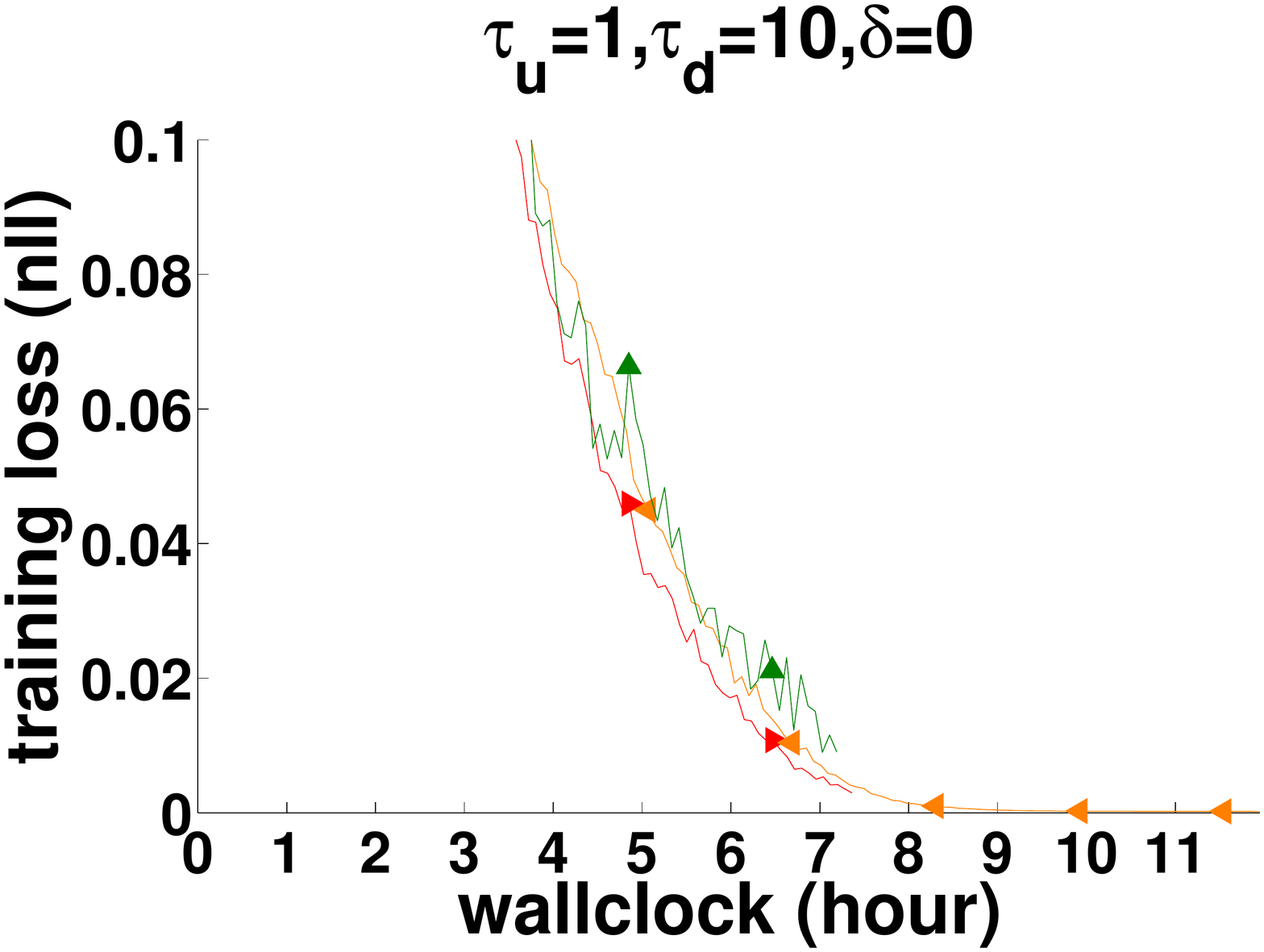} \\
\includegraphics[trim=0cm 0cm 0cm 0cm,clip,width = 3in]{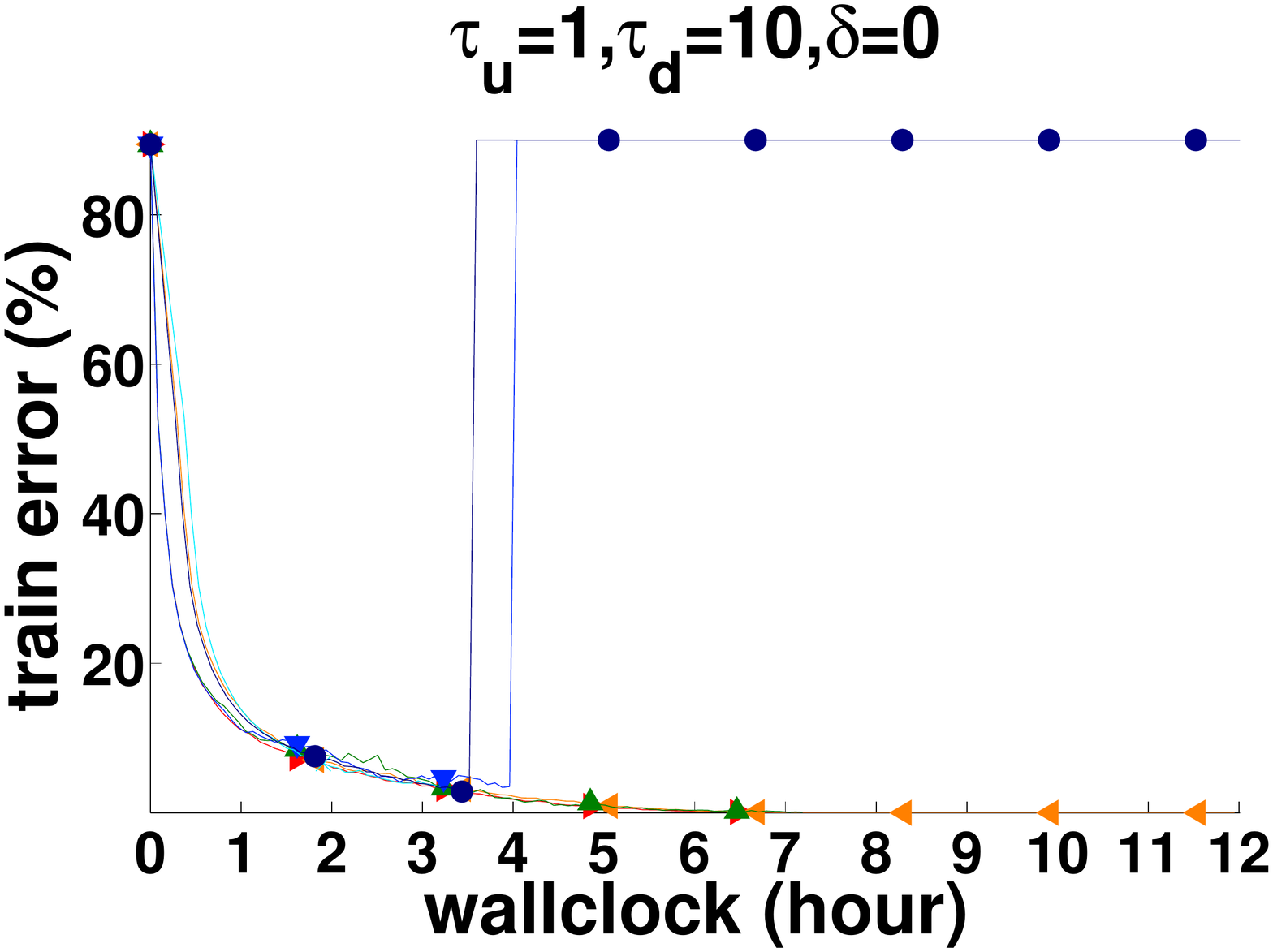}
\hspace{-0.3in}\includegraphics[trim=0cm 0cm 0cm 0cm,clip,width = 3in]{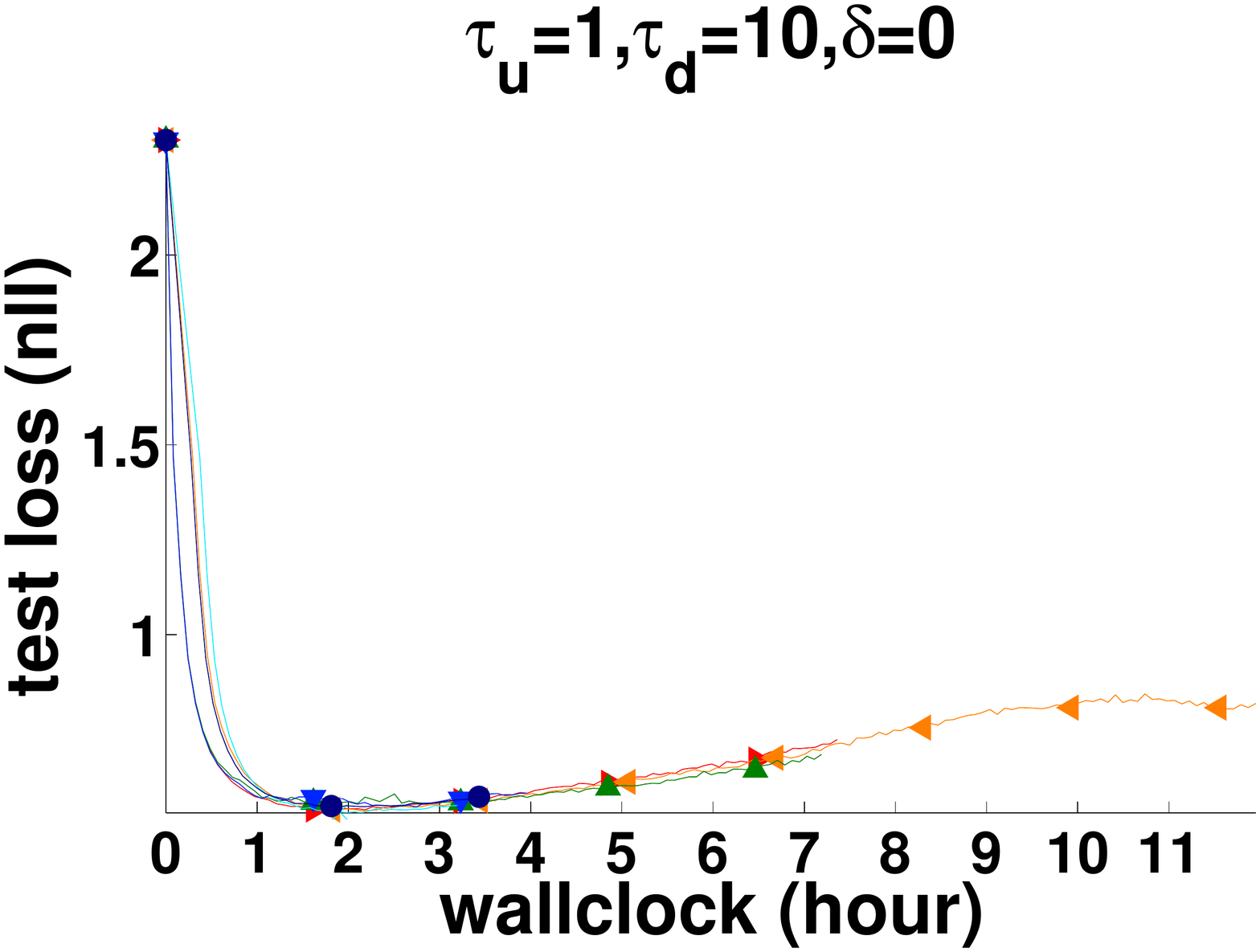} \\
\includegraphics[trim=0cm 0cm 0cm 0cm,clip,width = 3in]{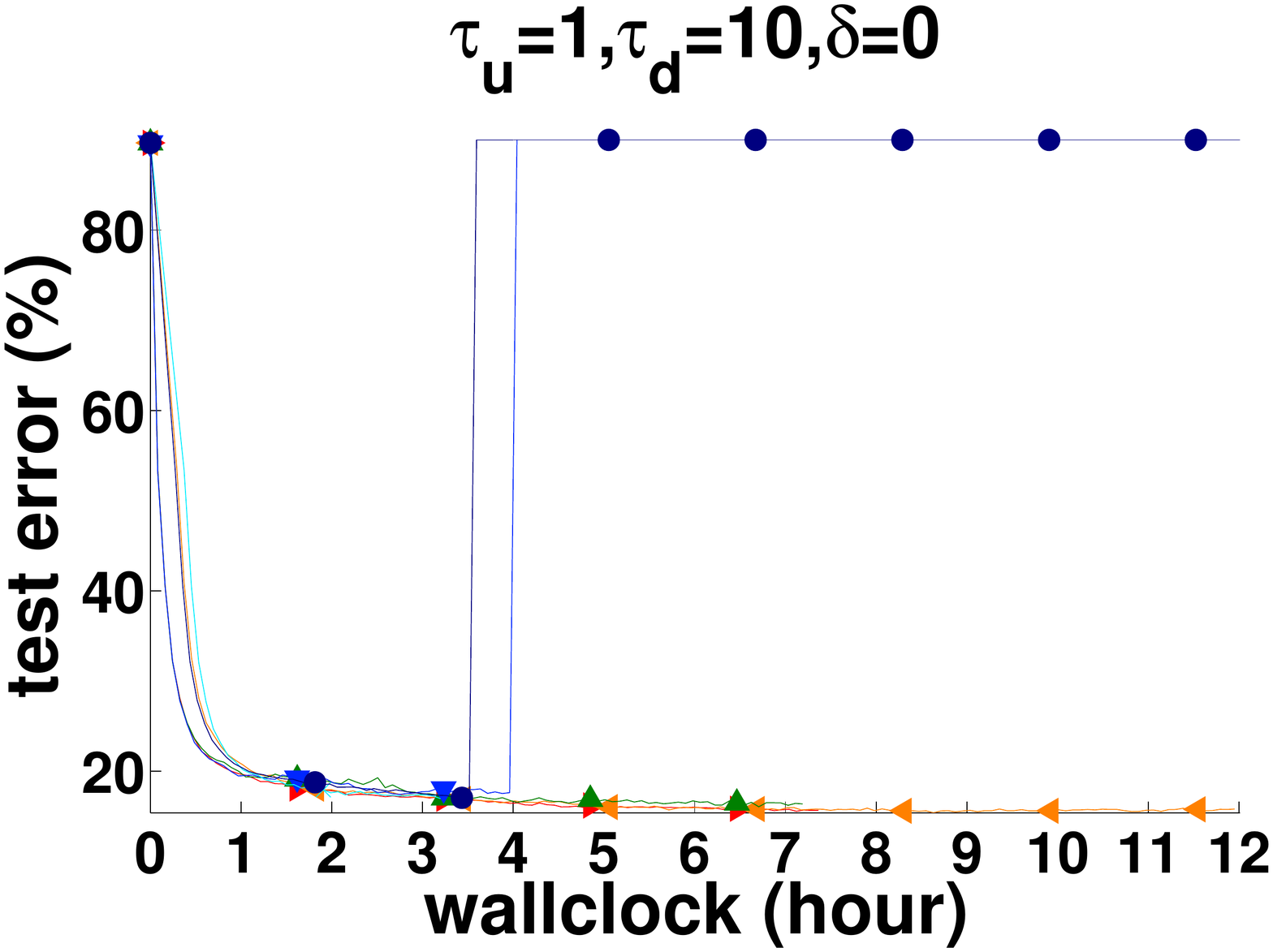}
\hspace{-0.3in}\includegraphics[trim=0cm 0cm 0cm 0cm,clip,width = 3in]{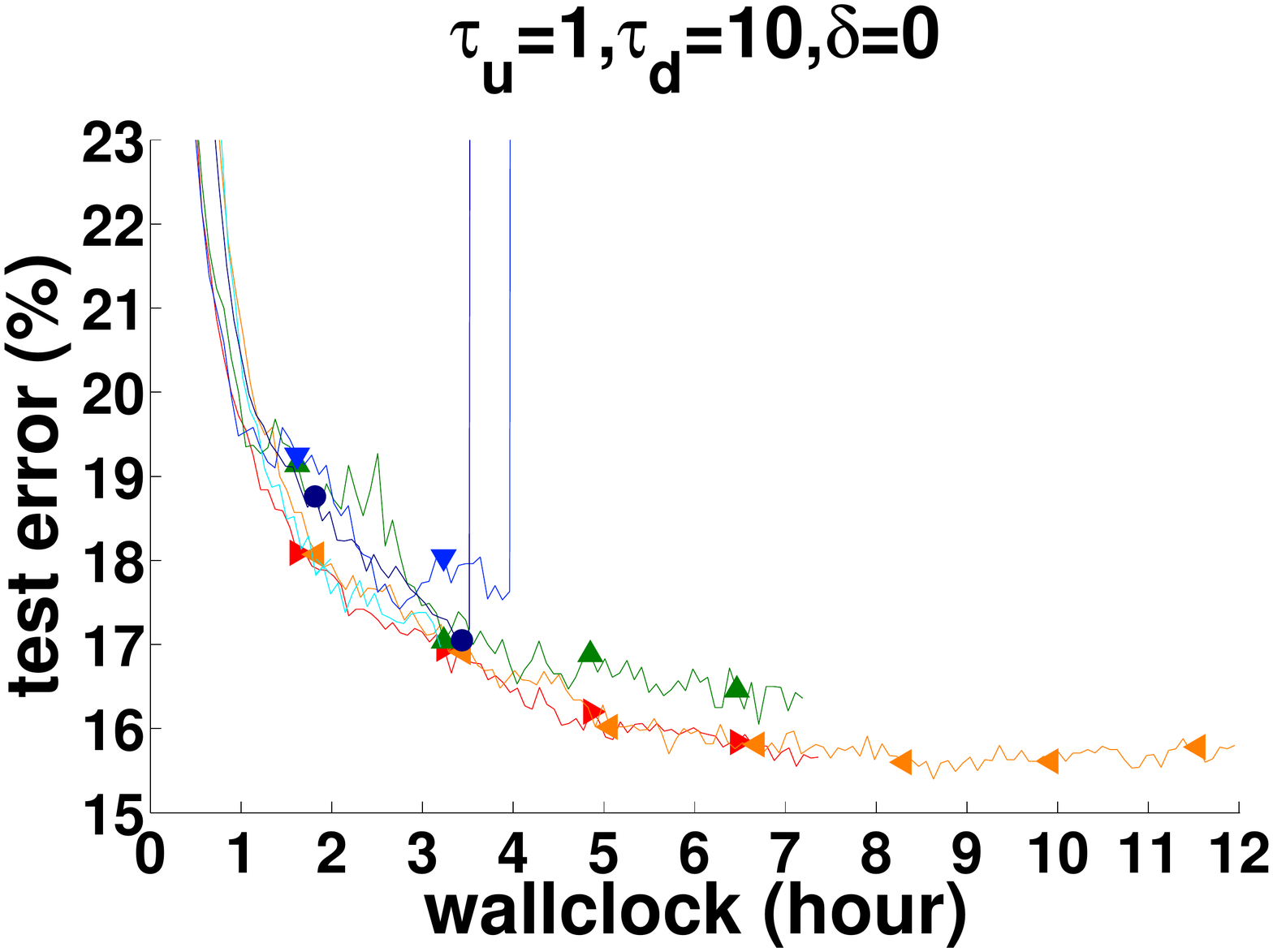}\\
\caption{\textit{EASGD Tree} on \textit{CIFAR-lowrank} using the second communication scheme with $\tau_{u}=1$ and $\tau_{d}=10$. Training loss and error, test loss and error of the root node versus a wallclock time. We run this experiment six times independently (labeled with a,b,c,d,e,f) with a same random initialization. $p=256,d=16,\eta=5e-2,\alpha=0.9/(d+1)$. The momentum rate $\delta=0$.
}
\label{fig:CIFARlr_mix124}
\end{figure}

\begin{figure}[p]
\center
\includegraphics[trim=0cm 0cm 0cm 0cm,clip,width = 3in]{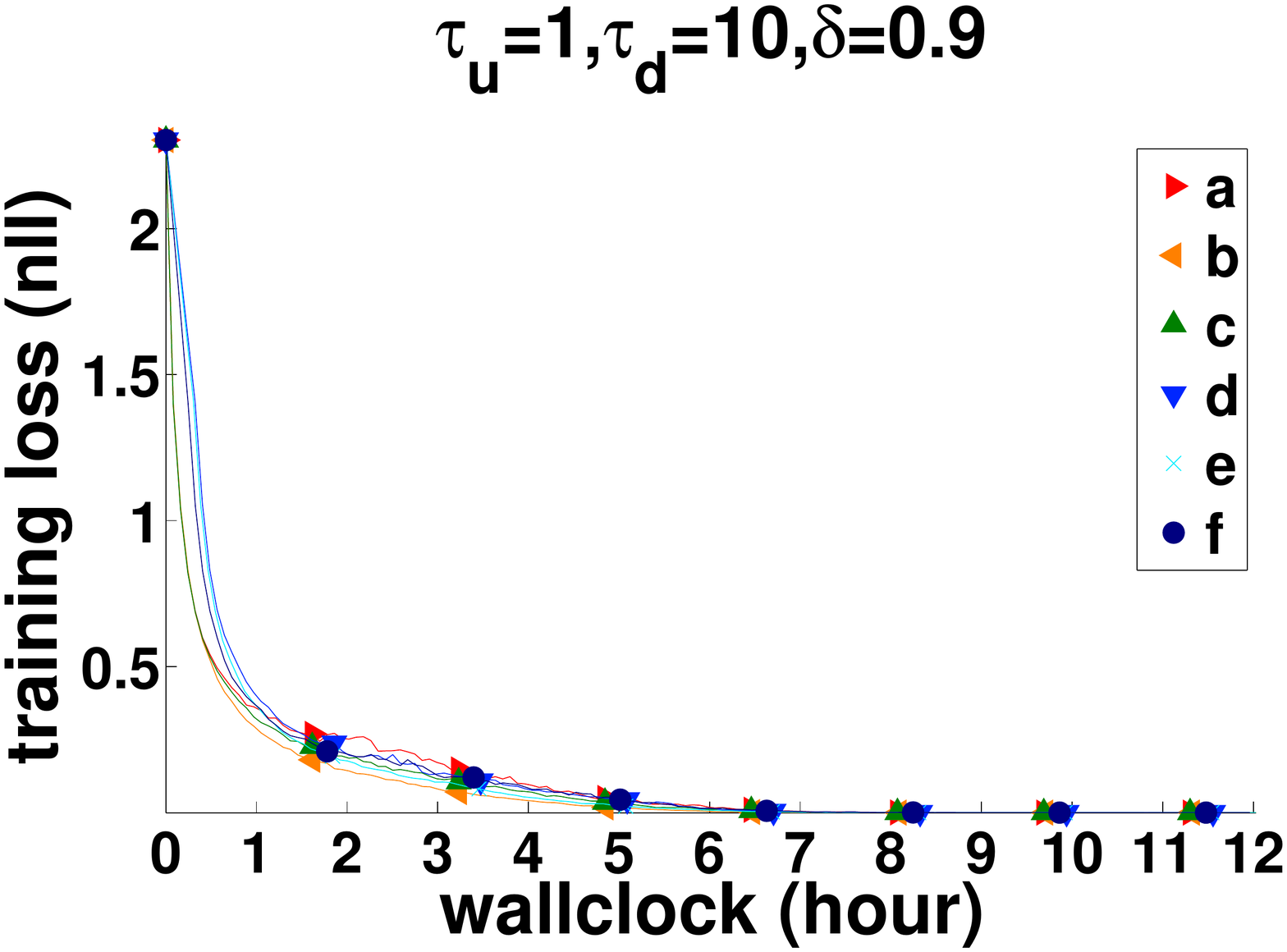}
\hspace{-0.3in}\includegraphics[trim=0cm 0cm 0cm 0cm,clip,width = 3in]{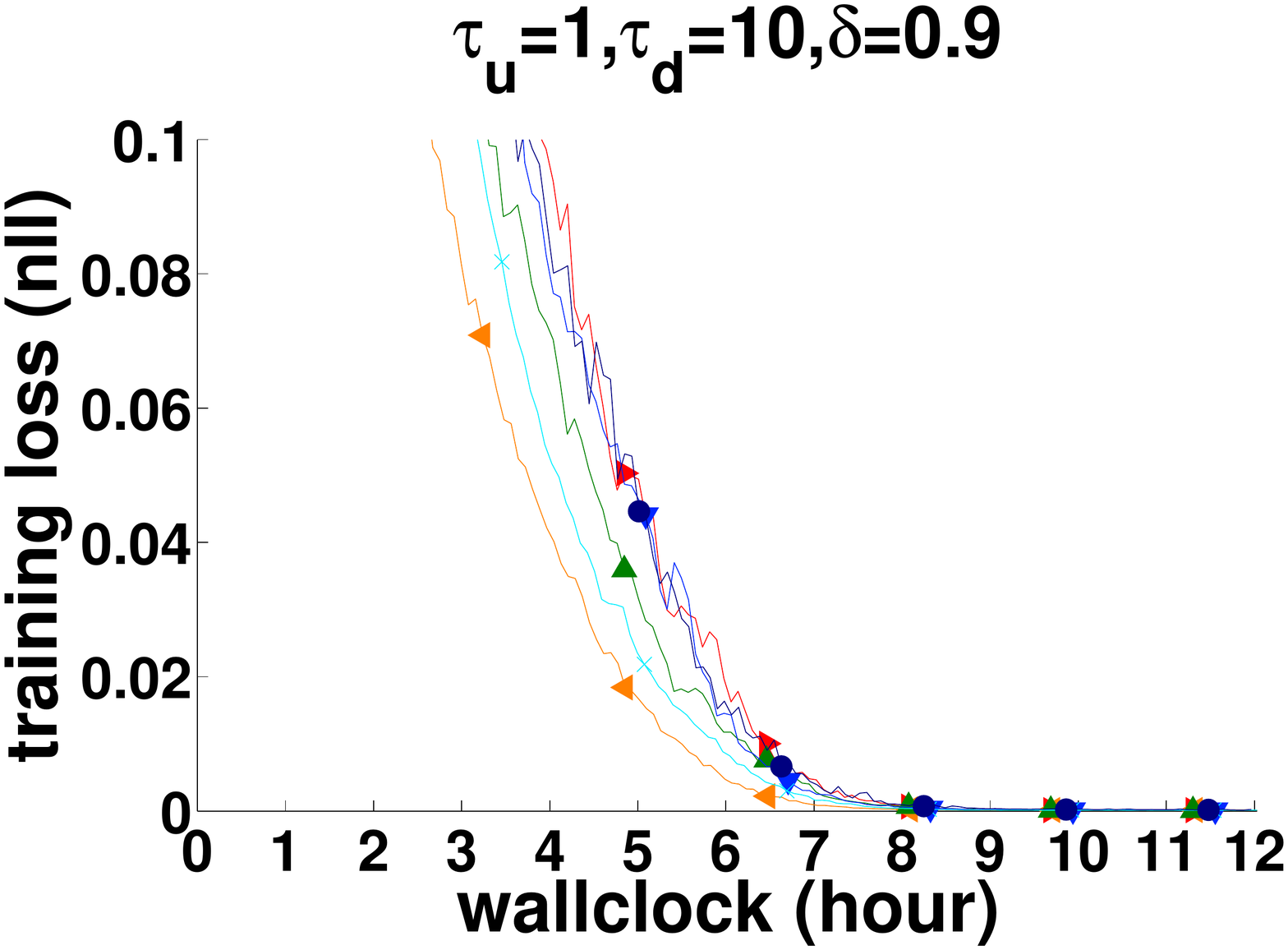} \\
\includegraphics[trim=0cm 0cm 0cm 0cm,clip,width = 3in]{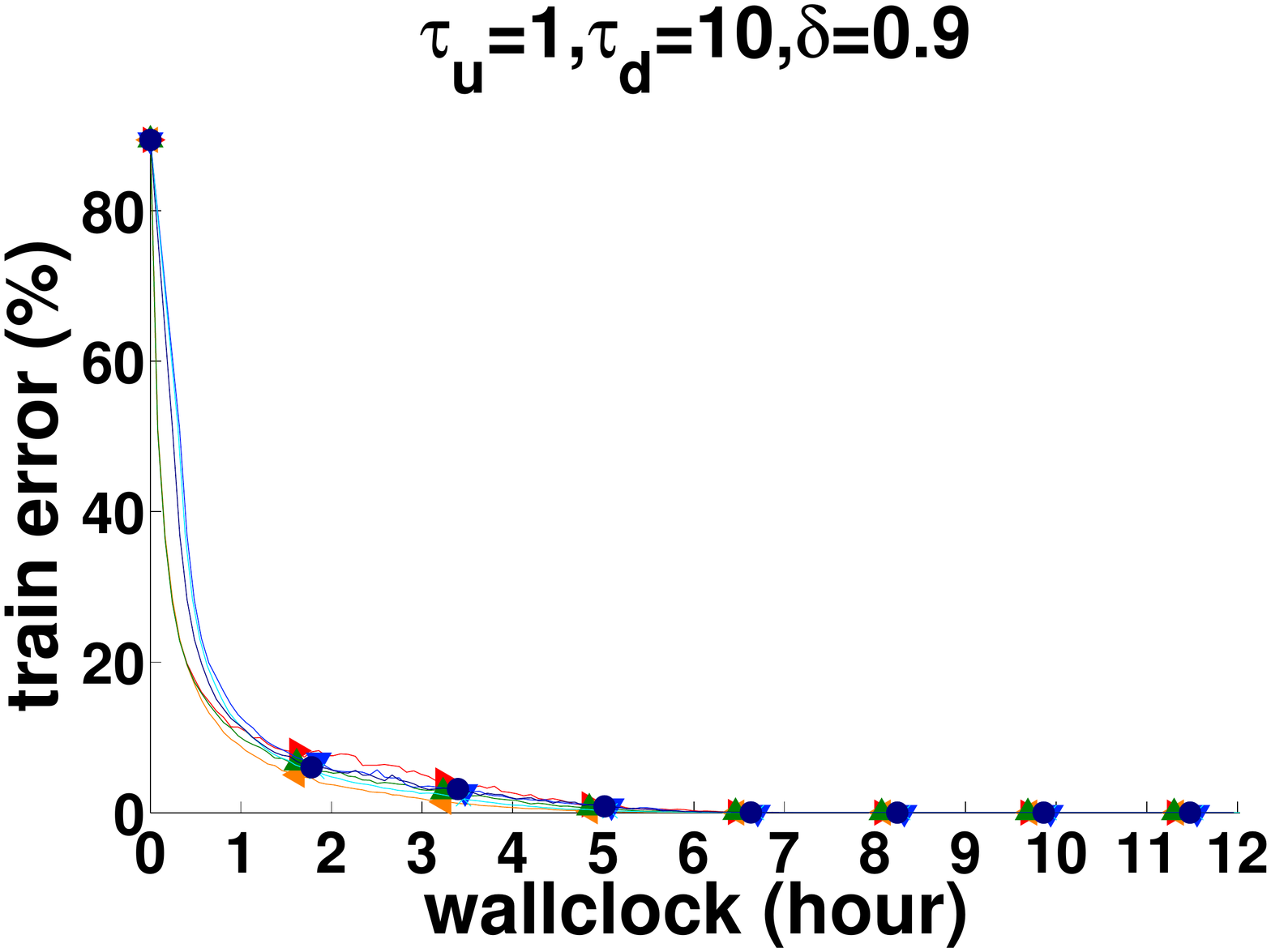}
\hspace{-0.3in}\includegraphics[trim=0cm 0cm 0cm 0cm,clip,width = 3in]{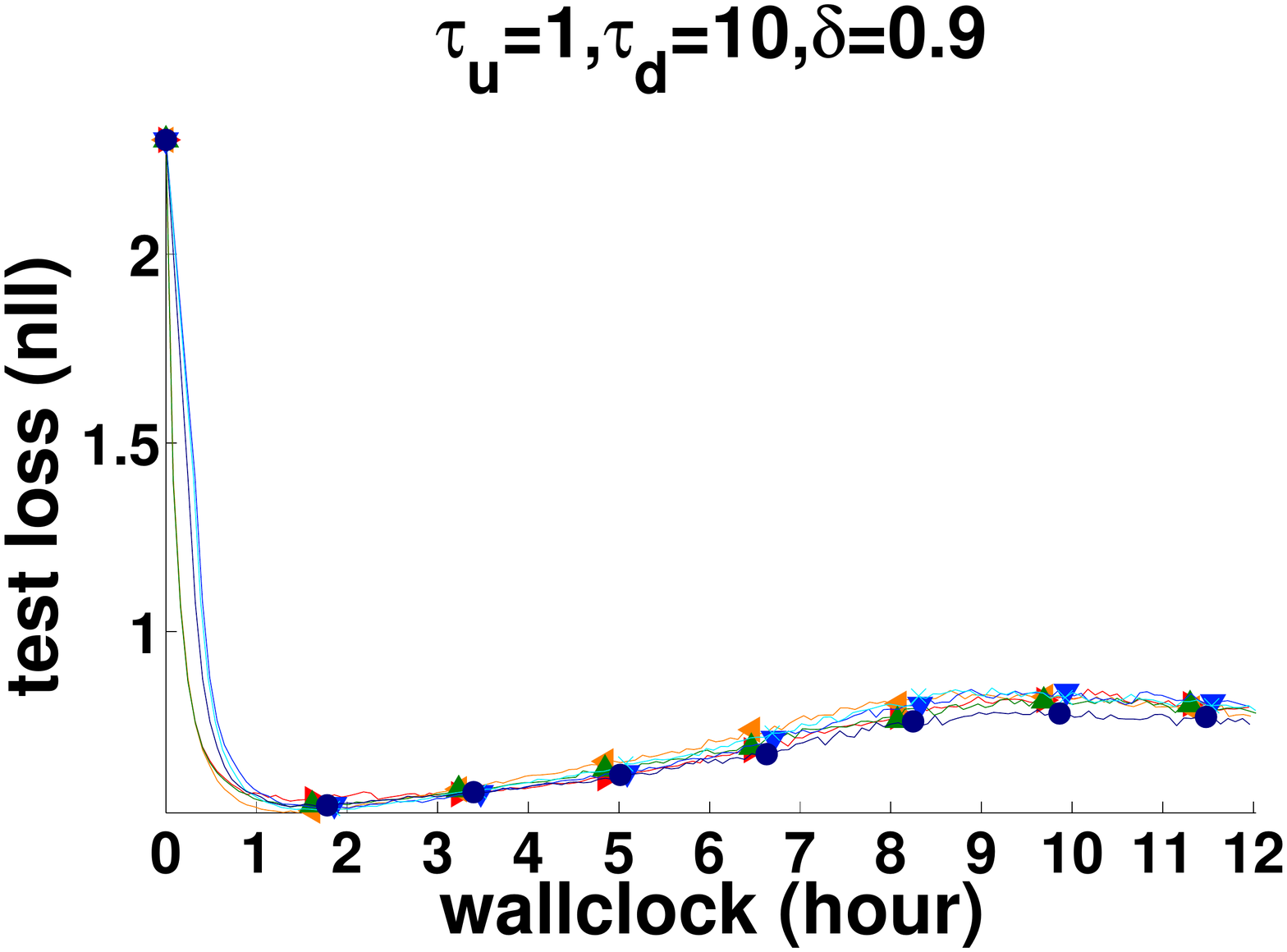} \\
\includegraphics[trim=0cm 0cm 0cm 0cm,clip,width = 3in]{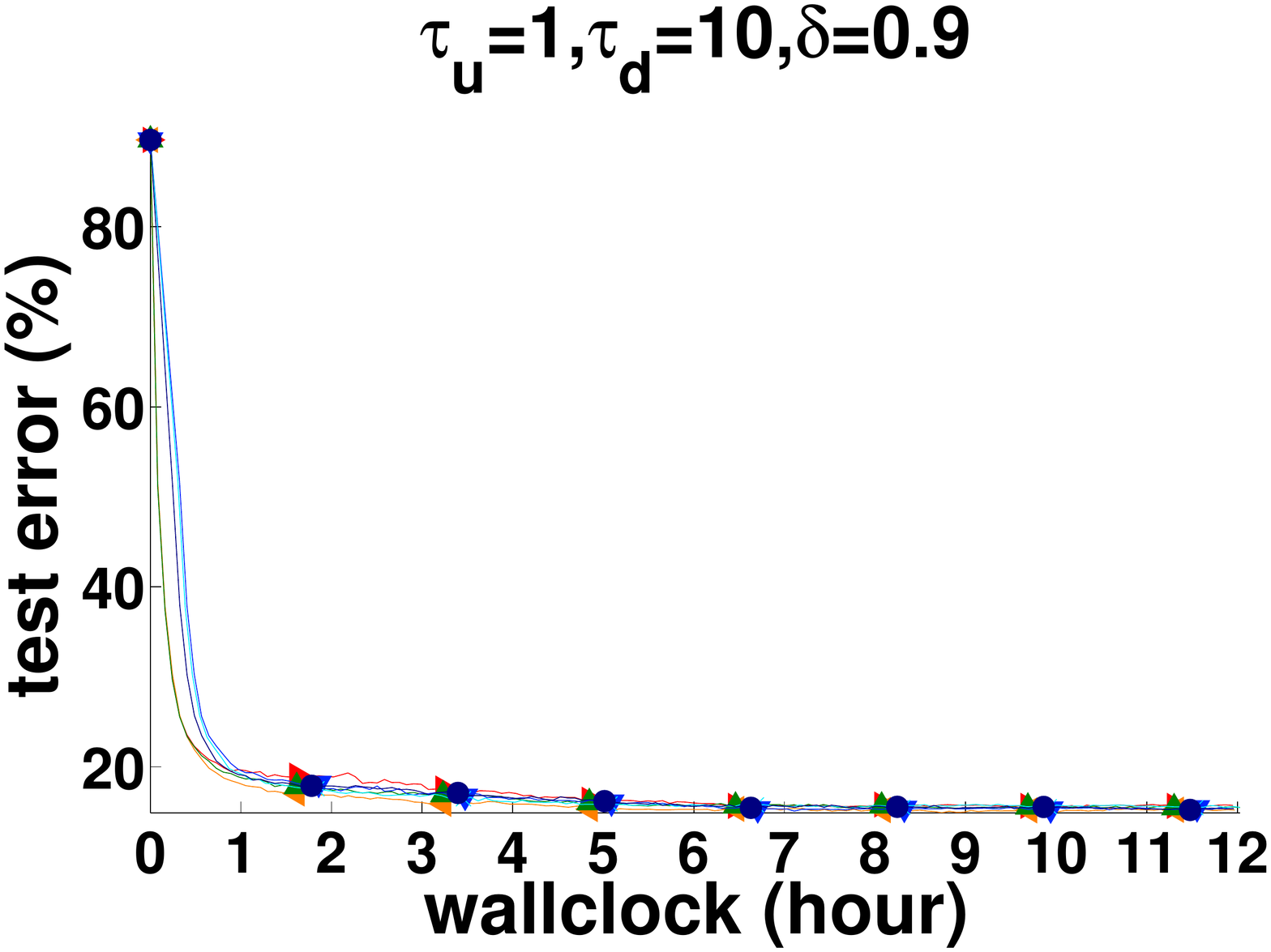}
\hspace{-0.3in}\includegraphics[trim=0cm 0cm 0cm 0cm,clip,width = 3in]{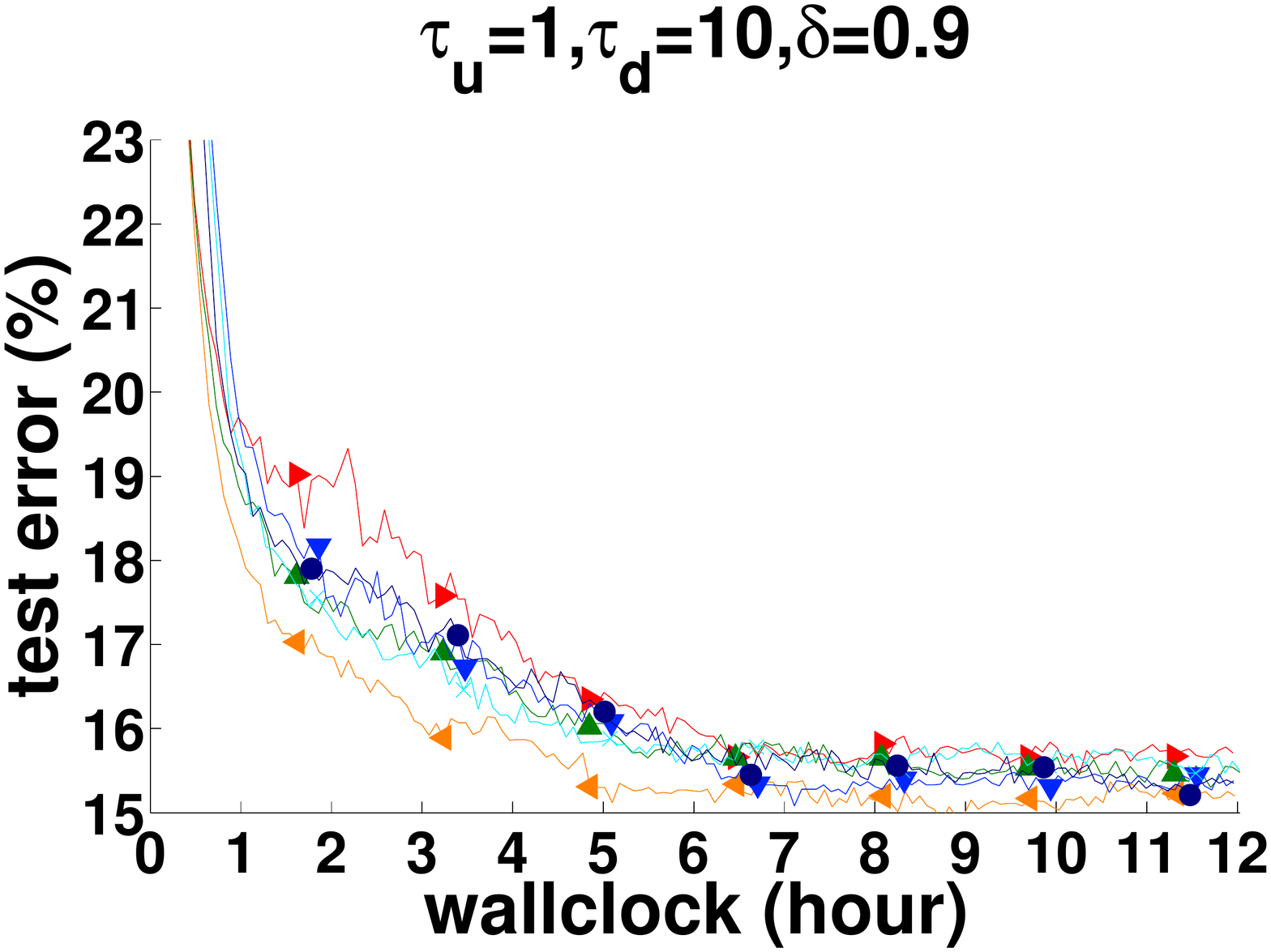}\\
\caption{\textit{EASGD Tree} on \textit{CIFAR-lowrank} using the second communication scheme with $\tau_{u}=1$ and $\tau_{d}=10$. Training loss and error, test loss and error of the root node versus a wallclock time. We run this experiment six times independently (labeled with a,b,c,d,e,f) with a same random initialization. $p=256,d=16,\eta=5e-3,\alpha=0.9/(d+1)$. The momentum rate $\delta=0.9$.
}
\label{fig:CIFARlr_mix125}
\end{figure}

\begin{figure}[p]
\center
\includegraphics[trim=0cm 0cm 0cm 0cm,clip,width = 3in]{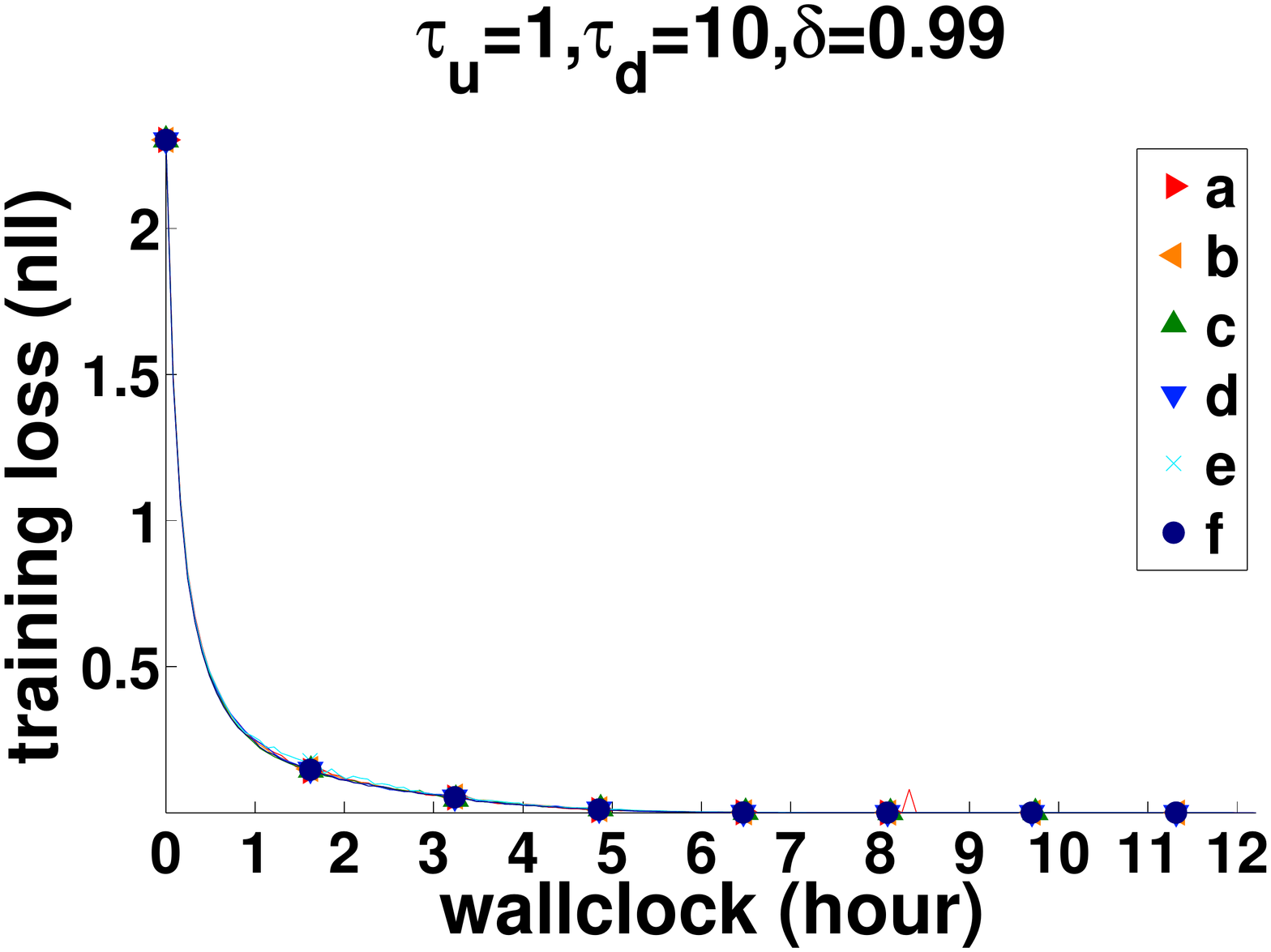}
\hspace{-0.3in}\includegraphics[trim=0cm 0cm 0cm 0cm,clip,width = 3in]{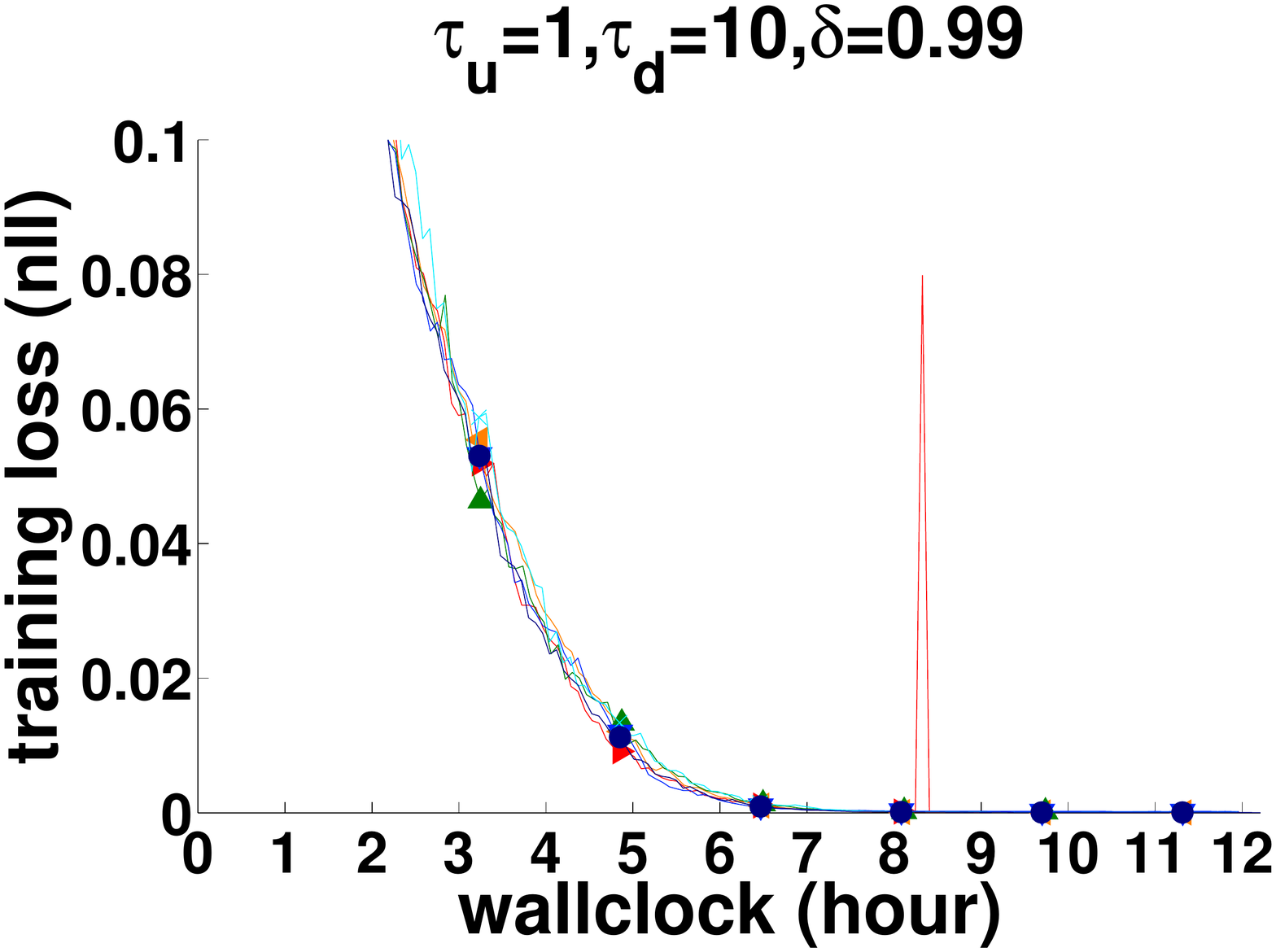} \\
\includegraphics[trim=0cm 0cm 0cm 0cm,clip,width = 3in]{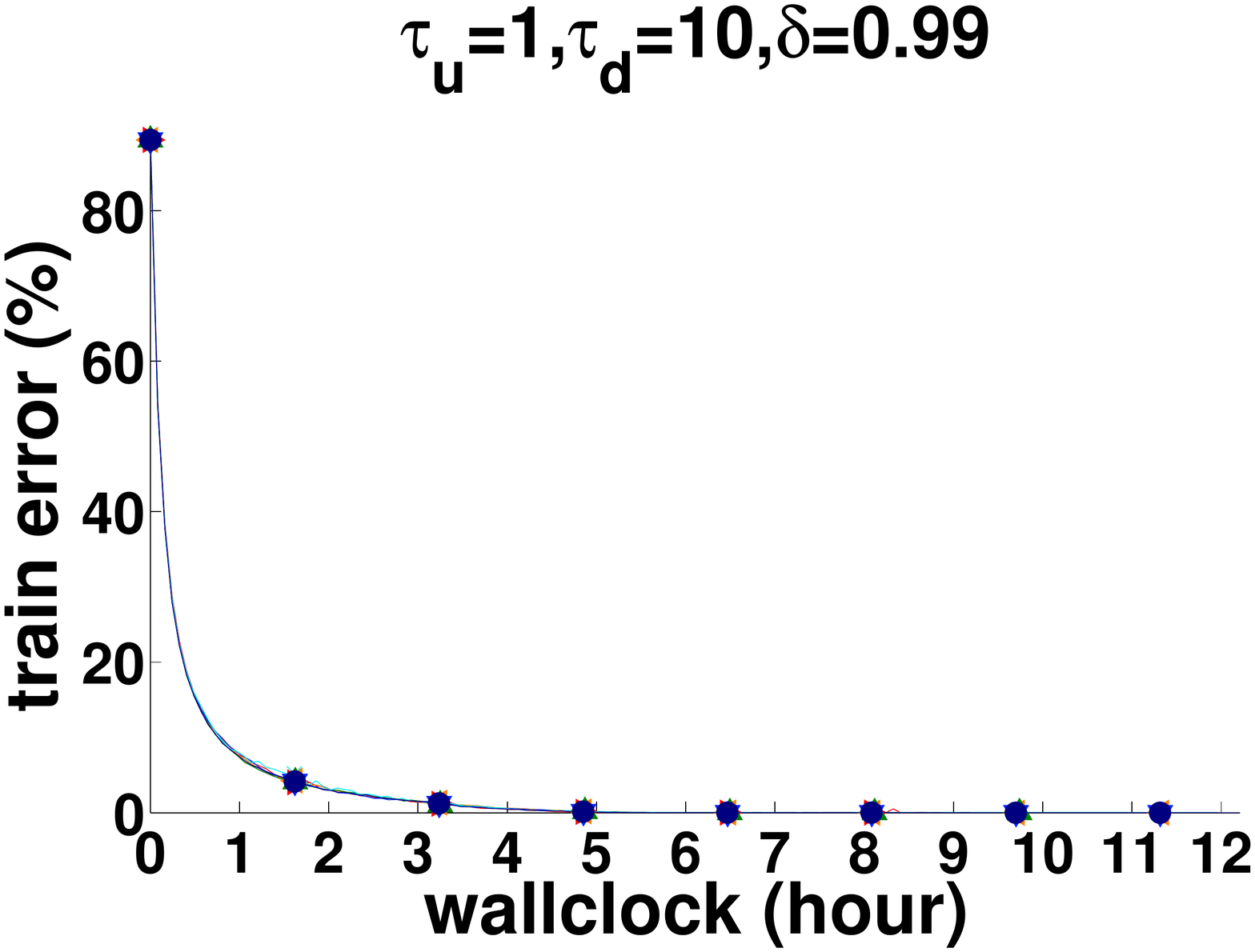}
\hspace{-0.3in}\includegraphics[trim=0cm 0cm 0cm 0cm,clip,width = 3in]{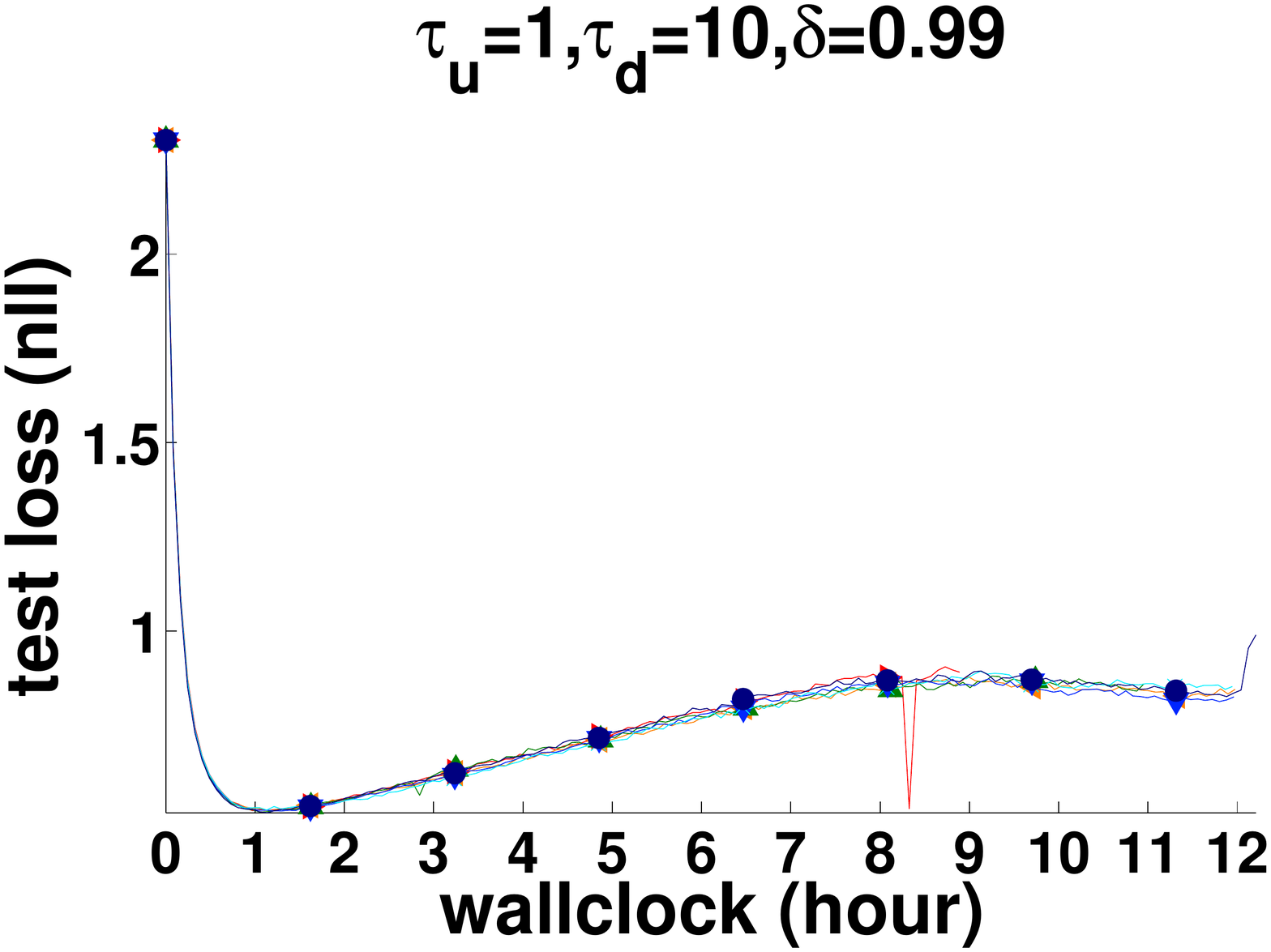} \\
\includegraphics[trim=0cm 0cm 0cm 0cm,clip,width = 3in]{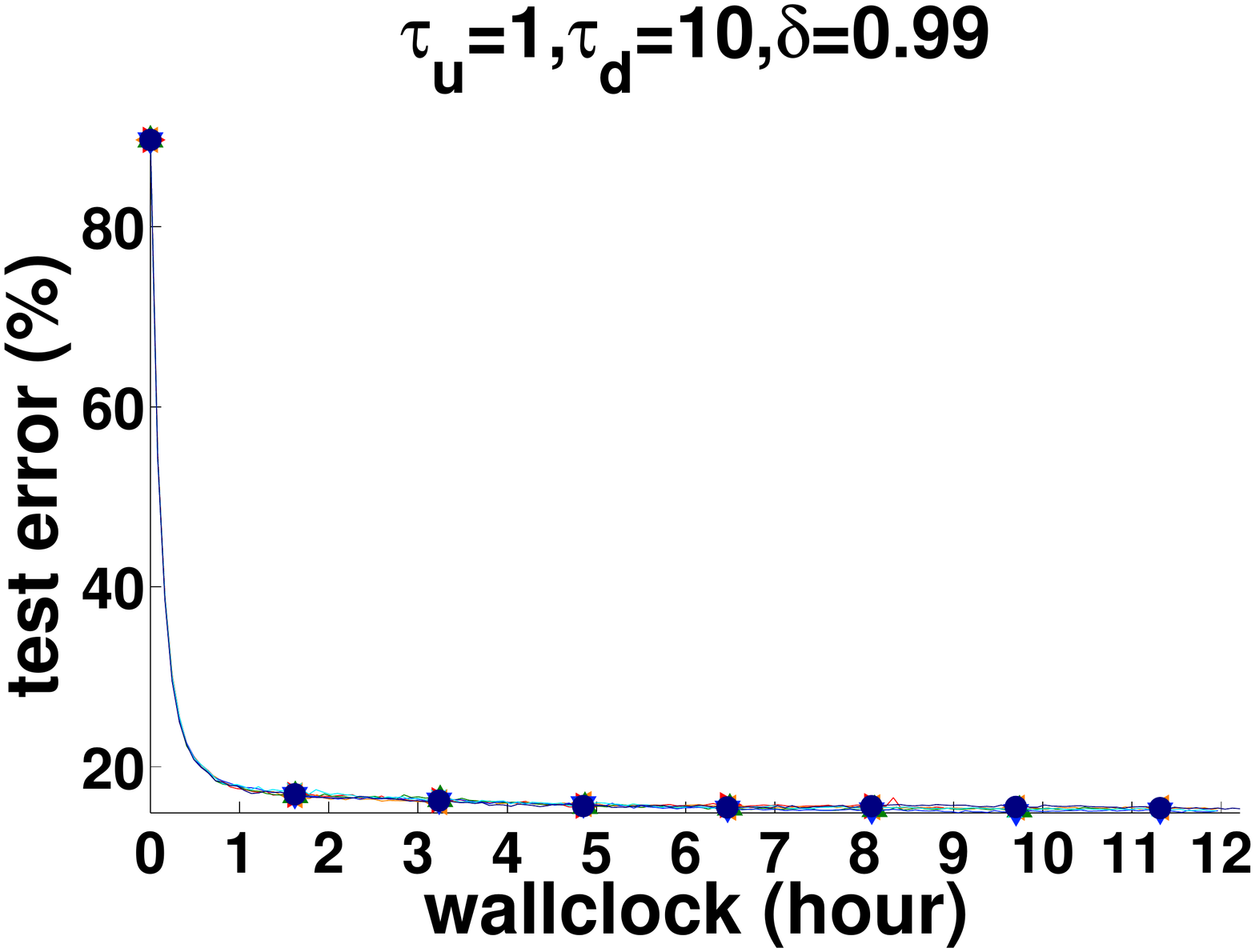}
\hspace{-0.3in}\includegraphics[trim=0cm 0cm 0cm 0cm,clip,width = 3in]{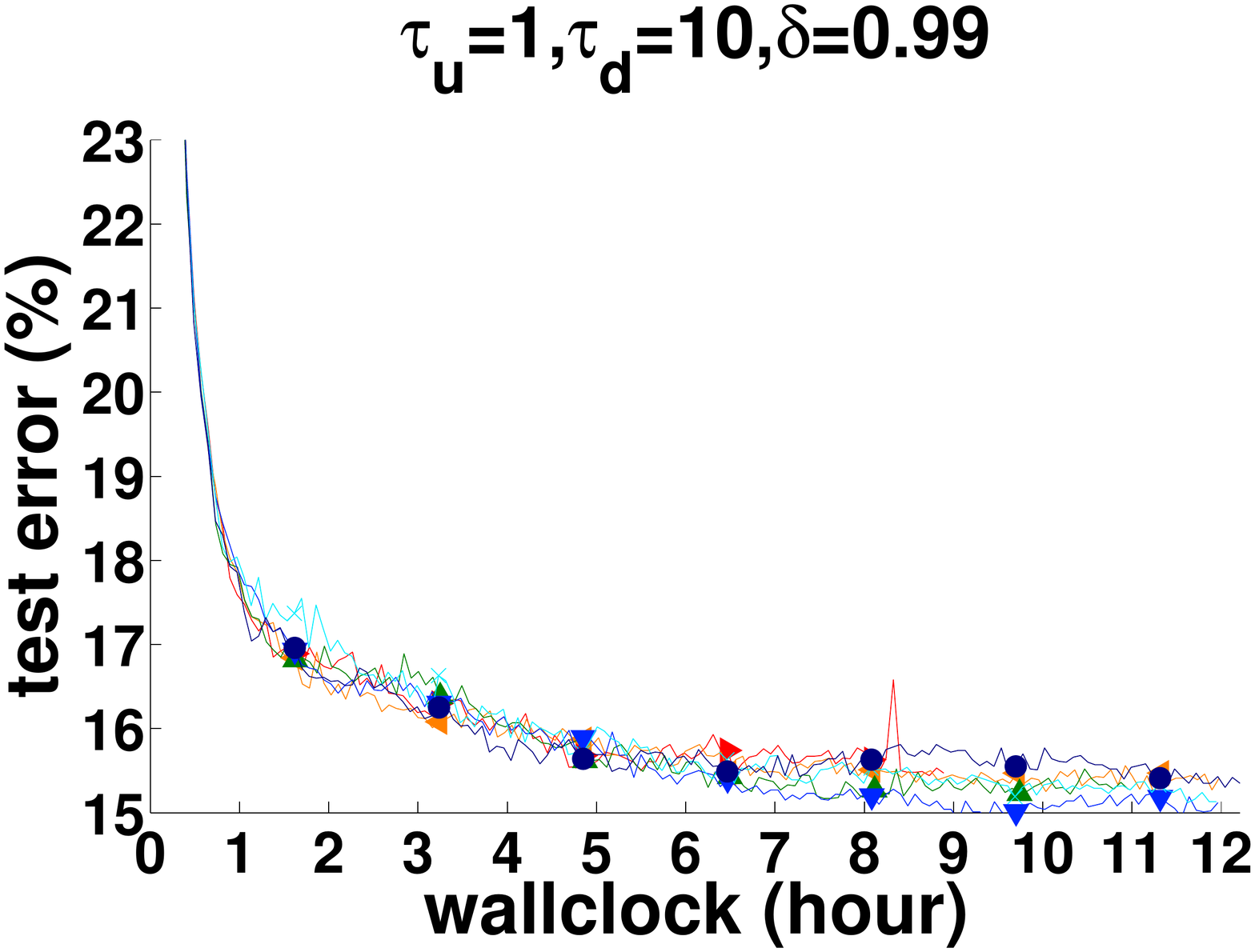}\\
\caption{\textit{EASGD Tree} on \textit{CIFAR-lowrank} using the second communication scheme with $\tau_{u}=1$ and $\tau_{d}=10$. Training loss and error, test loss and error of the root node versus a wallclock time. We run this experiment six times independently (labeled with a,b,c,d,e,f) with a same random initialization. $p=256,d=16,\eta=5e-4,\alpha=0.9/(d+1)$. The momentum rate $\delta=0.99$.
}
\label{fig:CIFARlr_mix126}
\end{figure}

\begin{figure}[p]
\center
\includegraphics[trim=0cm 0cm 0cm 0cm,clip,width = 3in]{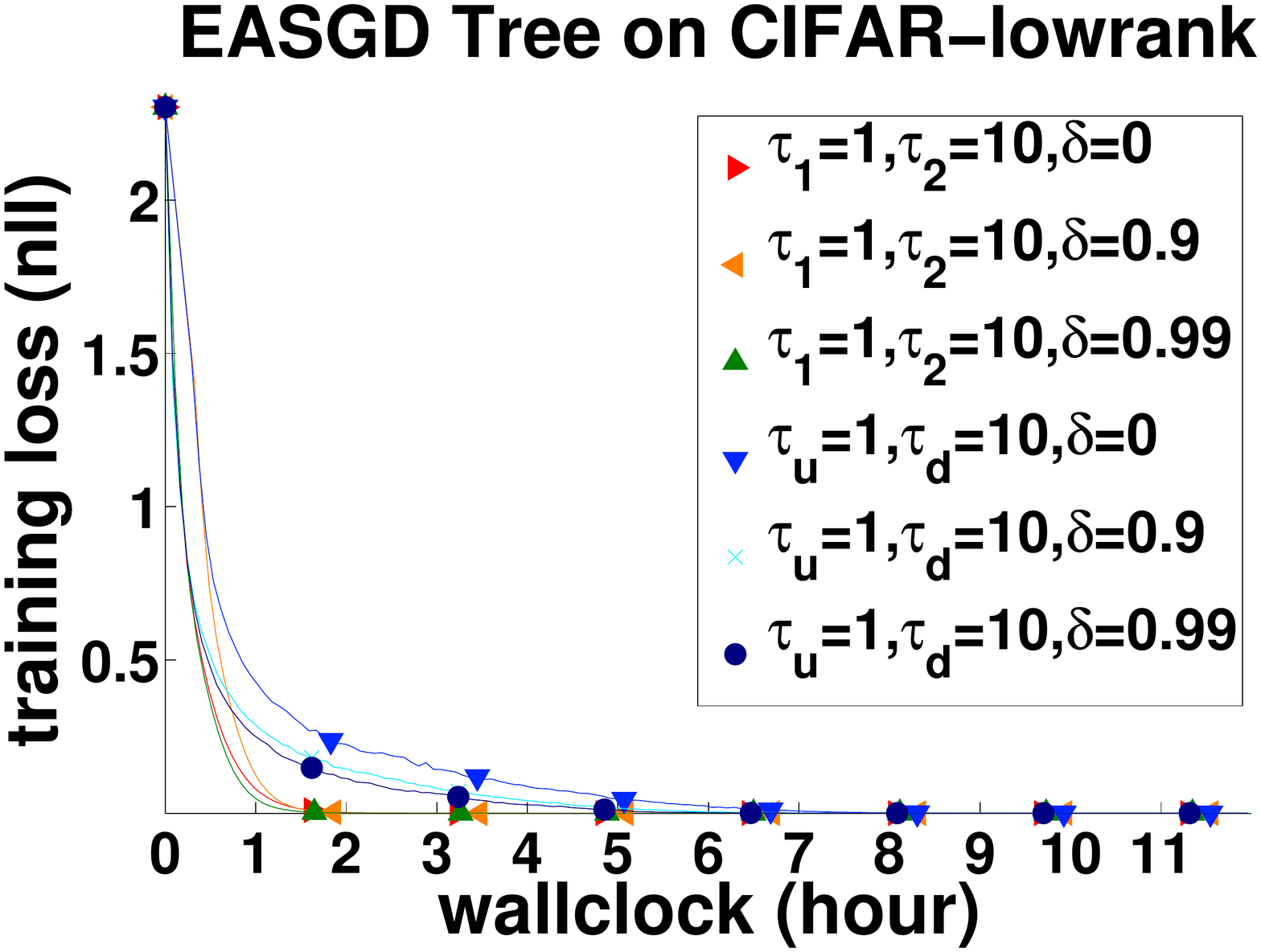}
\hspace{-0.3in}\includegraphics[trim=0cm 0cm 0cm 0cm,clip,width = 3in]{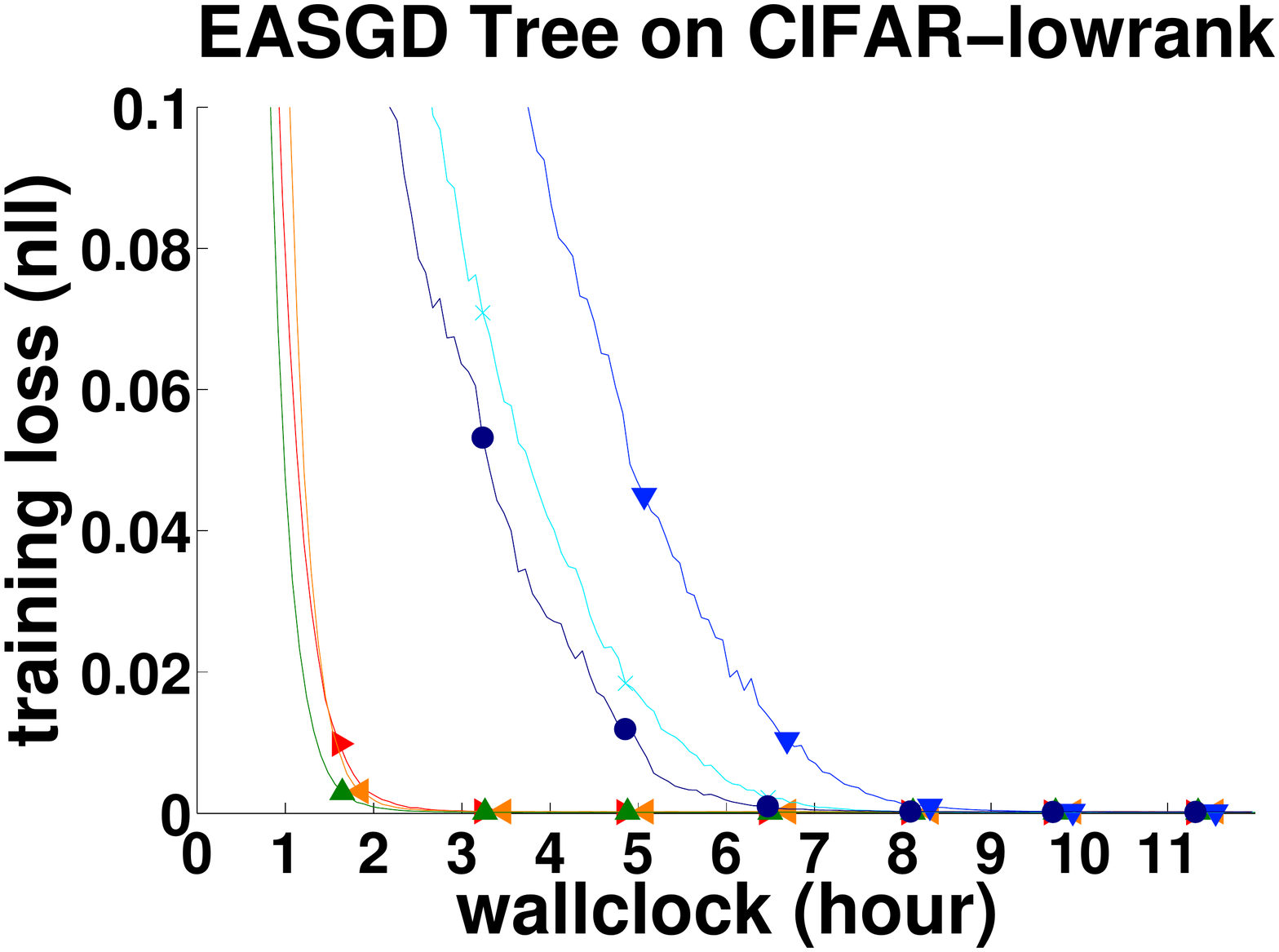} \\
\includegraphics[trim=0cm 0cm 0cm 0cm,clip,width = 3in]{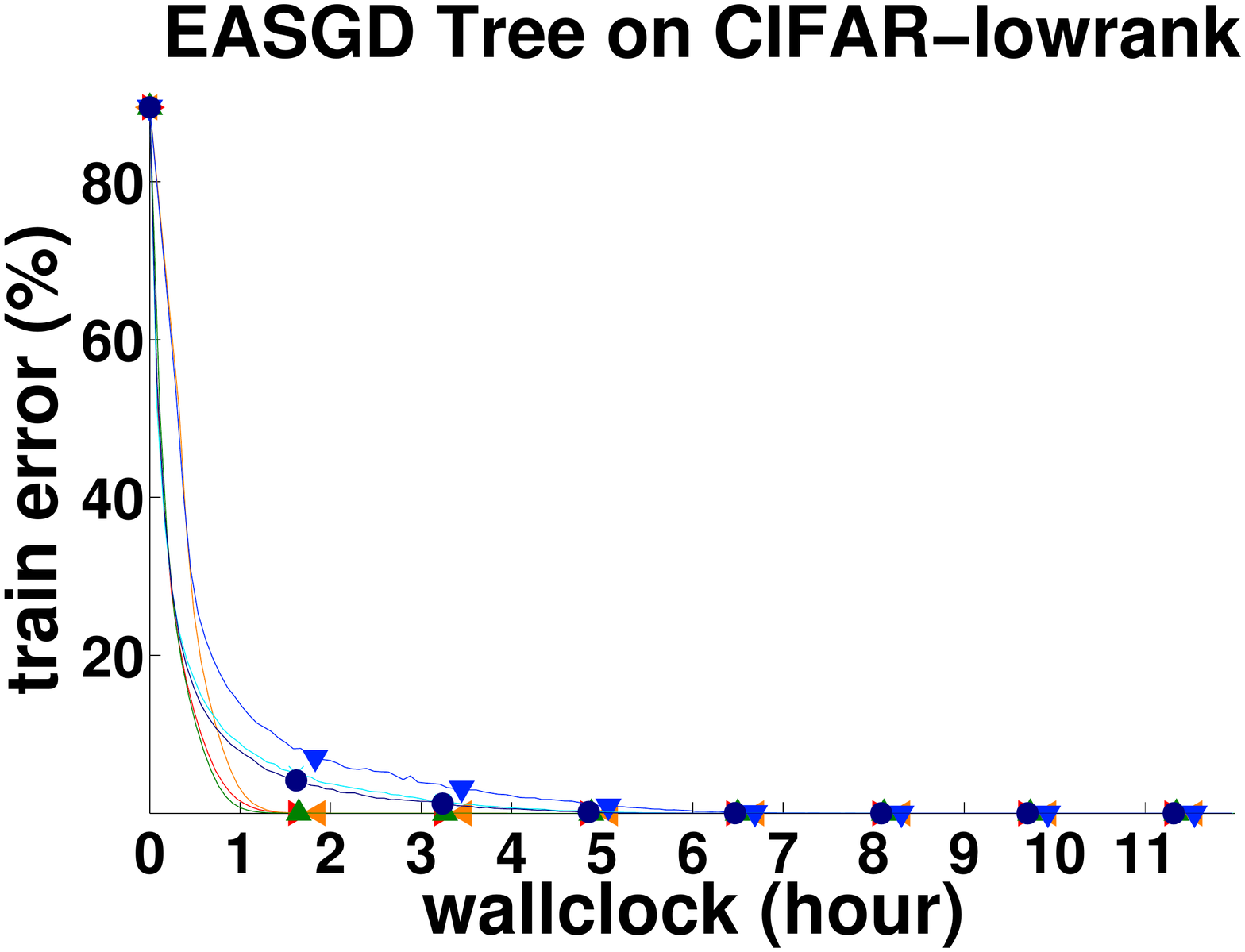}
\hspace{-0.3in}\includegraphics[trim=0cm 0cm 0cm 0cm,clip,width = 3in]{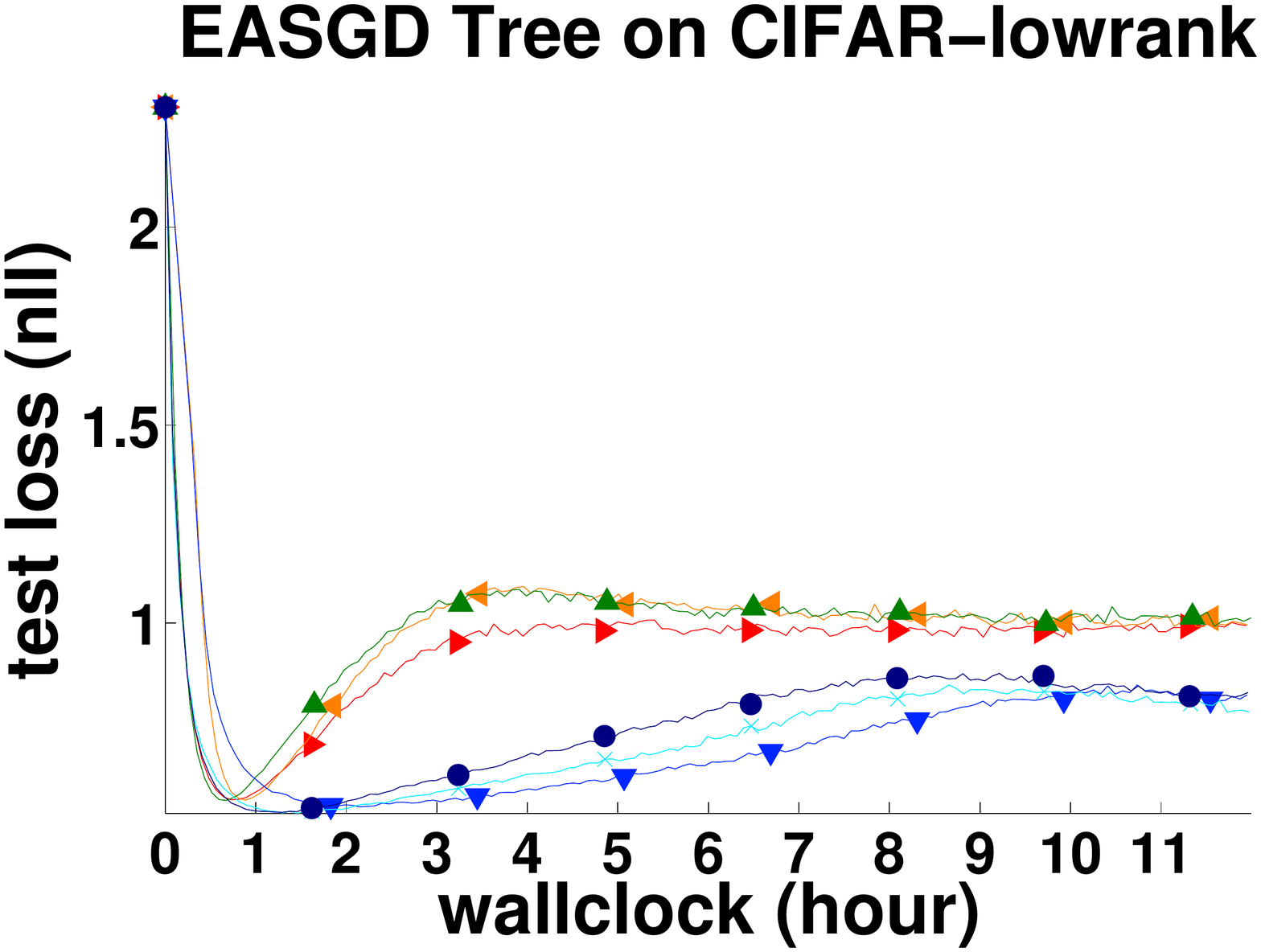} \\
\includegraphics[trim=0cm 0cm 0cm 0cm,clip,width = 3in]{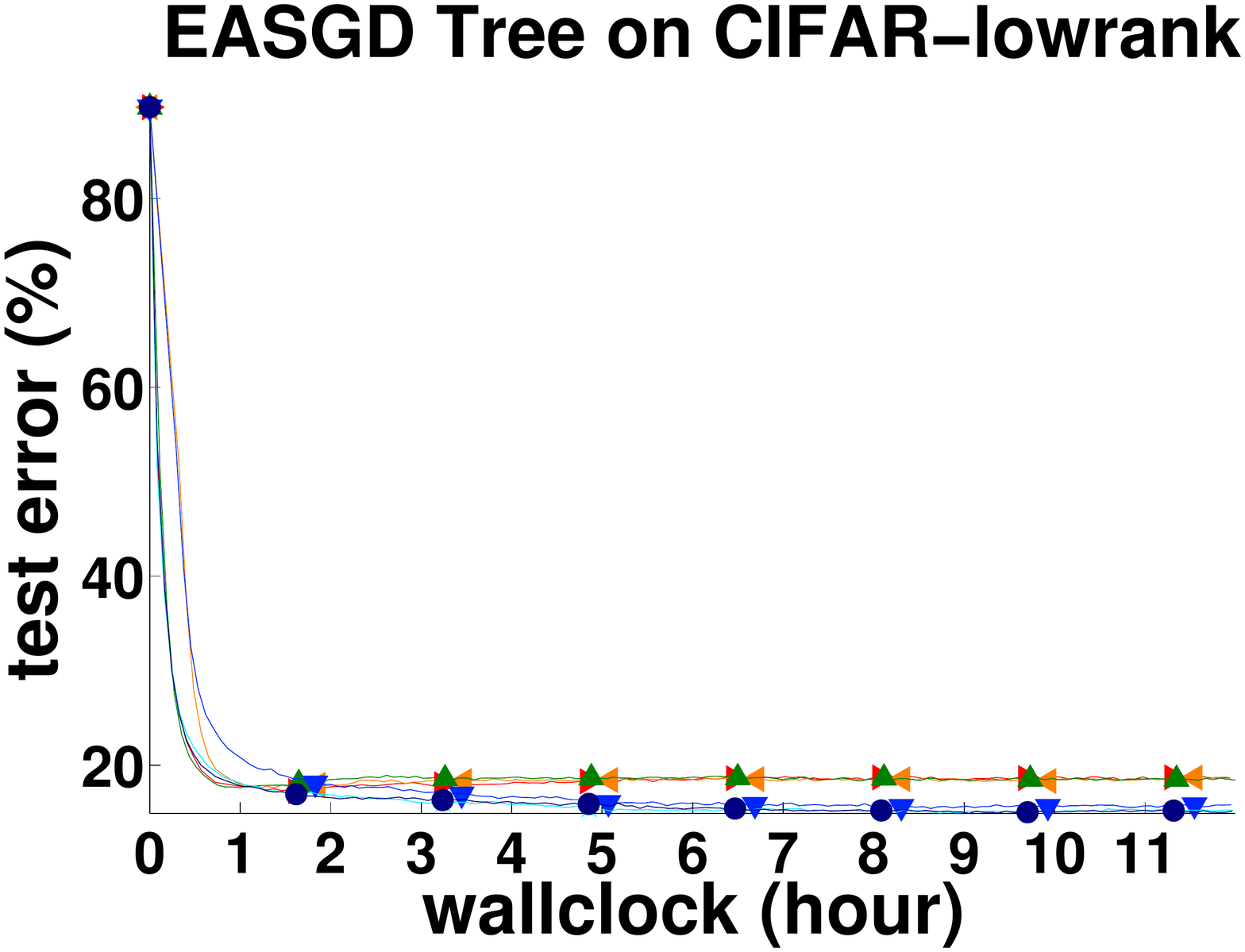}
\hspace{-0.3in}\includegraphics[trim=0cm 0cm 0cm 0cm,clip,width = 3in]{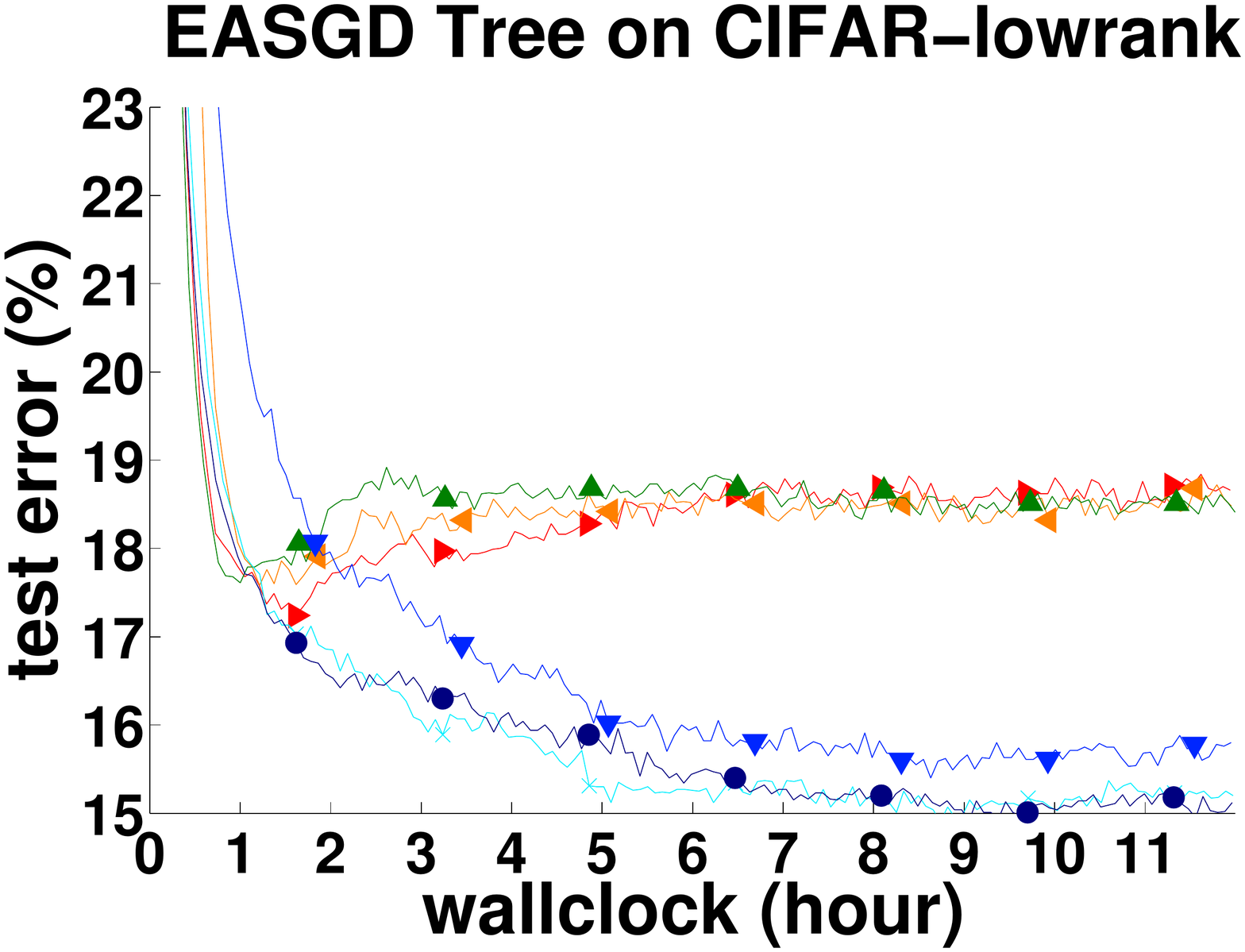}\\
\caption{\textit{EASGD Tree} on \textit{CIFAR-lowrank} using the first and second communication scheme with momentum. Training loss and error, test loss and error of the root node versus a wallclock time. The curve e in Figure~\ref{fig:CIFARlr_mix121}, the curve e in Figure~\ref{fig:CIFARlr_mix122}, the curve a in Figure~\ref{fig:CIFARlr_mix123}, the curve b in Figure~\ref{fig:CIFARlr_mix124}, the curve b in Figure~\ref{fig:CIFARlr_mix125} and the curve d in Figure~\ref{fig:CIFARlr_mix126} are selected and plotted.
}
\label{fig:CIFARlr_mb16}
\end{figure}

\begin{figure}[p]
\center
\includegraphics[trim=0cm 0cm 0cm 0cm,clip,width = 3in]{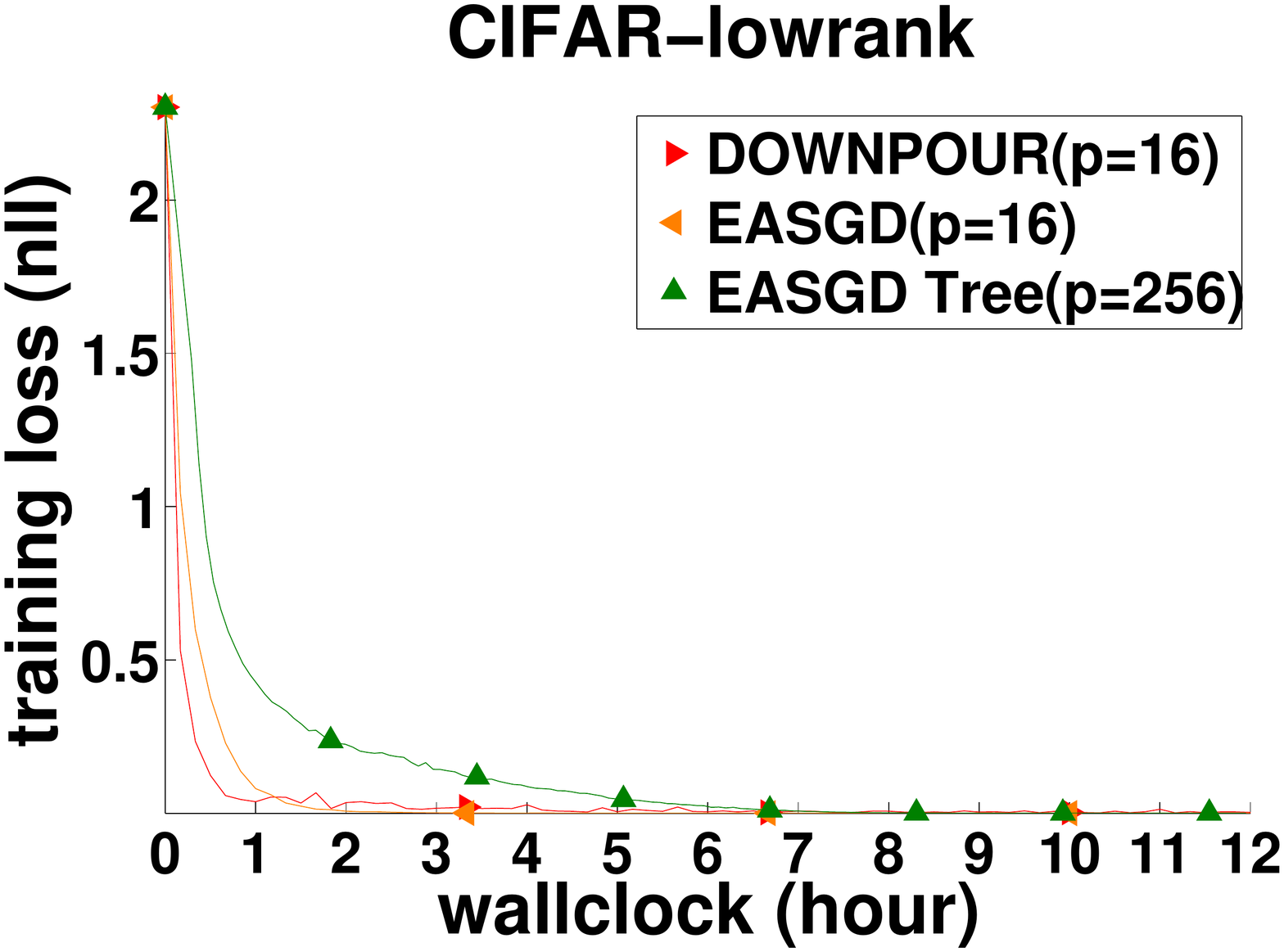}
\hspace{-0.3in}\includegraphics[trim=0cm 0cm 0cm 0cm,clip,width = 3in]{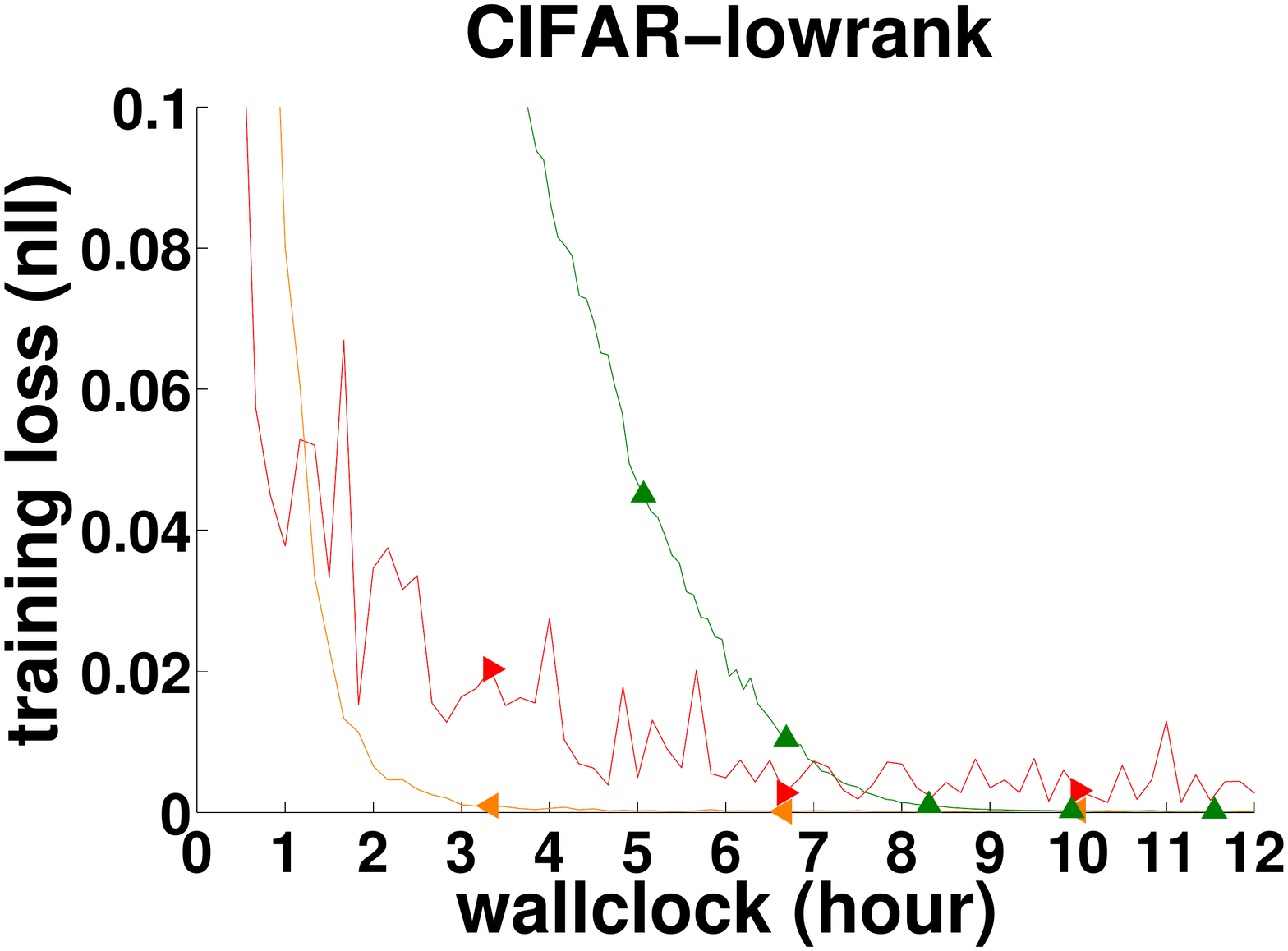} \\
\includegraphics[trim=0cm 0cm 0cm 0cm,clip,width = 3in]{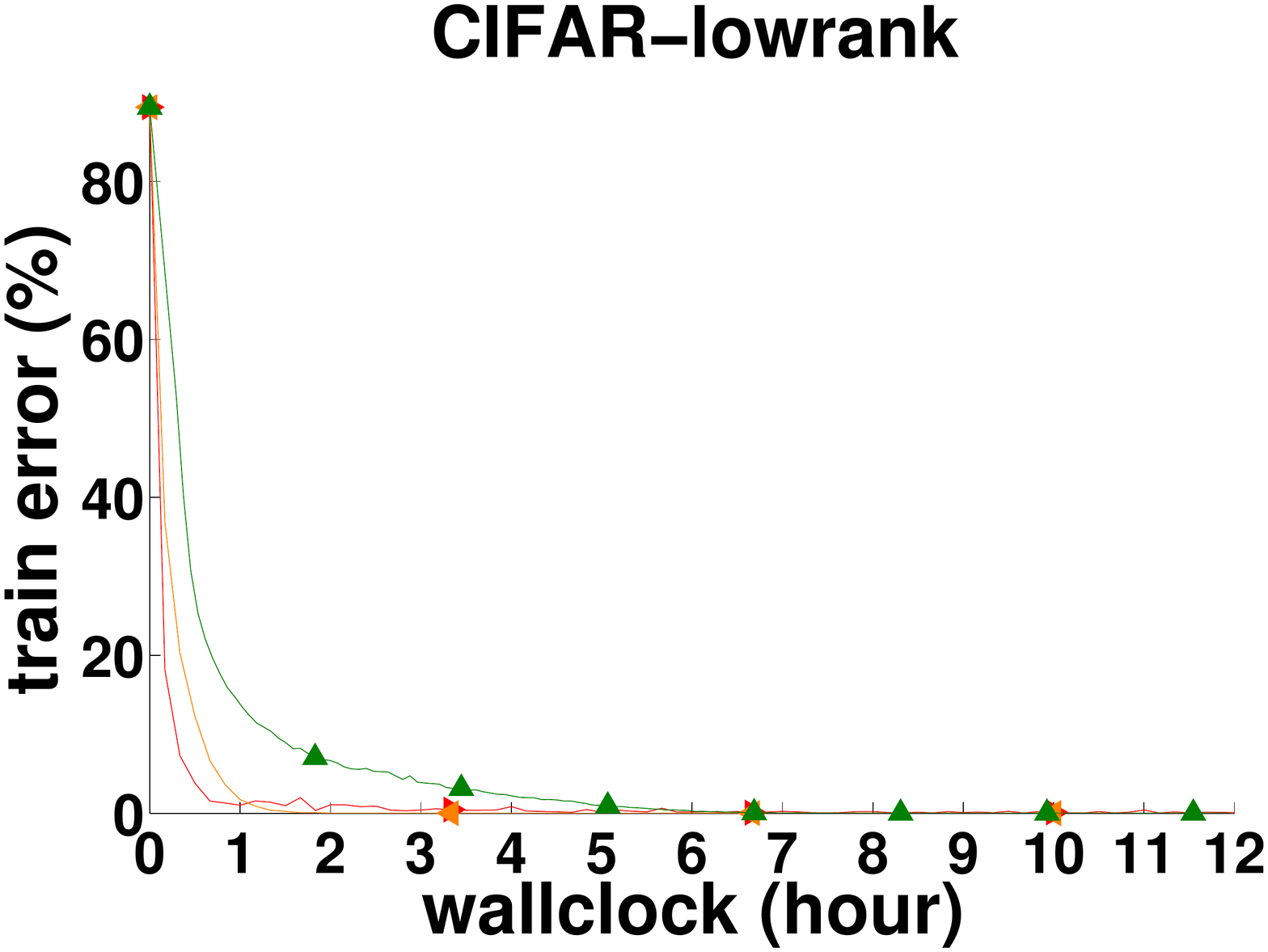}
\hspace{-0.3in}\includegraphics[trim=0cm 0cm 0cm 0cm,clip,width = 3in]{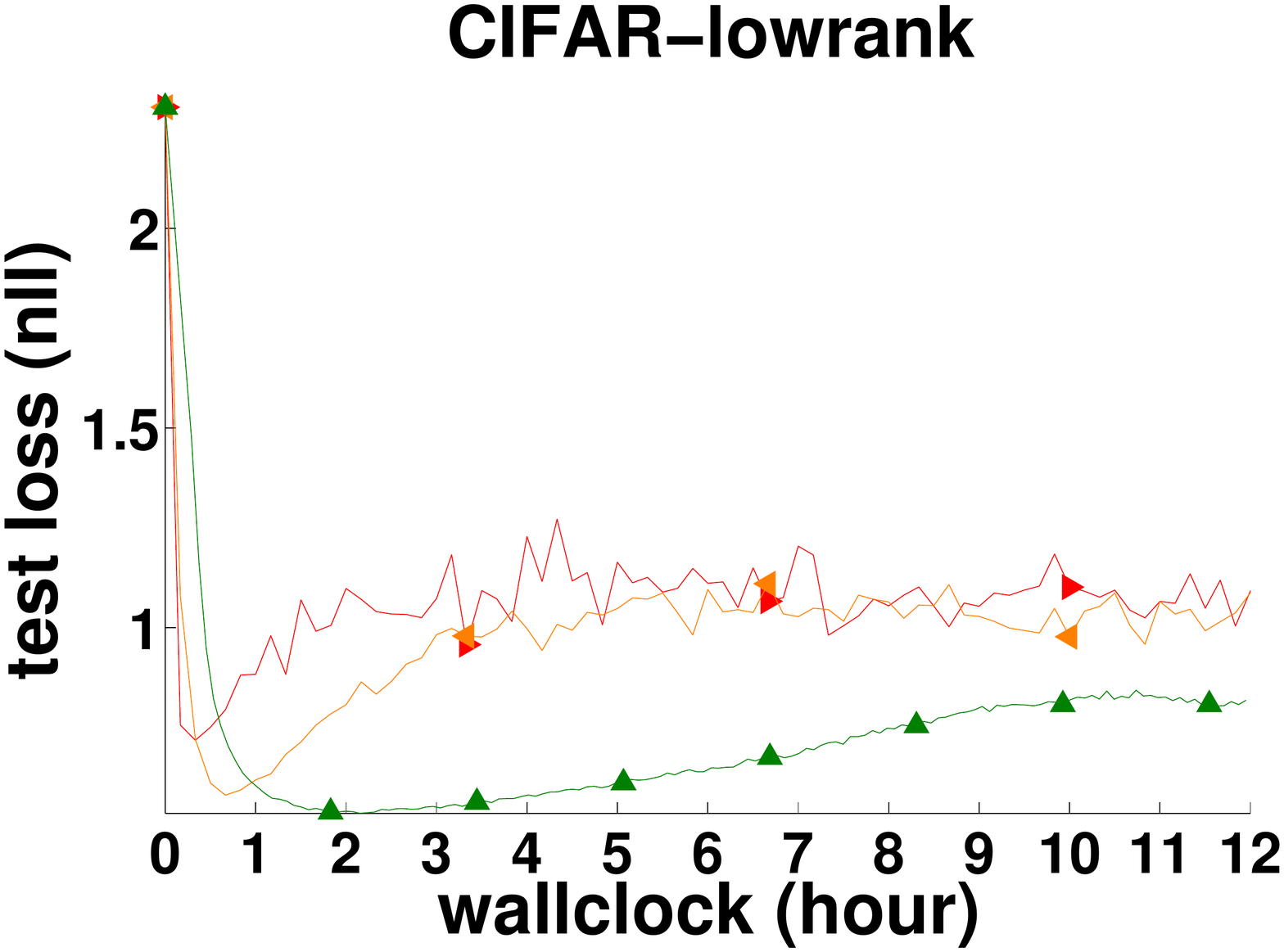} \\
\includegraphics[trim=0cm 0cm 0cm 0cm,clip,width = 3in]{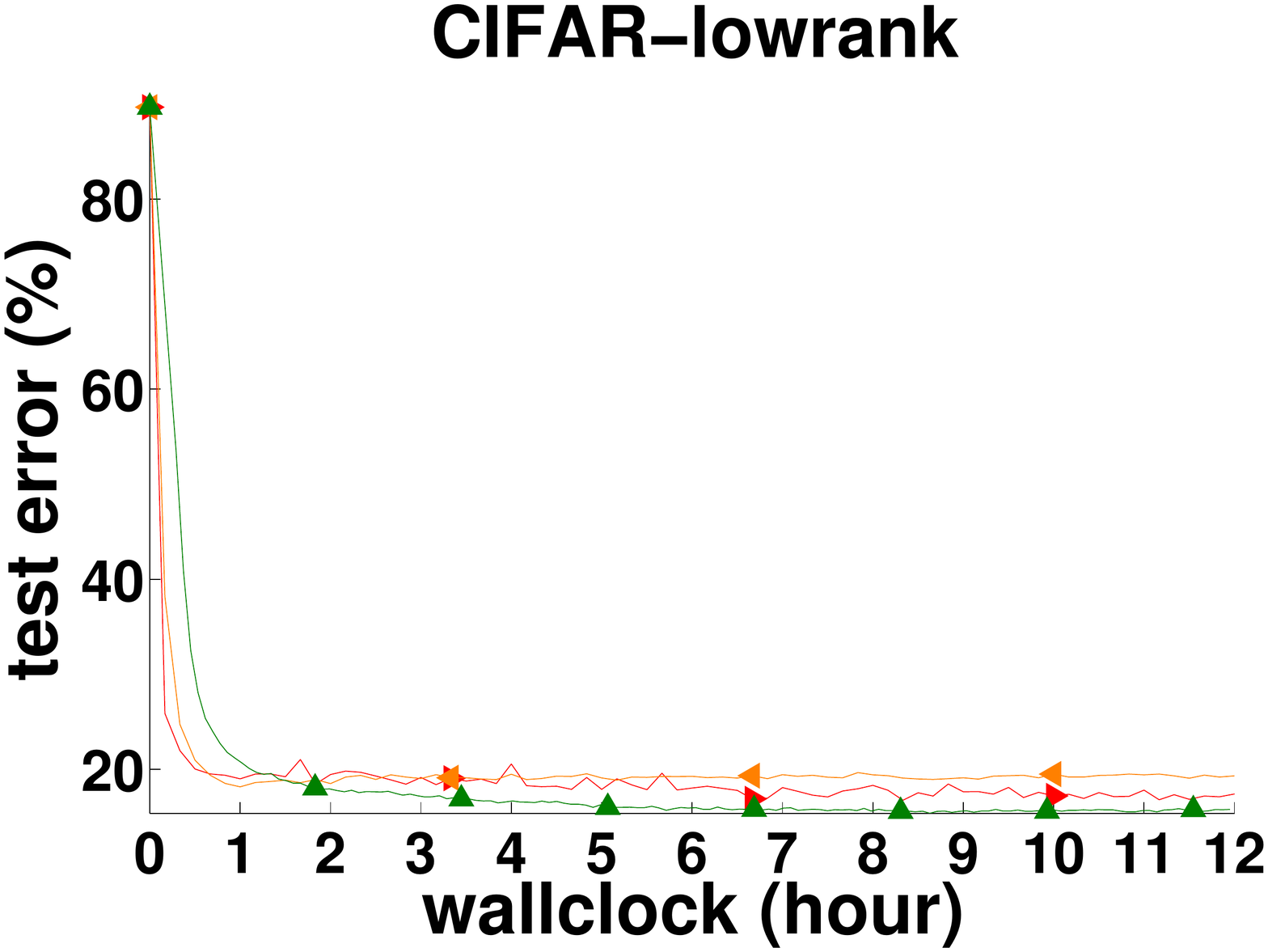}
\hspace{-0.3in}\includegraphics[trim=0cm 0cm 0cm 0cm,clip,width = 3in]{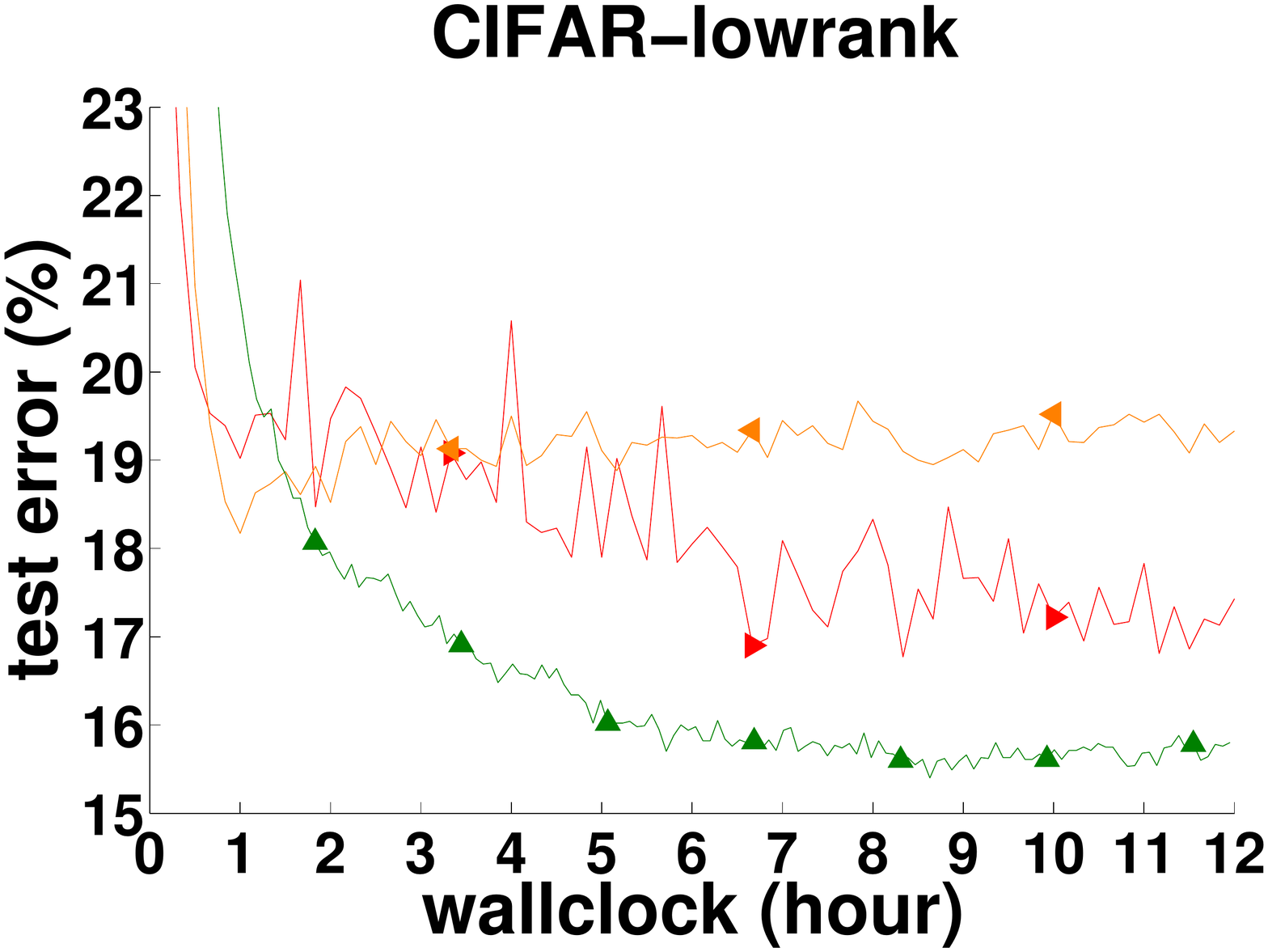}\\
\caption{Best test performance of \textit{DOWNPOUR} (p=16), \textit{EASGD} (p=16), and \textit{EASGD Tree} (p=256) on \textit{CIFAR-lowrank}. Training loss and error, test loss and error (of the center variable or the root node) versus a wallclock time. The momentum is not applied ($\delta=0$).
}
\label{fig:CIFARlr_final}
\end{figure}

Next we examine the effect of the momentum when using a mini-batch size of $16$. We use mini-batch as we have observed that the momentum can help more when the stochastic gradient variance is smaller (c.f. also Figure~\ref{fig:ch5_mul_msgd_sp2}). Moreover, it would be nice to see whether using a relatively small mini-batch size can still hurt the test performance. As in \textit{EAMSGD}, we can apply Nesterov's momentum to each of the leaf node during the gradient computation. Compared to the earlier results (without using mini-batch) in Figure~\ref{fig:CIFARlr_mix111} and~\ref{fig:CIFARlr_mix112}, we increase the total training time from $3$ hours to $12$ hours. This is because we do not have more CPU resources to parallelize the mini-batch computation. It takes 0.13sec/step in the minibatch case vs. 0.01sec/step in the case without using mini-batch.

In Figure~\ref{fig:CIFARlr_mix121},~\ref{fig:CIFARlr_mix122} and~\ref{fig:CIFARlr_mix123}, we report the results of the momentum in the first communication scheme. We see that without using momentum (i.e. $\delta=0$), the training process is still not very stable as before. Three out of the six runs have diverged.
Using momentum $\delta=0.9$ allows us to reduce the learning rate by a factor of ten (see also the discussion of Figure~\ref{fig:ch5_mul_msgd_sp2}), and gives a much more stable training process. Also they look almost identical, except for the two curves marked in e and f. This difference is due to a longer initialization of all the nodes. Using momentum $\delta=0.99$ is also very stable, though the results are more varied.

In Figure~\ref{fig:CIFARlr_mix124},~\ref{fig:CIFARlr_mix125} and~\ref{fig:CIFARlr_mix126}, we report the results of the momentum in the second communication scheme. Without using momentum is still not as stable. The curve a and c have stopped in the middle without any indication of divergence. But they did have diverged and ended with \textit{NaN}. Nevertheless, the curve b gives a smallest test error 15.4\%. The second communication scheme again leads to better test performance. Using momentum $\delta=0.9$ gives a smallest test error 14.91\% in curve b. Using momentum $\delta=0.99$ also gives a smallest test error 14.91\% in curve d. The most surprising observation is that the curve a has a large peak between hour 8 and 9. We have investigated the log and found that it is correlated with the network traffic.

We can now compare all the momentum results by selecting the best test performance curve in previous results. They are summarized in Figure~\ref{fig:CIFARlr_mb16}. It's very clear that the two communication schemes give very different performance. The first scheme gives better training performance, while the second scheme gives better test performance. The momentum no longer plays a significant role to determine the test performance as in \textit{EAMSGD}, but it is still helpful. To explain why still remains very open.

For completeness, we compare \textit{EASGD}, \textit{DOWNPOUR} with $p=16$, and the \textit{EASGD Tree} with $p=256$. Figure~\ref{fig:CIFARlr_final} shows the best test performance for each method, selected among a wide range of the learning rates.

\section{Unifying \textit{EASGD} and \textit{DOWNPOUR}}
\label{sec:unify}

It's beneficial to write down the updates of the \textit{DOWNPOUR} method (Algorithm~\ref{alg:asyncD} in Section~\ref{sec:addpcode}) in a synchronous scenario. For simplicity, let's assume that the communication period $\tau=1$. The global clock $t$ is shared among the local workers. Every step \textit{DOWNPOUR} performs the following updates to solve the original Problem~\ref{eq:original} in Chapter~\ref{sec:ch_intro}:
\begin{equation}
\begin{array}{r@{}l}
x^i_{t+1} &{}= \tilde{x}_t - \eta \nabla{f(\tilde{x}_t,\xi^i_t)}, \\
\tilde{x}_{t+1} &{}= \tilde{x}_{t} + \sum_{i=1}^p (x^i_{t+1} - \tilde{x}_t).
\end{array}
\label{eq:downpourS}
\end{equation}

Now compare with the \textit{EASGD} updates (from Equation~\ref{eq:movingaverage2} in Chapter $2$):
\begin{equation}
\begin{array}{r@{}l}
x^i_{t+1} &{}= x^i_t - \eta \nabla f( x_t^i,\xi^i_t) - \alpha (x^i_t - \tilde{x}_t), \\
\tilde{x}_{t+1} &{}= \tilde{x}_t + \frac{\beta}{p} \sum_{i=1}^p (x^i_t-\tilde{x}_t).
\label{eq:easgdS}
\end{array}
\end{equation}

We realize that it's possible to unify these two methods by drawing a parallel between the classical \textit{Jacobi} and \textit{Gauss-Seidel} methods. To motivate this connection, let's look back at the convergence result of \textit{EASGD}. Lemma~\ref{lem:lemm1} indicates that the divergence can happen if the condition~\ref{eq:condition} is not satisfied. One might wonder why there's such a stability condition. After all, \textit{EASGD} performs only the gradient descent and the moving averages. Each of these operations is contractive on its own! A short answer is that \textit{EASGD} is formalized in the \textit{Jacobi} form rather than the \textit{Gauss-Seidel} form in order to capture the delay of the parameter communication. We can transform the \textit{EASGD}'s \textit{Jacobi} form in Equation~\ref{eq:easgdS} to a \textit{Gauss-Seidel} form as follows:
\begin{equation}
\begin{array}{r@{}l}
x^i_{t+1/2} &{}= x^i_t - \alpha (x^i_t - \tilde{x}_t), \\
x^i_{t+1} &{}= x^i_{t+1/2} - \eta \nabla f( x_{t+1/2}^i,\xi^i_{t+1/2}) , \\
\tilde{x}_{t+1} &{}= \tilde{x}_t + \frac{\beta}{p} \sum_{i=1}^p (x^i_{t+1}-\tilde{x}_t).
\end{array}
\label{eq:easgdG}
\end{equation}

The first and the third step in Equation~\ref{eq:easgdG} is contractive if we choose $\alpha \in (0,2)$ and $\beta \in (0,2)$. Thus the stability condition~\ref{eq:condition} is too much a constraint in this case. Notice that the \textit{DOWNPOUR} method in Equation~\ref{eq:downpourS} is equivalent to the \textit{EASGD}'s \textit{Gauss-Seidel} form in Equation~\ref{eq:easgdG} with $\alpha=1$ and $\beta=p$. It is in this sense that \textit{DOWNPOUR} and \textit{EASGD} get connected! However, $\beta=p$ can be very large, how can it be stable?

For stability analysis, we consider again the quadratic case where $\nabla f(x,\xi) = x$. We can reduce the Equation~\ref{eq:easgdG} by introduing $\bar{x}_t = \frac{1}{p} \sum_{i=1}^p x^i_{t}$ as follows:
\begin{equation}
\begin{array}{r@{}l}
\bar{x}_{t+1/2} &{}= \bar{x}_t - \alpha ( \bar{x}_t - \tilde{x}_t), \\
\bar{x}_{t+1} &{}= \bar{x}_{t+1/2} - \eta \bar{x}_{t+1/2}, \\
\tilde{x}_{t+1} &{}= \tilde{x}_t + \beta (\bar{x}_{t+1}-\tilde{x}_t).
\end{array}
\label{eq:easgdGReduce}
\end{equation}

This is a linear system which can be written as $(\bar{x}_{t+1},\tilde{x}_{t+1})^T = M (\bar{x}_{t},\tilde{x}_{t})^T$, where
\begin{equation}
M = \left (
\begin{array}{cc}
1 & 0 \\
\beta & 1-\beta
\end{array} \right)
\left (
\begin{array}{cc}
1-\eta & 0 \\
0 & 1
\end{array} \right)
\left (
\begin{array}{cc}
1-\alpha & \alpha \\
0 & 1
\end{array} \right).
\label{eq:easgdGReduceM}
\end{equation}

We evaluate the two eigenvalues $z_1$ and $z_2$ of this matrix $M$, which should satisfy the equation $z^2 - 2b z + c =0$. One can check that
\begin{equation}
\begin{array}{r@{}l}
b &{}= \frac{1}{2} (2 - \beta - \eta - \alpha (1 - \beta)(1 - \eta)), \\
c &{}= (1-\eta)(1-\alpha)(1-\beta).
\end{array}
\label{eq:easgdGReduceMbc}
\end{equation}

We need a basic lemma to study the stability, i.e. when the absolute value of $z_1$ and $z_2$ are smaller than one.
\begin{lemma}
\label{lem:lemmS}
Let $z_1$ and $z_2$ be the two roots of the polynomial $z^2 - 2b z + c =0$ with real coefficients $b \in \mathbb{R}$ and $c \in \mathbb{R}$, then the sufficient and necessary condition for $|z_1| < 1$ and
$|z_2| < 1$ is
\begin{equation}
-1 < c < 1, \quad c > 2b -1, \quad c > -2b -1.
\label{eq:lemmScond}
\end{equation}
\end{lemma}
\begin{proof}
There are two cases to consider. The first one is the real-valued case, i.e. $b^2 > c$. The second one is the imaginary case, i.e. $b^2 \le c$. If $b^2 \le c$, then $|z_1| = |z_2| = c^2 < 1$ is equivalent to $-1 < c < 1$. If $b^2 > c$ then we can suppose $z_1 = b + \sqrt{b^2-c}$ and $z_2 = b - \sqrt{b^2-c}$. The condition $-1<z_2 < z_1<1$ is equivalent to $c > 2b -1$, $b<1$, $c>-2b-1$ and $b>-1$. Combining the condition from the two cases gives us the Equation~\ref{eq:lemmScond}.
\end{proof}

Applying the Lemma~\ref{eq:lemmScond} to the above case in Equation~\ref{eq:easgdGReduceMbc}, we obtain the stability condition for the reduced linear system~\ref{eq:easgdGReduce} as below,
\begin{equation*}
-1 < c < 1, \quad 0 < \beta \eta < 2(c+1),
\end{equation*}
where $c =(1-\eta)(1-\alpha)(1-\beta)$. It's easy to verify this since $2b = c + 1 - \beta \eta$.

We summarize the obtained stability condition (using Mathematica) in the following theorem.
\begin{theorem}
\label{lem:thmS}
Assume $\eta>0$, $\alpha>0$ and $\beta>0$, the stability condition for the linear system~\ref{eq:easgdGReduce} is equivalent to union of the following set of conditions:
\begin{itemize}
\item $0<\eta<1$, $0 < \beta < 1$ and $0 < \alpha < c_2$,
\item $0<\eta<1$, $\beta = 1$ and $ \alpha > 0$,
\item $0<\eta<1$, $1 < \beta \le 2$ and $0 < \alpha < c_3$,
\item $0<\eta<1$, $2 < \beta < 4/\eta$ and $c_2 < \alpha < c_3$,
\item $\eta=1$, $0 < \beta < 2$ and $ \alpha >0$,
\item $1< \eta < 2$, $0 < \beta < 1$ and $ 0 < \alpha <c_3$,
\item $1< \eta < 2$, $\beta =1$ and $ \alpha >0$,
\item $1< \eta < 2$, $1< \beta <2$ and $ 0<\alpha <c_2$,
\item $2 \le \eta \le 4$, $0< \beta <1$ and $c_2<\alpha <c_3$,
\item $ \eta > 4$, $0< \beta < 4/\eta$ and $c_2<\alpha <c_3$,
\end{itemize}
where \[c_2 = \frac{4 - 2 \beta - 2 \eta + \beta \eta}{2 - 2 \beta - 2 \eta + 2 \beta \eta}, \quad c_3 = \frac{-\beta - \eta + \beta \eta}{1 - \beta - \eta + \beta \eta}.\]
\end{theorem}

There is only one condition above which satisfies the \textit{DOWNPOUR} method when $p>2$. This condition is $0<\eta<1$, $2 < \beta < 4/\eta$ and $c_2 < \alpha < c_3$. It means that if $\beta=p$ is very large, we need to use quite small learning rate such that $0<\eta < 4/p$. Moreover, $\alpha$ has to stay very close to $1$ (i.e. forgetting all the local memory). This is because $c_2$ tends to $1$ as $\beta=p$ tends to infinity (assume $\beta \eta<4$ always holds); and $c_3$ tends to $1$ as well. For example, this condition does not hold for $\alpha=0.9$ and $\beta=16$, for any $\eta>0$. This is rather surprising: the stability condition for $\alpha=1$ and $\beta=16$ is $0<\eta<1/8$. However, there's no $\eta>0$ for $\alpha=0.9$ and $\beta=16$ to be stable. This suggest \textit{DOWNPOUR} is a very peculiar singleton in the class of the methods defined by Equation~\ref{eq:easgdG}. \textit{EASGD} has much more flexibilities as it operates in the region where $\beta \in (0,2)$. In practice, we have observed that the case $\beta=1,0<\eta<2,\alpha>0$ works well. It is very dangerous to operate in the region of $\eta>2$ when $\alpha$ is small.

\newpage
\chapter{Conclusion}
\label{chap:conclusions}

In this thesis, we focused on the problem of training large-scale deep learning models in a parallel and distributed computing environment.

Our starting point is a reformulation of the global variable consensus problem into an unconstrained optimization problem using a quadratic penalty between each local variable and the center variable. By reinterpreting the gradient of the quadratic penalty as an averaging process, we introduced the Elastic Averaging SGD (\textit{EASGD}) method. The \textit{EASGD} maintains the stochastic nature of the sequential \textit{SGD} method through a weak coupling between the local variables and the center variable. Each local variable is optimized using an \textit{SGD}-based method.
The averaging process attracts the local variable and the center variable toward each other so that all of them move toward a local optimum. The center variable can slowly track the spatial average of the local variables, thus having a smaller variance.
We then discussed how to extend the \textit{EASGD} method to the asynchronous scenario, and how to accelerate it using the momentum.

We proceeded to study the convergence rate the \textit{EASGD} method in terms of the center variable.
In the synchronous scenario, we discussed the variance reduction effect of the \textit{EASGD} method by increasing the number of local variables, for both quadratic and strongly-convex case. We also introduced a double averaging sequence in the spirit of the \textit{Polyak}'s averaging and showed that it is indeed asymptotically optimal. We then compared the stability region of the \textit{EASGD} method with the \textit{ADMM} method in the round-robin scheme, where we have found quite unusual instability region for the \textit{ADMM} method.

Having studied the (local) convergence property of the \textit{EASGD} method, we applied
its asynchronous extension to train deep learning models. We focused on the \textit{CIFAR-10} and \textit{ImageNet ILSVRC 2013} dataset for image classification in a supervised learning setting.
The (global) convergence behavior of the \textit{EASGD} method was simulated starting from a random initialization. We found that the \textit{EASGD} method accelerates the training of the baseline~\textit{SGD} method. Moreover, it is very communication efficient compared to the asynchronous \textit{SGD} method \textit{DOWNPOUR}. After combining with \textit{Nesterov}'s momentum method, we obtained \textit{EAMSGD} and it achieved even better test accuracy. We also discussed the tradeoff between the data communication and the parameter communication. In our cases, it turned out to save the total bandwidth a lot by using more data communication for less parameter communication (recall \textit{EASGD} method needs to sample the whole dataset in order to avoid the bias of the stochastic gradient).

Based on the empirical results we have obtained,
we asked what is the limit of the speedup by increasing the number of processors.
We studied first an additive noise model to capture the asymptotic behavior of various stochastic optimization method, then we proposed a multiplicative noise model to capture the initial behavior of these methods.
For the additive noise case, we first studied the \textit{SGD} method with mini-batch. We saw that increasing the mini-batch size (i.e. number of processors) can reduce the asymptotic variance, but it is not able to increase the convergence speed. We then studied the momentum \textit{SGD} in the \textit{Nesterov}'s form, and showed that faster convergence speed is possible, but that will lead to larger asymptotic variance. We have also generalised the \textit{EASGD} method by decoupling (de-symmetrize) the elastic averaging between the center variable and the local variables, and found that the \textit{EASGD} method can behavior like a momentum \textit{SGD} with a negative moving average rate in some optimal sense (i.e. the center variable pushes the local variables instead of pulling). For the multiplicative noise case, we have found that \textit{SGD} method with mini-batch can greatly improved the initial convergence speed by choosing the optimal (larger) learning rate. However, the momentum \textit{SGD} is not as effective in this case. The \textit{EASGD} method is also slower than \textit{SGD} with mini-batch, but its stability region is still improved by increasing the number of processors. We showed that given an infinite number of processors,
the improvement for stability region of \textit{EASGD} depends on the spread of the Gamma distribution of the input data. The larger the spread, the more the improvement.

Back to our starting point, we studied an interesting non-convex case for the \textit{EASGD} method based on the an unconstrained objective we have introduced. We have discussed the stability of the critical points. We found that when the quadratic penalty is too small (i.e. the coupling between the center variable and the local variable is too weak), the \textit{EASGD} method can introduce a stable local optimum trapped by a saddle point.

Knowing the theoretical limitation of the speedup, we scale up the \textit{EASGD} by using a tree structured network topology. We showed empirically that \textit{EASGD Tree} can further accelerate the training speed, but no longer in a linear speedup region. This is somehow predicted by our analysis in the multiplicative noise model. The surprising observation is \textit{EASGD Tree} can still yield better test accuracy when the local variables fluctuate further from the center variable. Moreover, inspired by the design of the \textit{EASGD Tree} in the asynchronous scenario, we carefully distinguished the difference between the \textit{Jacobi} form of \textit{EASGD} with its \textit{Gauss-Seidel} form, which in turn unified the \textit{EASGD} and the \textit{DOWNPOUR} method in the synchronous scenario.

We conclude with a few open problems from convex optimization, non-convex optimization, distributed and parallel computing, statistical and deep learning.

\begin{itemize}
\item The smoothing property of \textit{EASGD} for non-smooth convex optimization.
We have discussed the connection between
the quadratic penalty term in Equation~\ref{eq:penalty} and the Moreau-Yosida regularization.
Can one expect a good rate of convergence?

\item The connection between \textit{EAMSGD} and \textit{Katyusha}~\cite{allen2016katyusha}. \textit{Katyusha} is an accelerated variance reduction method. The speedup of its acceleration is dragged back by a center variable so as to achieve a smaller variance. This resembles the role of the center variable in \textit{EAMSGD}.
\item In the simulation of \textit{EASGD Tree},
we have seen that network congestion can occur. Is there a good way to
detect when the network becomes slow so as to avoid the potential congestion?
This is an adaptive learning rate problem. It is not very simple as we have seen in Chapter~\ref{sec:ch_limit} that the optimal moving rate can sometime be positive, but sometime be negative.
\item The convergence analysis of the \textit{EASGD Tree} is still missing. The basic tool~\cite{tsitsiklis1984distributed} based on the mean-field method (i.e. treating each node the same) seems to be too macroscopic to capture the global (e.g. multi-scale variance reduction) behavior of the \textit{EASGD Tree}.
\item The description of the asynchronous behavior using vector-clock~\cite{singhal1992efficient}. Although we have unified \textit{EASGD} and \textit{DOWNPOUR} in the synchronous scenario, it is still not clear how to measure their behavior in the asynchronous scenario and in the real-time workloads. The idea of using vector-clock from the distributed computing is worthing exploring in the future.
\item How non-convex is the energy landscape? A related question may be what to do when \textit{EASGD} is trapped by a saddle point. Also why there is a such a big difference between \textit{ASGD} and \textit{MVASGD} as we have studied in Figure~\ref{fig:ImageNet_supp}?
\item What is the role of the oscillation in deep learning to achieve better test performance?
We have observed that both
\textit{EAMSGD} and \textit{EASGD Tree} can achieve better test performance when the oscillation of the local variables is sufficiently large. Looking back to the Dropout~\cite{JMLR:v15:srivastava14a} and DropConnect~\cite{DBLPWanZZLF13} regularization, is there any deeper reason in common~\cite{sherratt2016oscillation}?
\end{itemize}

\newpage

% \newpage
% \cleardoublepage
% \phantomsection
% \addcontentsline{toc}{chapter}{Appendices}
% \input appendix
% \newpage

\bibliographystyle{ieee}
\cleardoublepage
\phantomsection
\addcontentsline{toc}{chapter}{Bibliography}
\bibliography{ref}

\end{document}